\documentclass[runningheads]{llncs}
\usepackage{graphicx}
\usepackage{textcomp} 
\usepackage{tikz}
\usepackage{comment}
\usepackage{amsmath,amssymb} 
\usepackage{color}

\usepackage[accsupp]{axessibility}  

\usepackage{cite}
\usepackage[width=122mm,left=12mm,paperwidth=146mm,height=193mm,top=12mm,paperheight=217mm]{geometry}

\usepackage{booktabs}
\usepackage{subcaption}
\usepackage{mathtools}
\usepackage{overpic}

\definecolor{orange}{rgb}{0.8, 0.6, 0.2}

\DeclareCaptionLabelFormat{andtable}{#1~#2  \&  \tablename~\thetable}

\begin{document}
\pagestyle{headings}
\mainmatter
\def\ECCVSubNumber{2921}  

\title{PRIF: Primary Ray-based Implicit Function} 

\titlerunning{PRIF: Primary Ray-based Implicit Function}
%
\author{Brandon Yushan Feng\inst{1} \and
Yinda Zhang\inst{2} \and
Danhang Tang\inst{2} \and
Ruofei Du\inst{2} \and
Amitabh Varshney\inst{1}}
\authorrunning{B. Feng et al.}
%
\institute{University of Maryland, College Park~\and~Google Research}
\maketitle

\begin{abstract}
We introduce a new implicit shape representation called Primary Ray-based Implicit Function (PRIF).
In contrast to most existing approaches based on the signed distance function (SDF) which handles spatial locations, our representation operates on oriented rays.
Specifically, PRIF is formulated to directly produce the surface hit point of a given input ray, without the expensive sphere-tracing operations, hence enabling efficient shape extraction and differentiable rendering. 
We demonstrate that neural networks trained to encode PRIF achieve successes in various tasks including single shape representation, category-wise shape generation, shape completion from sparse or noisy observations, inverse rendering for camera pose estimation, and neural rendering with color.
\end{abstract}

\definecolor{ToRephaseColor}{rgb}{1.0, 0.4, 0.0}
\definecolor{TodoColor}{rgb}{1.0, 0.0, 1.0}
\definecolor{GreenColor}{rgb}{0.0, 0.0, 1.0}
\definecolor{PurpleColor}{rgb}{0.5, 0.0, 0.5}
\definecolor{BlueColor}{rgb}{0.0, 0.0, 1.0}
\definecolor{BabyBlueColor}{rgb}{0.5, 0.5, 1}
\definecolor{RevisionFixedColor}{rgb}{0.6, 0.0, 0.4}
\definecolor{FinalColor}{rgb}{0.0, 0.0, 0.0}

\def\sectionautorefname{Section}%
\def\subsectionautorefname{section}%
\def\subsubsectionautorefname{section}%

\newcommand{\bmat}[1]{\begin{bmatrix}#1\end{bmatrix}}
\newcommand{\pmat}[1]{\begin{pmatrix}#1\end{pmatrix}}

\newcommand{\todo}[1]{{\color{TodoColor} [TODO: #1]}}
\newcommand{\torephrase}[1]{ {\color{ToRephaseColor} [ToRephrase:] #1} }
\newcommand{\du}[1]{\textcolor[rgb]{0.46,0.42,0.69}{{[Ruofei: #1]}}}
\newcommand{\ruofei}[1]{\textcolor[rgb]{0.46,0.42,0.69}{{[Ruofei: #1]}}}
\newcommand{\revis}[1]{{\color{BlueColor}#1}}
\newcommand{\yd}[1]{\textcolor[rgb]{1,0.6,0.1}{{[YZ: #1]}}}
\newcommand{\dt}[1]{\textcolor[rgb]{1,.1, 0.6}{{[DT: #1]}}}

\newcommand{\etal}{{\em et al.}}
\newcommand{\eg}{{\em e.g.}}
\newcommand{\degree}{\ensuremath{^{\circ}} }
\newcommand{\ceil}[1]{{\lceil #1 \rceil}}
\newcommand{\floor}[1]{{\lfloor #1 \rfloor}}

\newcommand{\Log}[1]{\log\left(#1\right)}
\newcommand{\NormTwo}[1]{\left\lVert#1\right\rVert_{2}}
\newcommand{\Norm}[1]{\left\lVert#1\right\rVert}
\newcommand{\mathbfit}[1]{\textbf{\textit{#1}}}
\newcommand{\Origin}{\mathbfit{O}}

\section{Introduction}
Learning an accurate and efficient geometric representation of a 3D object is an important problem for computer graphics, computer vision, and robotics.
Recent advances in machine learning have inspired a growing trend of implicit neural shape representations, where a neural network learns to predict the signed distance function (SDF) for an arbitrary location in the 3D space.
Moreover, in addition to the 3D location $(x, y, z)$, the neural SDF network may take in a latent vector that describes the object identity, thus enabling the generative modeling of multiple objects.
Such an implicit neural representation (INR) not only produces fine-grained geometry, but also enables a plethora of applications \cite{Liu2020DIST}, \eg, shape completion, pose estimation, via a differentiable rendering based optimization.

However, rendering and extracting the shape from a trained neural SDF network are computationally expensive and often limited to watertight shapes.
The direct approach to rendering from SDF requires sphere tracing, which needs access to the SDF values at multiple locations along each pixel ray~\cite{Hart1996Sphere}.
The indirect approach computes and stores the SDF values at predefined 3D grid points, from which the shape can be rendered with sphere-tracing or extracted as a polygon mesh through Marching Cubes~\cite{lorensen1987marching}.
Both cases demand a large number of network evaluations since they require sampling SDF values at locations far away from the surface.
Moreover, the shape quality is ultimately constrained by the converging criteria of sphere tracing or the 3D resolution of the grid.

\begin{figure}[t]
	\centering
	\includegraphics[width=\columnwidth]{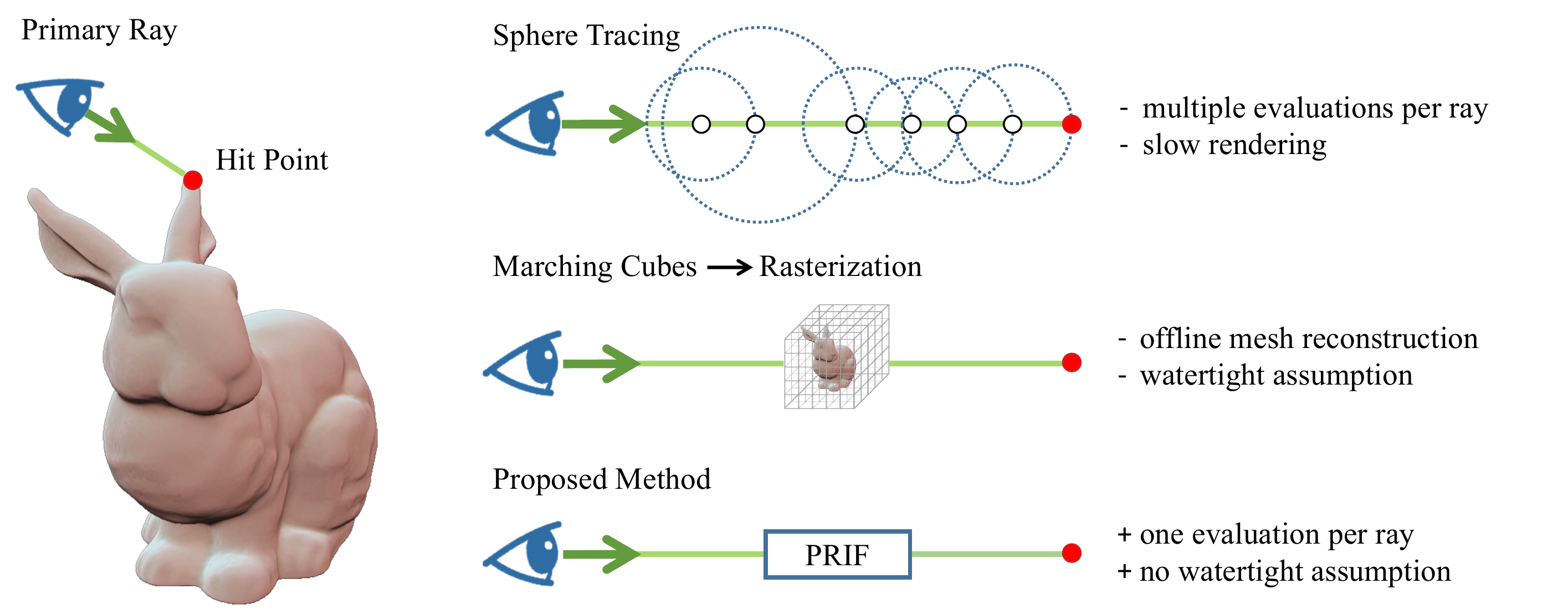}
	\caption{{\bf Overview.} Common neural shape representation methods learn the level-set functions implicitly describing the object geometry, such as the signed distance functions (SDF) and occupancy functions (OF). Rendering from these implicit networks requires either sphere tracing or rasterizing a separately extracted mesh. Sphere tracing is inefficient since it requires multiple network evaluations for a single ray. Rasterization induces a separate meshing step (often through Marching Cubes), which hinders end-to-end gradient propagation in differentiable applications, and the shape quality is ultimately restricted by the meshing algorithm's limitation, such as grid resolutions and shape watertightness. Our new representation, PRIF, directly maps each primary ray to its hit point. A network encoding PRIF is more efficient and convenient for rendering, since it requires only one evaluation for each ray, avoids the watertight constraint in conventional methods, and easily enables differentiable rendering.}
	\label{fig:IntroTeaser}
\end{figure}

In this work, we present a novel geometric representation that is efficient, accurate, and innately compatible for downstream tasks involving reconstruction and rendering.
We break away from the conventional point-wise implicit functions and propose to encode 3D geometry into a novel ray-based implicit function.
Specifically, our representation operates on the realm of oriented rays $r = (\textbf{p}_{r}, \textbf{d}_{r})$, where $\textbf{p}_{r} \in \mathbb{R}^{3}$ is the ray origin and $\textbf{d}_{r} \in \mathbb{S}^{2}$ is the normalized ray direction.
Unlike SDF that only outputs the distance to the nearest but undetermined surface point, we formulate our representation such that its output directly reveals the surface hit point of the input ray.
The rays whose surface intersection we care about are specifically known in computer graphics as \textit{primary} rays, whose origins are the rendering viewpoint, as opposed to \textit{secondary} rays that originate from the object surface \cite{Akenine2019Real}. Therefore, we name our representation as Primary Ray-based Implicit Function (PRIF).
In effect, a neural network trained to encode PRIF represents the manifestation of the object's geometry at any viewpoint that it is observed.

Modeling the object from the ray-based perspective, rather than 3D-point-based, has huge implications for efficient application in any task that involves rendering the object from a viewpoint.
In Section~\ref{section:experiments}, we show that PRIF outperforms common functions such as SDF and OF, and we further demonstrate successful applications of PRIF to various tasks using neural networks.

Properly formulating the rays is nontrivial.
While it may be intuitive to let PRIF output the distance from the ray origin $\textbf{p}_{r}$ to the hit point $\textbf{h}_{r}$, this formulation leads to ray aliasing - \textit{i.e.}, if we move $\textbf{p}_{r}$ along the ray direction $\textbf{d}_{r}$, the distance to the hit point $\textbf{h}_{r}$ needs to change.
However, the actual surface intersection point would not change to translation along the view direction.
Therefore, it is undesirable to have such a potential variance as it adds unnecessary complexity to the network output.

To avoid the aliasing, we reparametrize the ray $r$ by replacing the view position $\textbf{p}_{r}$ with  perpendicular foot $\textbf{f}_{r}$ from the coordinate origin~\Origin~to $r$.
Wherever we move $\textbf{p}_{r}$ along the ray direction $\textbf{d}_{r}$, the perpendicular foot $\textbf{f}_{r}$ stays the same.
Furthermore, we can easily define the surface hit point $\textbf{h}_{r}$ by a single scalar value: its distance from $\textbf{f}_{r}$.
Thus, we formulate PRIF so that it outputs this distance for an input ray.
More details are in Section~\ref{section:method}.

In summary, our main contributions are the followings:

\renewcommand{\labelitemi}{$\bullet$}
\begin{itemize}

\item We present PRIF, a novel formulation for geometric representation using ray-based neural networks.

\item We show that PRIF outperforms common level-set-based methods in shape representation accuracy and extraction speed.

\item We demonstrate that PRIF enables generative modeling, shape completion, and camera pose estimation.

\end{itemize}
\section{Related Work} \label{section:related}
We discuss prior art on neural representations for 3D shapes and scene rays.

\subsection{3D Shape Representations}
\subsubsection{Functional Representations.} Traditional 3D shape representations include polygon meshes, point clouds, and voxels.
In recent years, as deep neural networks achieve remarkable success on various vision-related tasks, there has been a growing interest in developing implicit neural representations (INRs) of 3D shapes.
Following seminal works~\cite{chen2019learning, park2019deepsdf, mescheder2019occupancy} showing successful applications of neural network to encode 3D shapes, many methods have been introduced to solve various vision and graphics tasks using INRs of 3D shapes~\cite{peng2020convolutional, Chibane2020ImplicitFI, Liu2020DIST, niemeyer2020differentiable, zakharov2020autolabeling, jiang2020sdfdiff, duan2020curriculum, sitzmann2020metasdf, saito2019pifu, xu2019disn, atzmon2020sal, chibane2020implicit, liu2019learning, driess2022learning, chen2021multiresolution, tang2020deep, simeonovdu2021ndf, bhatnagar2020combining, yang2021geometry}.

INRs usually use the multilayer perceptron (MLP) architecture to encode geometric information of a 3D shape by learning the mapping from a given 3D spatial point and a scalar value.
Typically, the output scalar value denotes either the occupancy at the given point, or the signed distance from the given point to the nearest point on the shape.
On one hand, networks that are trained to encode the occupancy function~\cite{mescheder2019occupancy} (OF) essentially learns the binary classification problem, where the output equals 0 if the point is empty, and equals 1 if occupied.
Therefore, the decision boundary where the network predictions equal to 0.5 represents the surface of the encoded shape.
On the other hand, for networks trained to encode the signed distance function~\cite{chen2019learning, park2019deepsdf} (SDF), the surface is represented by the decision boundary where the network predictions equal to 0.
A 3D surface determined in such fashions is also known as an isosurface, which is a level set of a continuous 3D function.
In practice, however, obtaining 3D meshes from these isosurfaces extracted from INRs still requires an additional meshing step often through the Marching Cubes algorithm.

In this paper, we introduce a new shape representation which is not determined by an isosurface.
Instead, our representation encodes a 3D shape by learning the PRIF associated with the shape.
Outputs of such a function directly correspond to points on the surface, and the shape can be extracted without an additional meshing step that could inject inaccuracies to the final representation.

\subsubsection{Global \textit{v.s.} Local Representations}
The initial works of INRs for 3D shapes inspire many techniques to improve its the rendering efficiency~\cite{Liu2020DIST, yariv2021volume} and representation quality~\cite{sitzmann2020implicit, tancik2020fourier}. 
Among many techniques, a common thread is the idea of spatial partitions which manifest in two main approaches.
One approach divides the surfaces of shapes into different local patches, reducing the difficulty of globally fitting a complex surface with a single network~\cite{tretschk2020patchnets, genova2019learning, jiang2020local, genova2020local, chabra2020deep, Paschalidou2021NeuralPL}.
Another approach divides the 3D volume into small local regions (often based on the spatial octree structure), and then train INRs to encode the geometric information within each local region~\cite{takikawa2021neural, yao20213d, chen2020bsp, deng2020cvxnet, martel2021acorn, lindell2021bacon, muller2022instant, saragadam2022miner}.

In this paper, we only focus on global representations, where a shape is represented by a single network without any spatial partitions.
While the aforementioned works largely focus on improving the performance of INRs based on SDF, our main contribution is a new functional representation in place of isosurface-based representations like SDF.
Nonetheless, the idea of spatial partitions explored in previous works has the potential to improve the performance of our currently global representation, and we see local specializations as a promising future direction to explore.

\subsection{Ray-based Neural Networks}
Our method is closely related to an emerging concept called Ray-based Neural Networks.
Rays are a common construct from computer graphics, and can be denoted by a tuple consisting of a 3D point (ray origin) and a 3D direction (ray direction).
Due to their closeness to rendering 3D scenes, rays have become a central component in the problem formulation of many recent works using neural networks to model 3D scenes~\cite{Mildenhall2020NeRF, sitzmann2019scene}.
Feng~\etal~\cite{feng2021signet} and Attal~\etal~\cite{attal2021learning} have further demonstrated that for front-facing light field scenes, MLPs can be trained to accurately map camera rays to their observed colors in highly detailed real-world scenes, but their networks only consider rays restricted within two parallel planes.
Most recently, Sitzmann~\etal~\cite{sitzmann2021light} successfully train a MLP to encode the observed color of unrestricted rays with arbitrary origins and directions, and the key is to parametrize rays using the Plucker coordinates~\cite{JiaPlucker2020}.

We similarly adopt rays as input to the neural network.
However, instead of the Plucker parametrization which replaces the ray position by its moment vector about the origin, we replace the moment vector by the perpendicular foot between the ray and the origin. 
As discussed in Sec.~\ref{section:method}, our formulation allows the network to simply produce an affine transformation to its input.

\begin{figure}[t]
	\centering
	\includegraphics[width=\columnwidth]{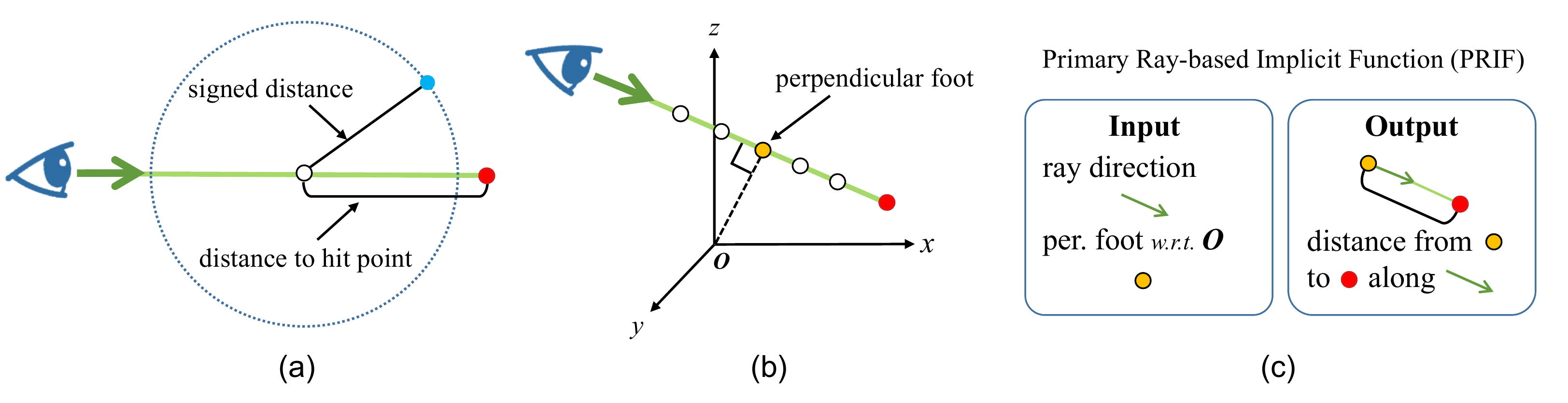}
	\caption{{\bf Representation.} (a) Signed distance at a sampling position (white) reveals the sphere (blue dots) where its nearest surface point (blue) exists, but that may be irrelevant when we really want to know the hit point (red) \textit{along a specific direction}. Thus, multiple samples are often required. (b) Our new representation uses only one sample (yellow) along the ray to obtain its surface hit point. The sampling position is the perpendicular foot between the given ray and the coordinate system's origin~\Origin. (c) The proposed function takes in the ray's direction and its sampling point, and returns the distance from that point to the actual surface hit point. We train neural networks to encode geometry by learning this new function, and we demonstrate the capability of our trained networks in various tasks involving shape representation and rendering.}
	\label{fig:IllustratePRIF}
\end{figure}

\section{Method} \label{section:method}
In this section, we introduce our new representation for 3D shapes. We first discuss the motivation behind ray parametrization used in prior works, and then we present our new formulation specifically designed for shape representations. 
\subsection{Background}
Sitzmann~\etal~\cite{sitzmann2021light} encode light fields by training an MLP $\Phi_{\phi}$ with parameters $\phi$ on a set of observed rays to learn the mapping of $r = (\textbf{p}_{r}, \textbf{d}_{r}) \to c_{r}$, where $c_{r}$ denotes the color of the observed radiance.
However, naively concatenating $\textbf{p}_{r}$ and $\textbf{d}_{r}$ as input to the network is not ideal due to ray aliasing.
If we move the position of a ray $r$ along the ray direction $\textbf{d}_{r}$, we would obtain a ray $r^{\prime} = (\textbf{p}_{r^{\prime}}, \textbf{d}_{r})$ that is really an aliased version of $r$. There is no guarantee that the trained network would produce the same output for different aliases $r^{\prime}$ of the ray $r$.

To resolve ray aliasing, Sitzmann~\etal~\cite{sitzmann2021light} reparametrize the ray $r = (\textbf{p}_{r}, \textbf{d}_{r})$ into Plucker coordinates as $r = (\textbf{m}_{r}, \textbf{d}_{r})$, where $\textbf{m}_{r} = \textbf{p}_{r} \times \textbf{d}_{r}$ is also known as the moment vector of $\textbf{p}_{r}$ about the origin $\Origin$. Plucker coordinates represent all oriented rays in space without singularity or special cases, and they are invariant to changes in the ray position along the ray direction.
To better understand this property, consider moving the ray position to some other point $\textbf{p}_{r}^{\prime}$ at a fixed ray direction $\textbf{d}_{r}$. Then, for a certain $\lambda \in \mathbb{R}$, $\textbf{p}_{r}^{\prime} = \textbf{p}_{r} - \lambda \textbf{d}_{r}$, and
\begin{align*}
    \textbf{p}_{r}^{\prime} \times \textbf{d}_{r} &= (\textbf{p}_{r} - \lambda \textbf{d}_{r}) \times \textbf{d}_{r} \\
    &= \textbf{p}_{r} \times \textbf{d}_{r} - \lambda\mathbf{0} \\
    &= \textbf{p}_{r} \times \textbf{d}_{r}. \addtocounter{equation}{1}\tag{\theequation}
\end{align*}
Therefore, $\textbf{m}_{r} = \textbf{p}_{r} \times \textbf{d}_{r}$ is invariant to any change of $\lambda$ along $\textbf{d}_{r}$.

\subsection{Describing Geometry with Perpendicular Foot}
Our goal is to train a neural network to encode the mapping from a ray to its hit point on the 3D shape's surface.
Although replacing ray position with the moment vector $\textbf{m}_{r}$ allows Sitzmann et al.~\cite{sitzmann2021light} to train networks to encode light fields, it is hard to geometrically relate a ray's moment vector to its surface hit point.
We propose an alternative way to parameterize a ray as input to the network. which has an intuitive and intrinsic relationship to its hit point.

Specifically, we consider the perpendicular foot $\textbf{f}_{r}$ between the ray $r$ and the coordinate system's origin $\Origin$, which may be computed by
\begin{equation}
    \textbf{f}_{r} = \textbf{d}_{r} \times ( \textbf{p}_{r} \times \textbf{d}_{r}).
\end{equation}
Similar to $\textbf{m}_{r}$ in Plucker coordinates, $\textbf{f}_{r}$ is also invariant to changing the ray position along the ray direction.
Specifically, let $\textbf{p}_{r}^{\prime}$ be the translated ray position defined as before, we can then write $\textbf{p}_{r^{\prime}} = \textbf{p}_{r} - \lambda \textbf{d}_{r}$, and 
\begin{align*}
    \textbf{f}_{r^{\prime}} &= \textbf{d}_{r} \times ( \textbf{p}_{r^{\prime}} \times \textbf{d}_{r}) \\
     &= \textbf{d}_{r} \times ( (\textbf{p}_{r} - \lambda \textbf{d}_{r}) \times \textbf{d}_{r} ) \\
    &= \textbf{d}_{r} \times ( \textbf{p}_{r} \times \textbf{d}_{r} - \lambda\mathbf{0} ) \\
    &= \textbf{d}_{r} \times ( \textbf{p}_{r} \times \textbf{d}_{r} ) \\
    &= \textbf{f}_{r} \addtocounter{equation}{1}\tag{\theequation}
\end{align*}
In other words, $\textbf{f}_{r}$ is invariant to moving $\textbf{p}_{r}$ along the direction $\textbf{d}_{r}$.

As a result, we can represent any ray $r = (\textbf{f}_{r}, \textbf{d}_{r})$, and we can further establish the following relationship for each ray $r$:
\begin{equation}
\textbf{h}_{r} = s_{r} \cdot \textbf{d}_{r} + \textbf{f}_{r},
\end{equation}
where $s_{r} \in \mathbb{R}$ denotes the signed displacement between the ray's hit point $\textbf{h}_{r}$ and its perpendicular foot $\textbf{f}_{r}$ \textit{w.r.t.} the world  origin $\Origin$.

To encode a 3D shape, we propose the mapping function $(\textbf{f}_{r}, \textbf{d}_{r}) \to s_{r}$, which we call Primary Ray-based Implicit Function (PRIF).
Different than SDF or OF, which implicitly encodes the geometry through the distance from any given point to its nearest surface point \textit{in any direction}, PRIF operates on oriented points with \textit{a specific ray direction}.

In practice, we train an MLP to learn
\begin{equation}
\Phi(\textbf{f}_{r}, \textbf{d}_{r}) = s_{r}.
\end{equation}
With this new representation, we are able to represent the surface hit point of a ray with a single value.
In effect, our objective is equivalent to finding a simple affine transformation $f(\textbf{x}) = \textbf{A} \textbf{x} + \textbf{b}$, with the input $\textbf{x} = \textbf{d}_{r}$, $\textbf{A} = s_{r} \mathit{I}_{3}$, and $\textbf{b} = \textbf{f}_{r}$.
We also avoid a major limitation in previous sphere-tracing-based methods, which is having to sample multiple points and perform multiple network evaluations to obtain a hit point.

\subsection{Background Mask}
For 3D functions like SDF or OF, every 3D point has a well-established scalar value denoting the distance to surface or the occupancy at that point.
In contrast, our function would take in rays that never intersect with the shape and therefore do not even have a hit point.

To address this issue, we let the network additionally produce
\begin{equation}
\Phi(\textbf{f}_{r}, \textbf{d}_{r}) = a_{r},
\end{equation}
where $a_{r} \in [0, 1]$ denotes the probability in which the ray $r$ hits the foreground. We compute the cross-entropy loss
\begin{equation}
\mathcal{L}_{a} = \sum_{r}
-a_{r}^{gt}\Log{a_{r}} - \left(1 - a_{r}^{gt}\right) \Log{1 - a_{r}},
\end{equation}
where $a_{r}^{gt} = 0$ for background rays and $a_{r}^{gt} = 1$ for foreground rays.

For the signed displacement $s_{r}$, we supervise the learning by computing its absolute difference to the ground truth as $\mathcal{L}_{s} = \sum_{r \in \mathcal{F}} \Norm{s_{r} - s_{r}^{gt}}$, given the set of foreground rays $\mathcal{F}$.
As a result, the total loss function to train our network is 
$\mathcal{L} = \mathcal{L}_{a} + \mathcal{L}_{s}$
and is averaged among all rays in a training batch

\subsection{Outlier Points Removal}
In rare cases where sharp surface discontinuities exist between two neighboring rays, the network would likely produce continuous predictions when interpolating between those two rays, resulting in undesirable outlier points.
Fortunately, since our network is fully differentiable, for each prediction $s_{r}$ we can compute its gradient with respect to the changes in ray position.
We discard all predictions that satisfy the threshold: 
$\Norm{ \frac{\partial s_{r}}{\partial \textbf{p}_{r}} } \geq \delta.$
In our experiments, $\delta$ is set equal to 5.

\section{Experiments} \label{section:experiments}
In this section, we first verify the efficacy of neural PRIF for shape representation.
Then, we show verious applications achieved by using PRIF as the underlying neural shape representation. 
Note that the scope of our experiments is to compare these functional shape representations \textit{encoded by neural networks}.

\begin{figure}[t]
	\centering
	\begin{subfigure}{23.5mm}
		\centering
		\includegraphics[width=23.5mm]{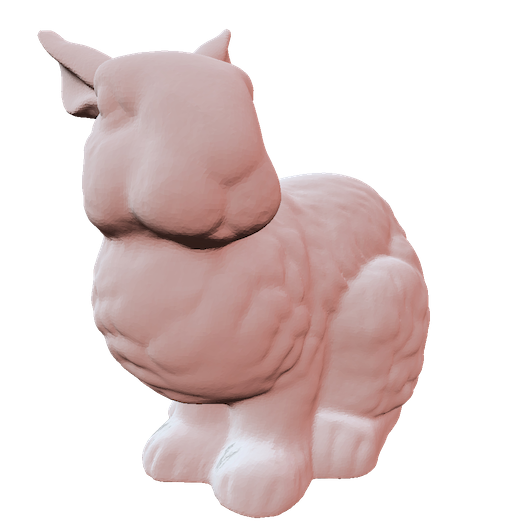}
	\end{subfigure}%
	~
	\begin{subfigure}{23.5mm}
		\centering
		\includegraphics[width=23.5mm]{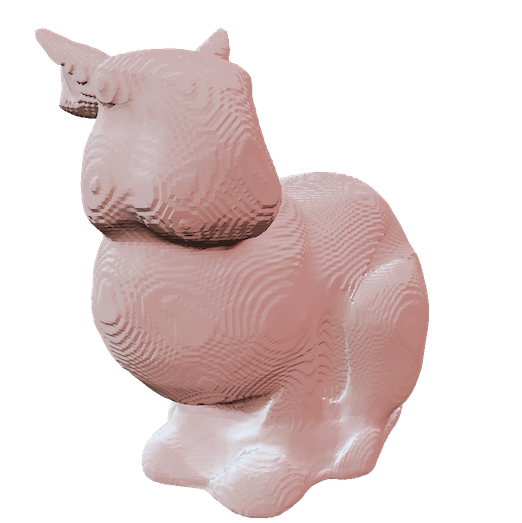}
	\end{subfigure}%
	~
	\begin{subfigure}{23.5mm}
		\centering
		\includegraphics[width=23.5mm]{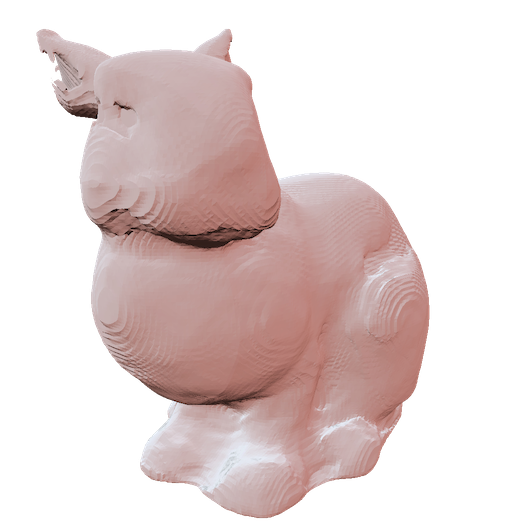}
	\end{subfigure}%
	~
	\begin{subfigure}{23.5mm}
		\centering
		\includegraphics[width=23.5mm]{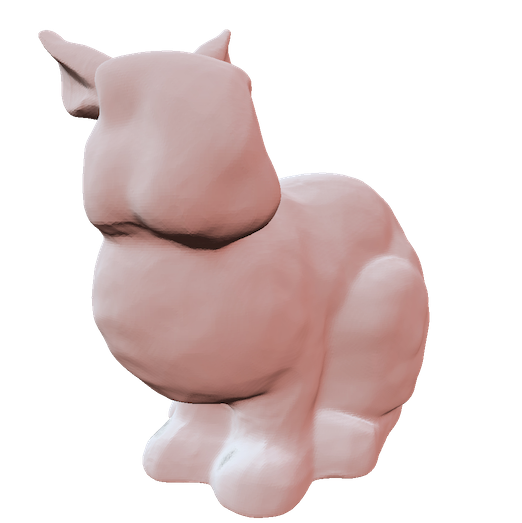}
	\end{subfigure}%
	~
	\begin{subfigure}{23.5mm}
		\centering
		\includegraphics[width=23.5mm]{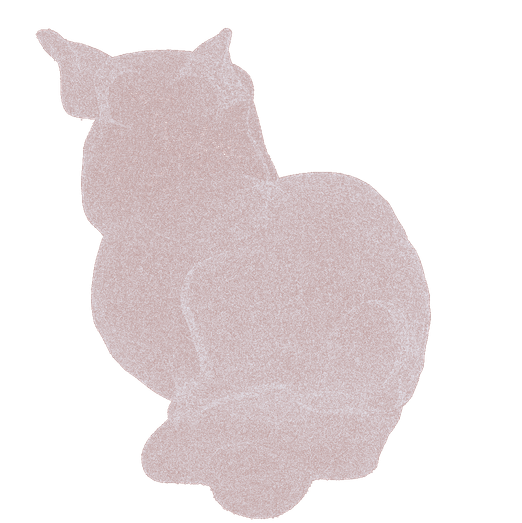}
	\end{subfigure}

	\begin{subfigure}{23.5mm}
		\centering
		\includegraphics[width=23.5mm]{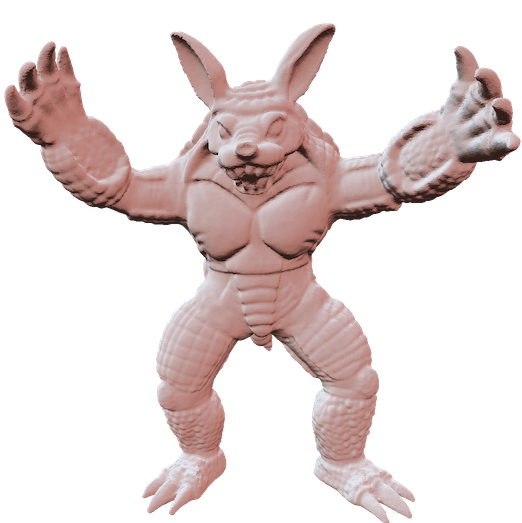}
	\end{subfigure}%
	~
	\begin{subfigure}{23.5mm}
		\centering
		\includegraphics[width=23.5mm]{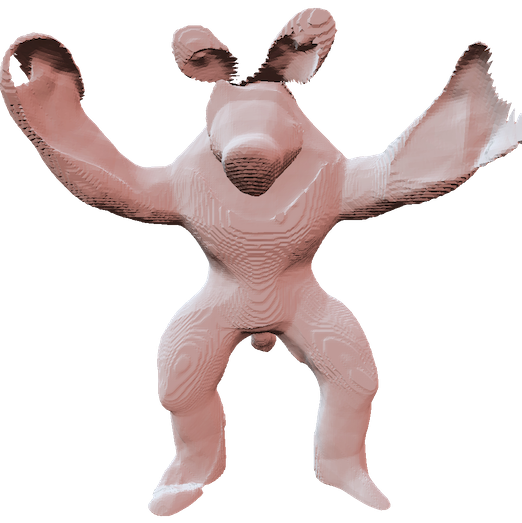}
	\end{subfigure}%
	~
	\begin{subfigure}{23.5mm}
		\centering
		\includegraphics[width=23.5mm]{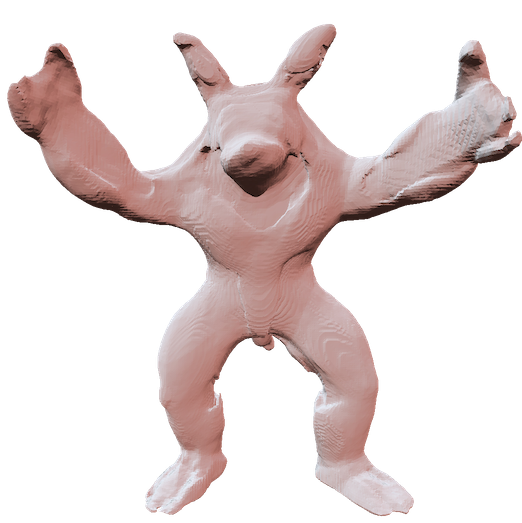}
	\end{subfigure}%
	~
	\begin{subfigure}{23.5mm}
		\centering
		\includegraphics[width=23.5mm]{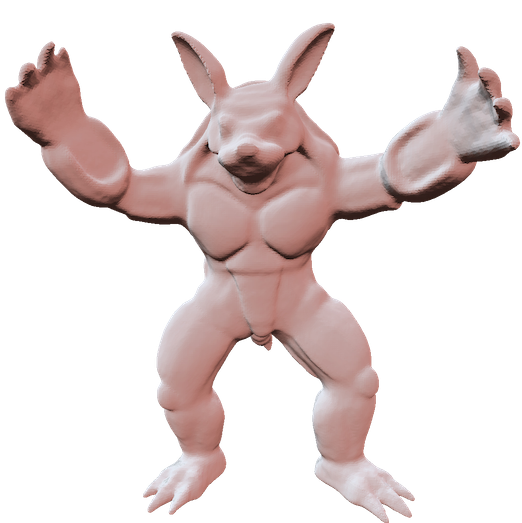}
	\end{subfigure}%
	~
	\begin{subfigure}{23.5mm}
		\centering
		\includegraphics[width=23.5mm]{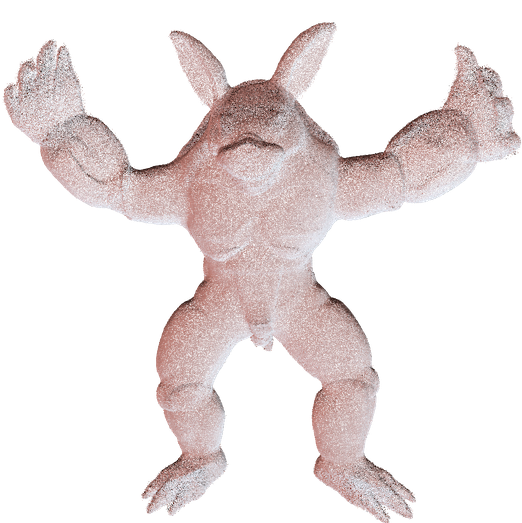}
	\end{subfigure}

	\begin{subfigure}{23.5mm}
		\centering
		\includegraphics[width=23.5mm]{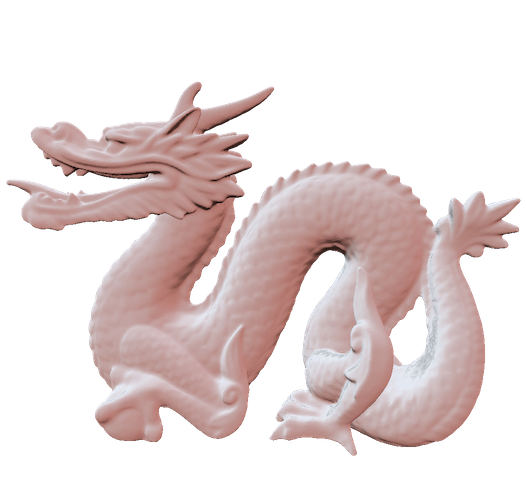}
		$\underbracket[0pt][1.0mm]{\hspace{\linewidth}}_%
    {\substack{\vspace{-5mm}\\ \colorbox{white}{Reference}}}$
	\end{subfigure}%
	~
	\begin{subfigure}{23.5mm}
		\centering
		\includegraphics[width=23.5mm]{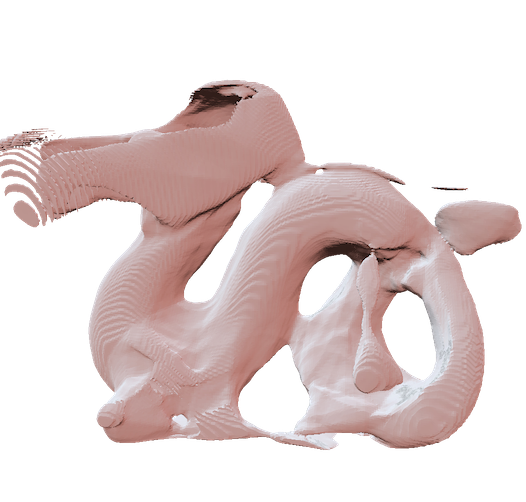}
		$\underbracket[0pt][1.0mm]{\hspace{\linewidth}}_%
    {\substack{\vspace{-5mm}\\ \colorbox{white}{OF}}}$
	\end{subfigure}%
	~
	\begin{subfigure}{23.5mm}
		\centering
		\includegraphics[width=23.5mm]{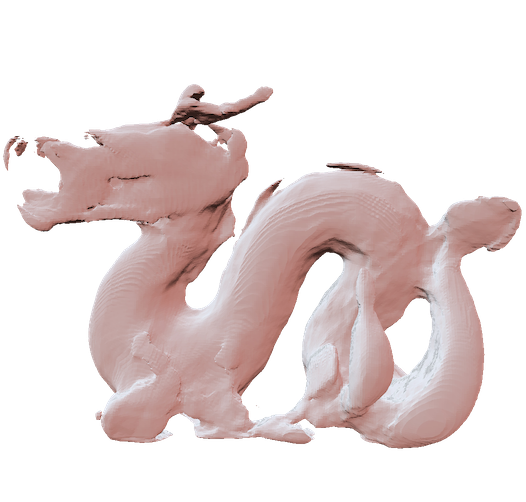}
		$\underbracket[0pt][1.0mm]{\hspace{\linewidth}}_%
    {\substack{\vspace{-5mm}\\ \colorbox{white}{SDF}}}$
	\end{subfigure}%
	~
	\begin{subfigure}{23.5mm}
		\centering
		\includegraphics[width=23.5mm]{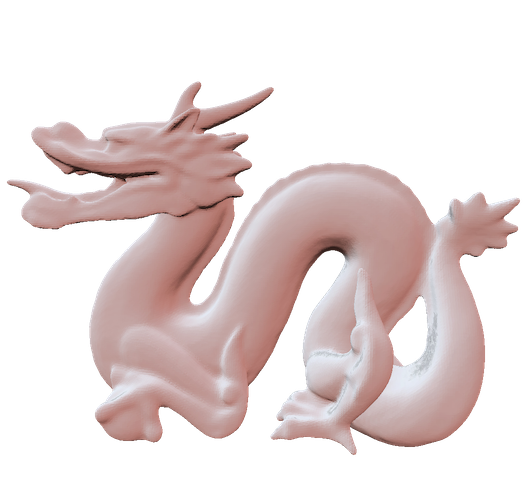}
		$\underbracket[0pt][1.0mm]{\hspace{\linewidth}}_%
    {\substack{\vspace{-5mm}\\ \colorbox{white}{Ours - Mesh}}}$
	\end{subfigure}%
	~
	\begin{subfigure}{23.5mm}
		\centering
		\includegraphics[width=23.5mm]{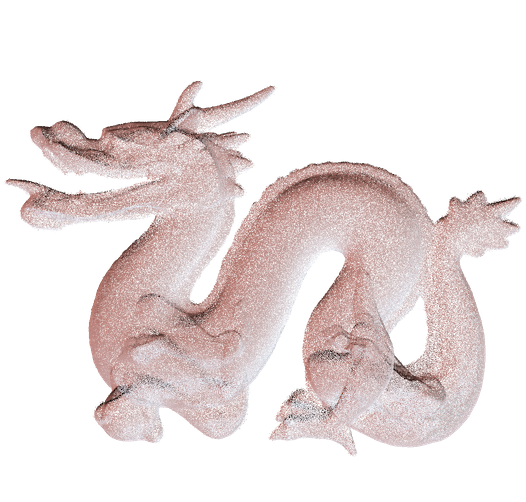}
		$\underbracket[0pt][1.0mm]{\hspace{\linewidth}}_%
    {\substack{\vspace{-5mm}\\ \colorbox{white}{Ours - Points}}}$
	\end{subfigure}
	\caption{{\bf Single Shape.} To examine the representation capability of PRIF, we train networks with the same architecture to encode the occupancy function (OF), signed distance function (SDF), and our proposed function (PRIF). Here, we visualize the extracted shapes from the trained neural representations. For OF and SDF, we follow the convention and extract the shapes by Marching Cubes. Our method directly outputs hit points, and we also apply the point-based Screened Poisson algorithm and present the resulting mesh for comparison.}
	\label{fig:SingleShape}
\end{figure}

\subsection{Single Shape Representation}
We select five models (\textit{Armadillo}, \textit{Bunny}, \textit{Buddha}, \textit{Dragon}, \textit{Lucy}) from the Stanford 3D Scanning Repository~\cite{turk1994zippered, curless1996volumetric} and train a neural network to fit the PRIF for each 3D model.
We also fit the signed distance function (SDF) and the 3D occupancy function (OF) for these models.
For a fair comparison between these functions, we adopt the same network architecture as Park et al.~\cite{park2019deepsdf}, containing eight layers with 512 hidden dimensions and ReLU activation.

Compared to SDF and OF which are trained on individual spatial points $(x, y, z)$, PRIF requires an inherently different strategy to generate the training data since it takes in individual rays.
In our experiments, for each 3D model, we select 50 virtual camera locations oriented towards the origin, and we capture $200\times 200$ rays at each location.
For SDF and OF, we follow Park et al.~\cite{park2019deepsdf} and sample $500,000$ points with more aggressive sampling near the object surface.

For all three functions, we train the neural representation for 100 epochs with the learning rate initialized as $10^{-4}$ and decayed to $10^{-7}$ with a cosine annealing strategy.
After training, the SDF- and OF-based shape representation are obtained by evaluating the neural network at uniform $256^3$ volume grid and extracted using Marching Cubes.
On the other hand, with our PRIF-based representation, we can evaluate the neural network at those virtual camera rays in the training set and directly obtain a dense set of surface points.

\begin{table}[t]
\centering
\begin{tabular}{c c c c c c}
\toprule
Method & Armadillo & Bunny & Buddha & Dragon & Lucy \\ \midrule
SDF &  1.905$\vert$1.260 & 1.717$\vert$1.147   &   6.119$\vert$2.258   &  5.184$\vert$1.946 &  3.387$\vert$1.417   \\ 
OF &  4.805$\vert$1.624  & 1.704$\vert$1.133   &  17.279$\vert$3.113   &    19.577$\vert$3.014  &   3.396$\vert$1.427    \\ 
\midrule
PRIF & \textbf{0.978}$\vert$\textbf{0.706}  & \textbf{1.169}$\vert$\textbf{0.835}  &  \textbf{1.443}$\vert$\textbf{0.821}   &   \textbf{1.586}$\vert$\textbf{0.913}  &   \textbf{0.846}$\vert$\textbf{0.519}   \\ 
\bottomrule
\\
\end{tabular}
\caption{Quantitative results on single shape representation on 3D models from the Stanford 3D Scanning Repository. The left and right numbers represent the mean and median CD (multiplied by $10^{-4}$). After extracting shapes from each representation, 30,000
points are sampled for evaluation.}
\label{tab:single_shape}
\end{table}

To evaluate the representation quality, for SDF- and OF-based representations, we first follow conventions~\cite{park2019deepsdf} and sample $8,192$ points on the mesh extracted with Marching Cubes.
For the point set produced by the PRIF-based representation, we apply the point-based meshing algorithm Screened Poisson~\cite{kazhdan2013screened} in MeshLab~\cite{cignoni2008meshlab} and then sample $8,192$ points from the reconstructed mesh.
Then, we obtain $8,192$ Poisson-disk samples of the ground truth surface points from the original 3D model.
Finally, we compute the mean and median Chamfer Distance (CD) between the ground truth point set and point sets sampled from those three representations.
In Table~\ref{tab:single_shape} and Fig.~\ref{fig:SingleShape}, we provide quantitative and qualitative comparisons among the three representations.
PRIF significantly outperform SDF and OF in accurately preserving the fine details of the 3D shapes.

\begin{figure}[!ht]
	\centering

	\begin{subfigure}{23.0mm}
		\centering
		\includegraphics[width=23.0mm]{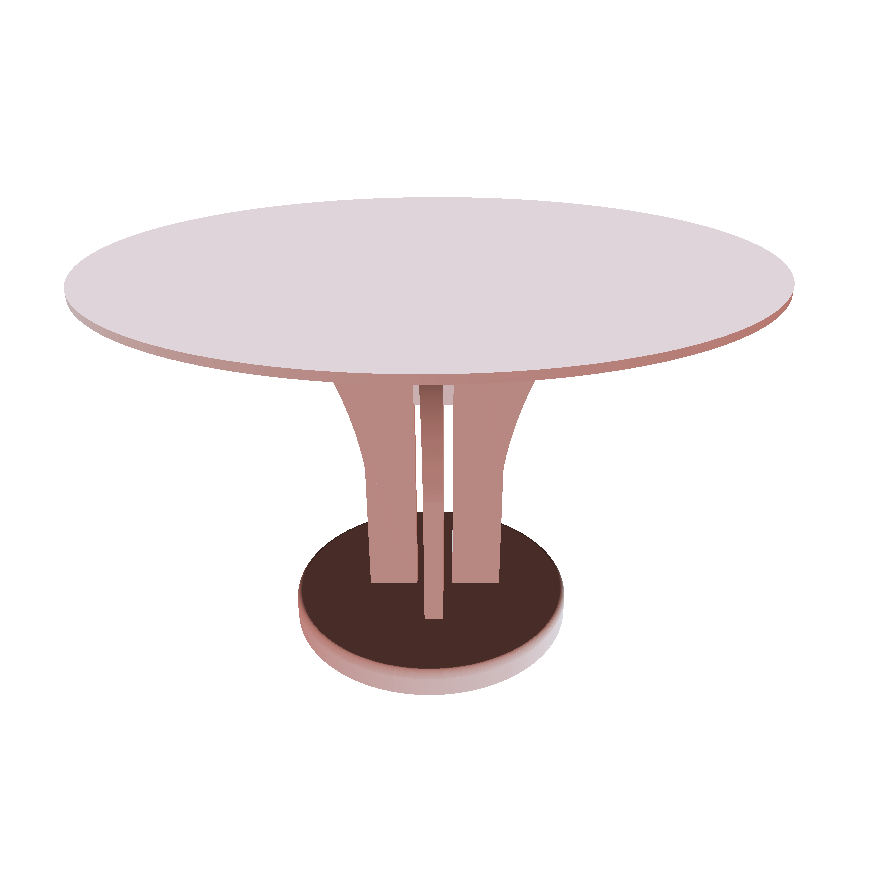}
	\end{subfigure}%
	~
	\begin{subfigure}{23.0mm}
		\centering
		\includegraphics[width=23.0mm]{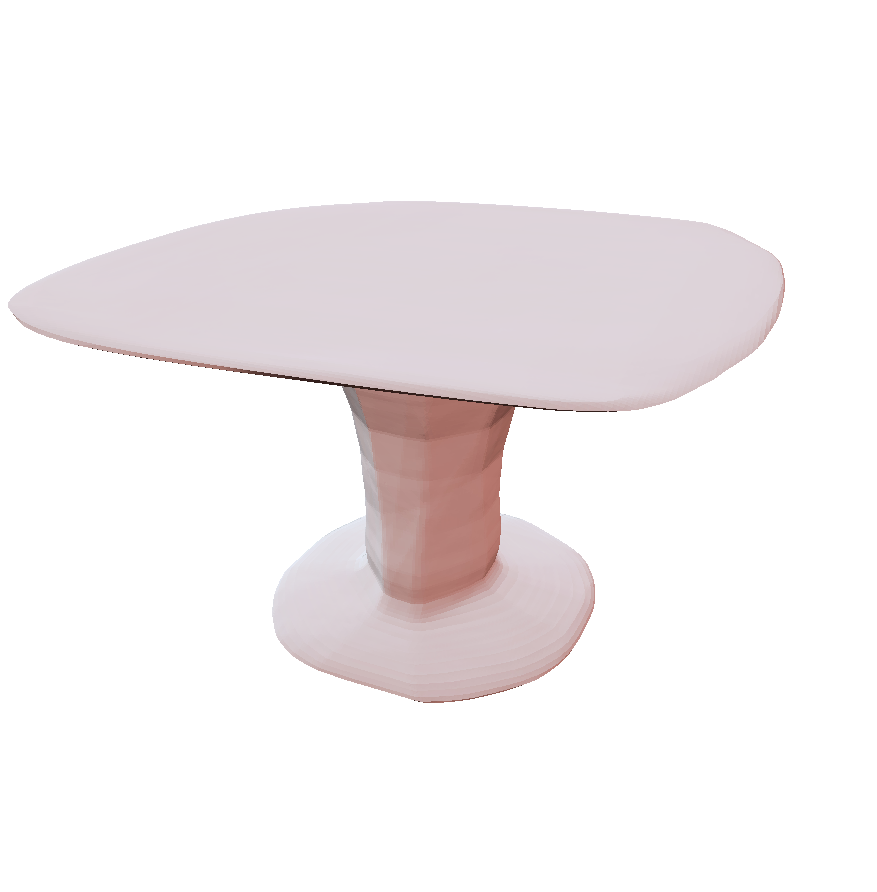}
	\end{subfigure}%
	~
	\begin{subfigure}{23.0mm}
		\centering
		\includegraphics[width=23.0mm]{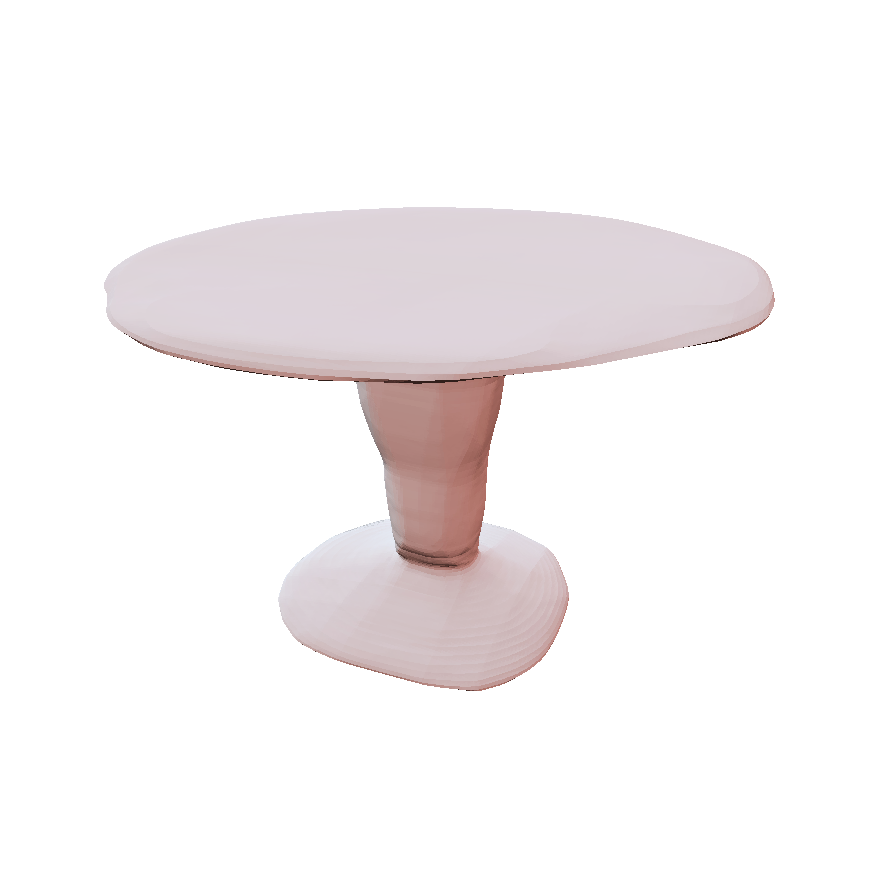}
	\end{subfigure}%
	~
	\begin{subfigure}{23.0mm}
		\centering
		\includegraphics[width=23.0mm]{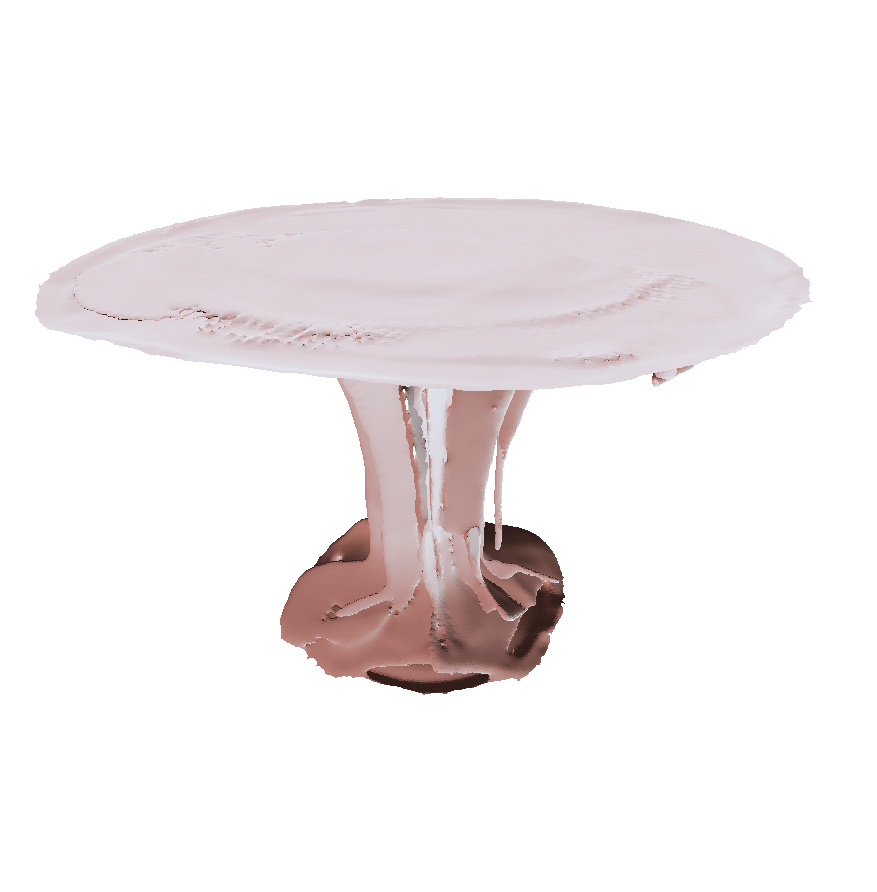}
	\end{subfigure}%
	~
	\begin{subfigure}{23.0mm}
		\centering
		\includegraphics[width=23.0mm]{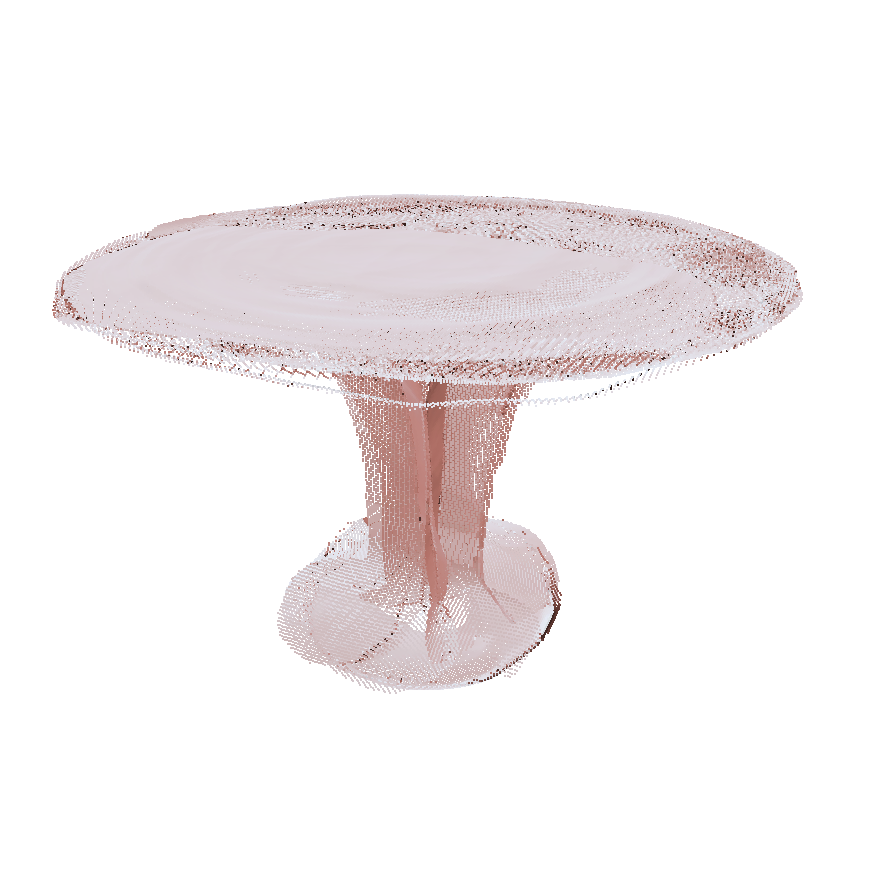}
	\end{subfigure}

	\begin{subfigure}{23.0mm}
		\centering
		\includegraphics[width=23.0mm]{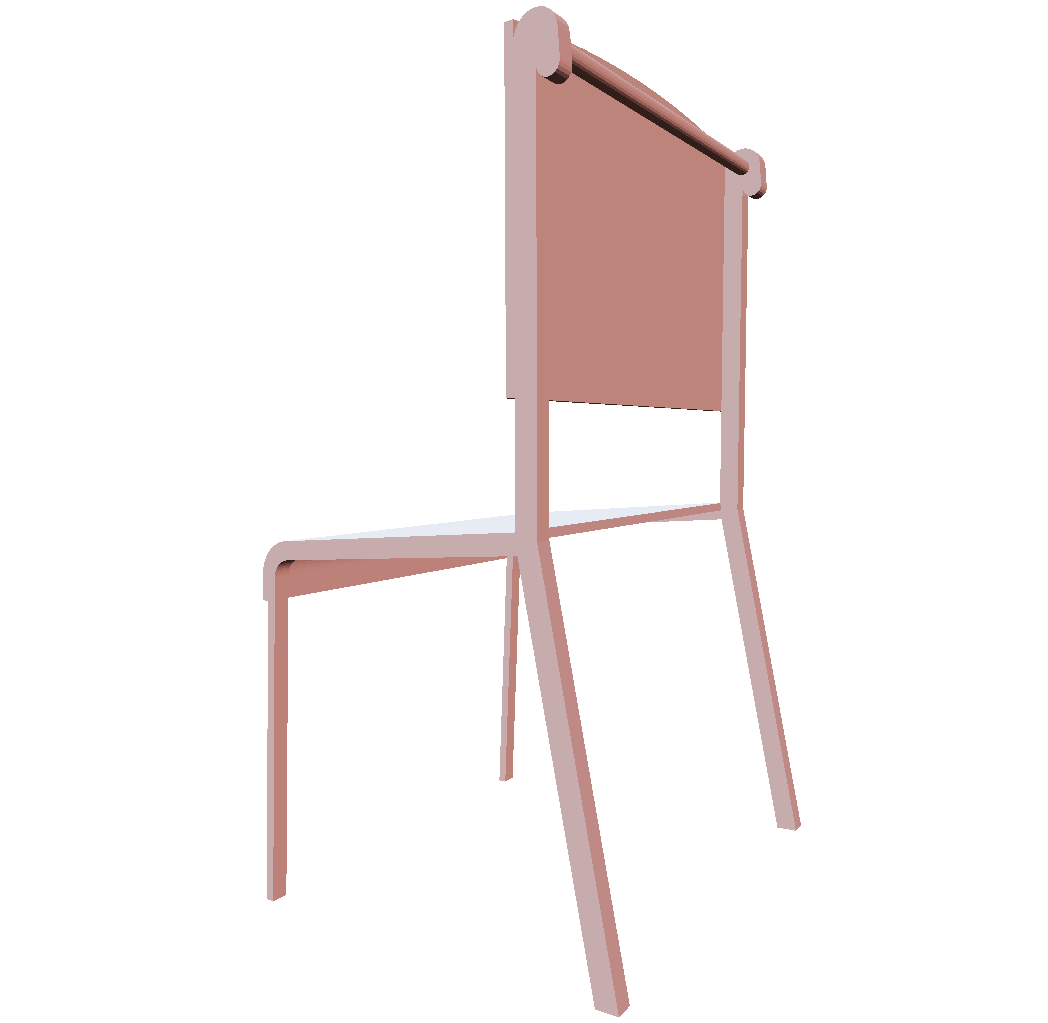}
	\end{subfigure}%
	~
	\begin{subfigure}{23.0mm}
		\centering
		\includegraphics[width=23.0mm]{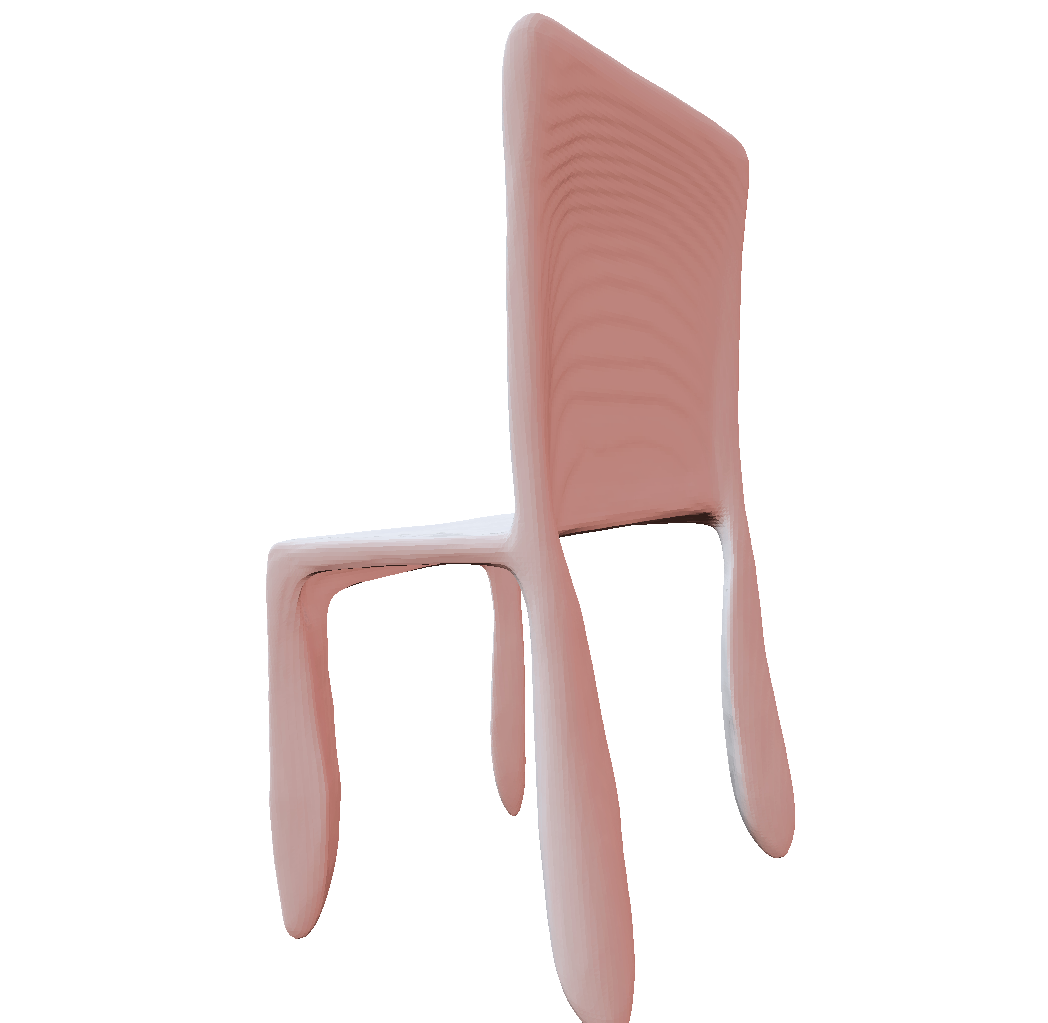}
	\end{subfigure}%
	~
	\begin{subfigure}{23.0mm}
		\centering
		\includegraphics[width=23.0mm]{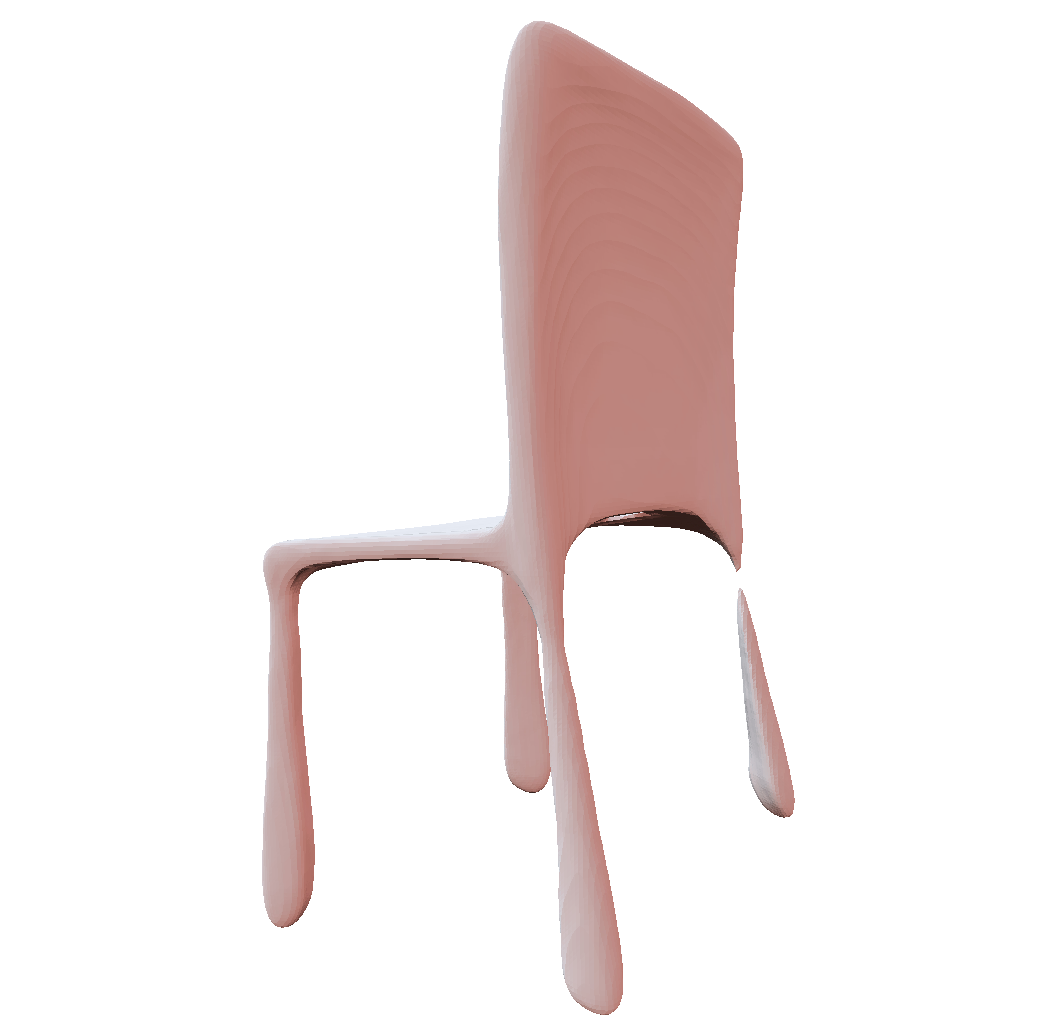}
	\end{subfigure}%
	~
	\begin{subfigure}{23.0mm}
		\centering
		\includegraphics[width=23.0mm]{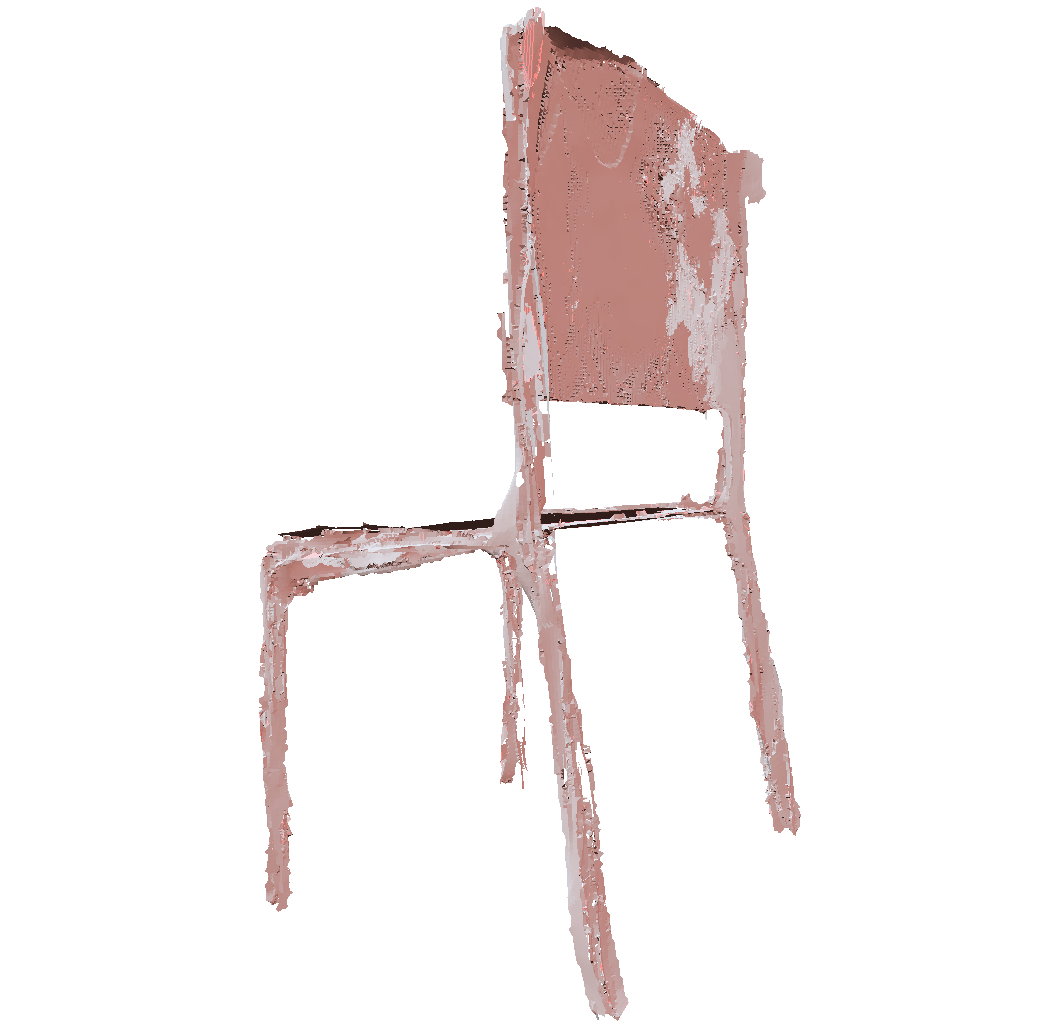}
	\end{subfigure}%
	~
	\begin{subfigure}{23.0mm}
		\centering
		\includegraphics[width=23.0mm]{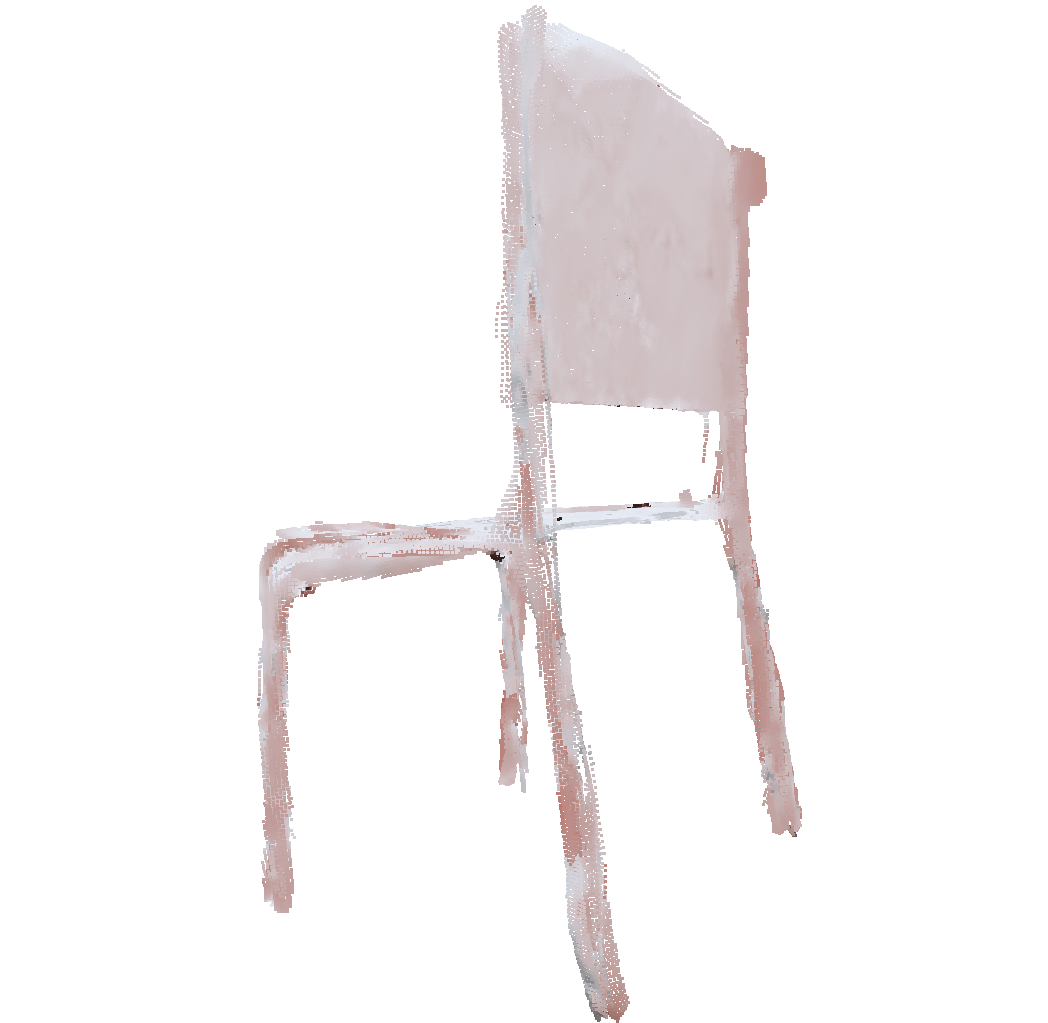}
	\end{subfigure}

	\begin{subfigure}{23.0mm}
		\centering
		\includegraphics[width=23.0mm]{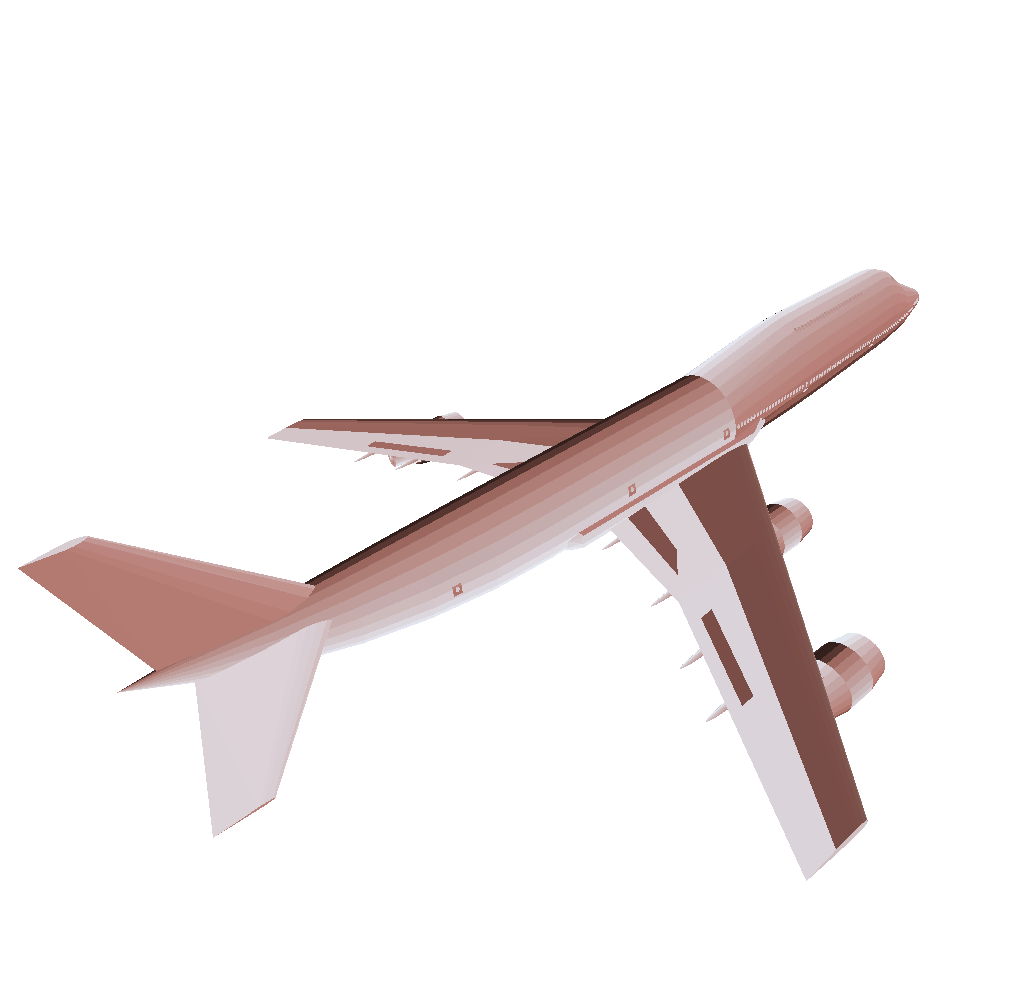}
		$\underbracket[0pt][1.0mm]{\hspace{\linewidth}}_%
    {\substack{\vspace{-5mm}\\ \colorbox{white}{Reference}}}$
	\end{subfigure}%
	~
	\begin{subfigure}{23.0mm}
		\centering
		\includegraphics[width=23.0mm]{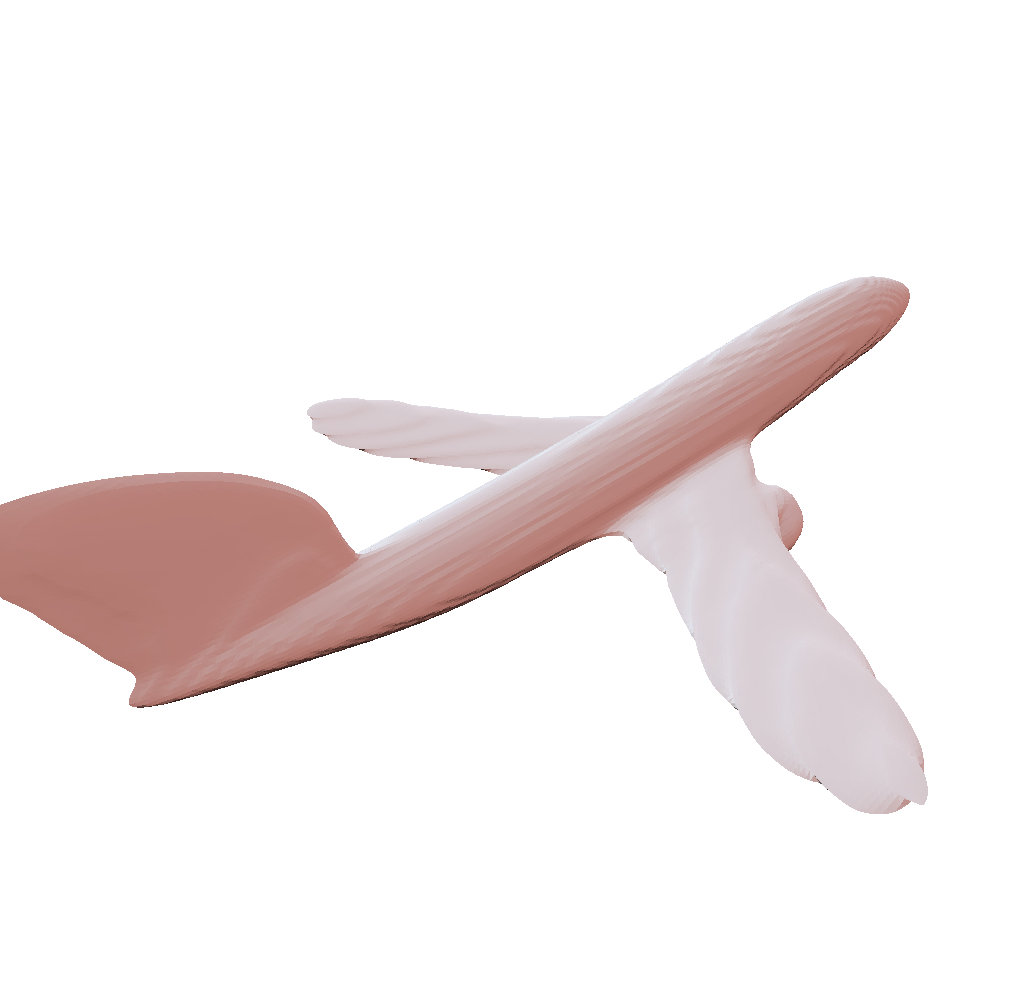}
		$\underbracket[0pt][1.0mm]{\hspace{\linewidth}}_%
    {\substack{\vspace{-5mm}\\ \colorbox{white}{OF}}}$
	\end{subfigure}%
	~
	\begin{subfigure}{23.0mm}
		\centering
		\includegraphics[width=23.0mm]{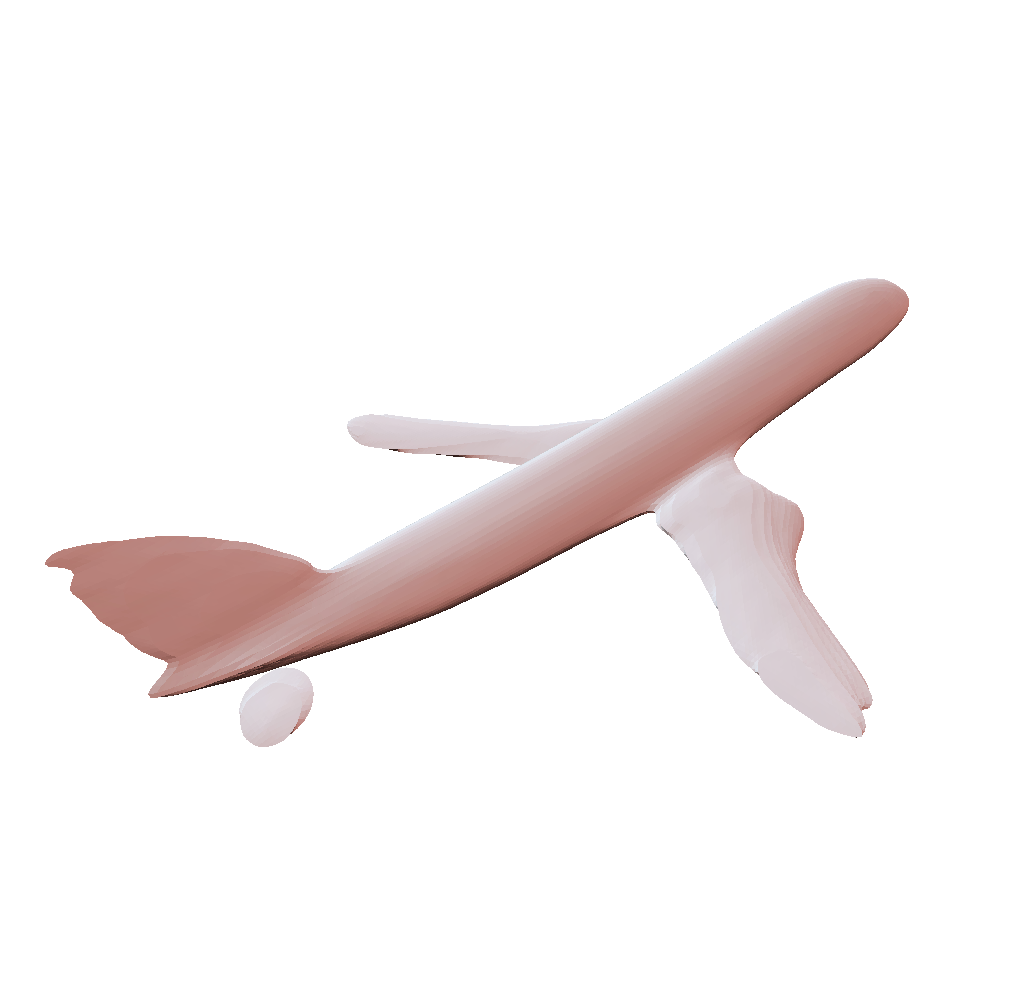}
		$\underbracket[0pt][1.0mm]{\hspace{\linewidth}}_%
    {\substack{\vspace{-5mm}\\ \colorbox{white}{SDF}}}$
	\end{subfigure}%
	~
	\begin{subfigure}{23.0mm}
		\centering
		\includegraphics[width=23.0mm]{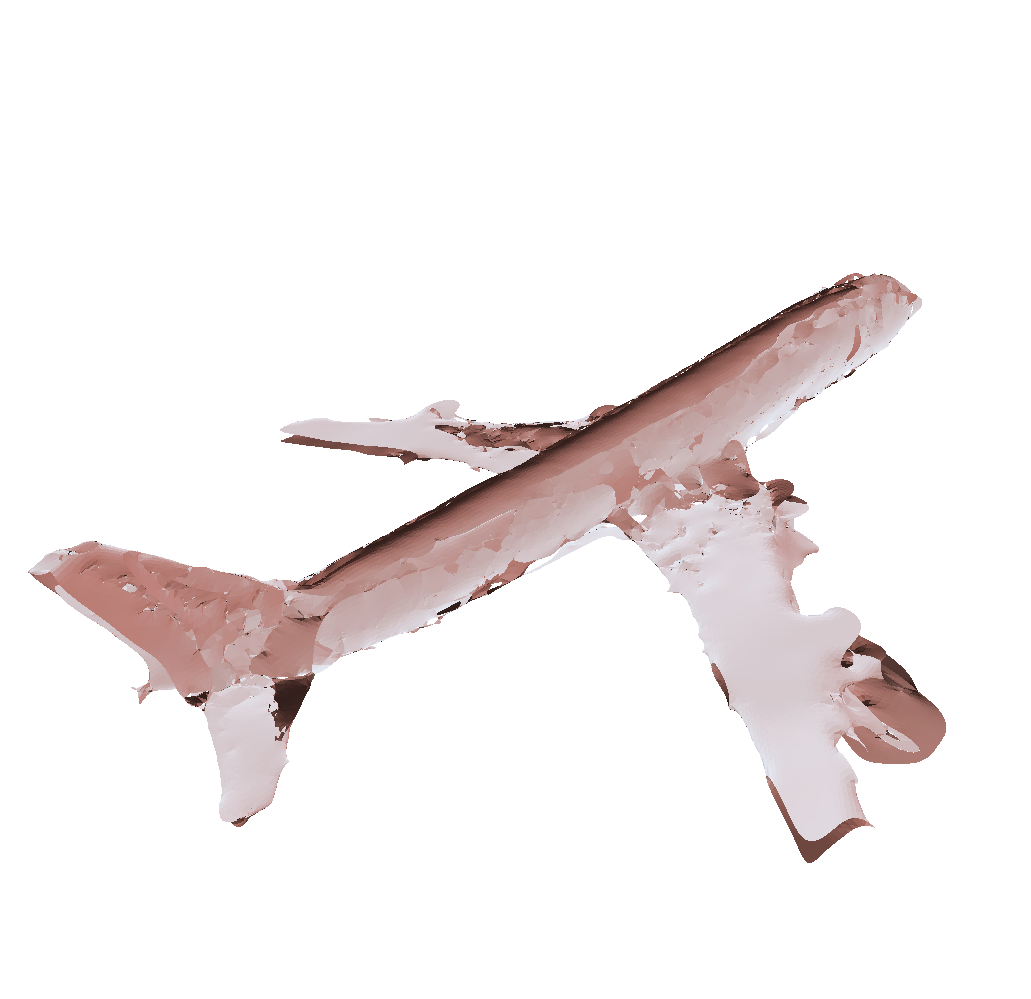}
		$\underbracket[0pt][1.0mm]{\hspace{\linewidth}}_%
    {\substack{\vspace{-5mm}\\ \colorbox{white}{Ours - Mesh}}}$
	\end{subfigure}%
	~
	\begin{subfigure}{23.0mm}
		\centering
		\includegraphics[width=23.0mm]{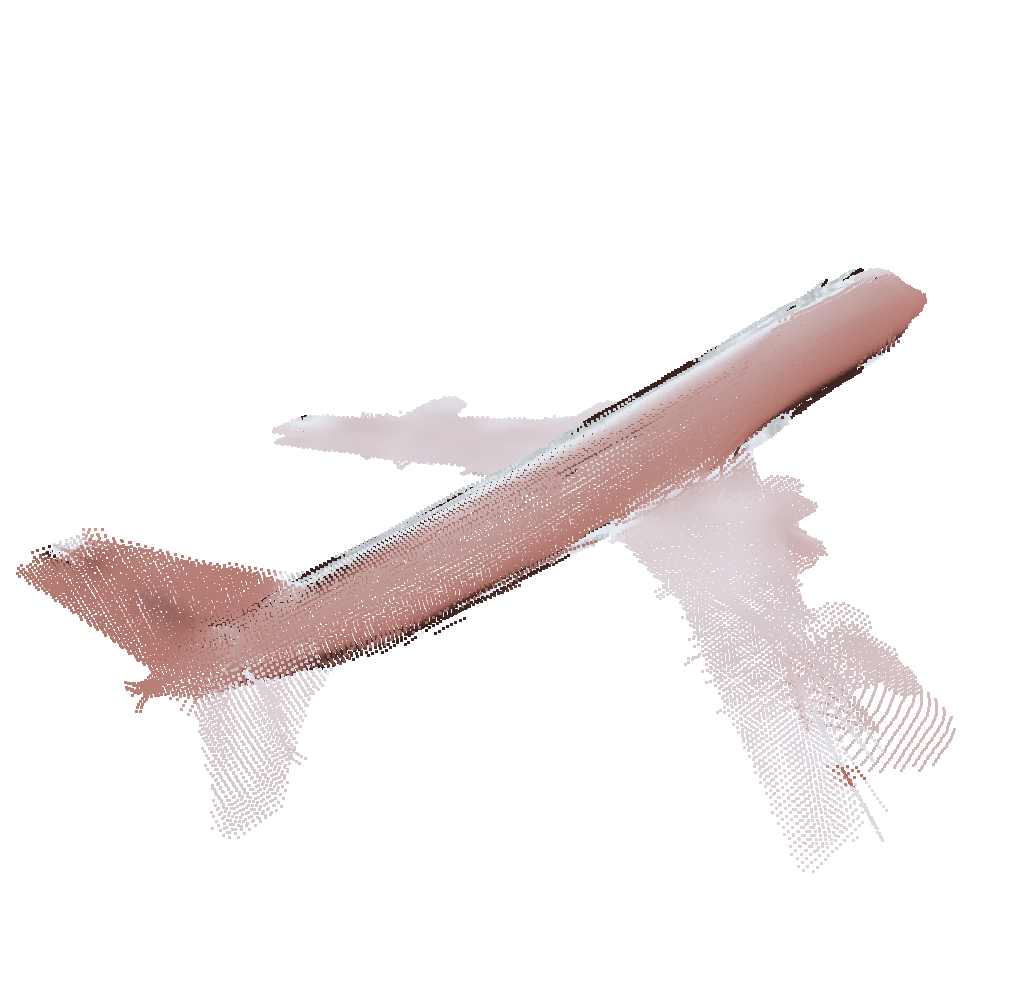}
		$\underbracket[0pt][1.0mm]{\hspace{\linewidth}}_%
    {\substack{\vspace{-5mm}\\ \colorbox{white}{Ours - Points}}}$
	\end{subfigure}

	\caption{{\bf Shape Generation.} We further examine the representation power on generative modeling over multiple shapes. After training the neural networks to learn shapes from a category, we use auto-decoding to find latent vectors that best represents novel shapes within this category. Here, we visualize the extracted shapes from networks trained to encode the three different functions. Note that $\textit{Points}$ are the immediate output of the PRIF networks.}
	\label{fig:Generation}
\end{figure}

\begin{table}[!ht]
\centering
\scalebox{0.92}{
\begin{tabular}{c c c c c c c}
\toprule
Method & Car & Chair & Table & Plane & Lamp & Sofa \\ \midrule
SDF &  2.315$\vert$0.495 & 2.649$\vert$0.407   &   7.213$\vert$0.441   &  2.728$\vert$0.170  &   32.571$\vert$3.475   &   6.427$\vert$0.218    \\ 
OF &  2.820$\vert$0.587  & 4.589$\vert$0.835   &   6.427$\vert$1.296  &   2.999$\vert$0.169  &   143.377$\vert$145.753  &   12.672$\vert$0.184     \\ 
\midrule
PRIF &  \textbf{1.961}$\vert$\textbf{0.347} &  \textbf{0.982}$\vert$\textbf{0.267}   &   \textbf{4.532}$\vert$\textbf{0.315}   &  \textbf{0.389}$\vert$\textbf{0.125} &    \textbf{3.276}$\vert$\textbf{0.534}   &   \textbf{1.236}$\vert$\textbf{0.222}   \\ 
\bottomrule
\\
\end{tabular}
}
\caption{Quantitative results on generative representation on unseen 3D shapes of six categories from ShapeNetCore. The left and right numbers represent the mean and median CD ($\times 10^{-3}$) averaged over the test set. After extracting shapes from each representation, 30,000
points are sampled for evaluation.}
\label{tab:generative}
\end{table}
\normalsize

\begin{figure}[!ht]
	\begin{subfigure}{26.5mm}
		\centering
		\includegraphics[width=26.5mm]{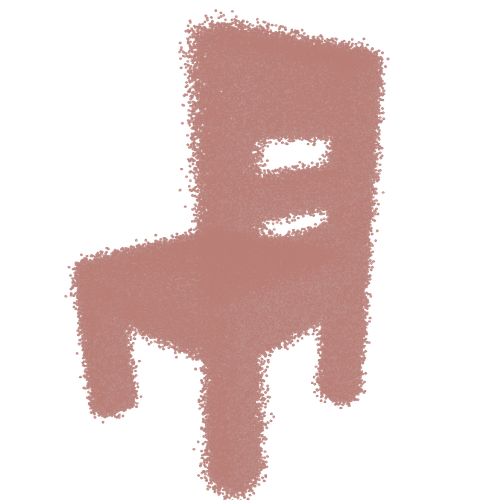}
	\end{subfigure}%
	~
	\begin{subfigure}{26.5mm}
		\centering
		\includegraphics[width=26.5mm]{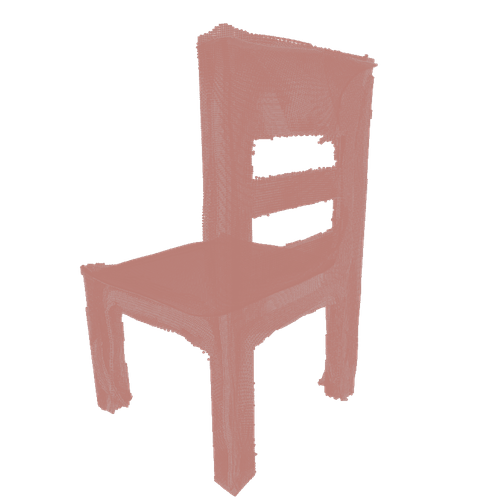}
	\end{subfigure} 
	~
	\begin{subfigure}{26.5mm}
		\centering
		\includegraphics[width=26.5mm]{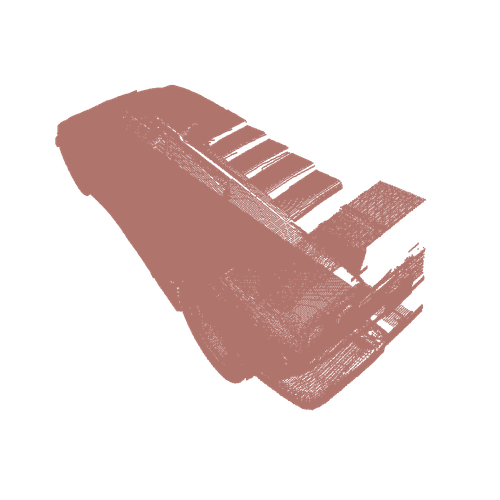}
	\end{subfigure} 
	~
	\begin{subfigure}{26.5mm}
		\centering
		\includegraphics[width=26.5mm]{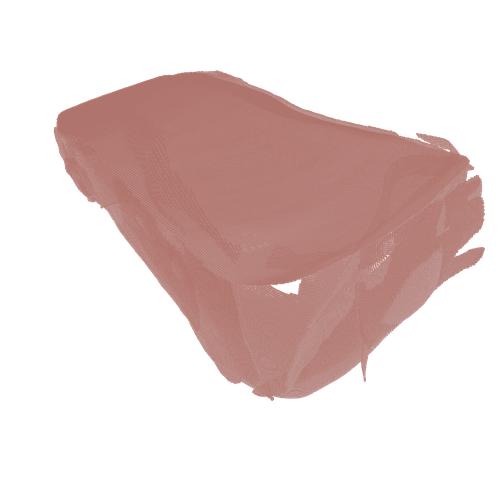}
	\end{subfigure} 

	\begin{subfigure}{26.5mm}
		\centering
		\includegraphics[width=26.5mm]{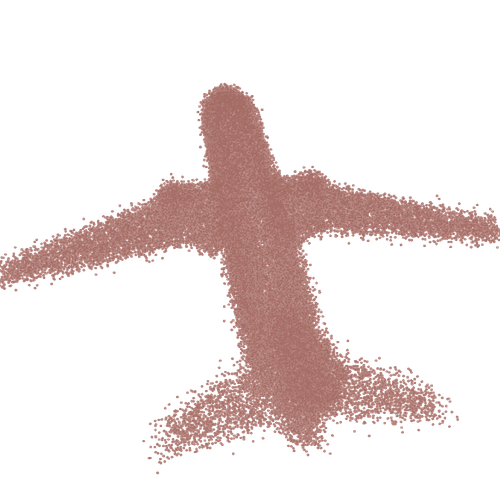}\vspace{-7mm}
	$\underbracket[0pt][1.0mm]{\hspace{\linewidth}}_%
    {\substack{ \colorbox{white}{Noisy Data}}}$
	\end{subfigure}%
	~
	\begin{subfigure}{26.5mm}
		\centering
		\includegraphics[width=26.5mm]{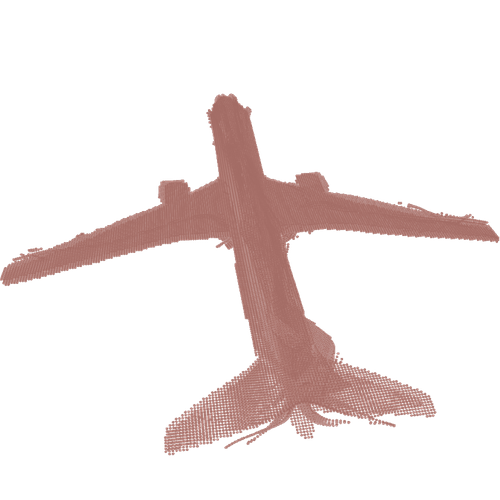}\vspace{-7mm}
	$\underbracket[0pt][1.0mm]{\hspace{\linewidth}}_%
    {\substack{ \colorbox{white}{Reconstruction}}}$
	\end{subfigure} 
	~
	\begin{subfigure}{26.5mm}
		\centering
		\includegraphics[width=26.5mm]{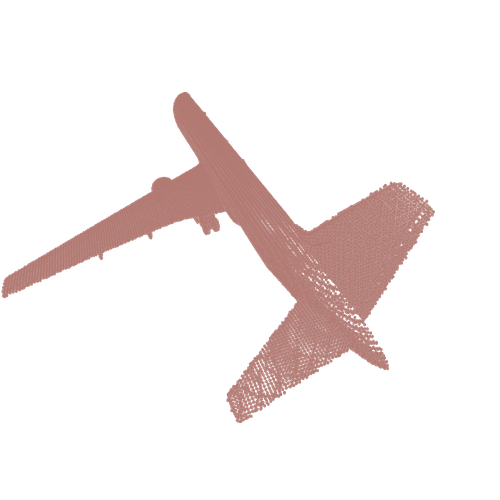}\vspace{-7mm}
	$\underbracket[0pt][1.0mm]{\hspace{\linewidth}}_%
    {\substack{ \colorbox{white}{Incomplete Data}}}$
	\end{subfigure} 
	~
	\begin{subfigure}{26.5mm}
		\centering
		\includegraphics[width=26.5mm]{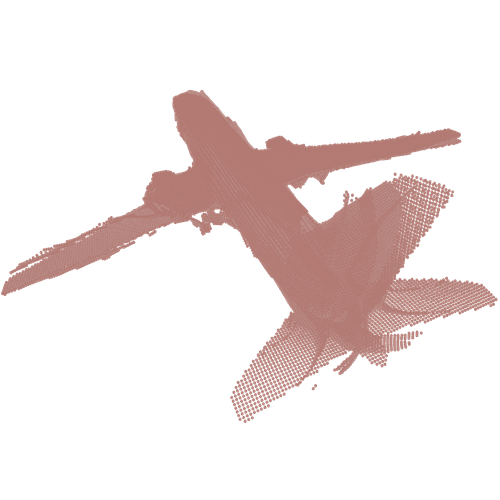}\vspace{-7mm}
	$\underbracket[0pt][1.0mm]{\hspace{\linewidth}}_%
    {\substack{ \colorbox{white}{Reconstruction}}}$
	\end{subfigure} 
	\caption{{\bf Noisy and Incomplete Observations.} Networks trained for shape generation can reconstruct unseen shapes in challenging scenarios where only noisy or incomplete observations are available. Here we visualize the observations and reconstructions as raw point cloud data.}
	\label{fig:ShapeCompletion}
\end{figure}

\subsection{Shape Generation}
Having established the representation power of PRIF for single shapes, we now examine its capability for generative shape modeling.
We adopt the strategy from Park~\etal~\cite{park2019deepsdf} and enable multi-shape generation from a single network by concatenating a latent code for each object with the original network input.

In line with Park~\etal~\cite{park2019deepsdf}, we select five categories of 3D objects from ShapeNetCore~\cite{shapenet2015}.
Within each category, we select the first 180 objects as the training set and the next 20 as the testing set.
We maintain the same network architecture and training schedule when training neural networks to fit OF, SDF, and PRIF.
We evaluate the trained networks on unseen shapes from the test set using auto-decoding~\cite{park2019deepsdf} and provide quantitative and qualitative comparisons in Table~\ref{tab:generative} and Fig.~\ref{fig:Generation}.
The PRIF-based representation extends its success from representing single shapes and remains effective in the generative task.

\begin{figure}[!t]
    \centering
    \hspace{-5mm}
	\begin{subfigure}{.47\textwidth}
	\centering
	\includegraphics[height=0.75\linewidth]{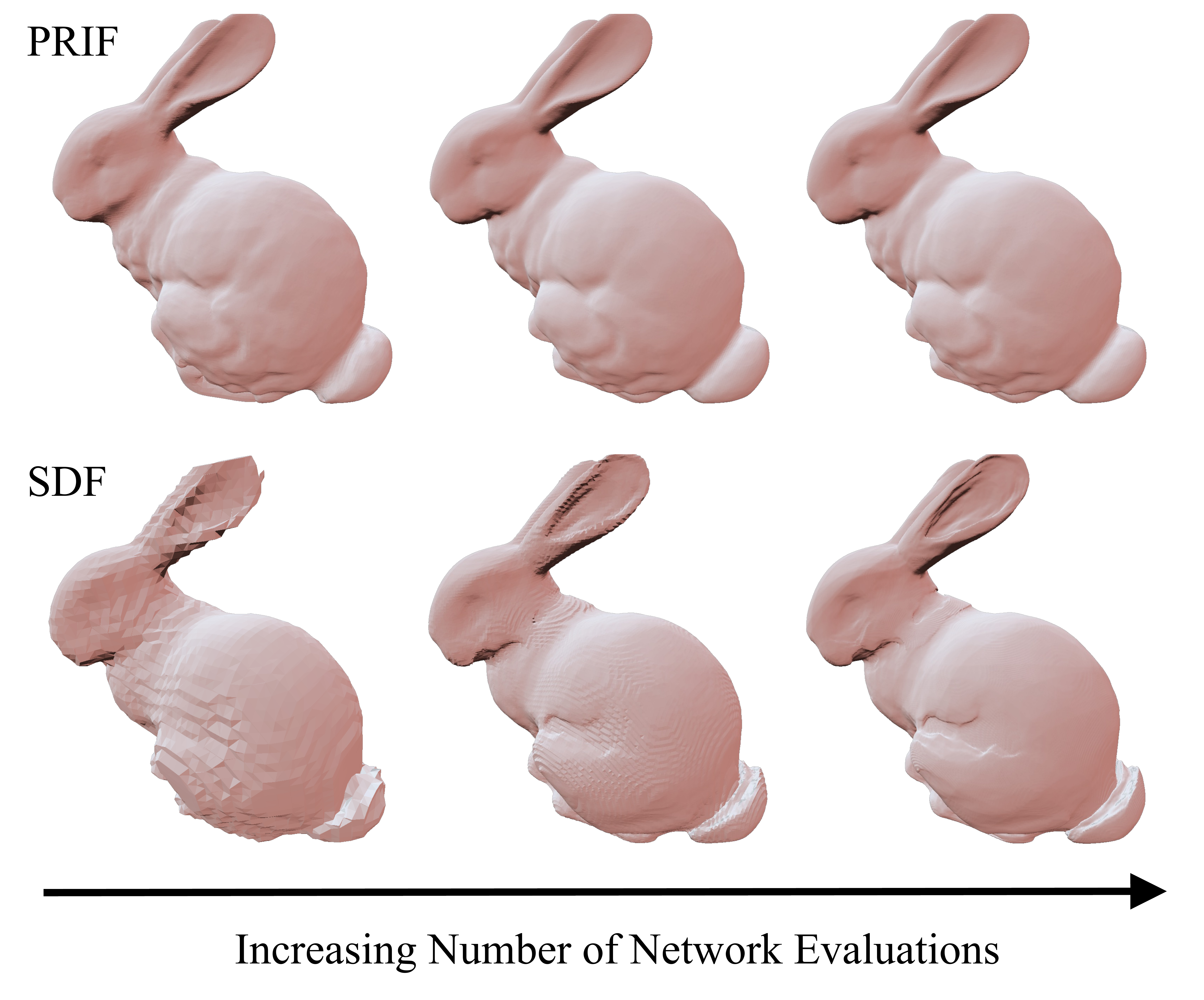}
	\end{subfigure}%
	\begin{subfigure}{.45\textwidth}
		\centering
		\includegraphics[height=0.75\linewidth]{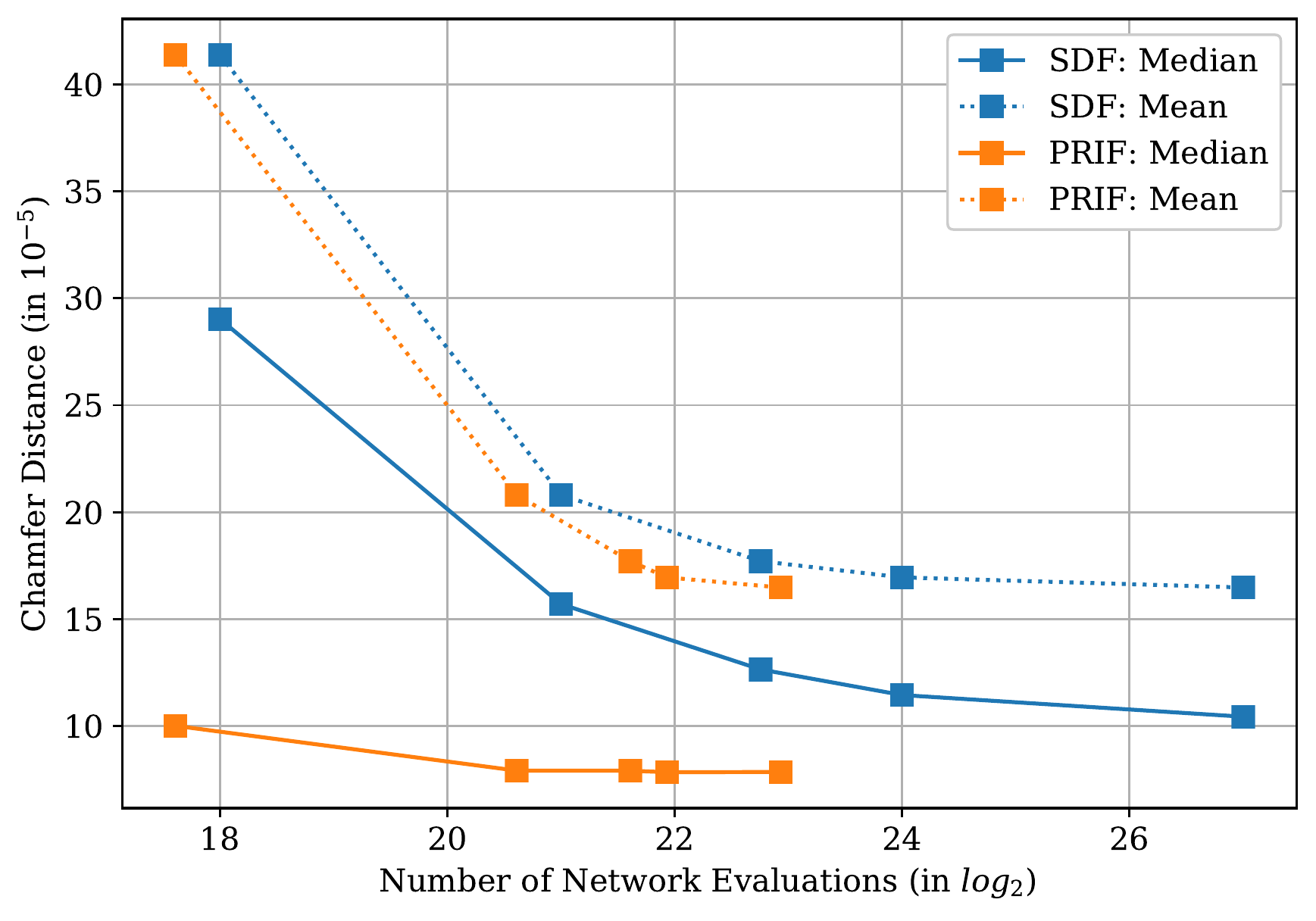}
	\end{subfigure}
    \caption{{\bf Complexity Analysis.} We evaluate the trained SDF network at grid resolutions of $64^3$, $128^3$, $192^3$, $256^3$, $512^3$ and obtain final meshes by Marching Cubes. We evaluate the trained PRIF network at resolutions of $5\times200^{2}$, $10\times 400^2$, $20\times 400^2$, $25\times 400^2$, $50\times 400^2$ (\# of Cameras $\times$ Resolution), and we apply Screened Poisson~\cite{kazhdan2013screened} to the output hit points to obtain the final mesh. The reconstructed mesh quality is measured by mean and median CD ($\times 10^{-5}$). The PRIF network achieves better quality with fewer evaluations.}
    \label{fig:Ablation}
\end{figure}

\begin{figure}[!ht]
    \vspace{-2mm}
	\begin{subfigure}{27mm}
		\centering
		\includegraphics[width=27mm]{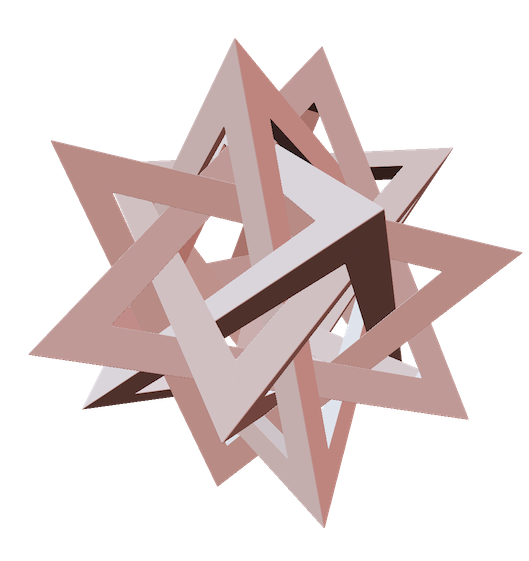}\vspace{-7mm}
	$\underbracket[0pt][1.0mm]{\hspace{\linewidth}}_%
    {\substack{ \colorbox{white}{Ground Truth}}}$
	\end{subfigure}%
	~
	\begin{subfigure}{27mm}
		\centering
		\includegraphics[width=27mm]{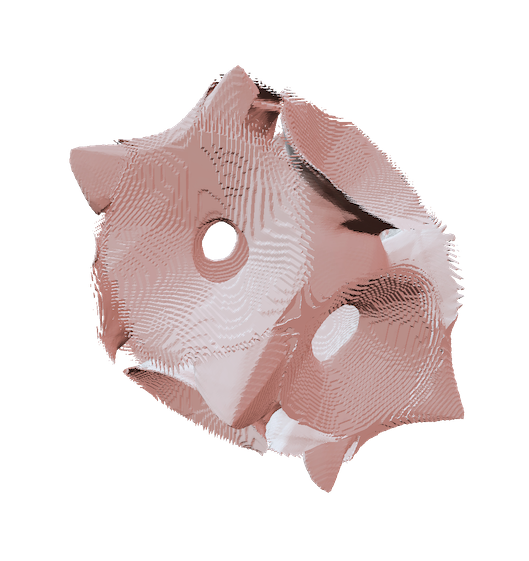}\vspace{-7mm}
	$\underbracket[0pt][1.0mm]{\hspace{\linewidth}}_%
    {\substack{ \colorbox{white}{OF}}}$
	\end{subfigure} 
	~
	\begin{subfigure}{27mm}
		\centering
		\includegraphics[width=27mm]{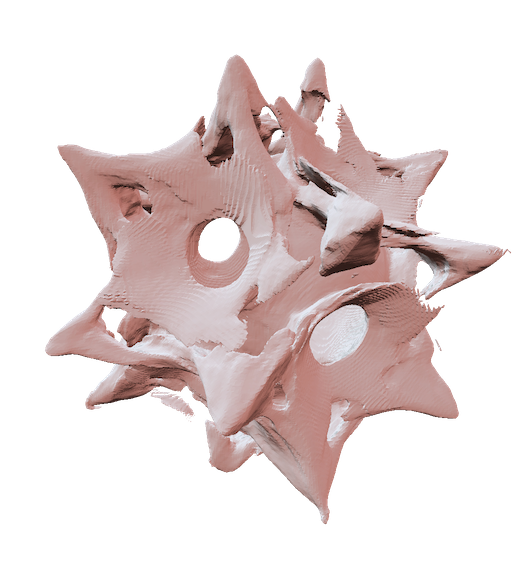}\vspace{-7mm}
	$\underbracket[0pt][1.0mm]{\hspace{\linewidth}}_%
    {\substack{ \colorbox{white}{SDF}}}$
	\end{subfigure} 
	~
	\begin{subfigure}{27mm}
		\centering
		\includegraphics[width=27mm]{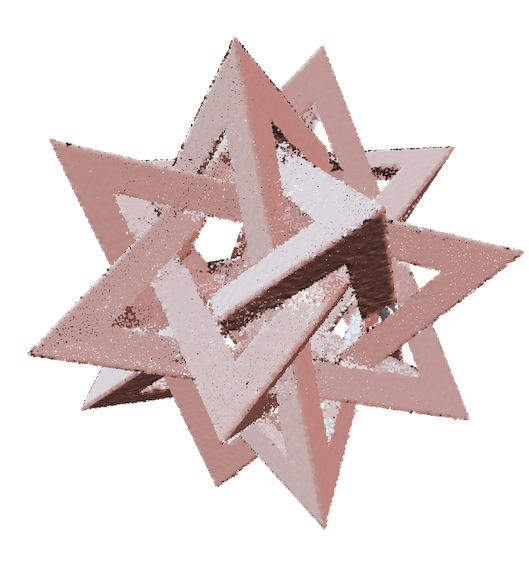}\vspace{-7mm}
	$\underbracket[0pt][1.0mm]{\hspace{\linewidth}}_%
    {\substack{ \colorbox{white}{Ours - Points}}}$
	\end{subfigure} 
	\caption{{\bf Stress Testing.} We test on a Tetrahedron grid that is self-intersecting and non-watertight. We obtain the SDF and OF values with the scanning method~\cite{KleinebergFW20}, and extract the mesh by Marching Cubes. While the level-set representations fail as expected, our method reliably preserves the shape.}
	\label{fig:Tetrahedron}
\end{figure}

\begin{figure}[!ht]
\centering
\includegraphics[height=27mm]{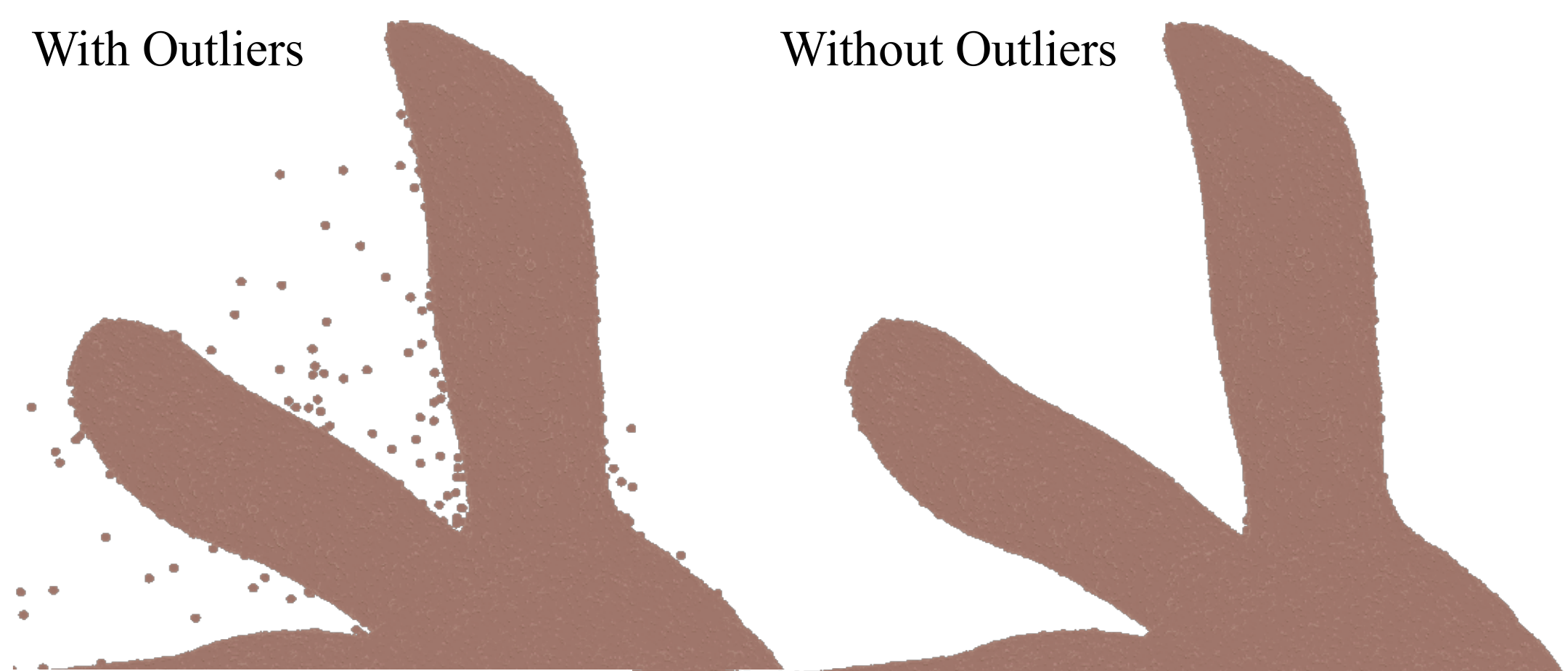}
\quad
\centering
\begin{tabular}[b]{c c c}
\toprule
Setup & Trained & Novel \\ \midrule
Proposed & \textbf{12.07}$\vert$\textbf{7.95} & \textbf{20.81}$\vert$\textbf{12.05}\\ 
$\textbf{f}_{r} \Rightarrow$ $\textbf{m}_{r}$ &  12.15$\vert$8.05  & 28.38$\vert$13.53 \\ 
$\textbf{f}_{r} \Rightarrow$ $\textbf{p}_{r}$ &  1.16$\vert$0.77  & 1924.4$\vert$1067.7\\
w/ Outliers &  12.14$\vert$8.06  & 21.32$\vert$12.05\\ 
\bottomrule
\end{tabular}

\captionlistentry[table]{A table beside a figure}
\captionsetup{labelformat=andtable}
\caption{{\bf Ablations.} {\it Left}: Impact of outlier removal described in Sec.~\ref{section:method}. While the outliers have little impact on the quantitative metrics since they are scant, removing them improves visual consistency. {\it Right}: Mean and median CD ($\times 10^{-5}$) of network prediction on trained rays and novel rays under different setup. Notice that changing $\textbf{f}_{r} \Rightarrow$ $\textbf{p}_{r}$ causes ray aliasing discussed in Sec.~\ref{section:method}, leading the network to overfit on trained rays and collapse on novel rays.}
\label{fig:AblationOutlier}
\end{figure}

\subsection{Shape Denoising and Completion}
As depth sensors have become available on mobile and AR/VR devices, there are an array of applications for persistent geometric reconstructions~\cite{Du2020DepthLab}. Given real-world data of sparse and noisy observations, the capability of generative modeling could enable shape recovery from noisy and incomplete point clouds. 


In Fig.~\ref{fig:ShapeCompletion}, we demonstrate the denoising and completion ability of a trained generative PRIF network.
Specifically, after training a PRIF network on an object category as described before, we provide the network with an unseen object's point cloud observations, which are either incomplete or contain noise.

\subsection{Analysis and Ablations}
\subsubsection{Complexity Analysis.} 
Shapes are commonly extracted from networks encoding SDF or OF using meshing algorithms like Marching Cubes.
This meshing step requires evaluating the network at many 3D grid sampling points, and the grid resolution affects the final quality of the mesh.
In our case, we can extract the shapes by rendering the scene from multiple positions and change the number of evaluations by varying the number and resolution of virtual cameras.
For a more objective comparison against prior arts that require meshing, we further apply Screened Poisson to mesh our output hit points.
In Fig.~\ref{fig:Ablation}, we analyze the trade-off between the number of network evaluations and the reconstructed mesh quality measured in mean and median CD.
Evidently, PRIF produces reconstruction with a better quality with fewer network evaluations.

\begin{figure}[!ht]
	\centering
	\begin{subfigure}{19mm}
		\centering
		\includegraphics[width=19mm]{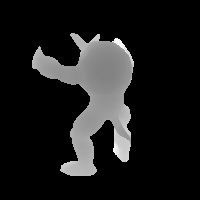}
	\end{subfigure}%
	~
	\begin{subfigure}{80mm}
		\centering
		\includegraphics[width=19mm]{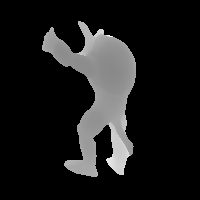}%
		~
		\includegraphics[width=19mm]{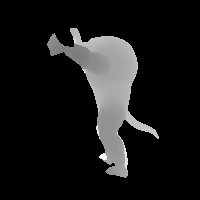}%
		~
		\includegraphics[width=19mm]{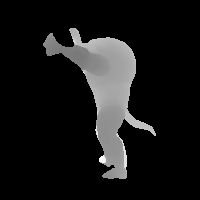}%
		~
		\includegraphics[width=19mm]{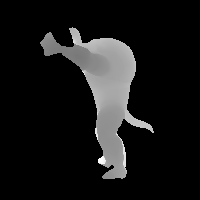}
	\end{subfigure}%
	~
	\begin{subfigure}{19mm}
		\centering
		\includegraphics[width=19mm]{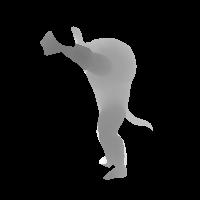}
	\end{subfigure} 
	
	\begin{subfigure}{19mm}
		\centering
		\includegraphics[width=19mm]{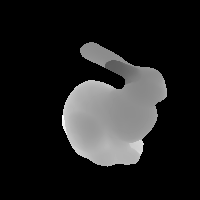}
		\label{fig:pose_bunny1}
		$\underbracket[0pt][1.0mm]{\hspace{\linewidth}}_%
    {\substack{\vspace{-5mm}\\ \colorbox{white}{Initial}}}$
	\end{subfigure}%
	~
	\begin{subfigure}{80mm}
		\centering
		\includegraphics[width=19mm]{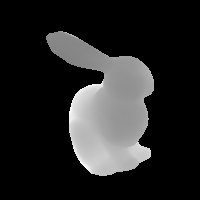}%
		~
		\includegraphics[width=19mm]{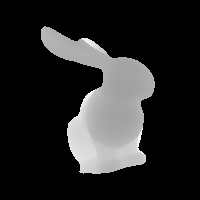}%
		~
		\includegraphics[width=19mm]{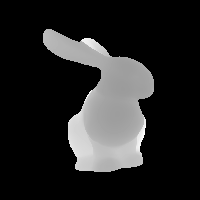}%
		~
		\includegraphics[width=19mm]{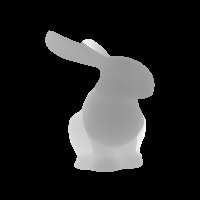}
		\label{fig:pose_bunny2}
    $\underbracket[1pt][1mm]{\hspace{\linewidth}}_%
    {\substack{\vspace{-5mm}\\ \colorbox{white}{Optimization Progress}}}$
	\end{subfigure}%
	~
	\begin{subfigure}{19mm}
		\centering
		\includegraphics[width=19mm]{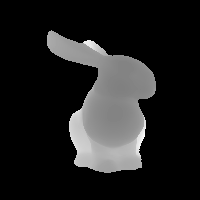}
		\label{fig:pose_bunny_gt}
		$\underbracket[0pt][1.0mm]{\hspace{\linewidth}}_%
    {\substack{\vspace{-5mm}\\ \colorbox{white}{Target}}}$
	\end{subfigure}%
	\caption{{\bf Learning Camera Poses.} Starting from the \textit{Initial} camera pose, we optimize the learnable pose parameters based on the difference between PRIF output rendered at the current pose and the PRIF output rendered at an unknown \textit{Target} pose. Here, we render the PRIF output as depth images. PRIF-based rendering successfully facilitates the differentiable optimization progress as the camera pose gradually converges to the correct \textit{Target} pose.}
	\label{fig:CameraPose}
\end{figure}

\begin{figure}[t]
	\centering
	\begin{subfigure}{27mm}
		\centering
		\includegraphics[width=27mm]{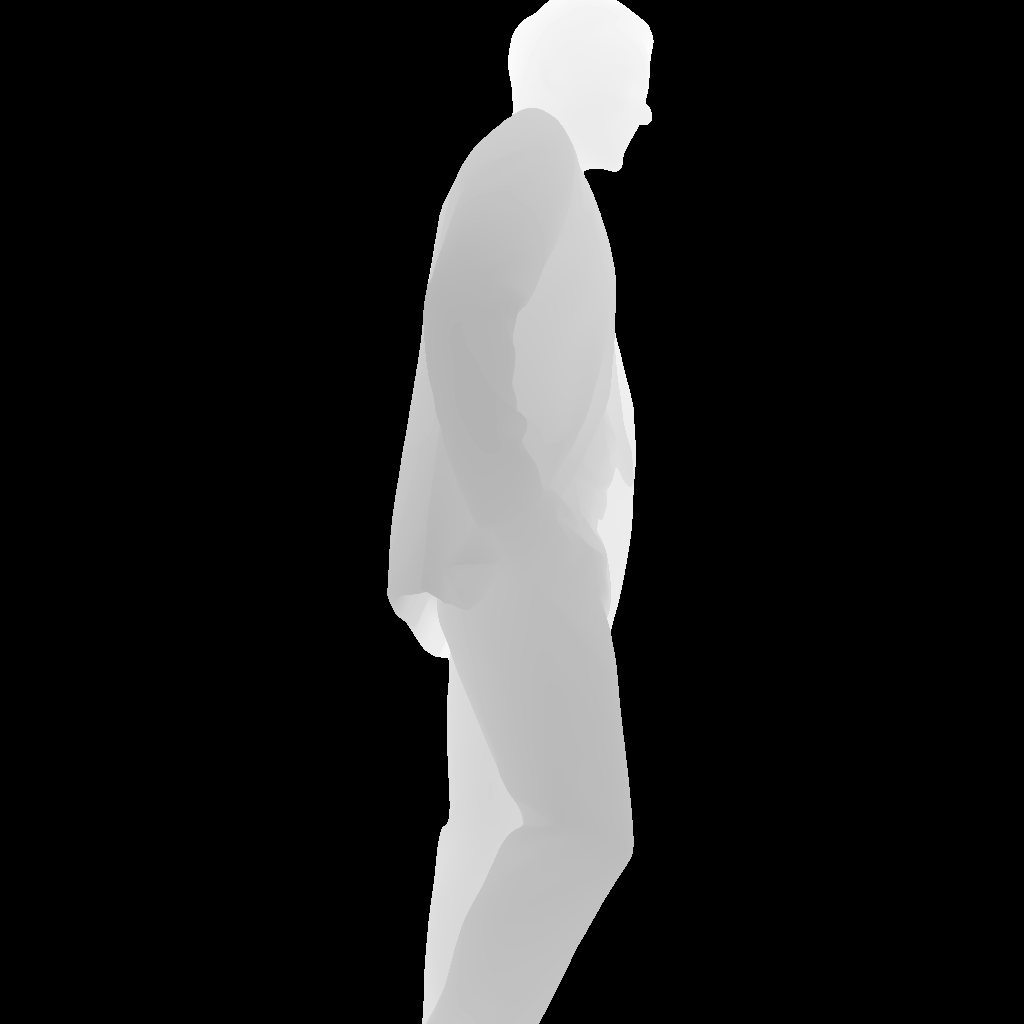}
	\end{subfigure}%
	~
	\begin{subfigure}{27mm}
		\centering
		\includegraphics[width=27mm]{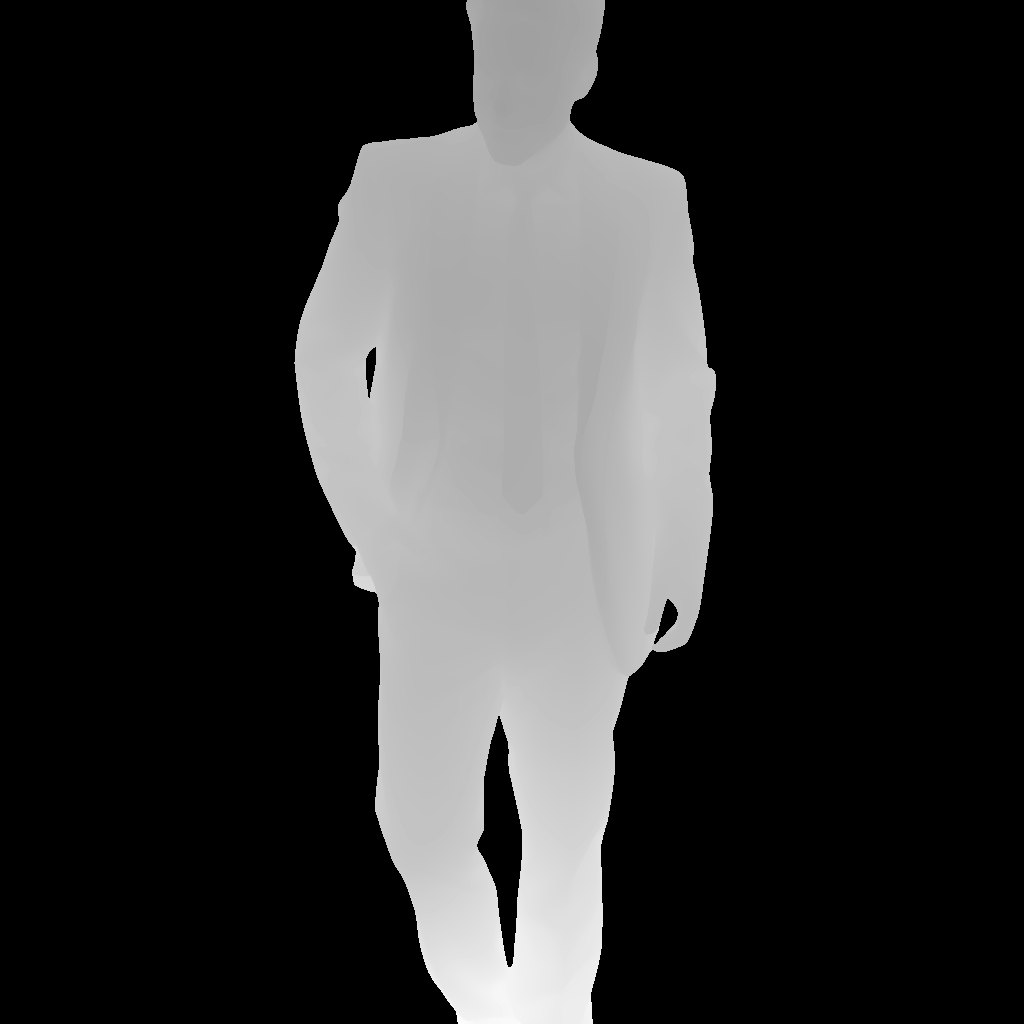}
	\end{subfigure}%
	~
	\begin{subfigure}{27mm}
		\centering
		\includegraphics[width=27mm]{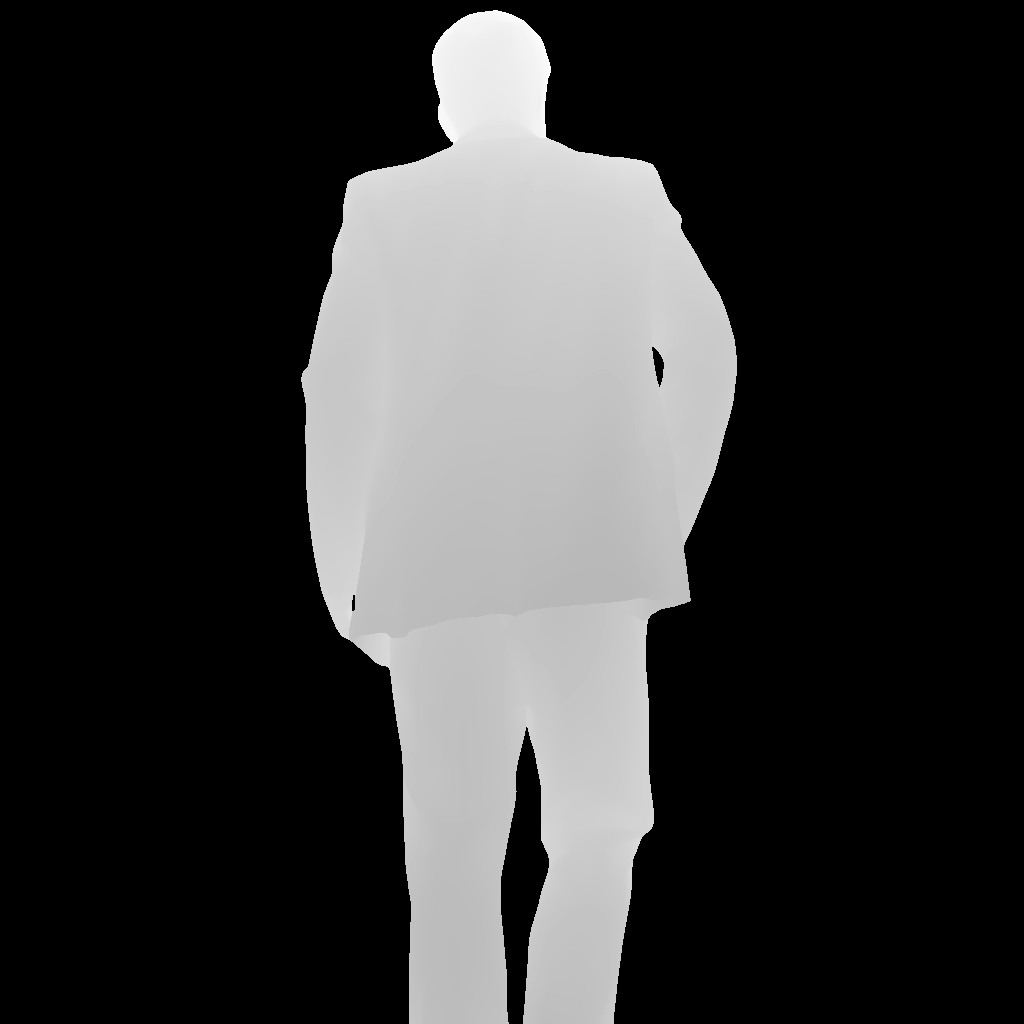}
	\end{subfigure}%
	~
	\begin{subfigure}{27mm}
		\centering
		\includegraphics[width=27mm]{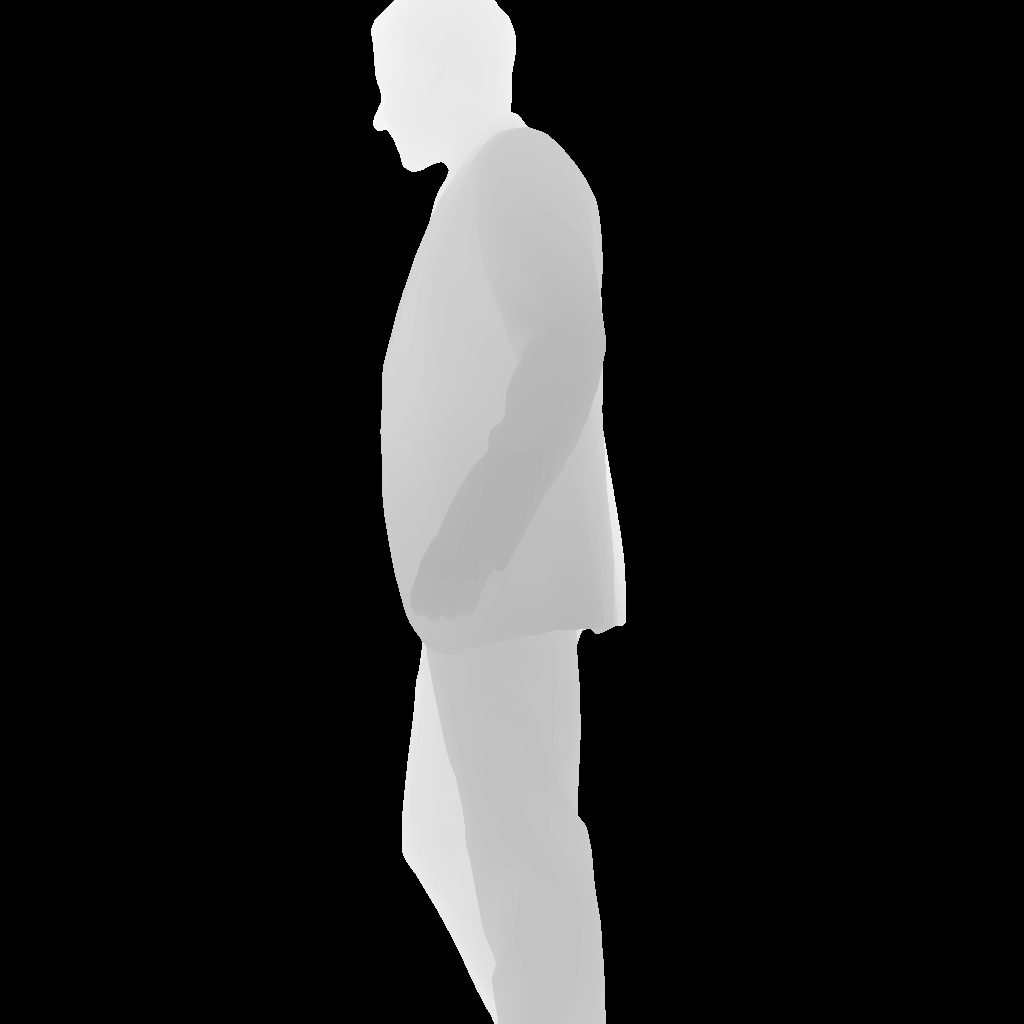}
	\end{subfigure}

	\begin{subfigure}{27mm}
		\centering
		\includegraphics[width=27mm]{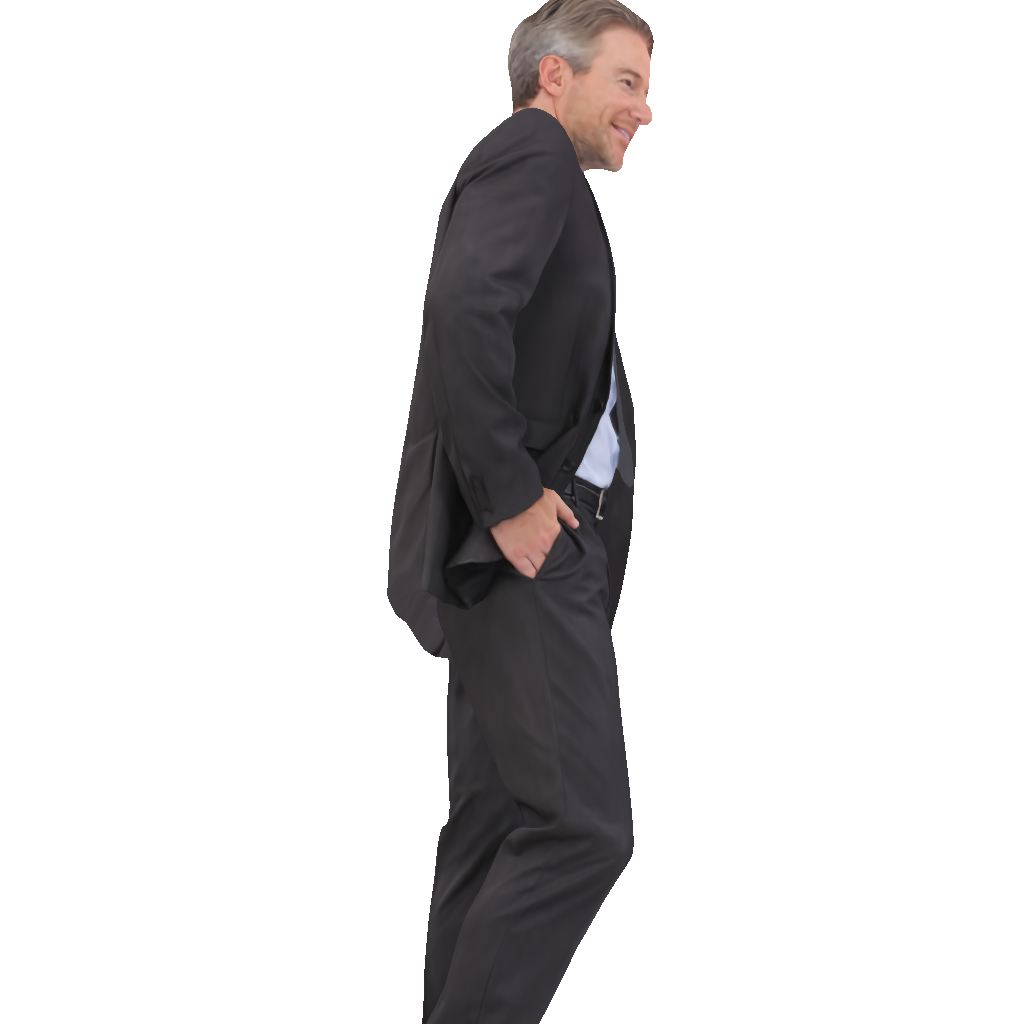}
	\end{subfigure}%
	~
	\begin{subfigure}{27mm}
		\centering
		\includegraphics[width=27mm]{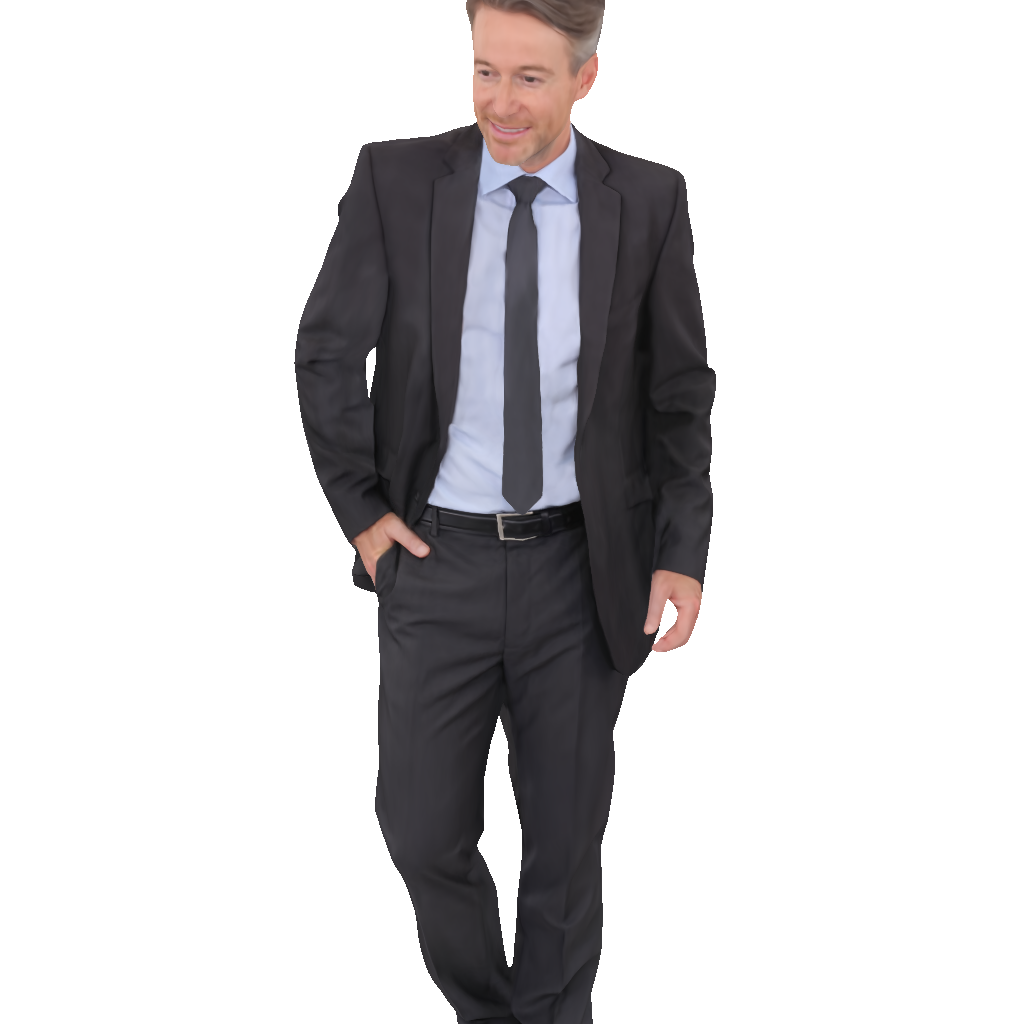}
	\end{subfigure} 
	~
	\begin{subfigure}{27mm}
		\centering
		\includegraphics[width=27mm]{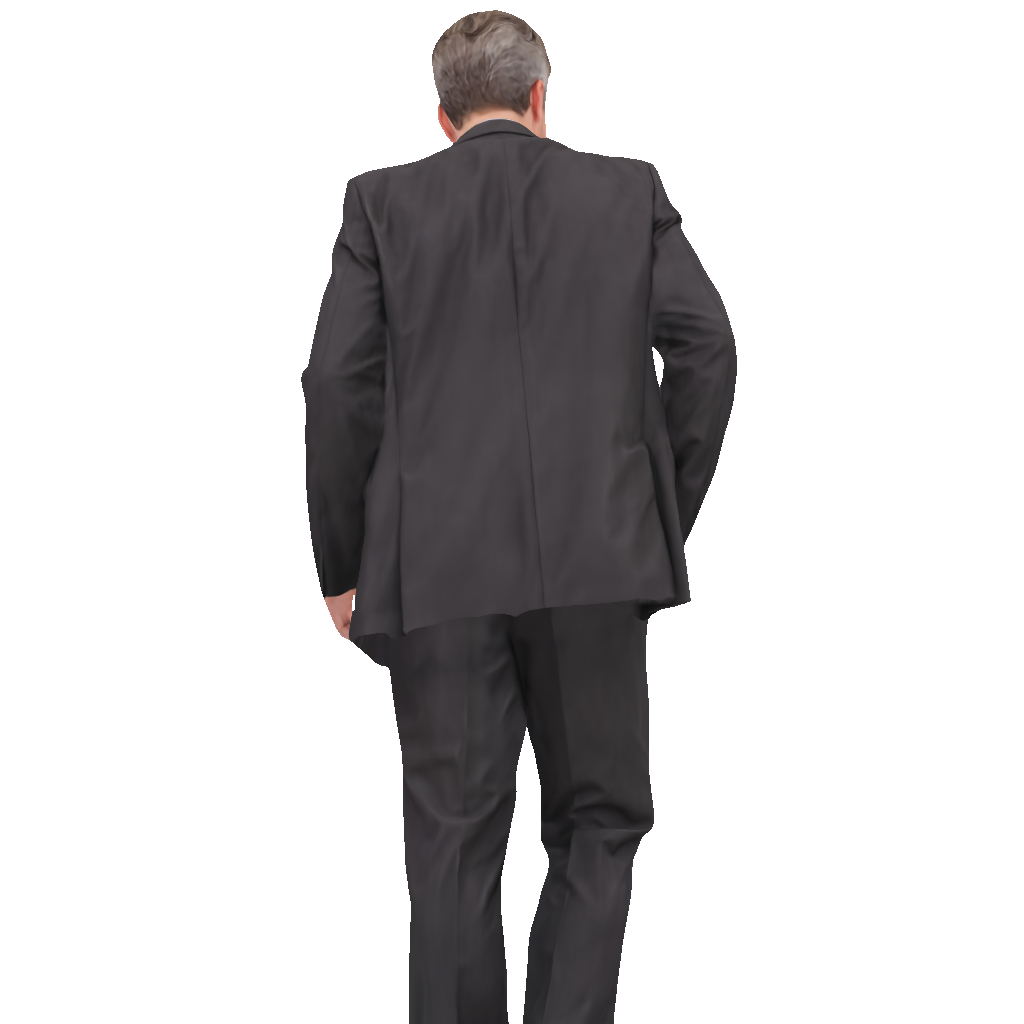}
	\end{subfigure} 
	~
	\begin{subfigure}{27mm}
		\centering
		\includegraphics[width=27mm]{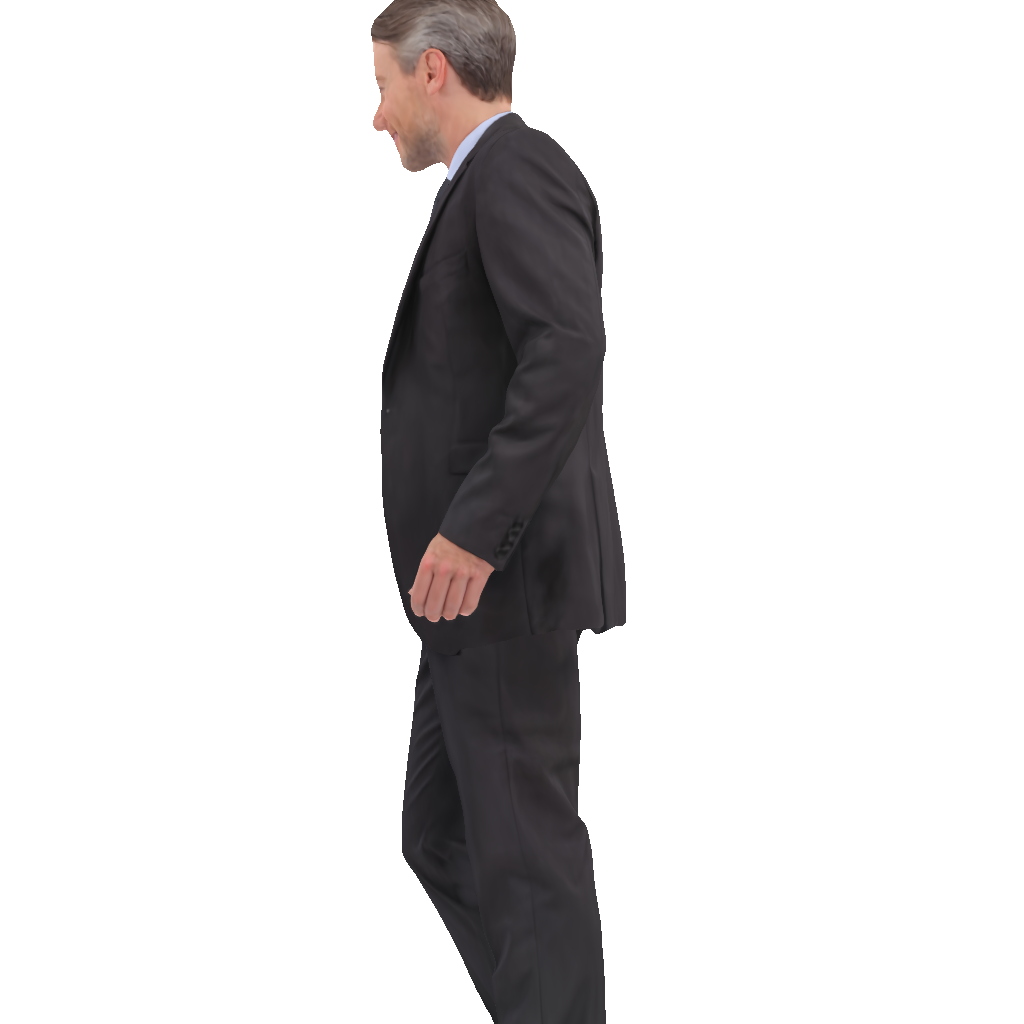}
	\end{subfigure} 
	\caption{{\bf Incorporating Color Rendering.} After training the PRIF network to encode the geometry (rendered as depth images in the first row), we can further train a second-stage network that takes in the hit point produced by PRIF and produces its corresponding color (rendered in the second row).}
	\label{fig:RGBRender}
\end{figure}

\subsubsection{Stress Testing.} 
Fig.~\ref{fig:Tetrahedron} shows an example of encoding a self-intersecting and non-watertight shape, which is expected to be challenging for neural networks trained to encode conventional functional representations like SDF and OF.
In contrast, the shape is well preserved by the network trained to encode our proposed functional representation PRIF.
\subsubsection{Ablations.}
We present ablation studies validating the effectiveness of our proposed techniques in Fig.~\ref{fig:AblationOutlier} and Table 3.
The results verify that: 1) not reparametrizing $r = (\textbf{p}_{r}, \textbf{d}_{r})$ leads to severe overfitting on the training data, and 2) replacing the Plucker moment vector $\textbf{m}_{r}$ with our proposed $\textbf{f}_{r}$ improves the encoding quality, possibly because the network only has to learn a simple affine transform matrix as mentioned in Sec. 3.2.

\subsection{Further Applications}
\subsubsection{Learning Camera Poses.}
Evaluating the PRIF network on some input camera rays is essentially performing differentiable rendering of the underlying scene represented by the network.
With a PRIF network trained on a 3D shape, given a silhouette image at an unknown camera pose, our task is to recover that camera pose based on the silhouette image.
In this task, the weights of the PRIF network are frozen, and the only learnable parameters are the camera pose matrix.
At each iteration, we render the scene through the PRIF network with camera rays defined by the current estimate of the camera pose, and we adjust the estimate based on the silhouette difference between the observed and rendered images.
In Fig.~\ref{fig:CameraPose}, we present the rendered image during the optimization steps. 
The estimated camera pose gradually converges to the correct solution as the rendered image becomes more similar to the observed image.

\subsubsection{Neural Rendering with Color.}
The PRIF network behaves as a geometry renderer since it effectively returns the surface hit point given a viewing ray.
Here we show that the PRIF network can be further extended to rendering color at the surface.
We select a 3D model of human from the Renderpeople dataset~\cite{RenderPeople} and virtually capture the hit points and RGB colors at 25 locations surrounding the model.
We then train the PRIF network to fit the geometric shape, similar to the single shape experiments.
Finally, we extend the network to produce the observed RGB color given the input ray while fixing its hit point prediction.
Fig.~\ref{fig:RGBRender} shows the rendered results of geometry and appearance of the 3D model.

\subsubsection{Limitation.}
This paper is focused on 3D shape encoding and decoding through hit points of rays from known views.
PRIF can render at novel views by sampling new rays, but to guarantee consistent novel view results would likely require further modifications such as multi-view consistency loss or denser training views.

\section{Conclusion}
We propose PRIF, a new 3D shape representation based on the relationship between a ray and its perpendicular foot with the origin.
We demonstrate that neural networks can successfully encode PRIF to achieve accurate shape representations.
With this new representation, we avoid multi-sample sphere tracing and obtain the hit point with a single network evaluation.
Neural networks trained to encode PRIF inherit such advantages and can represent shapes more accurately than common neural shape representations using the same network architecture.
We further extend the neural PRIF networks to enable various downstream tasks including generative shape modeling, shape denoising and completion, camera pose optimization, and color rendering.
Promising future directions include using spatial partitions to improve the network accuracy, jointly learning the geometry and view-dependent color directly from images, speeding up network training, and real-time inference for robotic applications.

%
%
\bibliographystyle{splncs04}
\bibliography{prif}

\clearpage
\section{Supplementary Material}

\subsection{Training Setup}
\subsubsection{Optimization.} In all of our experiments, we train the MLPs in PyTorch using the Adam optimizer with the default parameters. We supervise the network based on the mean squared error loss between reconstructed color and ground truth color. 
We adopt the cosine annealing learning rate scheduler, with initial learning rate as $10^{-4}$ and final learning rate as $10^{-7}$. We randomly shuffled all the training samples and divided them into batches of size equal to 1,024. 
To ensure reproducibility, we set the random seed as 0. \\
On a machine with NVIDIA GeForce RTX 2080 TI, training 10 epochs takes around 4 minutes for the SDF and OF condition, and 13 minutes for PRIF.

\subsubsection{Network Architecture.} We set all networks to have 10 layers with residual connections, where the 8 intermediate layers having a dimension size of $512 \times 512$.
We apply layer normalization after all MLP layers except the final one.
When we further extend the network to model color or background mask, we simply invoke a second network with the same architecture, as network compression rate or total parameter count is not the main focus of the current work.

\subsection{Runtime Discussions}
\subsubsection{Rendering Speed.}
The formulation of PRIF inherently allows us to bypass the inefficient multi-sample issue of the previous methods that adopt level-set functional representations.
To more quantitatively measure the speed-up, we measure the number of network evaluations (Queries) and the actual time to render a $512 \times 512$ depth image from a trained neural representation on the same machine with an NVIDIA GeForce RTX 2080 Ti graphics card.
\begin{table}
\centering
\begin{tabular}{c c c c}
\toprule
Method & Steps & Queries & Time (s) \\ \midrule
DeepSDF + Sphere Tracing &  100 & 2,453,700 & 1.01070 \\ 
DeepSDF + DIST Acceleration &  100 & 382,325 & 0.30539 \\ 
PRIF &  \textbf{1} & \textbf{262,144} & \textbf{0.00578} \\ 
\bottomrule
\end{tabular}
\vspace{4pt}
\caption{Quantitative results on the cost to render a $512 \times 512$ depth image from trained neural shape representations. {\it Queries} refer to the number of network evaluations, and {\it Steps} refer to the maximum sphere tracing steps specified in the actual implementation. {\it Steps} = 100 is in line with the original DIST implementation from Liu~\etal~\cite{Liu2020DIST}.}
\label{tab:runtime_compare}
\end{table}
We first consider a baseline method where a trained DeepSDF network is used to perform naive sphere tracing.
We then consider an improved baseline where the DeepSDF-based  sphere tracing is sped up by various techniques proposed by Liu~\etal~\cite{Liu2020DIST}.
Both baselines are adopted from the implementation of Liu~\etal, and correspond to two conditions (trivial and recursive) in their implementation.

\subsubsection{Meshing Speed}
The overall meshing speed depends on {\it network inference} + {\it surface reconstruction}.
We examine the trade-off by decoding the same shape with PRIF and SDF trained with networks of equivalent capacity.
Our method spends \underline{1.01s} at inference, then \underline{2.65s} to obtain the triangle mesh from MeshLab (Poisson with normal estimation).
The SDF-based network (similar to DeepSDF) spends \underline{6.97s} for inference, then \underline{1.19s} for Marching Cubes.
We also note that many practical scenarios do not require the meshing process, such as camera pose registration and neural rendering.

\subsection{More Ablations}
\subsubsection{Comparison to Plücker Coordinate.}
We use the original Plücker coordinate for the shape generation task for the {\it Car} category. Below are the results (with mean/median Chamfer distance):
\begin{table}[!ht]
\centering
\scalebox{0.8}{
\begin{tabular}{c c c c c c c}
\toprule
Method & Car & Chair & Table & Plane & Lamp & Sofa \\ \midrule
Plain Plücker &  2.397$\vert$0.547 & 2.603$\vert$0.531   &   5.054$\vert$0.554   &  0.766$\vert$0.199  &   4.838$\vert$1.094   &   1.738$\vert$0.305    \\ 
Proposed &  \textbf{1.961}$\vert$\textbf{0.347} &  \textbf{0.982}$\vert$\textbf{0.267}   &   \textbf{4.532}$\vert$\textbf{0.315}   &  \textbf{0.389}$\vert$\textbf{0.125} &    \textbf{3.276}$\vert$\textbf{0.534}   &   \textbf{1.236}$\vert$\textbf{0.222}   \\
\bottomrule
\end{tabular}
}
\end{table}\\
Although the plain Plücker formulation works well for single shape representation (Section 5 Table 3), it performs much worse in the generative task.
We find the network trained with Plücker coordinates often fails to predict correct background masks, leading to many noisy points and worse performance overall.
Compared to the Plücker moment vector, our proposed reference points is better geometrically related to the surface hitpoint, which is our overarching goal of shape representation and rendering.

\subsubsection{Varying Model complexity.}
We examine the effect of varying the model capacity (by changing MLP layer width) for the generation task on shapes from the {\it Car} category.
\begin{table}[!h]
\centering
\scalebox{0.8}{
\begin{tabular}{c c c c c | c}
\toprule
\# Param (in Millions) & 2.96 & 2.23 & 1.58 & 1.35 & SDF (2.10)\\ \midrule
Mean$\vert$Median CD &  1.961$\vert$0.347 & 2.026$\vert$0.364   &   2.058$\vert$0.368  &  2.055$\vert$0.386 & 2.315$\vert$0.495 \\ 
\bottomrule
\end{tabular}
}
\end{table}\\
Results show we can decrease the number of parameters and still achieve better performance than SDF.

\subsubsection{Different Noise Levels.}
We perform additional denoising experiments (as Fig. 5 in the paper) and present the impact of different noise levels on the shape representation accuracy on the {\it Car} category (measured by Chamfer distance):
\begin{table}[!ht]
\centering
\scalebox{0.8}{
\begin{tabular}{c c c c c c}
\toprule
Noise Level & 0.01 & 0.02 & 0.04 & 0.08 & 0.1 \\ \midrule
Mean$\vert$Median CD & 1.929$\vert$0.368  &  1.931$\vert$0.428 & 2.005$\vert$0.590   &   2.471$\vert$0.951   &  2.842$\vert$1.123\\ 
\bottomrule
\end{tabular}
}
\end{table}

\subsection{Additional Results}

\begin{figure}[!ht]
    \centering
	\begin{subfigure}{27.5mm}
		\centering
		\includegraphics[width=27.5mm]{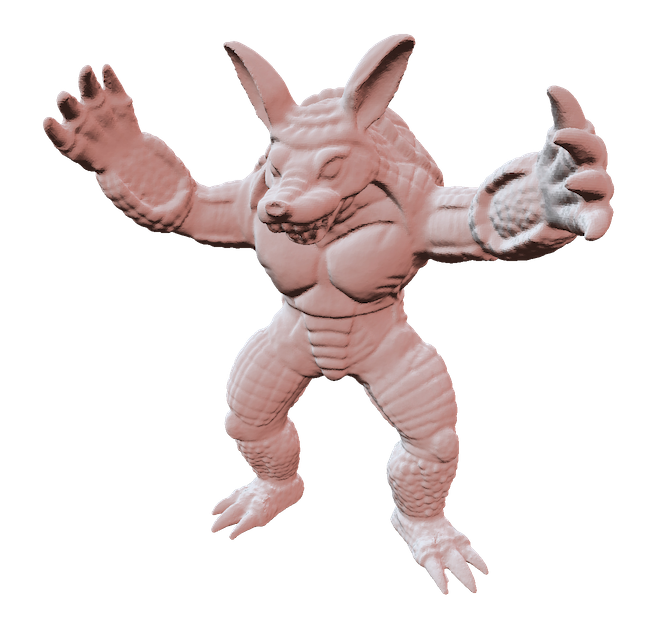}
	\end{subfigure}
	~
	\begin{subfigure}{27.5mm}
		\centering
		\includegraphics[width=27.5mm]{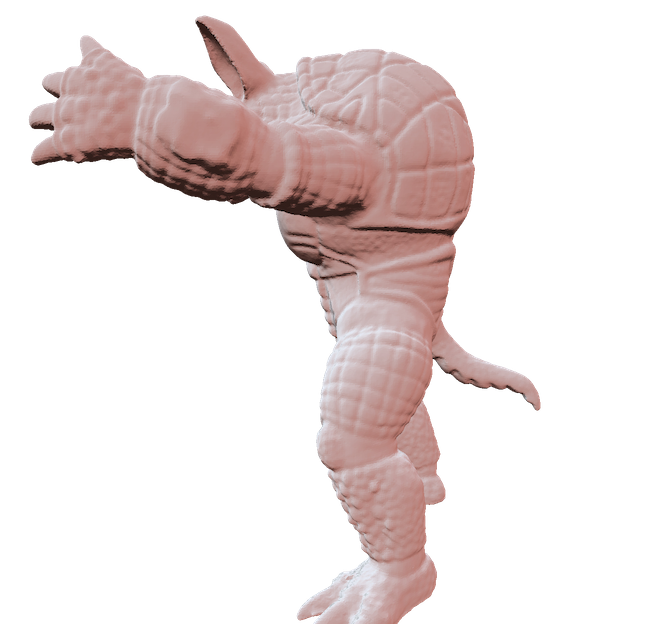}
	\end{subfigure} 
	~
	\begin{subfigure}{27.5mm}
		\centering
		\includegraphics[width=27.5mm]{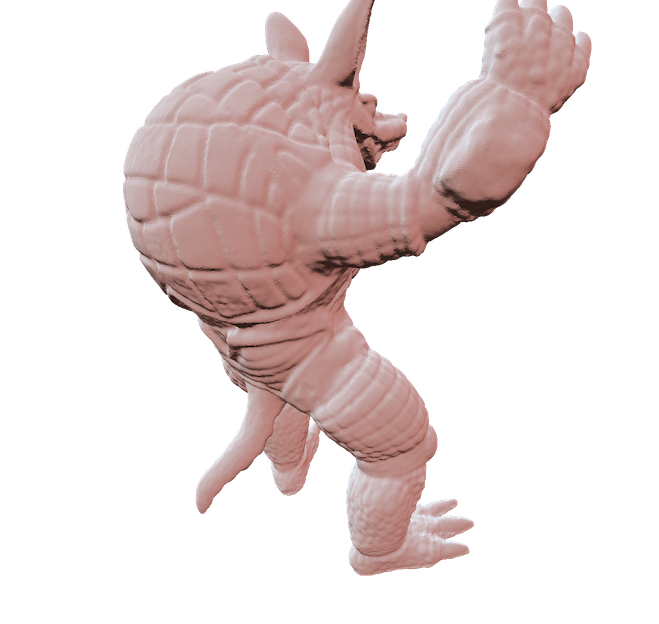}
	\end{subfigure}
	~
	\begin{subfigure}{27.5mm}
		\centering
		\includegraphics[width=27.5mm]{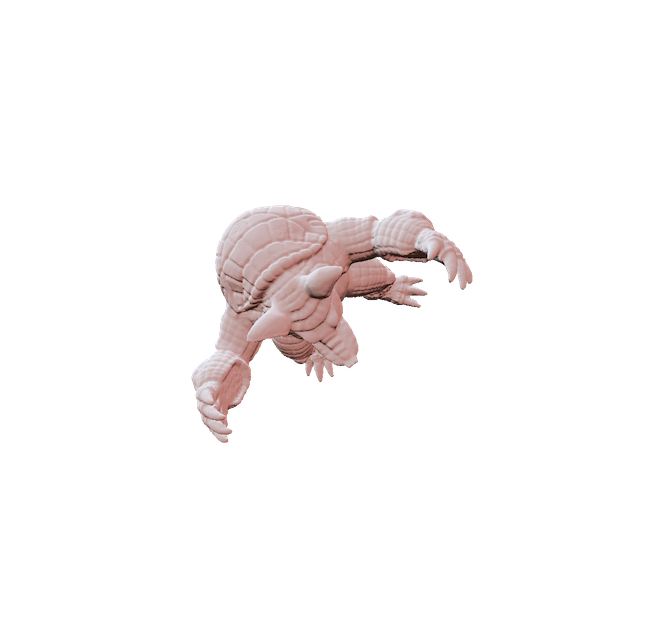}
	\end{subfigure}  \\
	\begin{subfigure}{27.5mm}
		\centering
		\includegraphics[width=27.5mm]{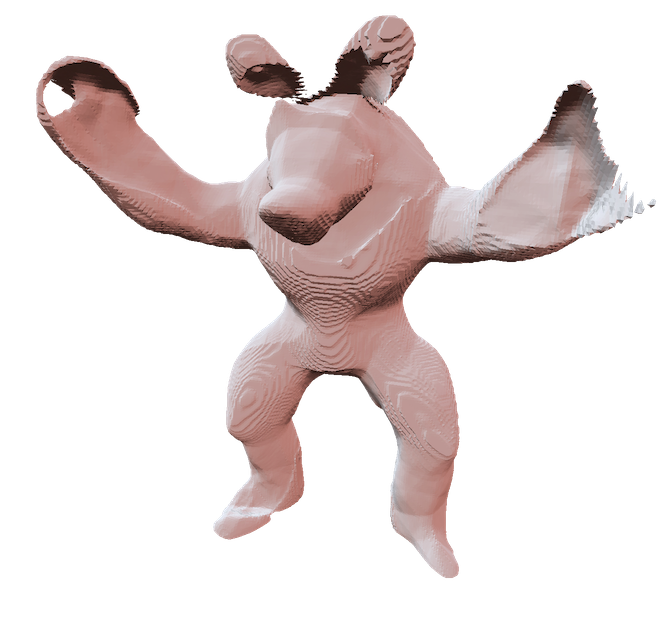}
	\end{subfigure}
	~
	\begin{subfigure}{27.5mm}
		\centering
		\includegraphics[width=27.5mm]{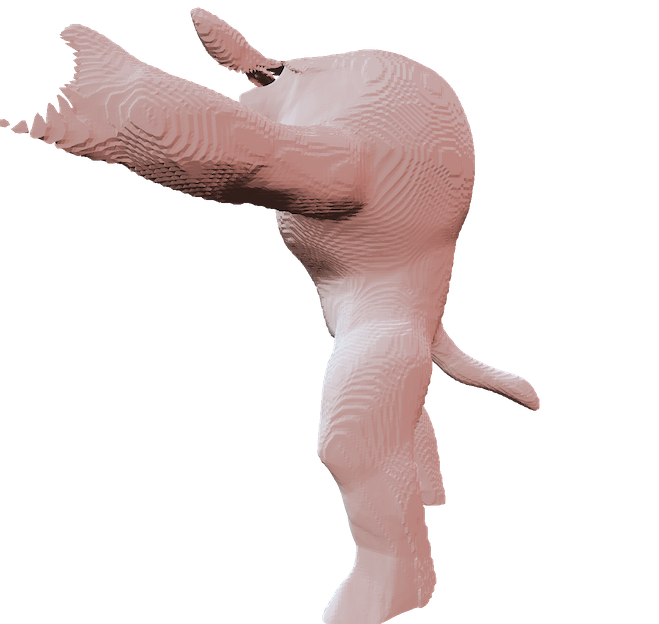}
	\end{subfigure} 
	~
	\begin{subfigure}{27.5mm}
		\centering
		\includegraphics[width=27.5mm]{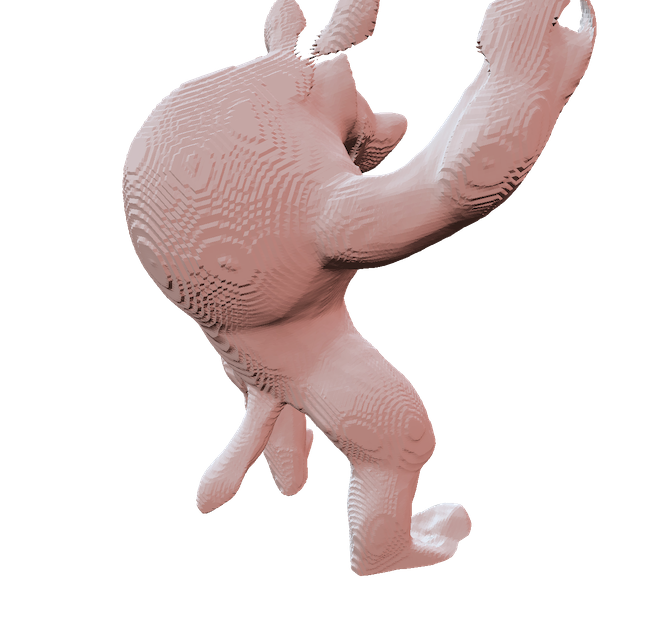}
	\end{subfigure} 
	~
	\begin{subfigure}{27.5mm}
		\centering
		\includegraphics[width=27.5mm]{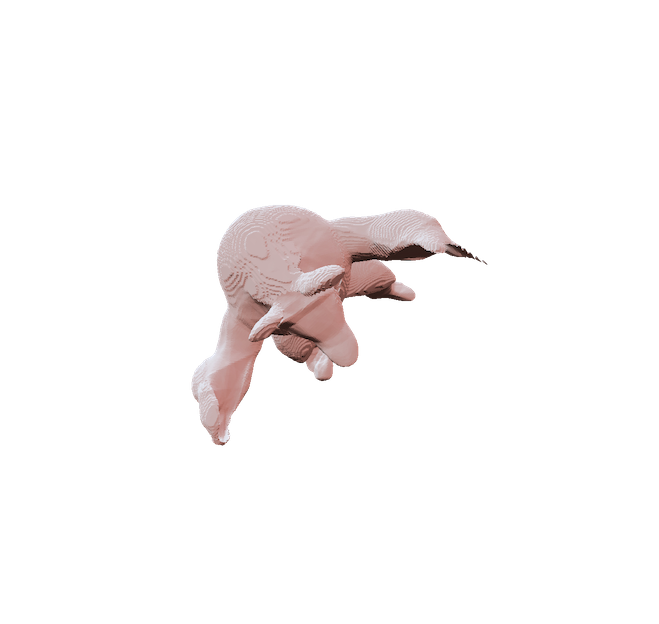}
	\end{subfigure}  \\
	\begin{subfigure}{27.5mm}
		\centering
		\includegraphics[width=27.5mm]{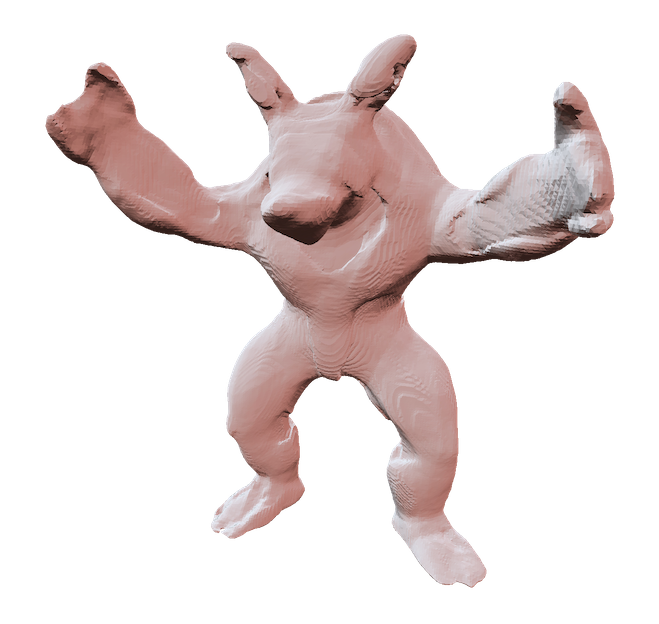}
	\end{subfigure}
	~
	\begin{subfigure}{27.5mm}
		\centering
		\includegraphics[width=27.5mm]{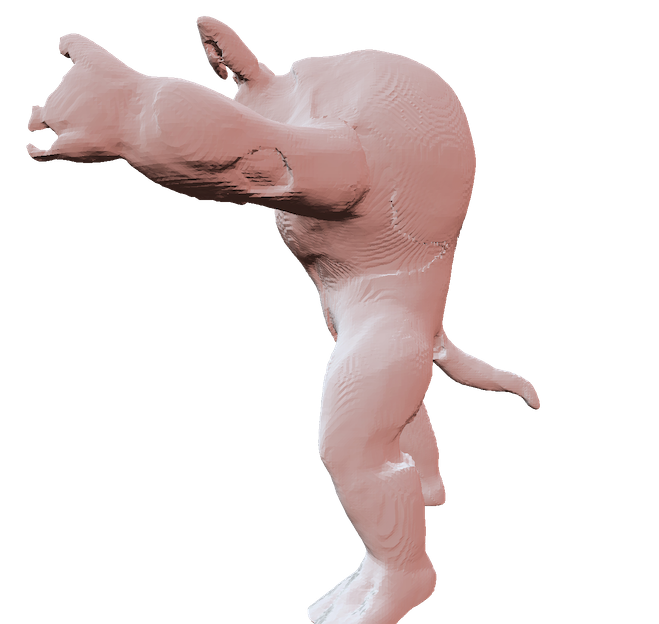}
	\end{subfigure} 
	~
	\begin{subfigure}{27.5mm}
		\centering
		\includegraphics[width=27.5mm]{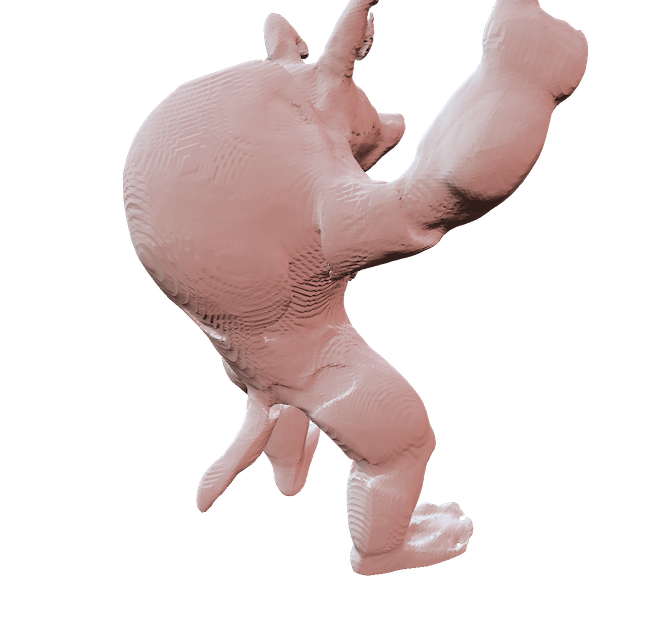}
	\end{subfigure} 
	~
	\begin{subfigure}{27.5mm}
		\centering
		\includegraphics[width=27.5mm]{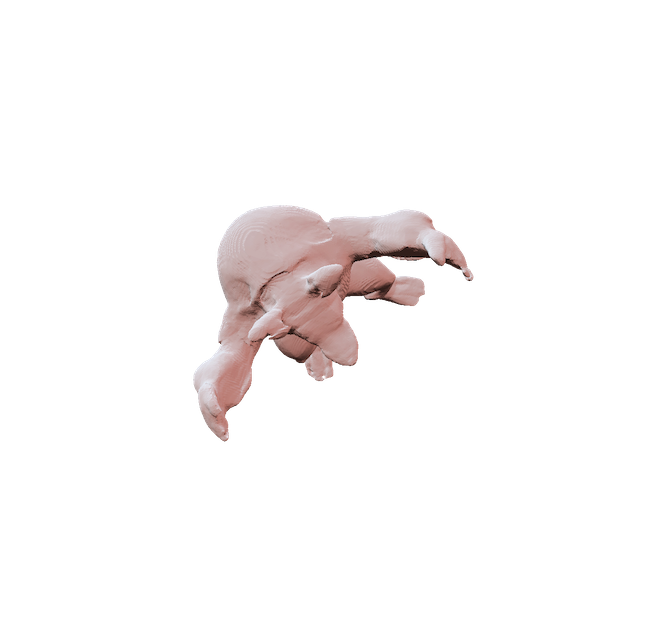}
	\end{subfigure} \\
	\begin{subfigure}{27.5mm}
		\centering
		\includegraphics[width=27.5mm]{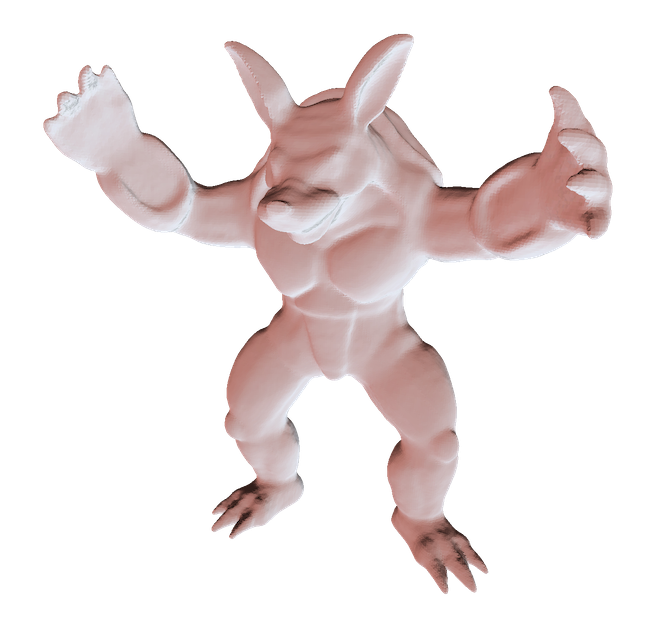}
	\end{subfigure}
	~
	\begin{subfigure}{27.5mm}
		\centering
		\includegraphics[width=27.5mm]{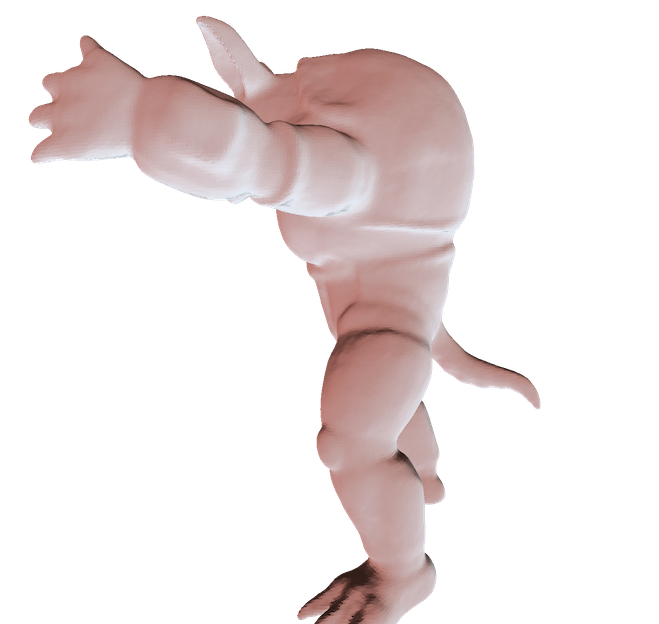}
	\end{subfigure} 
	~
	\begin{subfigure}{27.5mm}
		\centering
		\includegraphics[width=27.5mm]{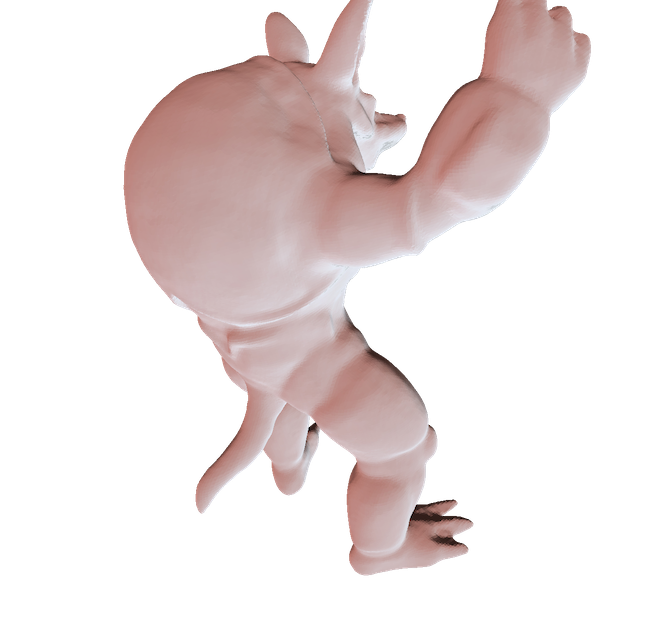}
	\end{subfigure} 
	~
	\begin{subfigure}{27.5mm}
		\centering
		\includegraphics[width=27.5mm]{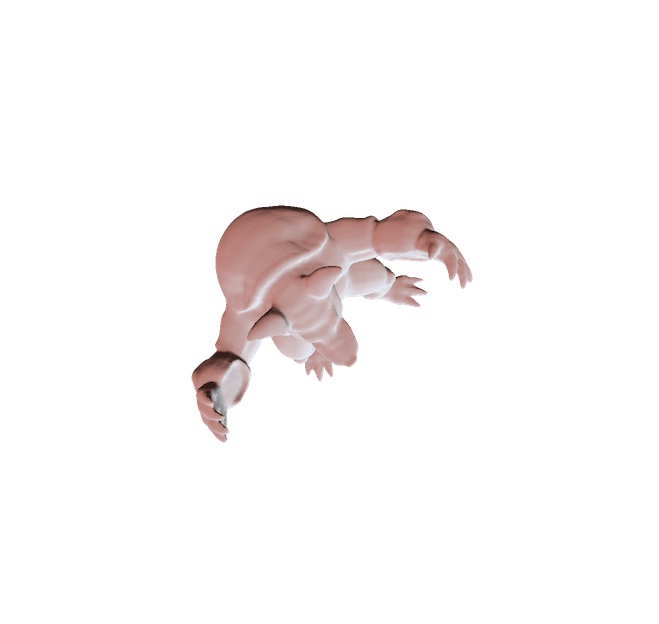}
	\end{subfigure} \\
	\caption{{\bf More Single Shape from Different Angles.} Results from Rows 1 to 4 correspond to {\bf Reference}, {\bf OF}, {\bf SDF}, {\bf PRIF - Mesh}.}
\end{figure}

\begin{figure}[!ht]
    \centering
	\begin{subfigure}{27.5mm}
		\centering
		\includegraphics[width=27.5mm]{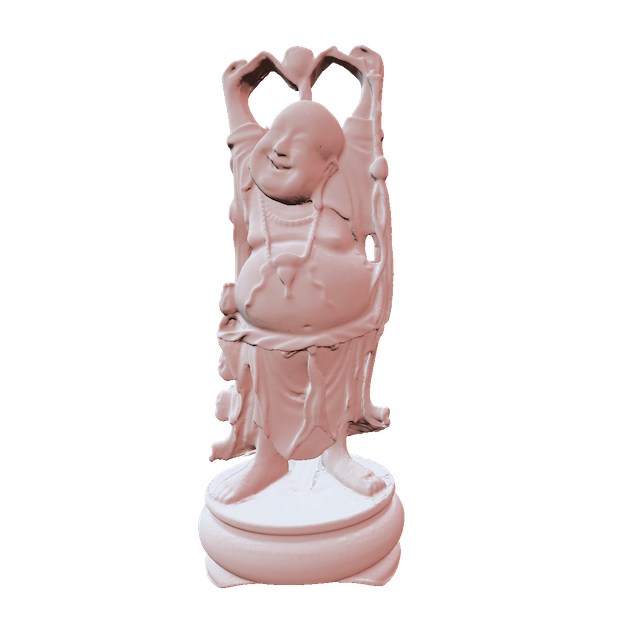}
	\end{subfigure}
	~
	\begin{subfigure}{27.5mm}
		\centering
		\includegraphics[width=27.5mm]{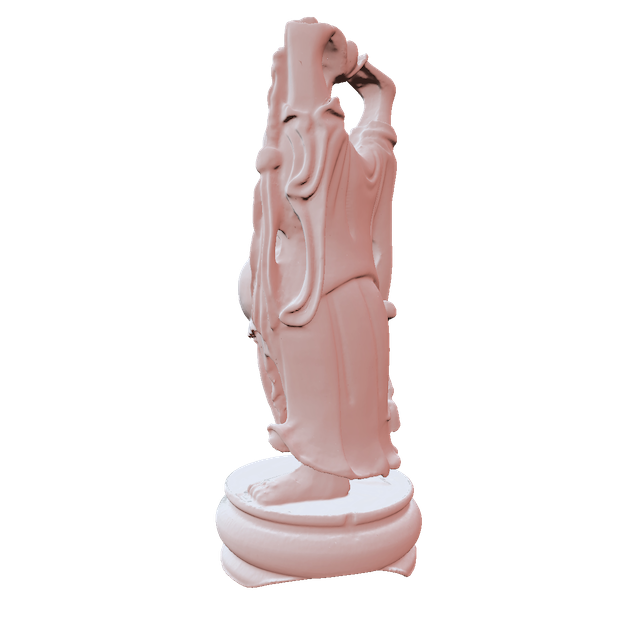}
	\end{subfigure} 
	~
	\begin{subfigure}{27.5mm}
		\centering
		\includegraphics[width=27.5mm]{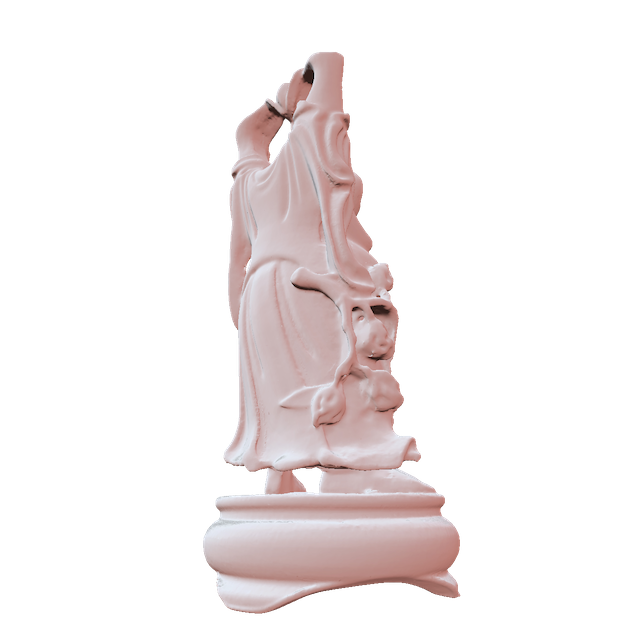}
	\end{subfigure}
	~
	\begin{subfigure}{27.5mm}
		\centering
		\includegraphics[width=27.5mm]{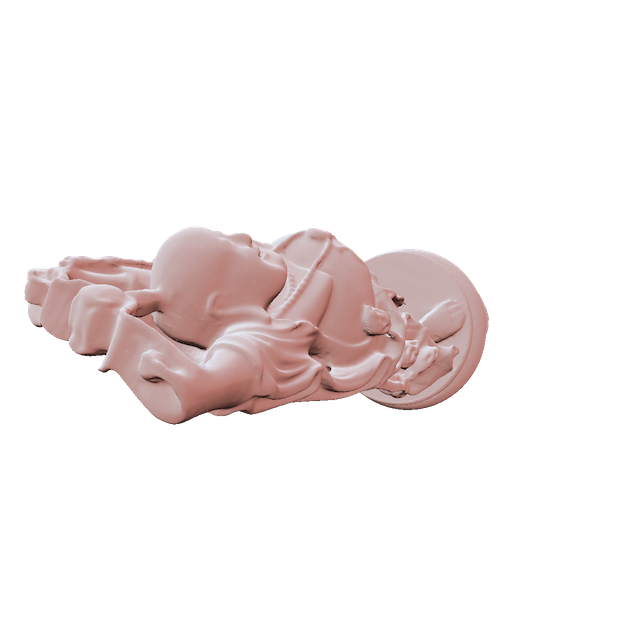}
	\end{subfigure}  \\
	\begin{subfigure}{27.5mm}
		\centering
		\includegraphics[width=27.5mm]{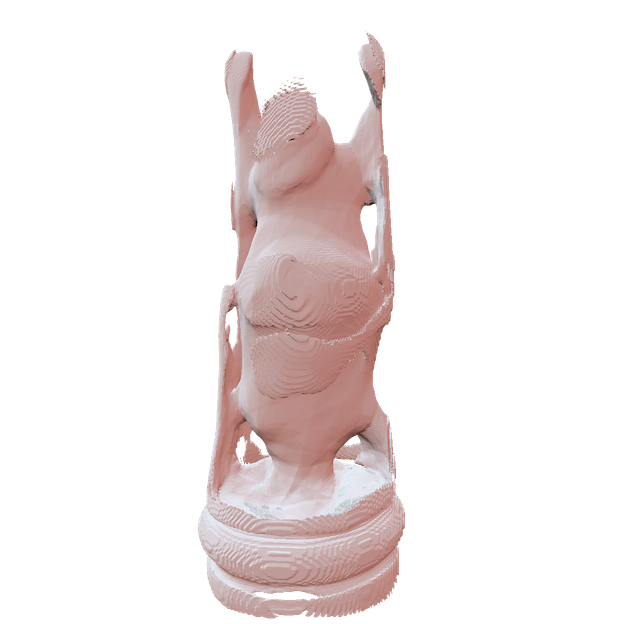}
	\end{subfigure}
	~
	\begin{subfigure}{27.5mm}
		\centering
		\includegraphics[width=27.5mm]{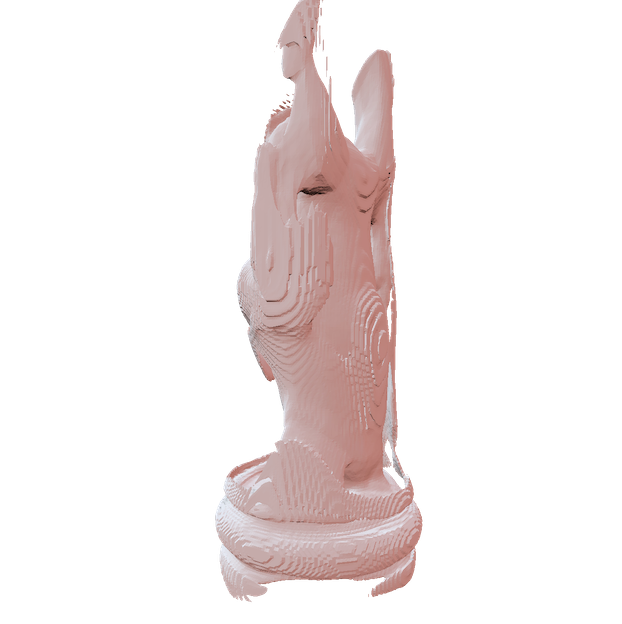}
	\end{subfigure} 
	~
	\begin{subfigure}{27.5mm}
		\centering
		\includegraphics[width=27.5mm]{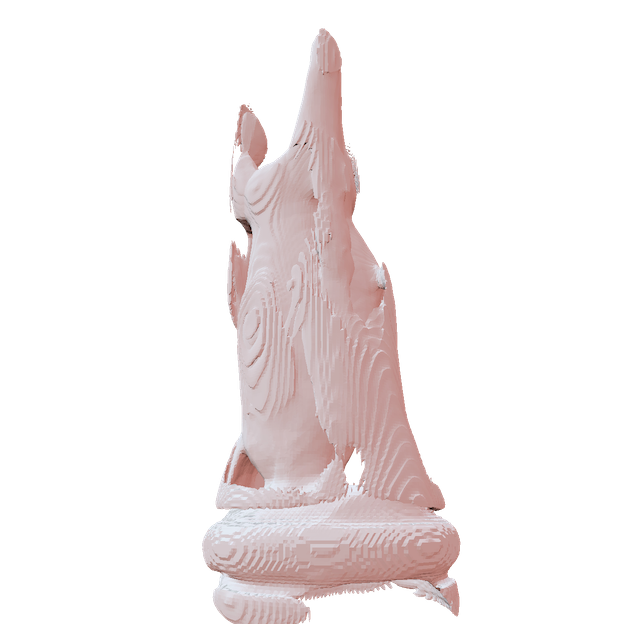}
	\end{subfigure} 
	~
	\begin{subfigure}{27.5mm}
		\centering
		\includegraphics[width=27.5mm]{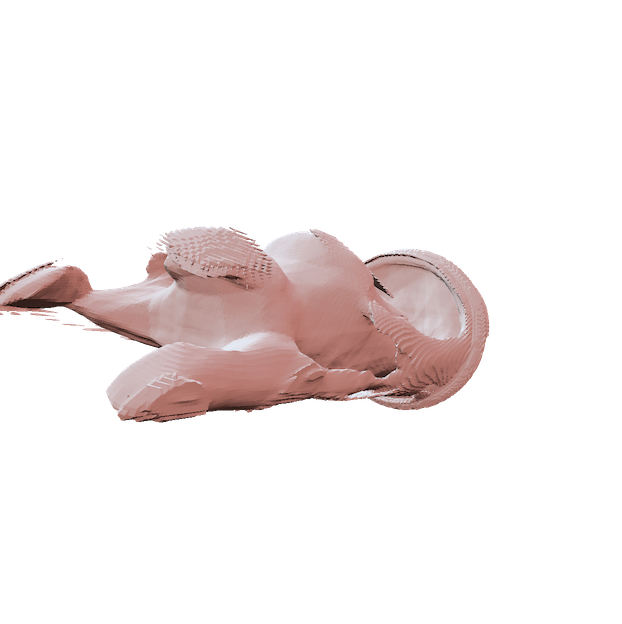}
	\end{subfigure}  \\
	\begin{subfigure}{27.5mm}
		\centering
		\includegraphics[width=27.5mm]{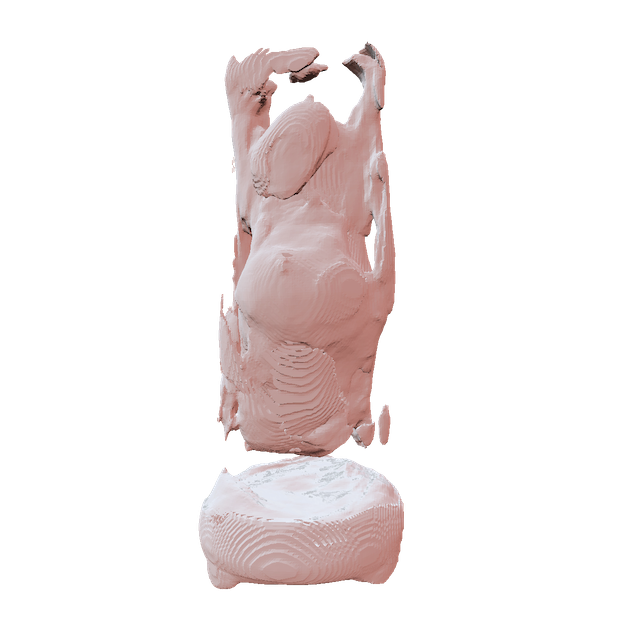}
	\end{subfigure}
	~
	\begin{subfigure}{27.5mm}
		\centering
		\includegraphics[width=27.5mm]{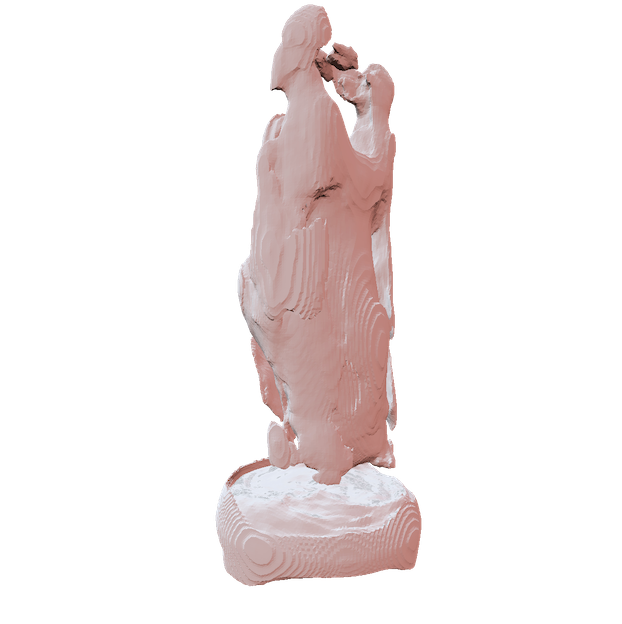}
	\end{subfigure} 
	~
	\begin{subfigure}{27.5mm}
		\centering
		\includegraphics[width=27.5mm]{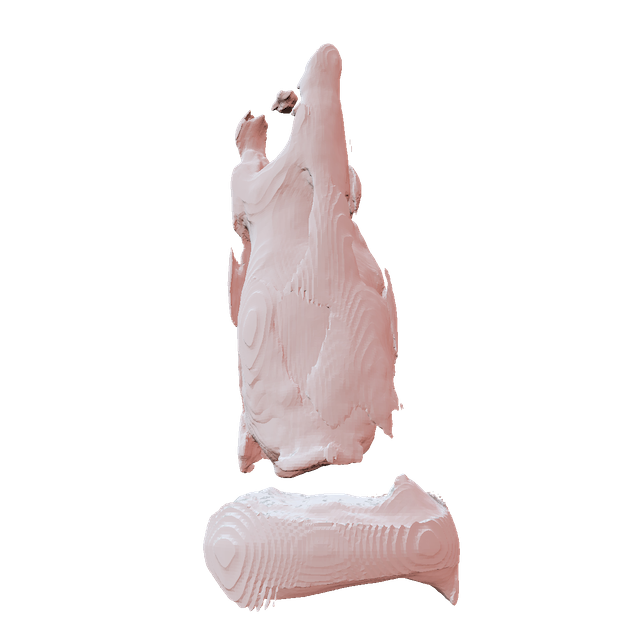}
	\end{subfigure} 
	~
	\begin{subfigure}{27.5mm}
		\centering
		\includegraphics[width=27.5mm]{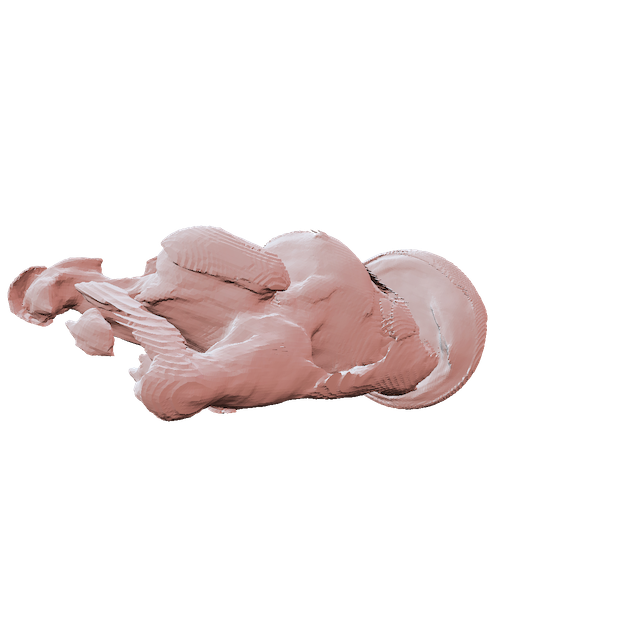}
	\end{subfigure} \\
	\begin{subfigure}{27.5mm}
		\centering
		\includegraphics[width=27.5mm]{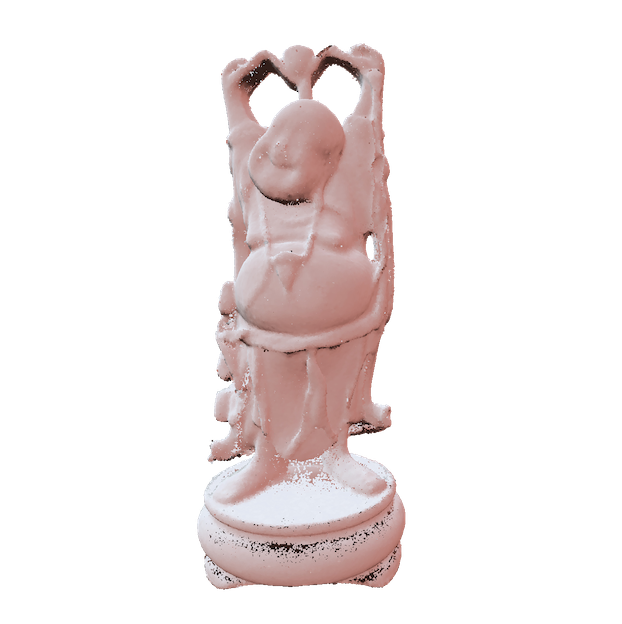}
	\end{subfigure}
	~
	\begin{subfigure}{27.5mm}
		\centering
		\includegraphics[width=27.5mm]{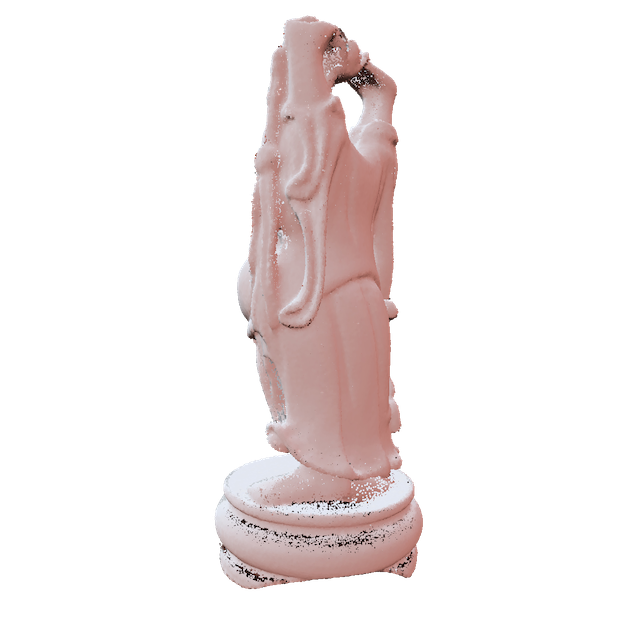}
	\end{subfigure} 
	~
	\begin{subfigure}{27.5mm}
		\centering
		\includegraphics[width=27.5mm]{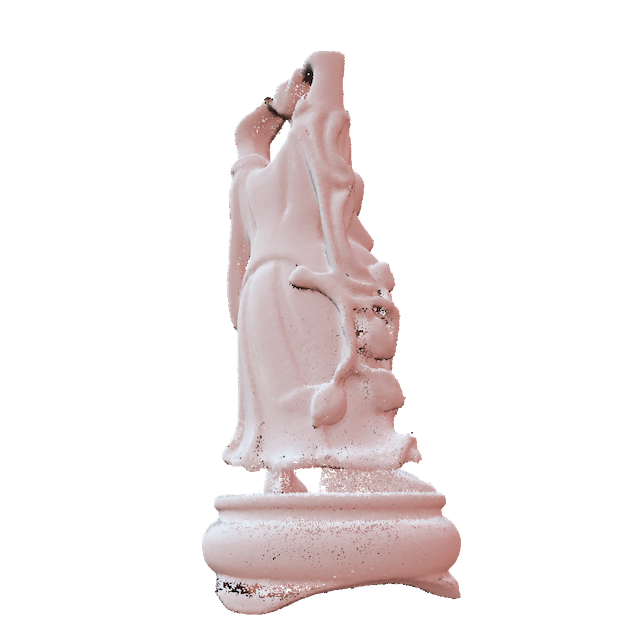}
	\end{subfigure} 
	~
	\begin{subfigure}{27.5mm}
		\centering
		\includegraphics[width=27.5mm]{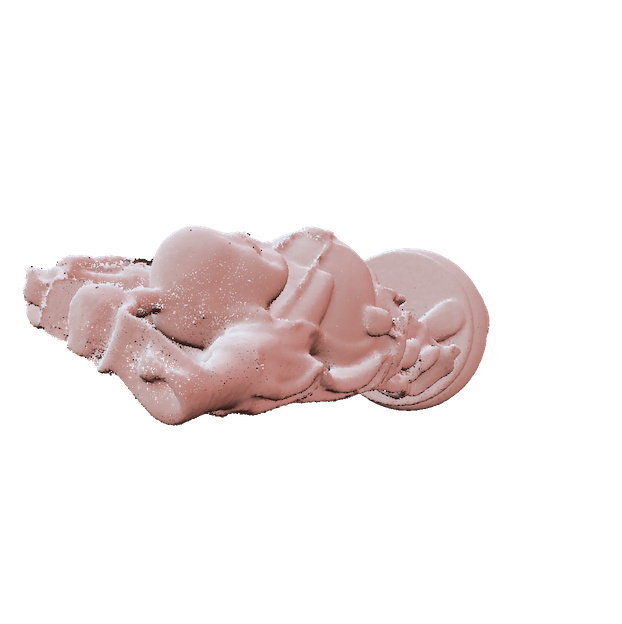}
	\end{subfigure} \\
	\caption{{\bf More Single Shape from Different Angles.} Results from Rows 1 to 4 correspond to {\bf Reference}, {\bf OF}, {\bf SDF}, {\bf PRIF - Mesh}.}
\end{figure}

\begin{figure}[!ht]
    \centering
	\begin{subfigure}{27.5mm}
		\centering
		\includegraphics[width=27.5mm]{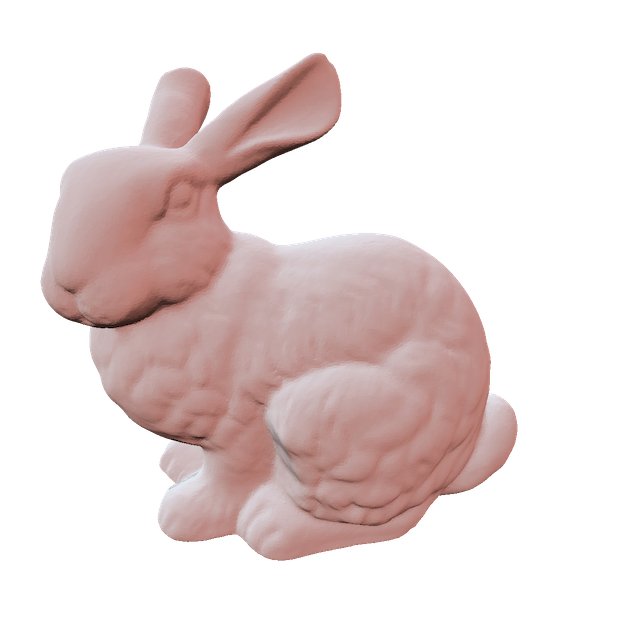}
	\end{subfigure}
	~
	\begin{subfigure}{27.5mm}
		\centering
		\includegraphics[width=27.5mm]{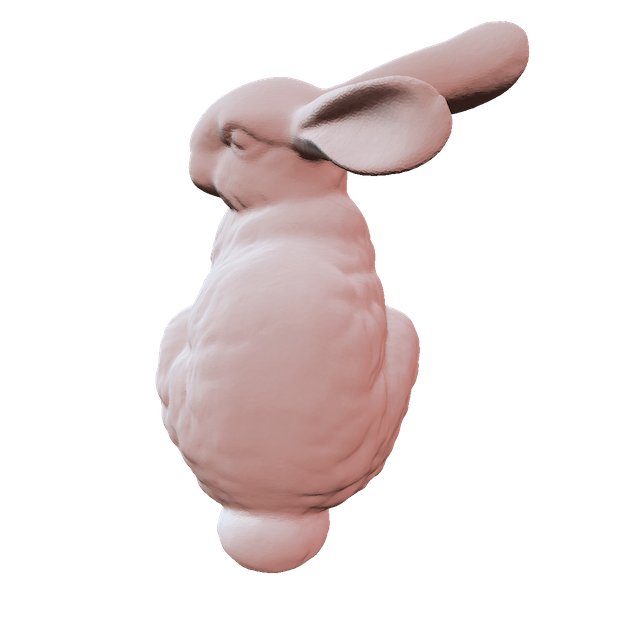}
	\end{subfigure} 
	~
	\begin{subfigure}{27.5mm}
		\centering
		\includegraphics[width=27.5mm]{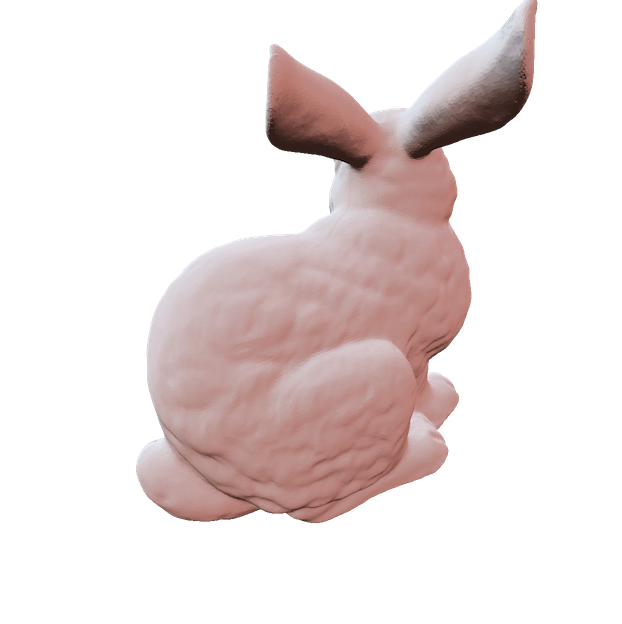}
	\end{subfigure}
	~
	\begin{subfigure}{27.5mm}
		\centering
		\includegraphics[width=27.5mm]{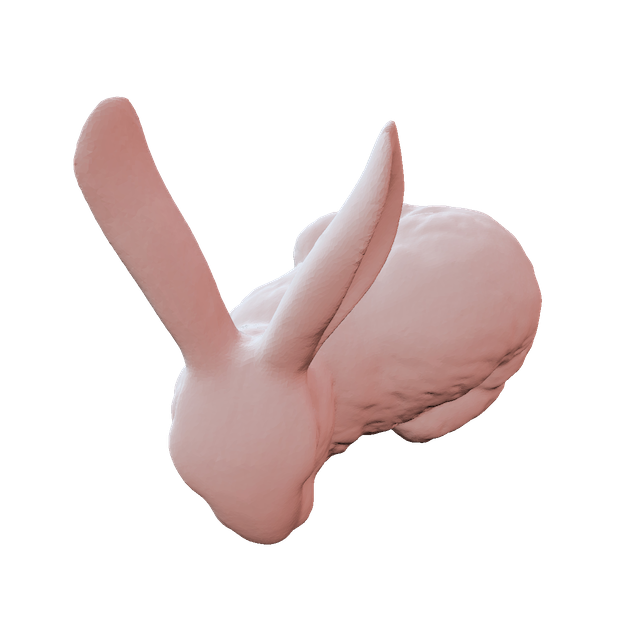}
	\end{subfigure}  \\
	\begin{subfigure}{27.5mm}
		\centering
		\includegraphics[width=27.5mm]{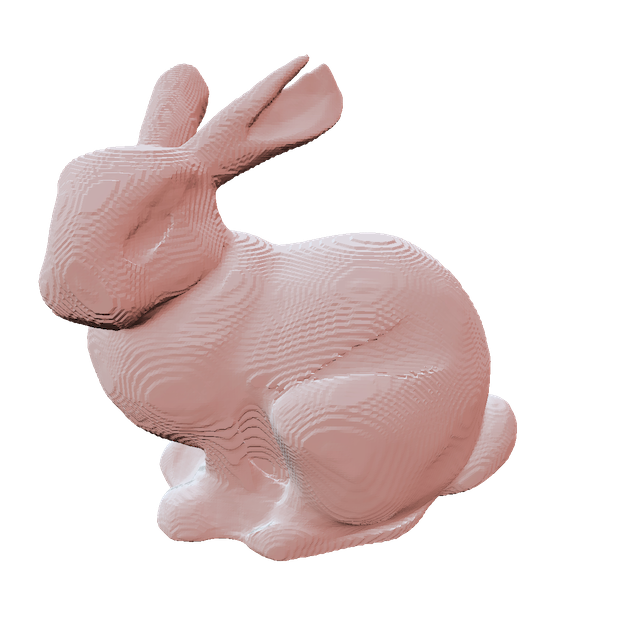}
	\end{subfigure}
	~
	\begin{subfigure}{27.5mm}
		\centering
		\includegraphics[width=27.5mm]{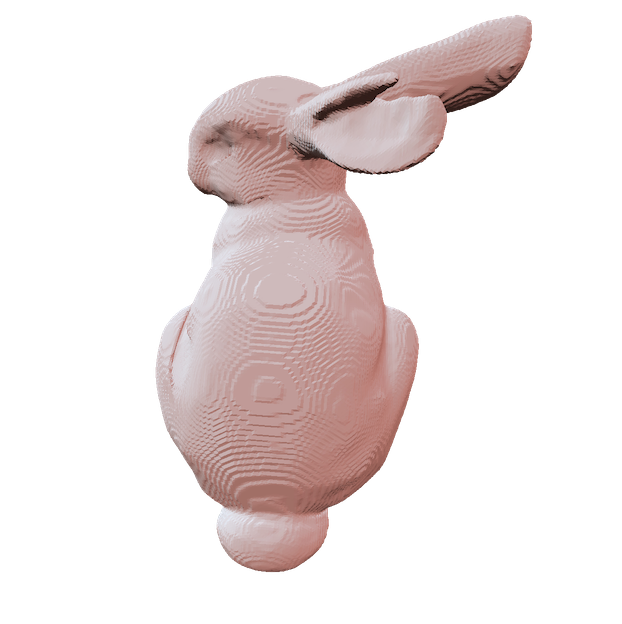}
	\end{subfigure} 
	~
	\begin{subfigure}{27.5mm}
		\centering
		\includegraphics[width=27.5mm]{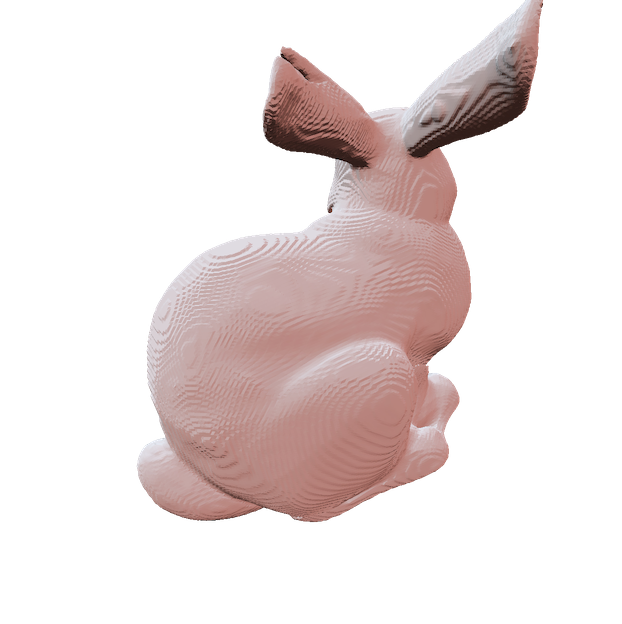}
	\end{subfigure} 
	~
	\begin{subfigure}{27.5mm}
		\centering
		\includegraphics[width=27.5mm]{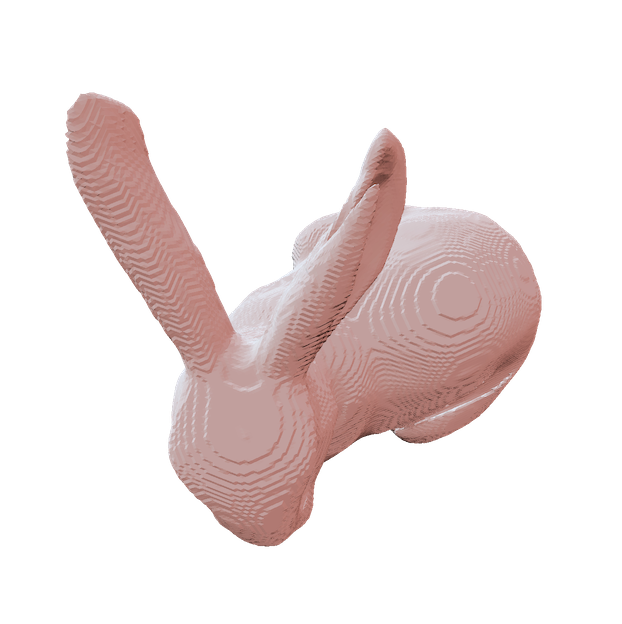}
	\end{subfigure}  \\
	\begin{subfigure}{27.5mm}
		\centering
		\includegraphics[width=27.5mm]{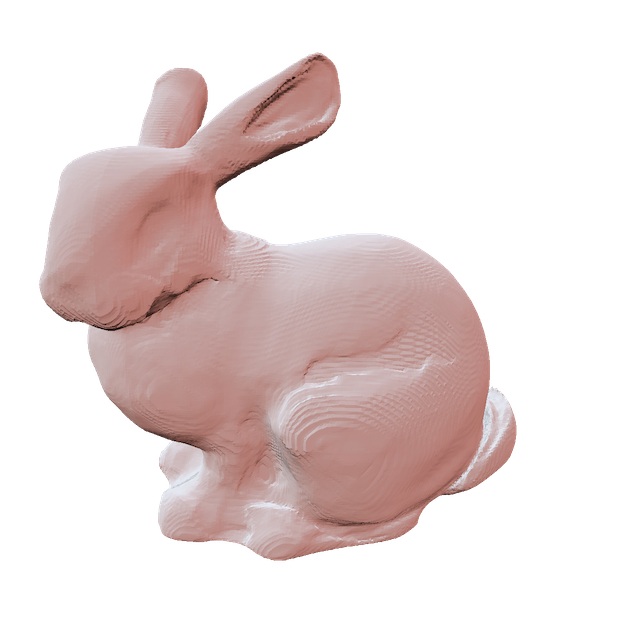}
	\end{subfigure}
	~
	\begin{subfigure}{27.5mm}
		\centering
		\includegraphics[width=27.5mm]{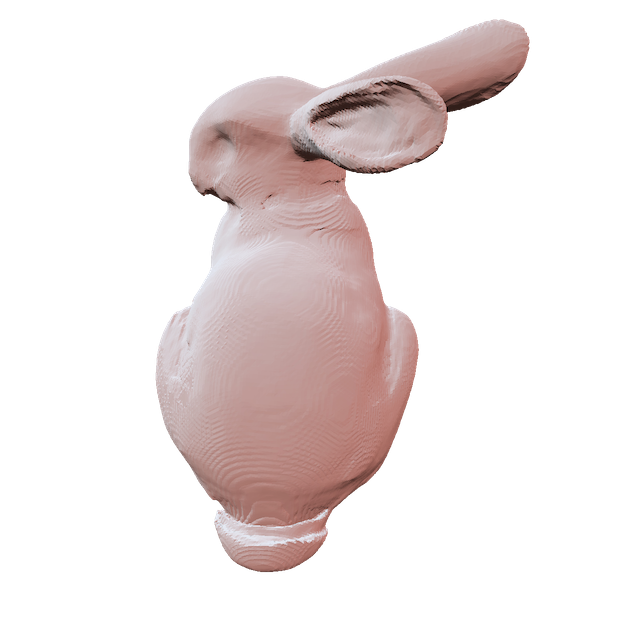}
	\end{subfigure} 
	~
	\begin{subfigure}{27.5mm}
		\centering
		\includegraphics[width=27.5mm]{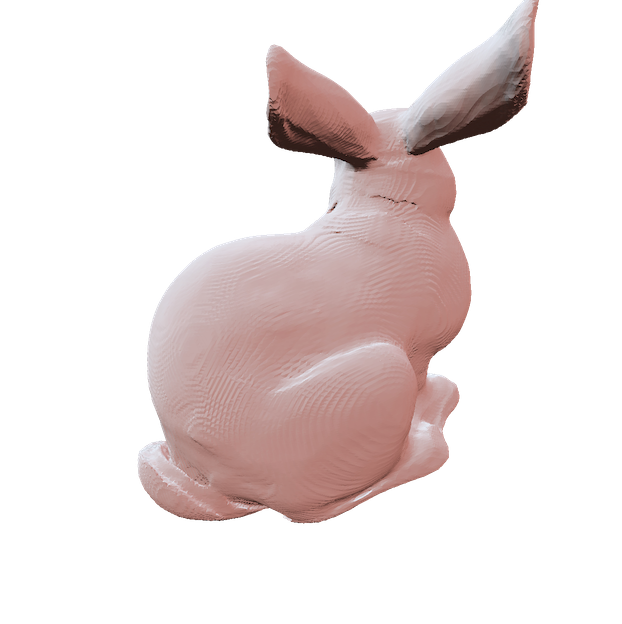}
	\end{subfigure} 
	~
	\begin{subfigure}{27.5mm}
		\centering
		\includegraphics[width=27.5mm]{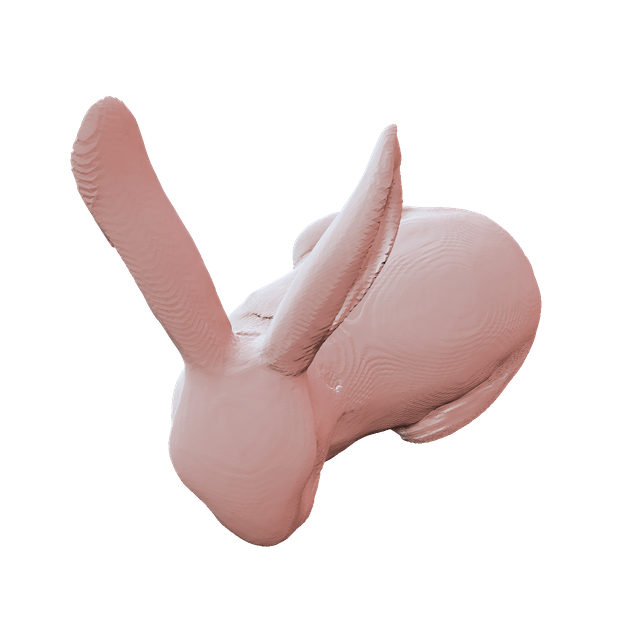}
	\end{subfigure} \\
	\begin{subfigure}{27.5mm}
		\centering
		\includegraphics[width=27.5mm]{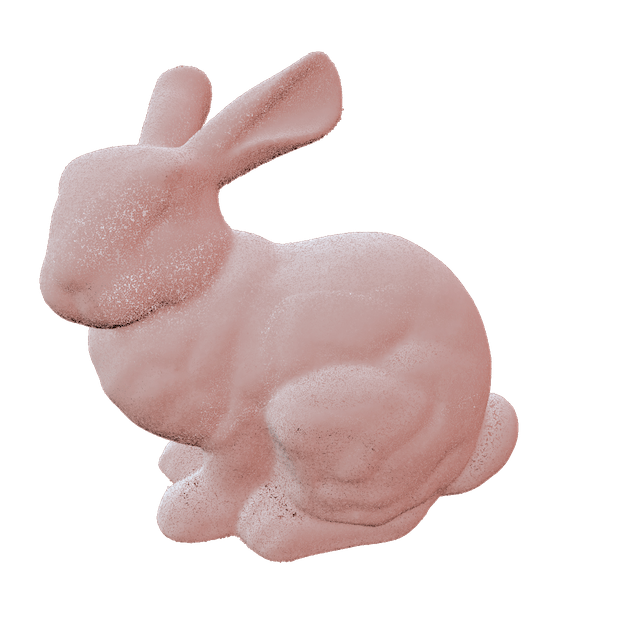}
	\end{subfigure}
	~
	\begin{subfigure}{27.5mm}
		\centering
		\includegraphics[width=27.5mm]{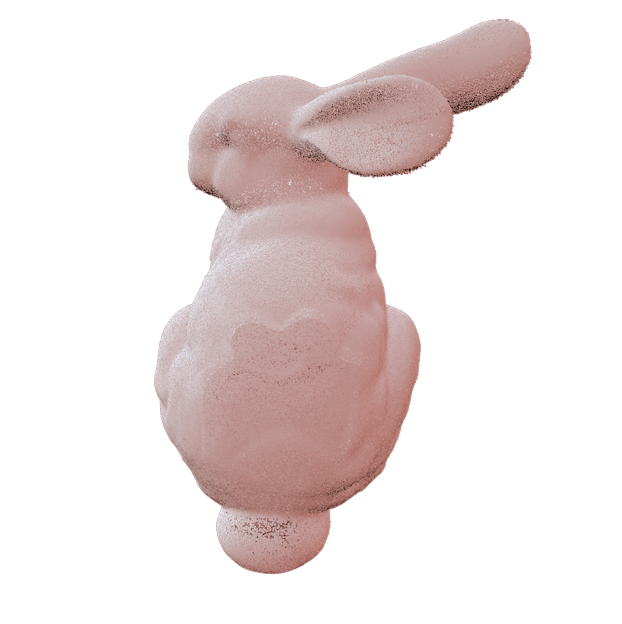}
	\end{subfigure} 
	~
	\begin{subfigure}{27.5mm}
		\centering
		\includegraphics[width=27.5mm]{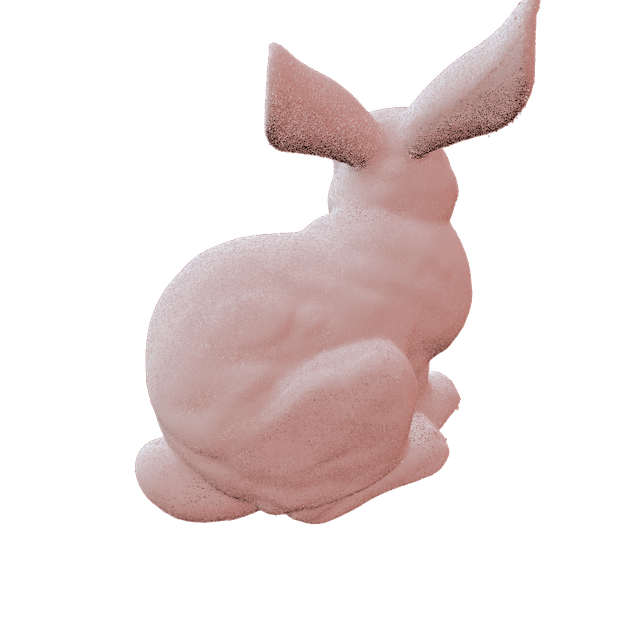}
	\end{subfigure} 
	~
	\begin{subfigure}{27.5mm}
		\centering
		\includegraphics[width=27.5mm]{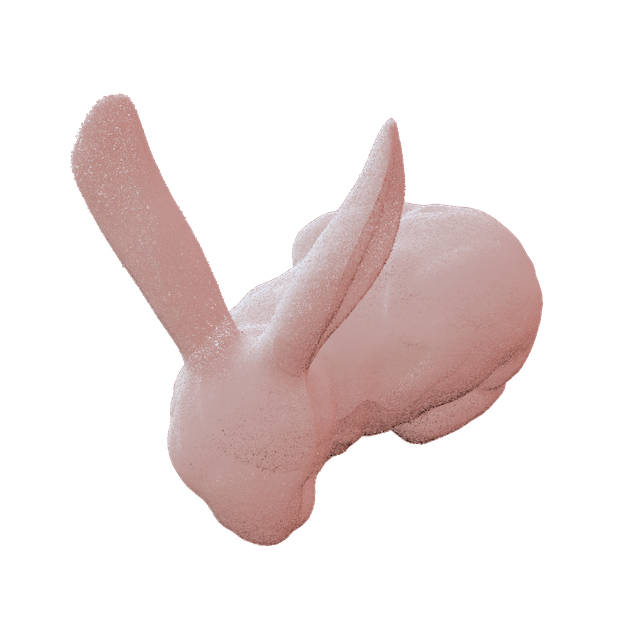}
	\end{subfigure} \\
	\caption{{\bf More Single Shape from Different Angles.} Results from Rows 1 to 4 correspond to {\bf Reference}, {\bf OF}, {\bf SDF}, {\bf PRIF - Mesh}.}
\end{figure}

\begin{figure}[!ht]
    \centering
	\begin{subfigure}{27.5mm}
		\centering
		\includegraphics[width=27.5mm]{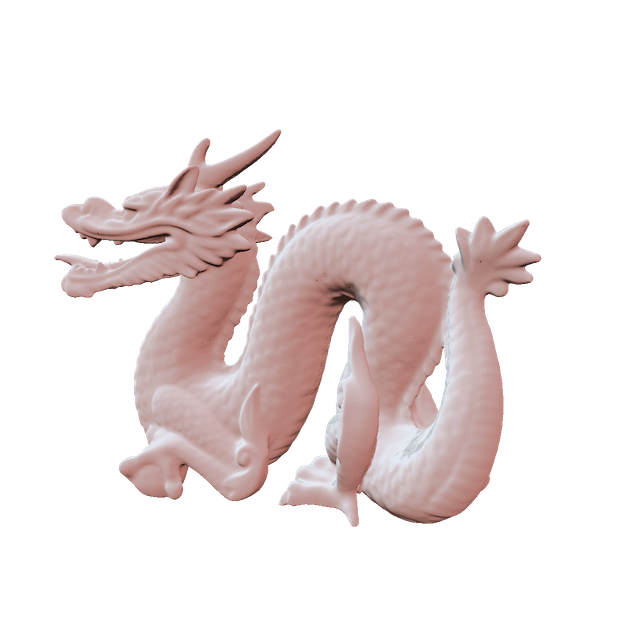}
	\end{subfigure}
	~
	\begin{subfigure}{27.5mm}
		\centering
		\includegraphics[width=27.5mm]{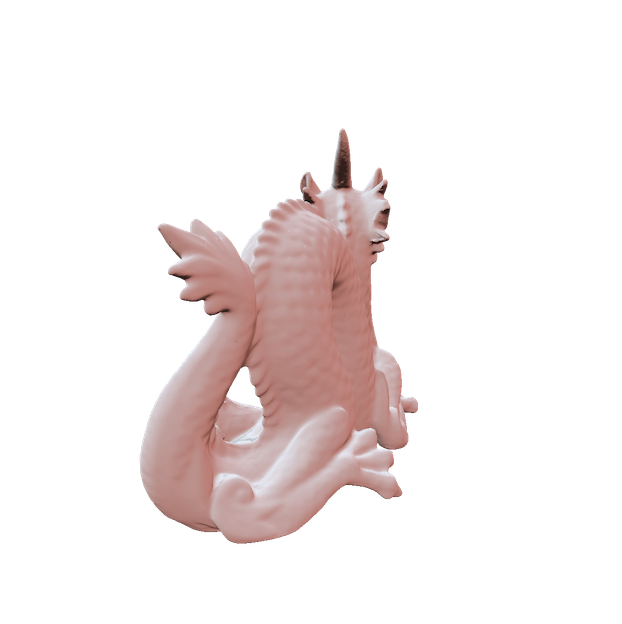}
	\end{subfigure} 
	~
	\begin{subfigure}{27.5mm}
		\centering
		\includegraphics[width=27.5mm]{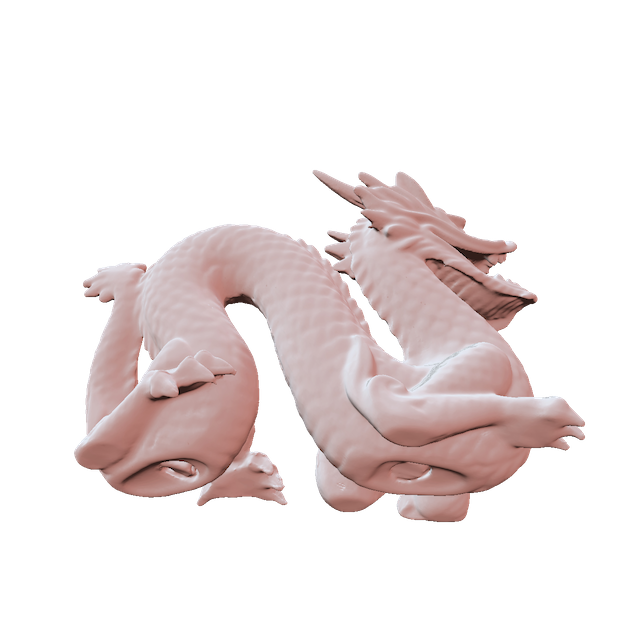}
	\end{subfigure}
	~
	\begin{subfigure}{27.5mm}
		\centering
		\includegraphics[width=27.5mm]{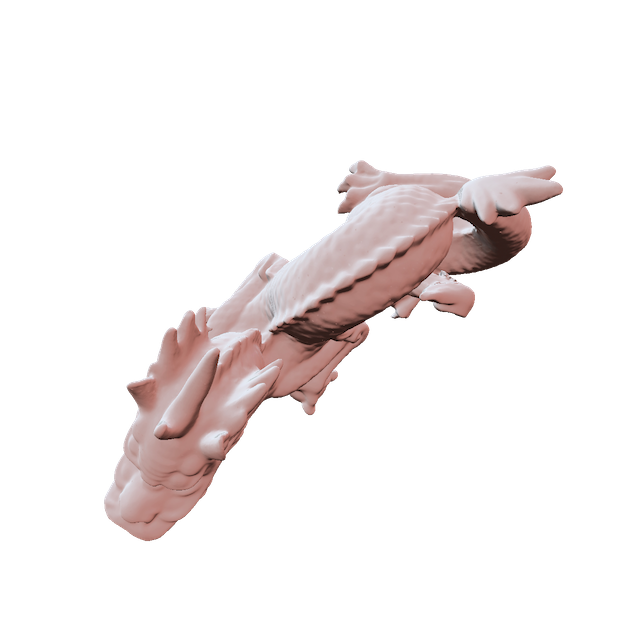}
	\end{subfigure}  \\
	\begin{subfigure}{27.5mm}
		\centering
		\includegraphics[width=27.5mm]{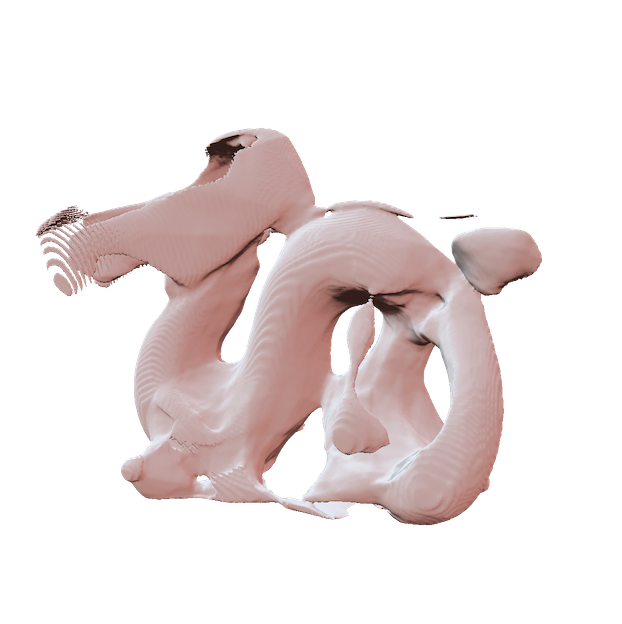}
	\end{subfigure}
	~
	\begin{subfigure}{27.5mm}
		\centering
		\includegraphics[width=27.5mm]{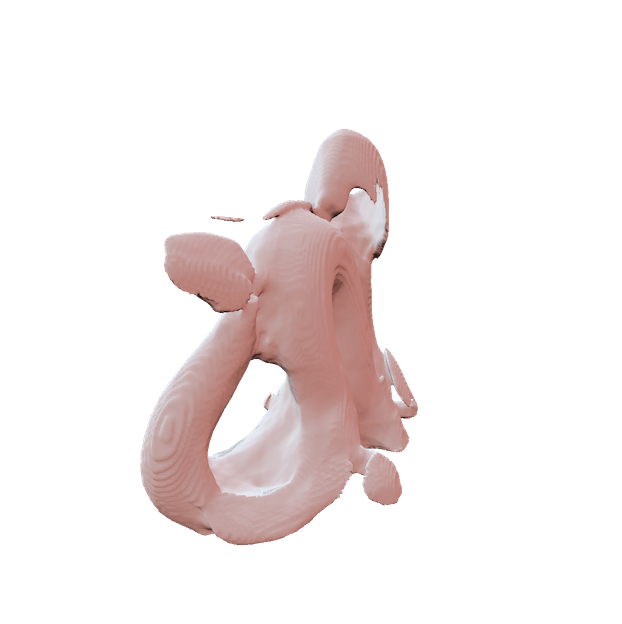}
	\end{subfigure} 
	~
	\begin{subfigure}{27.5mm}
		\centering
		\includegraphics[width=27.5mm]{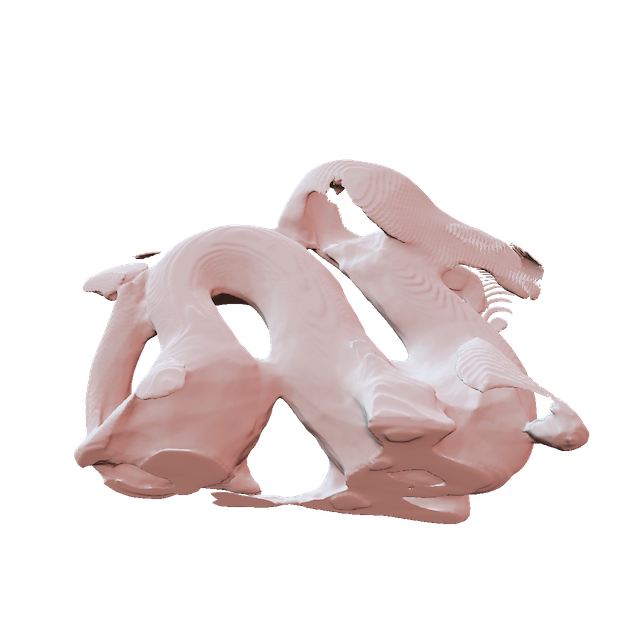}
	\end{subfigure} 
	~
	\begin{subfigure}{27.5mm}
		\centering
		\includegraphics[width=27.5mm]{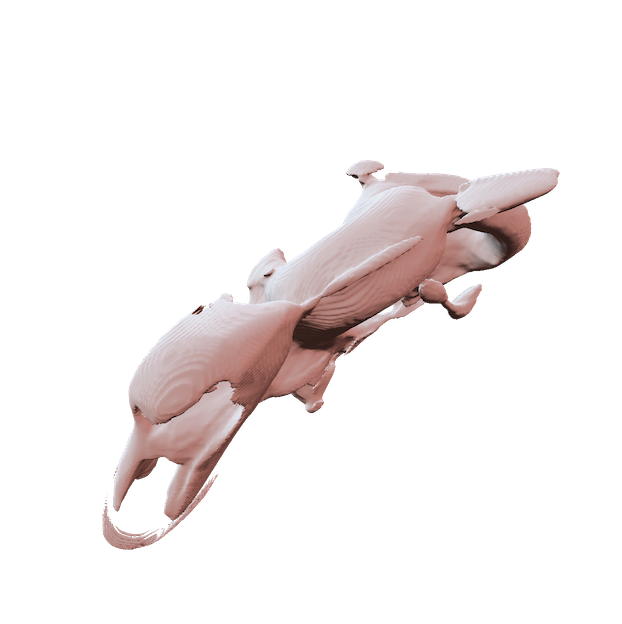}
	\end{subfigure}  \\
	\begin{subfigure}{27.5mm}
		\centering
		\includegraphics[width=27.5mm]{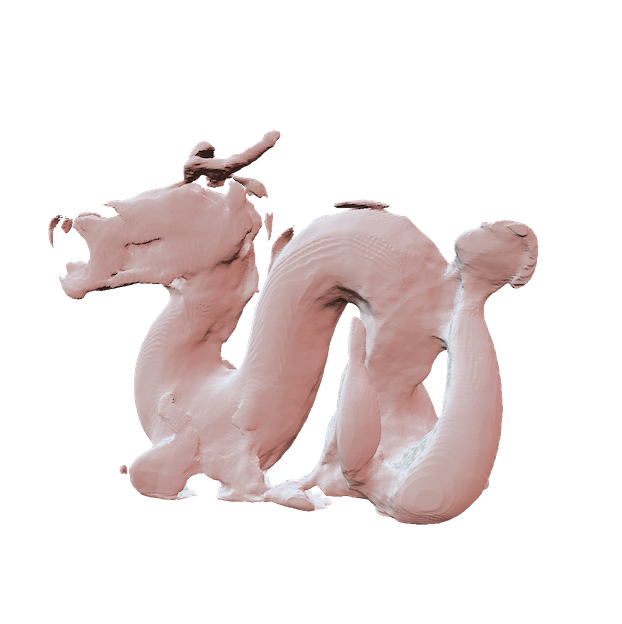}
	\end{subfigure}
	~
	\begin{subfigure}{27.5mm}
		\centering
		\includegraphics[width=27.5mm]{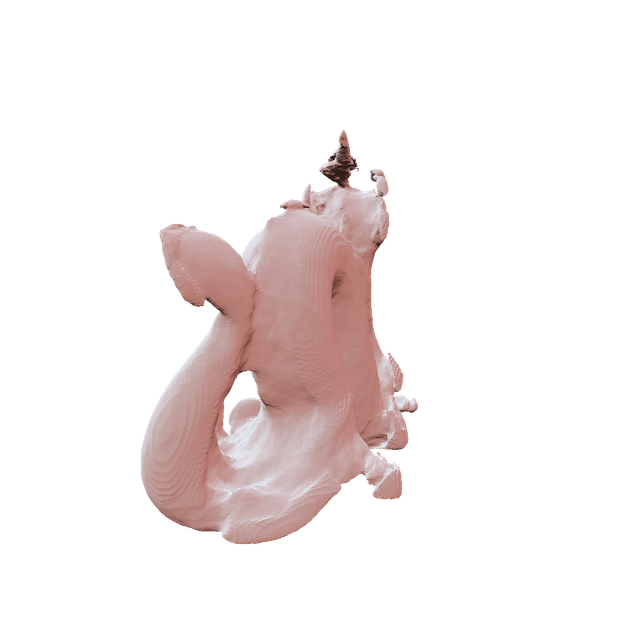}
	\end{subfigure} 
	~
	\begin{subfigure}{27.5mm}
		\centering
		\includegraphics[width=27.5mm]{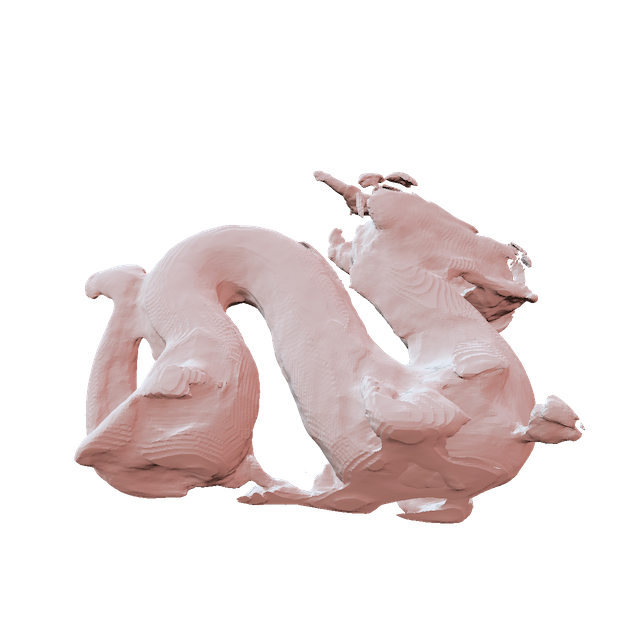}
	\end{subfigure} 
	~
	\begin{subfigure}{27.5mm}
		\centering
		\includegraphics[width=27.5mm]{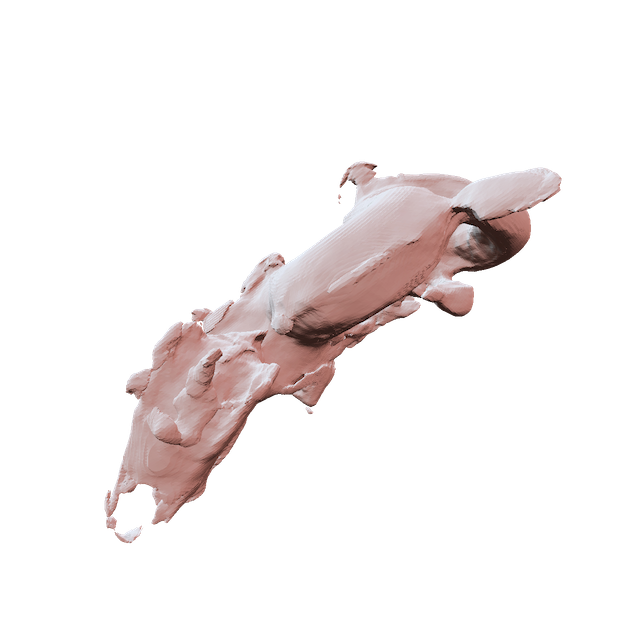}
	\end{subfigure} \\
	\begin{subfigure}{27.5mm}
		\centering
		\includegraphics[width=27.5mm]{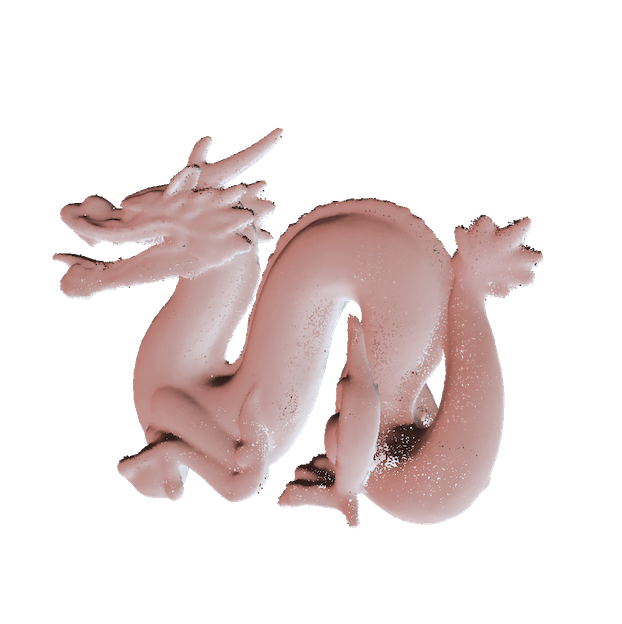}
	\end{subfigure}
	~
	\begin{subfigure}{27.5mm}
		\centering
		\includegraphics[width=27.5mm]{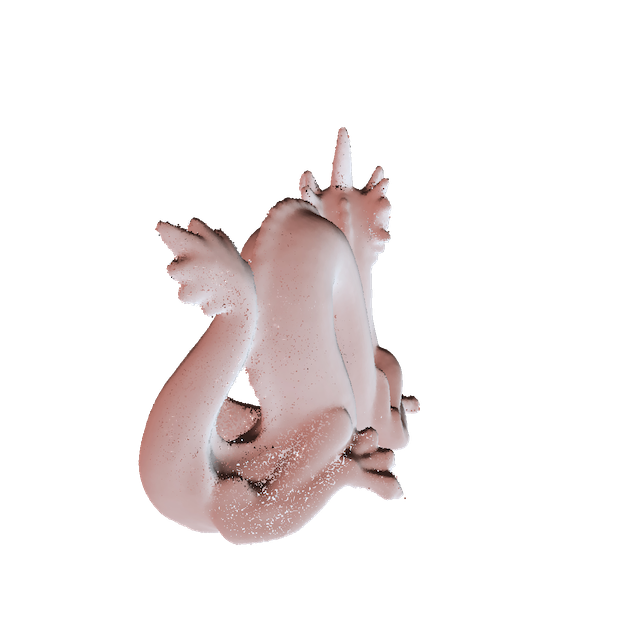}
	\end{subfigure} 
	~
	\begin{subfigure}{27.5mm}
		\centering
		\includegraphics[width=27.5mm]{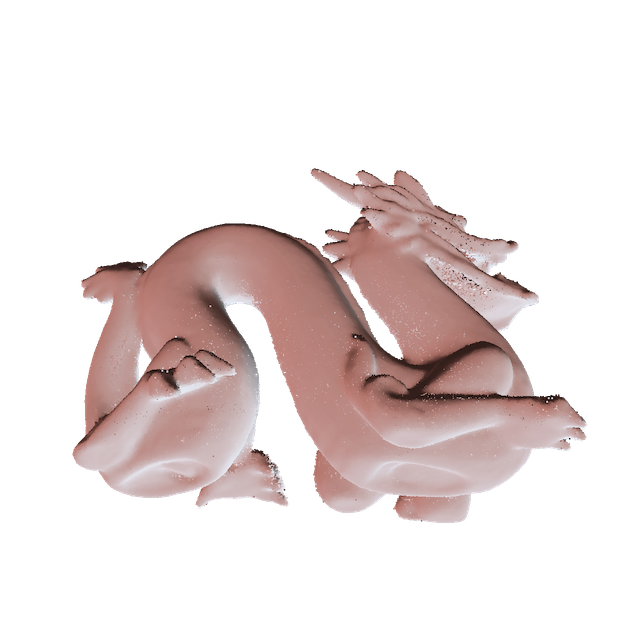}
	\end{subfigure} 
	~
	\begin{subfigure}{27.5mm}
		\centering
		\includegraphics[width=27.5mm]{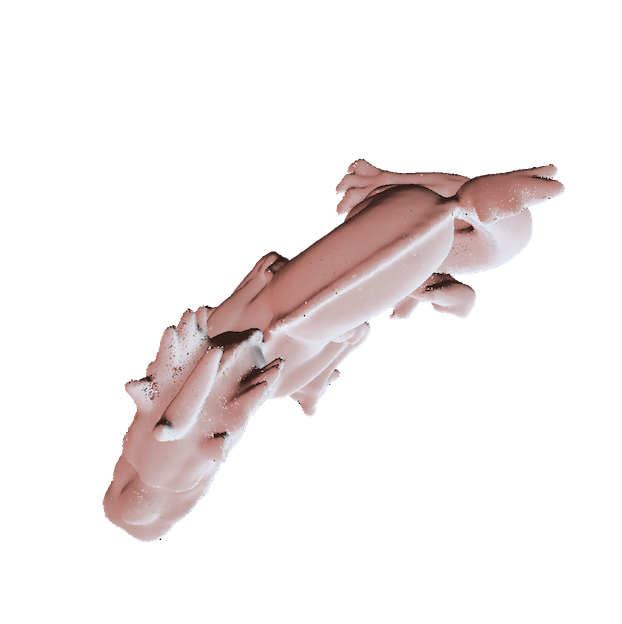}
	\end{subfigure} \\
	\caption{{\bf More Single Shape from Different Angles.} Results from Rows 1 to 4 correspond to {\bf Reference}, {\bf OF}, {\bf SDF}, {\bf PRIF - Mesh}.}
\end{figure}

\begin{figure}[!ht]
    \centering
	\begin{subfigure}{27.5mm}
		\centering
		\includegraphics[width=27.5mm]{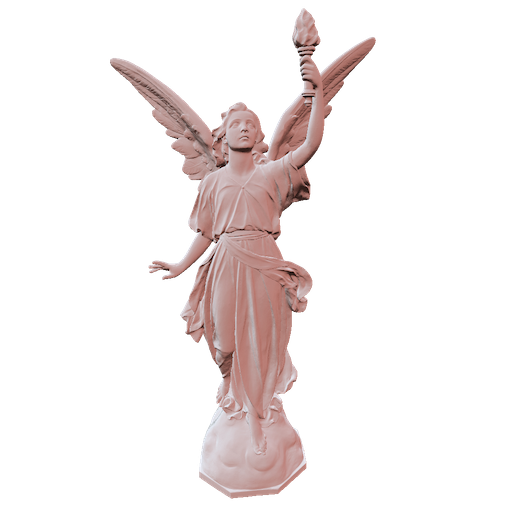}
	\end{subfigure}
	~
	\begin{subfigure}{27.5mm}
		\centering
		\includegraphics[width=27.5mm]{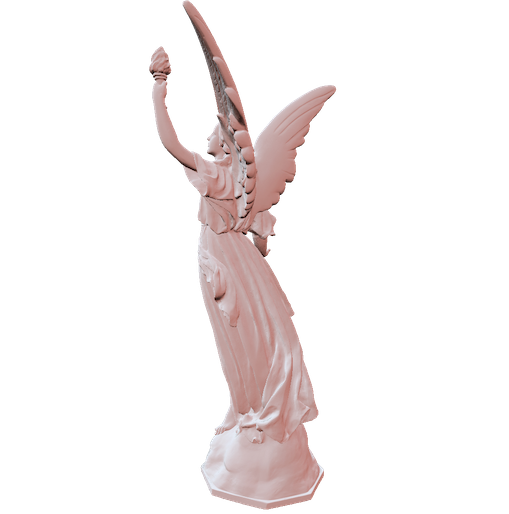}
	\end{subfigure} 
	~
	\begin{subfigure}{27.5mm}
		\centering
		\includegraphics[width=27.5mm]{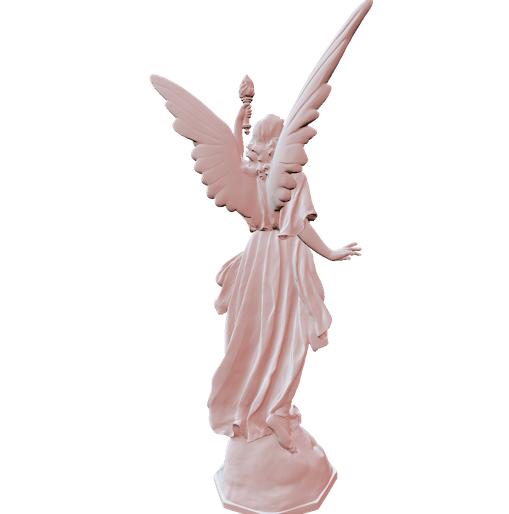}
	\end{subfigure}
	~
	\begin{subfigure}{27.5mm}
		\centering
		\includegraphics[width=27.5mm]{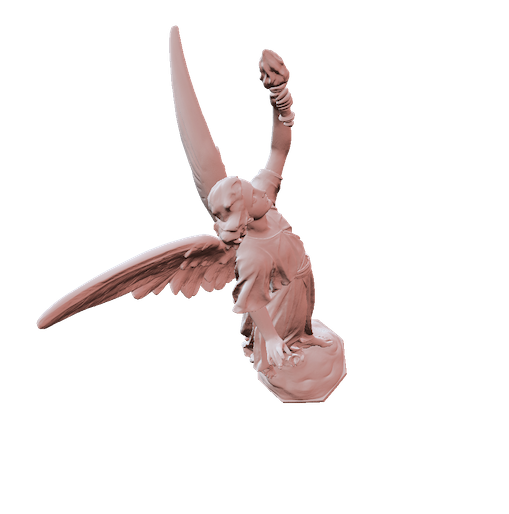}
	\end{subfigure}  \\
	\begin{subfigure}{27.5mm}
		\centering
		\includegraphics[width=27.5mm]{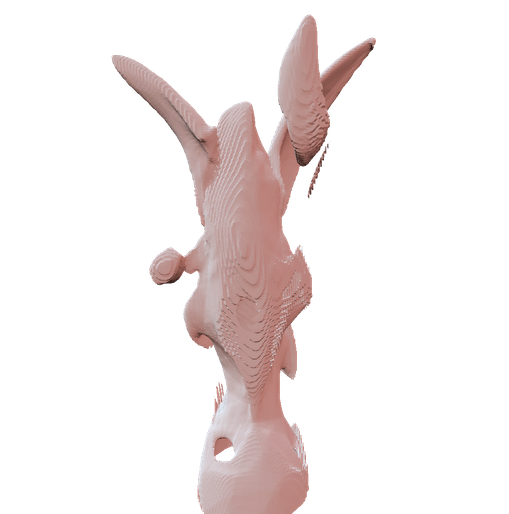}
	\end{subfigure}
	~
	\begin{subfigure}{27.5mm}
		\centering
		\includegraphics[width=27.5mm]{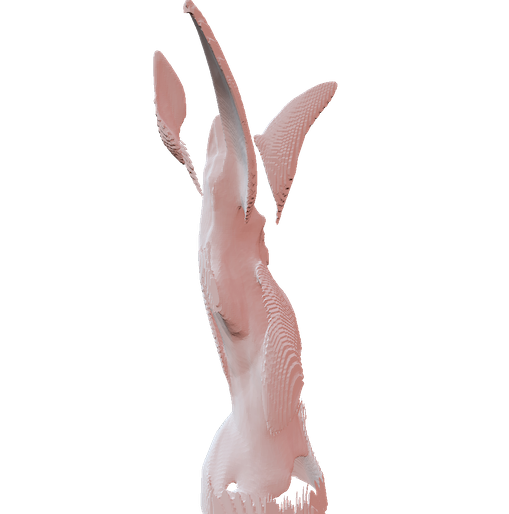}
	\end{subfigure} 
	~
	\begin{subfigure}{27.5mm}
		\centering
		\includegraphics[width=27.5mm]{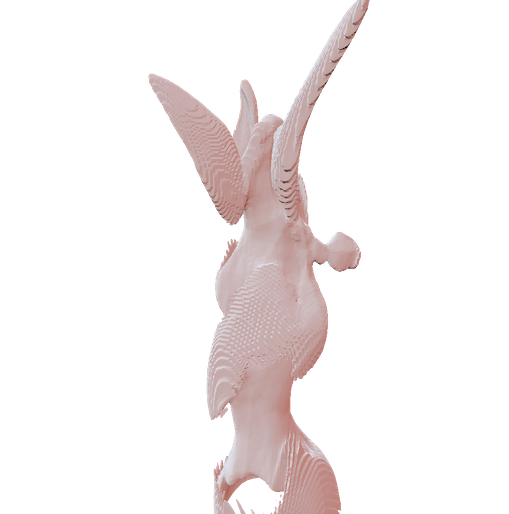}
	\end{subfigure} 
	~
	\begin{subfigure}{27.5mm}
		\centering
		\includegraphics[width=27.5mm]{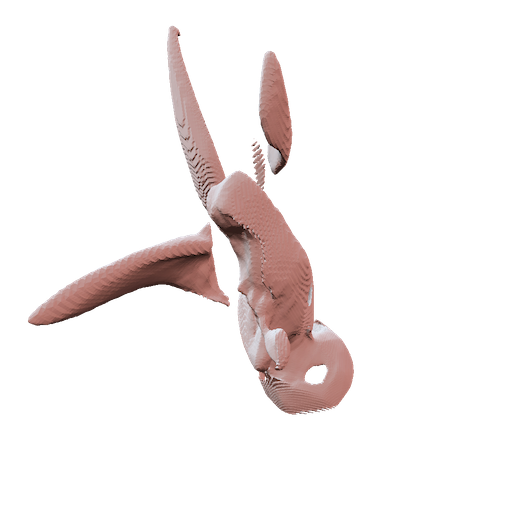}
	\end{subfigure}  \\
	\begin{subfigure}{27.5mm}
		\centering
		\includegraphics[width=27.5mm]{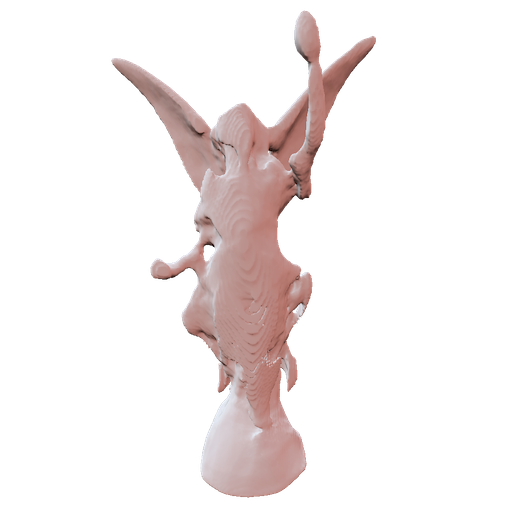}
	\end{subfigure}
	~
	\begin{subfigure}{27.5mm}
		\centering
		\includegraphics[width=27.5mm]{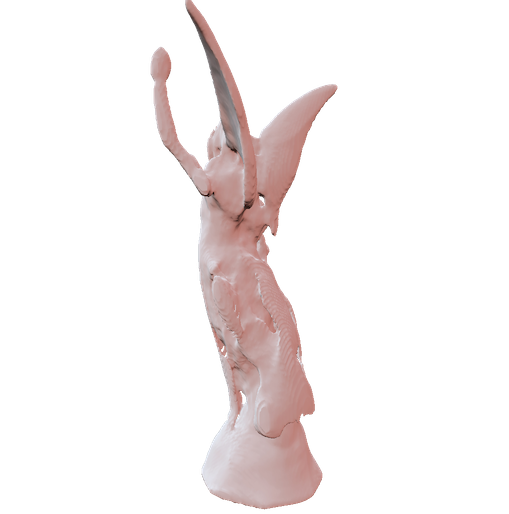}
	\end{subfigure} 
	~
	\begin{subfigure}{27.5mm}
		\centering
		\includegraphics[width=27.5mm]{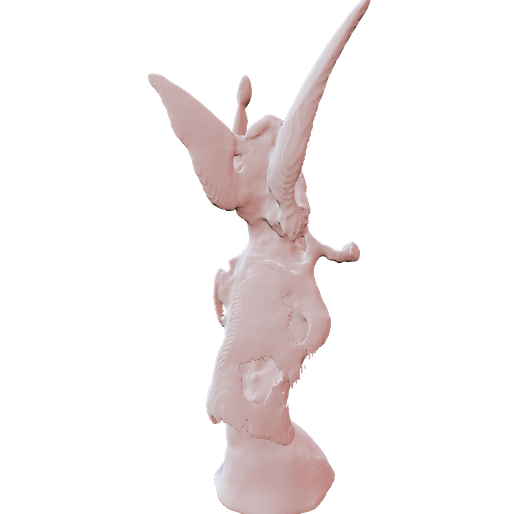}
	\end{subfigure} 
	~
	\begin{subfigure}{27.5mm}
		\centering
		\includegraphics[width=27.5mm]{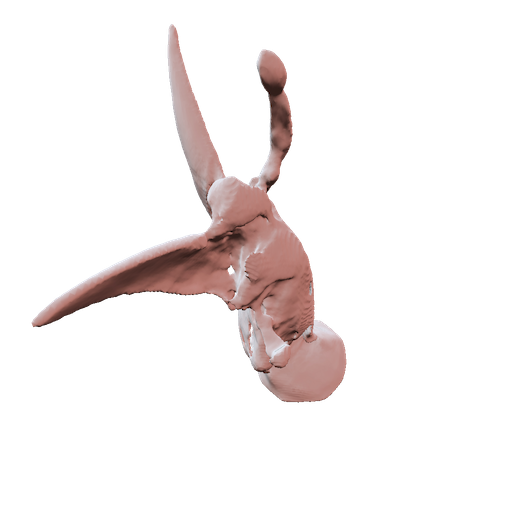}
	\end{subfigure} \\
	\begin{subfigure}{27.5mm}
		\centering
		\includegraphics[width=27.5mm]{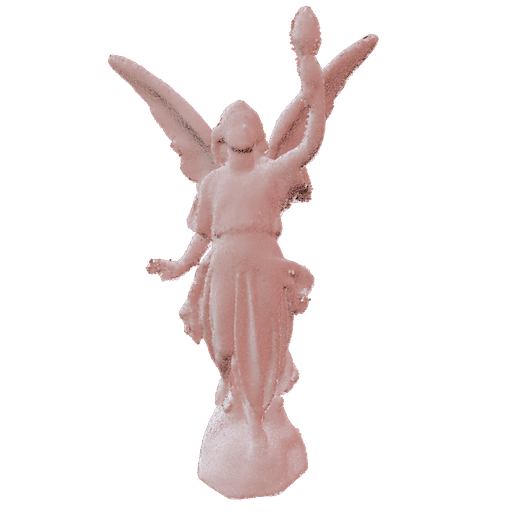}
	\end{subfigure}
	~
	\begin{subfigure}{27.5mm}
		\centering
		\includegraphics[width=27.5mm]{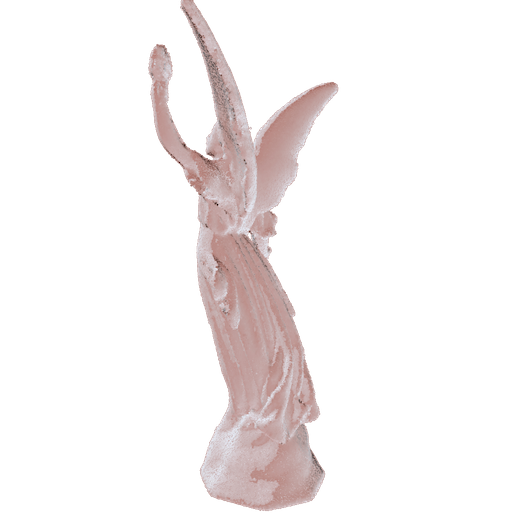}
	\end{subfigure} 
	~
	\begin{subfigure}{27.5mm}
		\centering
		\includegraphics[width=27.5mm]{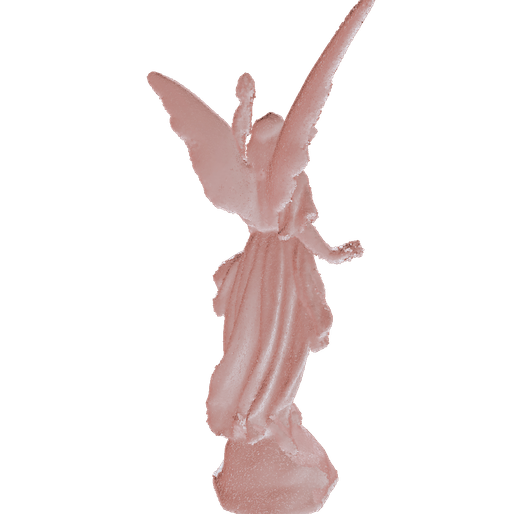}
	\end{subfigure} 
	~
	\begin{subfigure}{27.5mm}
		\centering
		\includegraphics[width=27.5mm]{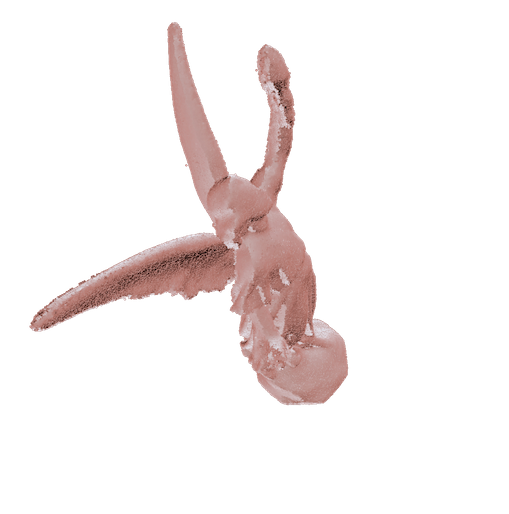}
	\end{subfigure} \\
	\caption{{\bf More Single Shape from Different Angles.} Results from Rows 1 to 4 correspond to {\bf Reference}, {\bf OF}, {\bf SDF}, {\bf PRIF - Mesh}.}
\end{figure}

\begin{figure}[!ht]
    \centering
	\begin{subfigure}{27.5mm}
		\centering
		\includegraphics[width=27.5mm]{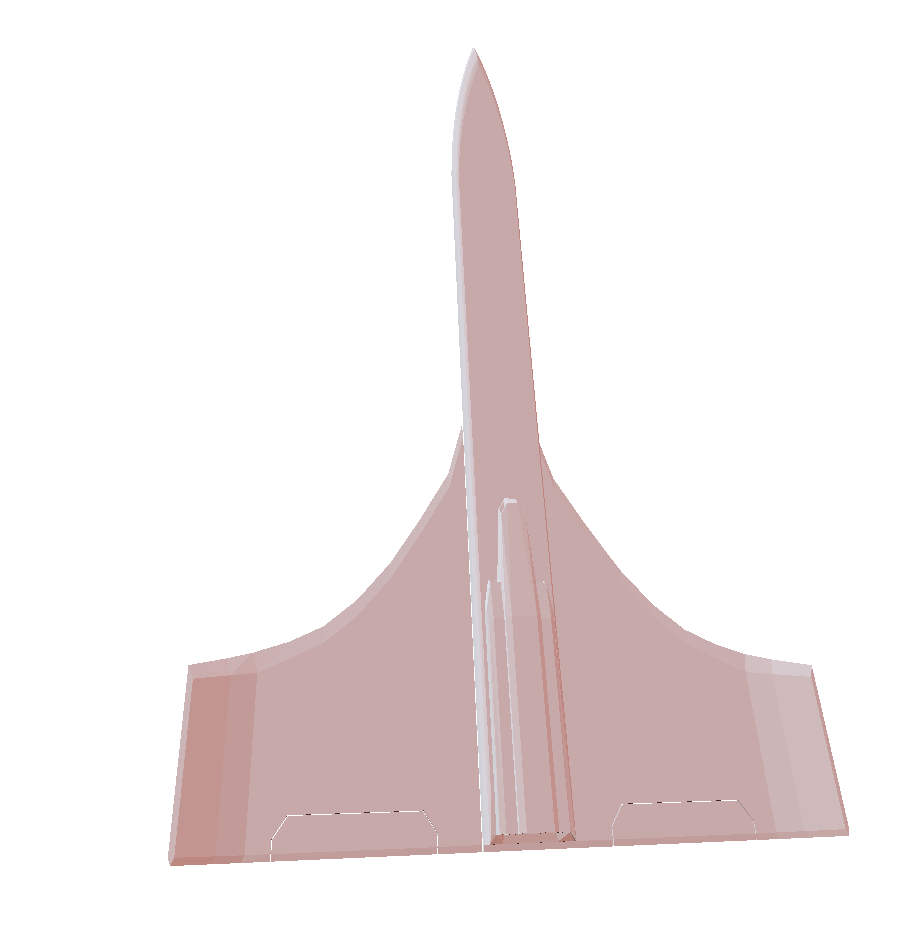}
	\end{subfigure}
	~
	\begin{subfigure}{27.5mm}
		\centering
		\includegraphics[width=27.5mm]{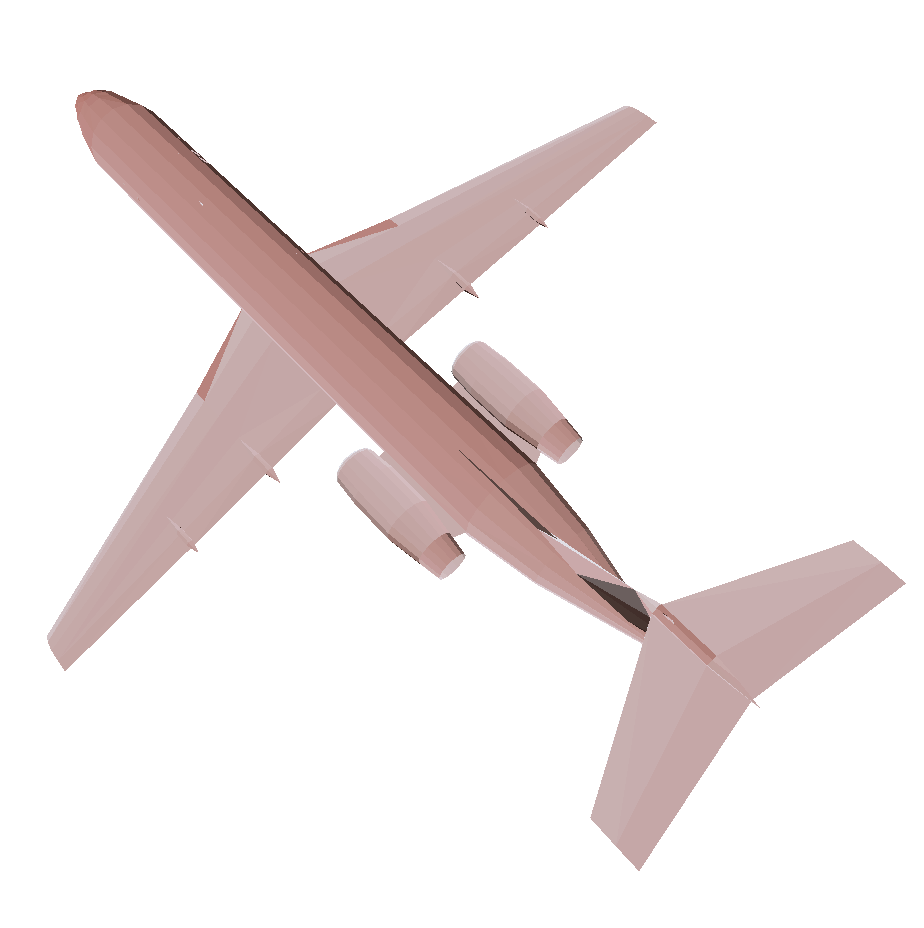}
	\end{subfigure} 
	~
	\begin{subfigure}{27.5mm}
		\centering
		\includegraphics[width=27.5mm]{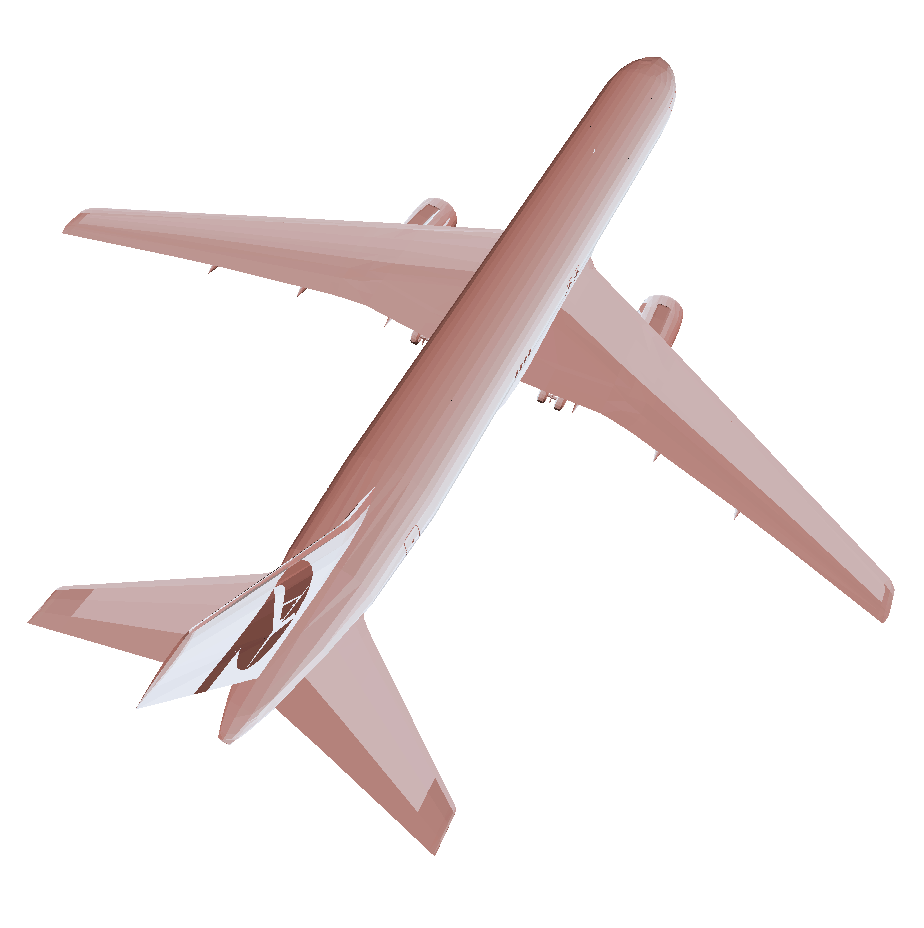}
	\end{subfigure}
	~
	\begin{subfigure}{27.5mm}
		\centering
		\includegraphics[width=27.5mm]{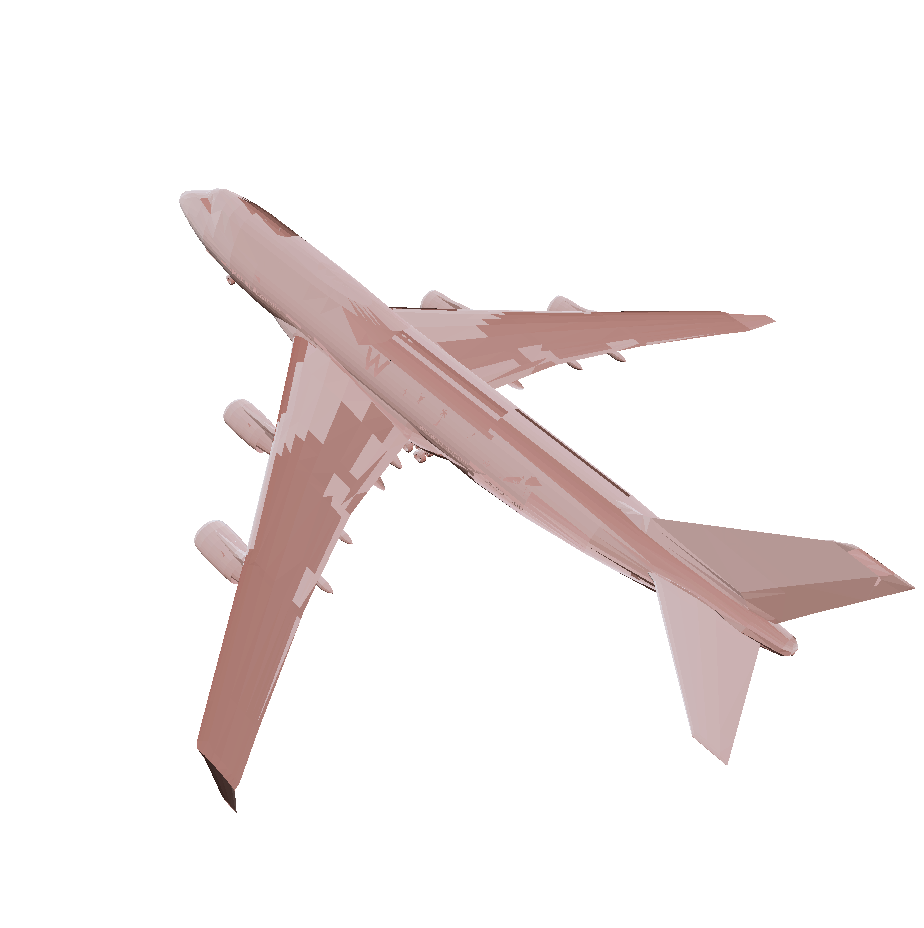}
	\end{subfigure}  \\
	\begin{subfigure}{27.5mm}
		\centering
		\includegraphics[width=27.5mm]{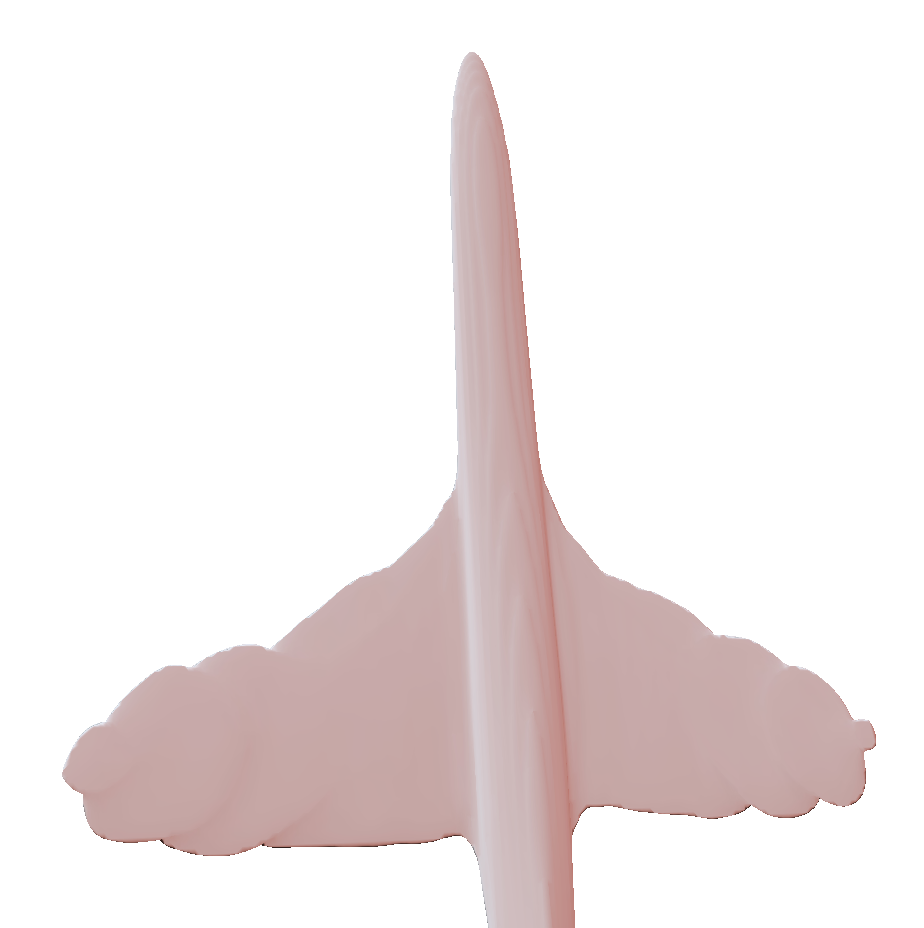}
	\end{subfigure}
	~
	\begin{subfigure}{27.5mm}
		\centering
		\includegraphics[width=27.5mm]{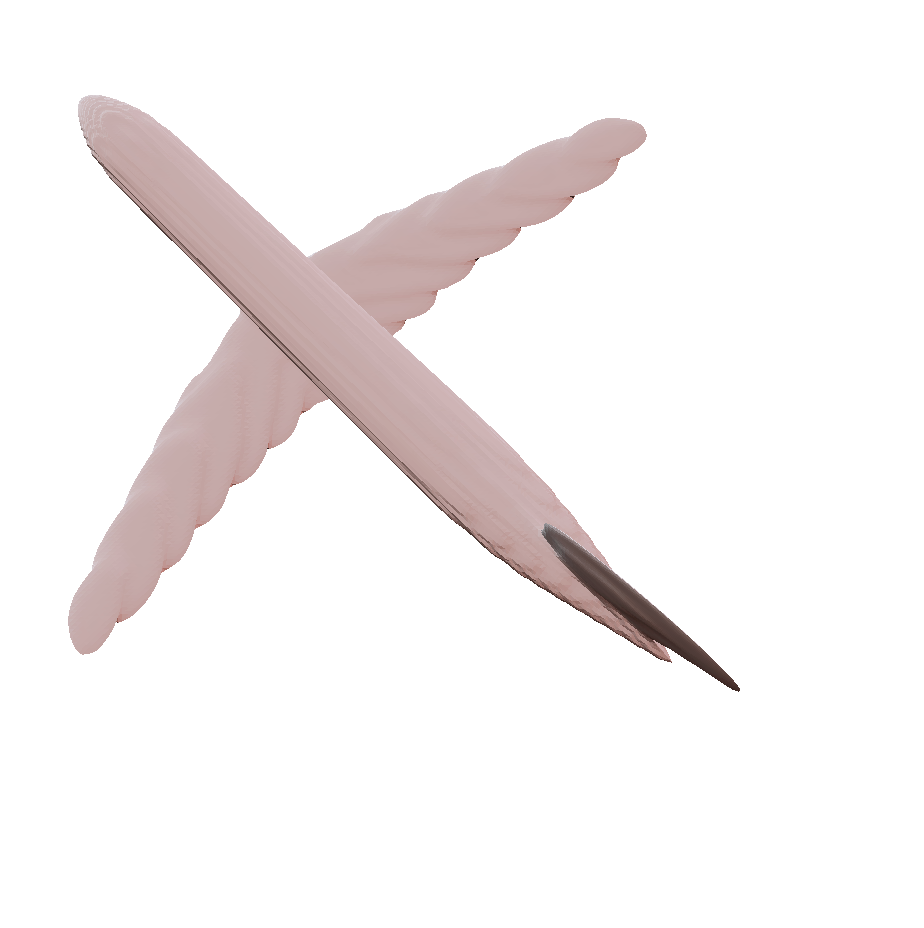}
	\end{subfigure} 
	~
	\begin{subfigure}{27.5mm}
		\centering
		\includegraphics[width=27.5mm]{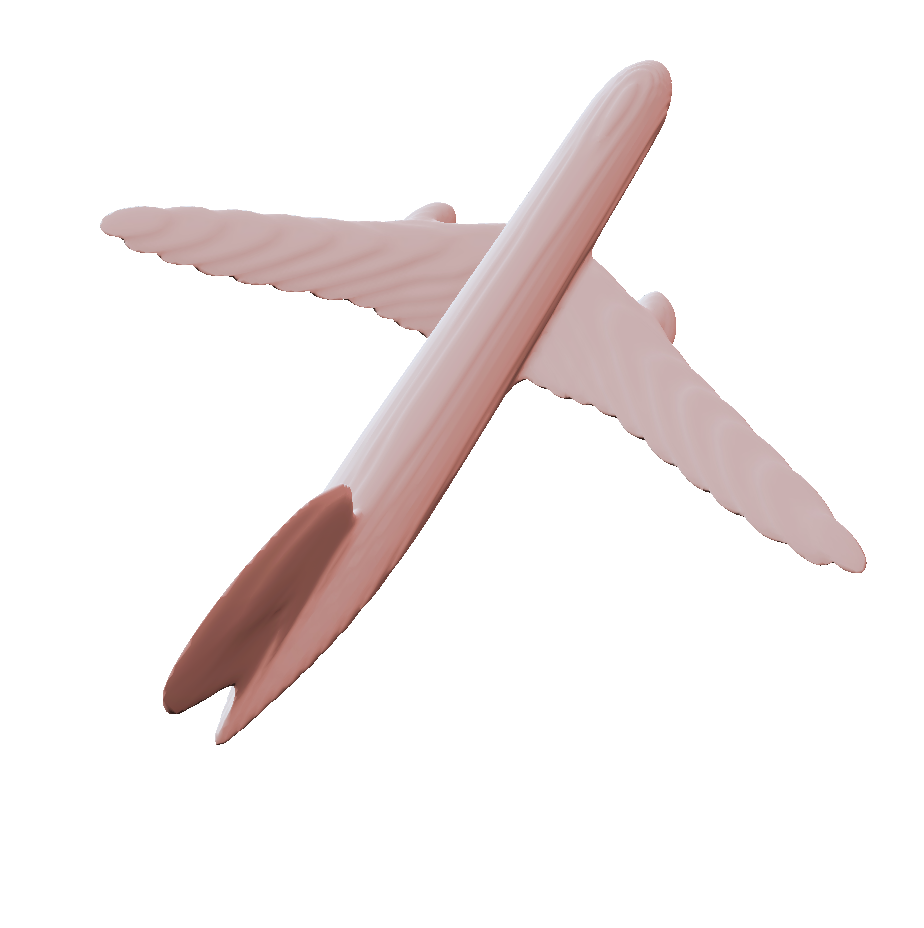}
	\end{subfigure} 
	~
	\begin{subfigure}{27.5mm}
		\centering
		\includegraphics[width=27.5mm]{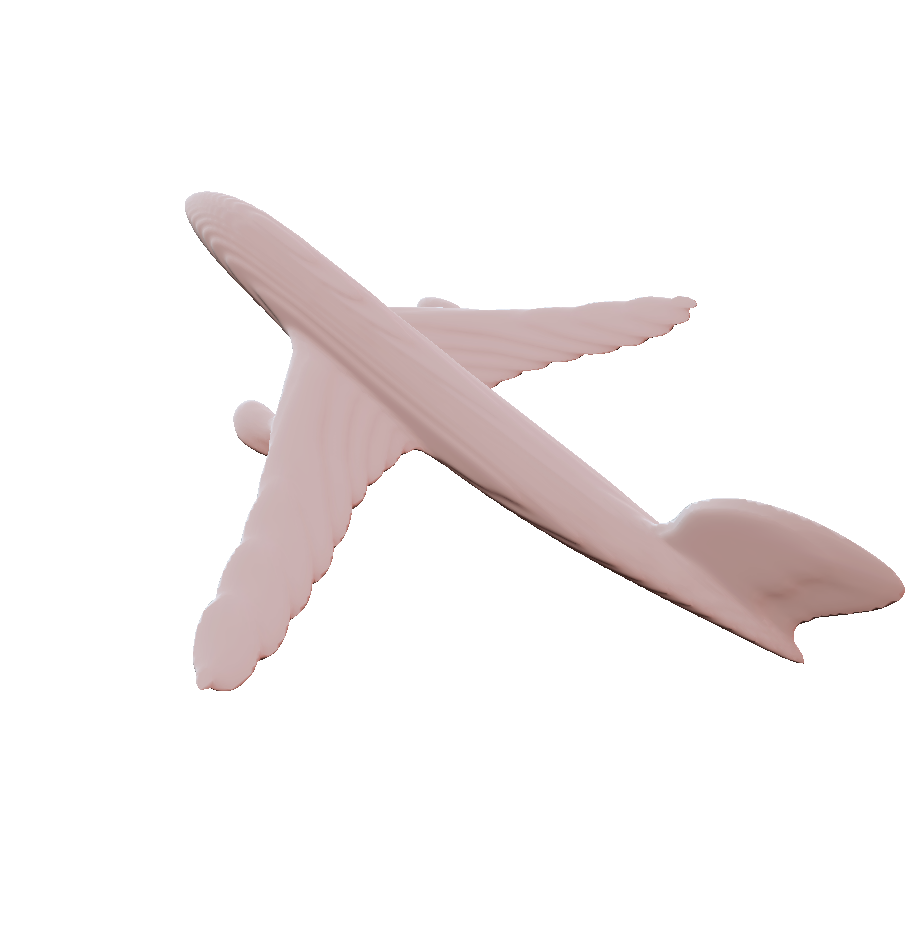}
	\end{subfigure}  \\
	\begin{subfigure}{27.5mm}
		\centering
		\includegraphics[width=27.5mm]{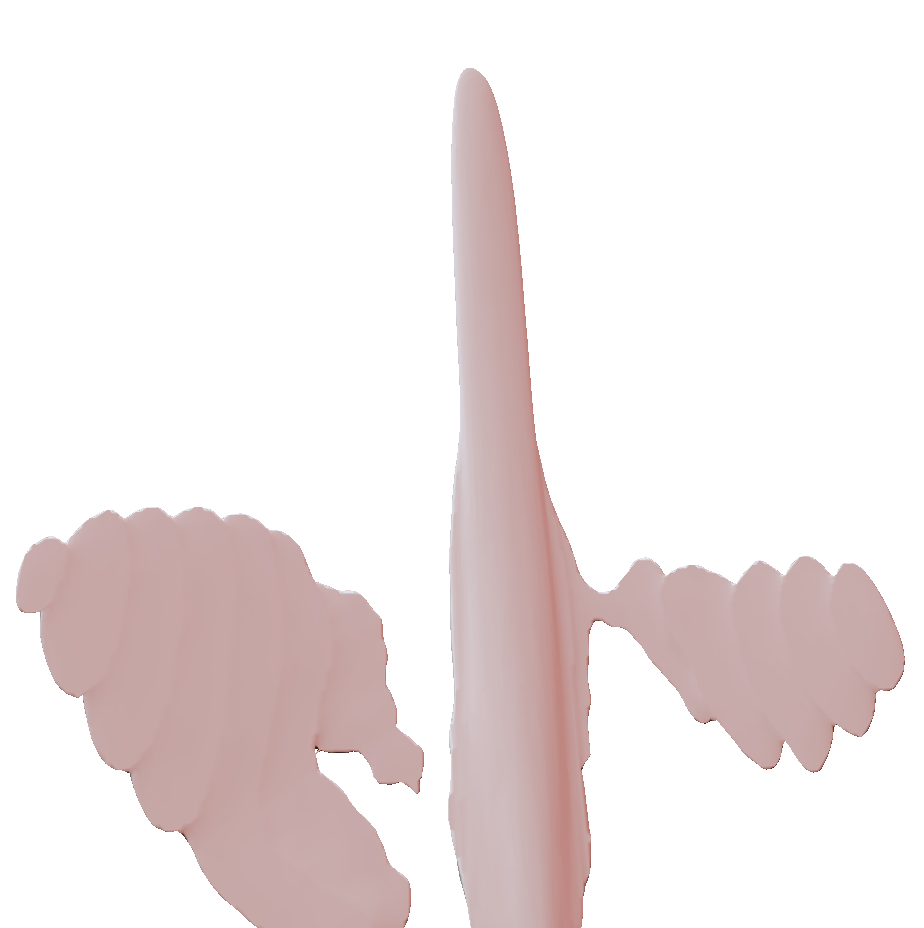}
	\end{subfigure}
	~
	\begin{subfigure}{27.5mm}
		\centering
		\includegraphics[width=27.5mm]{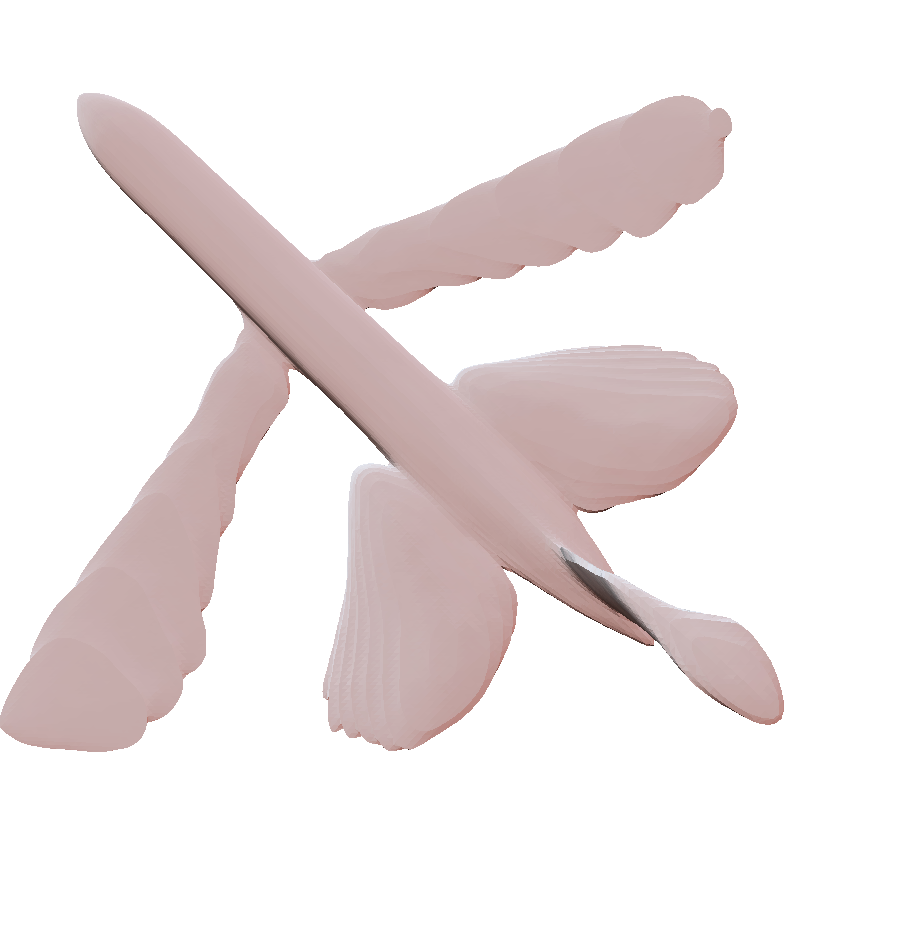}
	\end{subfigure} 
	~
	\begin{subfigure}{27.5mm}
		\centering
		\includegraphics[width=27.5mm]{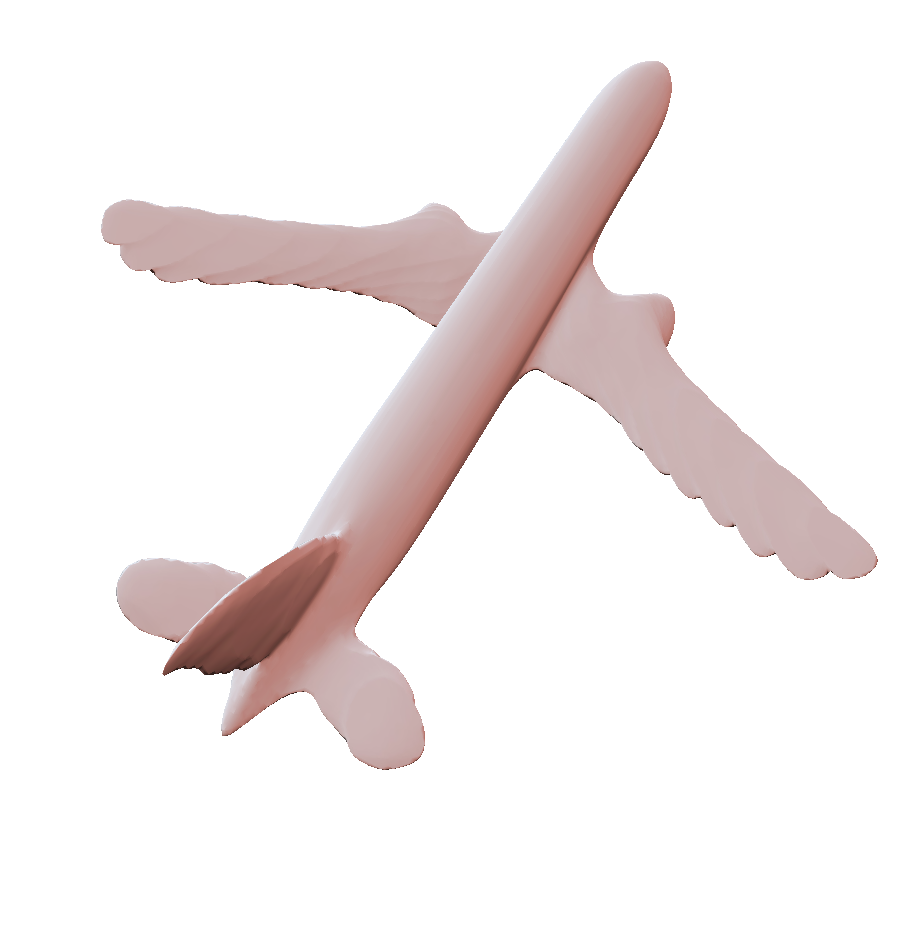}
	\end{subfigure} 
	~
	\begin{subfigure}{27.5mm}
		\centering
		\includegraphics[width=27.5mm]{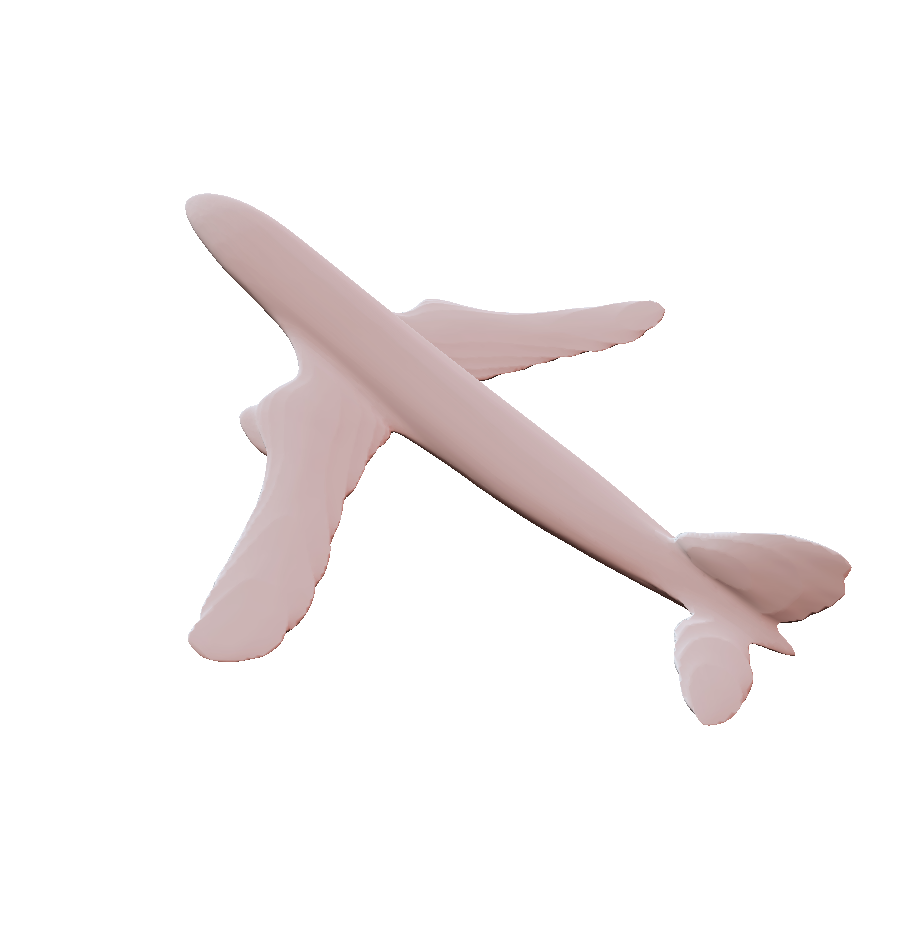}
	\end{subfigure} \\
	\begin{subfigure}{27.5mm}
		\centering
		\includegraphics[width=27.5mm]{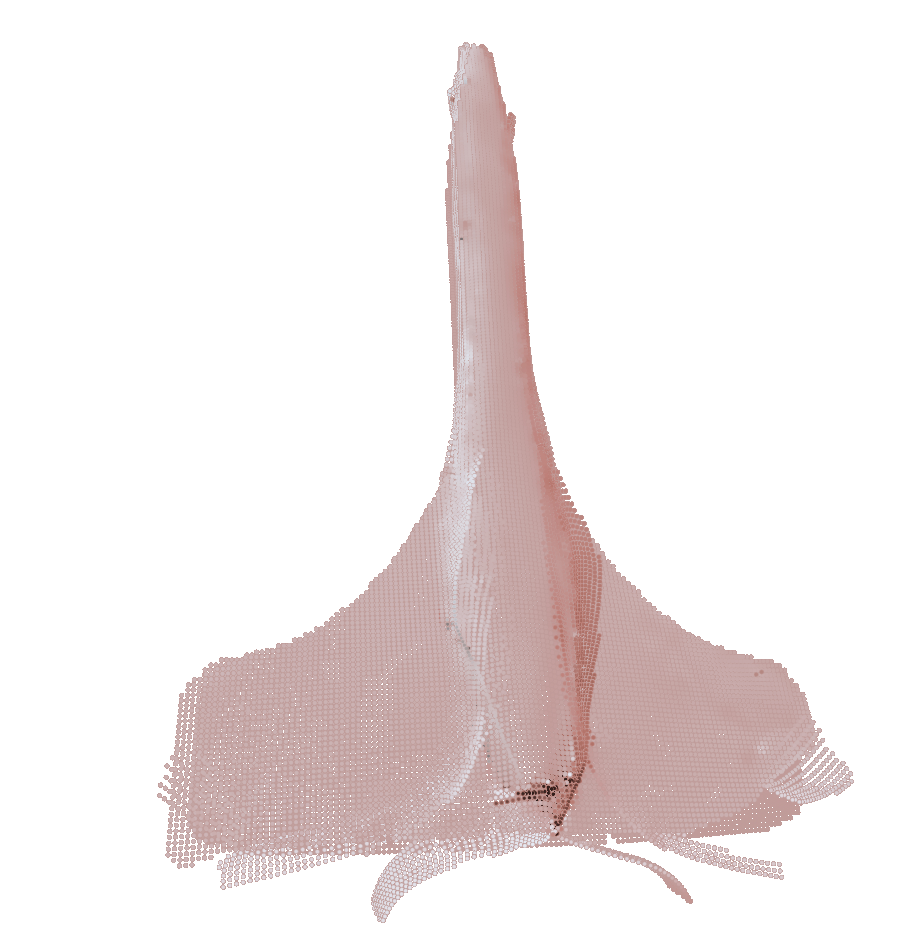}
	\end{subfigure}
	~
	\begin{subfigure}{27.5mm}
		\centering
		\includegraphics[width=27.5mm]{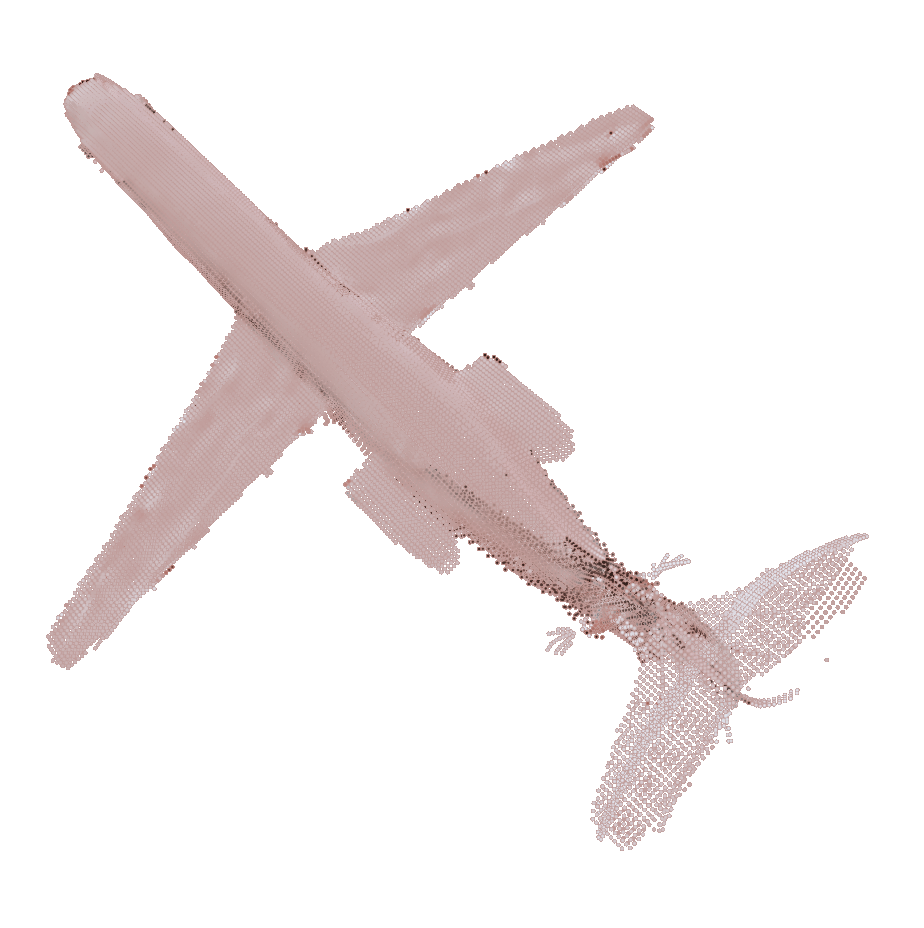}
	\end{subfigure} 
	~
	\begin{subfigure}{27.5mm}
		\centering
		\includegraphics[width=27.5mm]{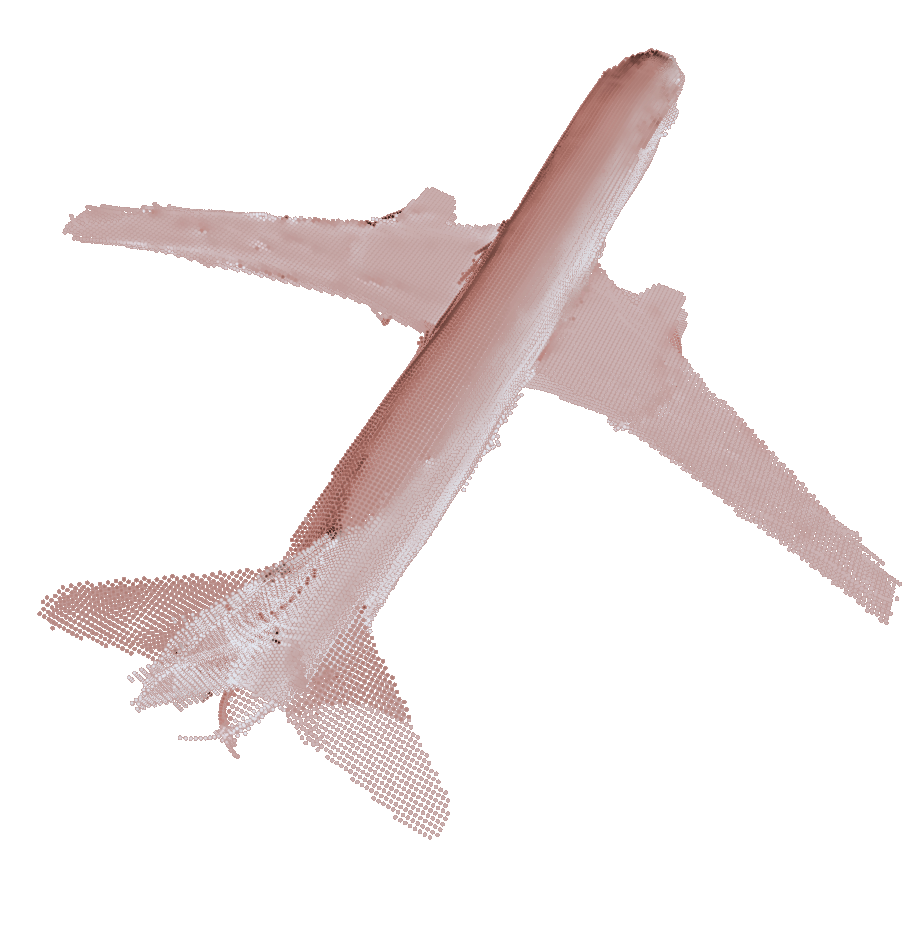}
	\end{subfigure} 
	~
	\begin{subfigure}{27.5mm}
		\centering
		\includegraphics[width=27.5mm]{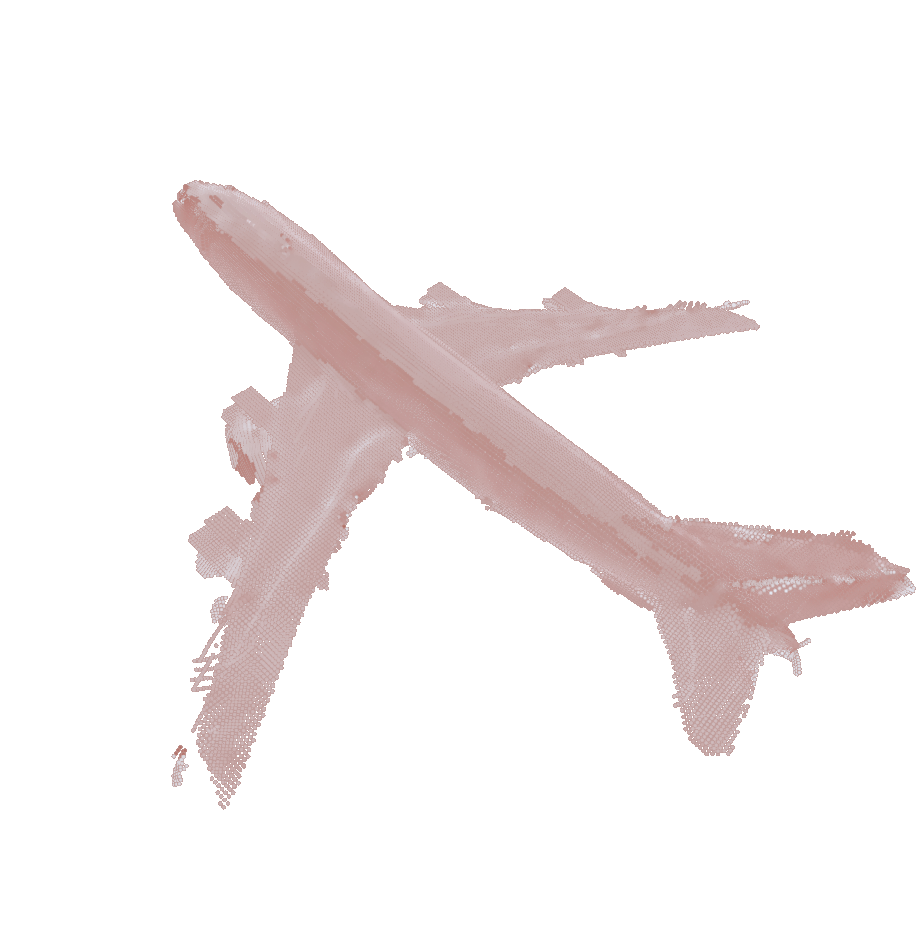}
	\end{subfigure} \\
	\caption{{\bf More Generative Results from Different Categories.} Results from Rows 1 to 4 correspond to {\bf Reference}, {\bf OF}, {\bf SDF}, {\bf PRIF - Mesh}.}
\end{figure}

\begin{figure}[!ht]
    \centering
	\begin{subfigure}{27.5mm}
		\centering
		\includegraphics[width=27.5mm]{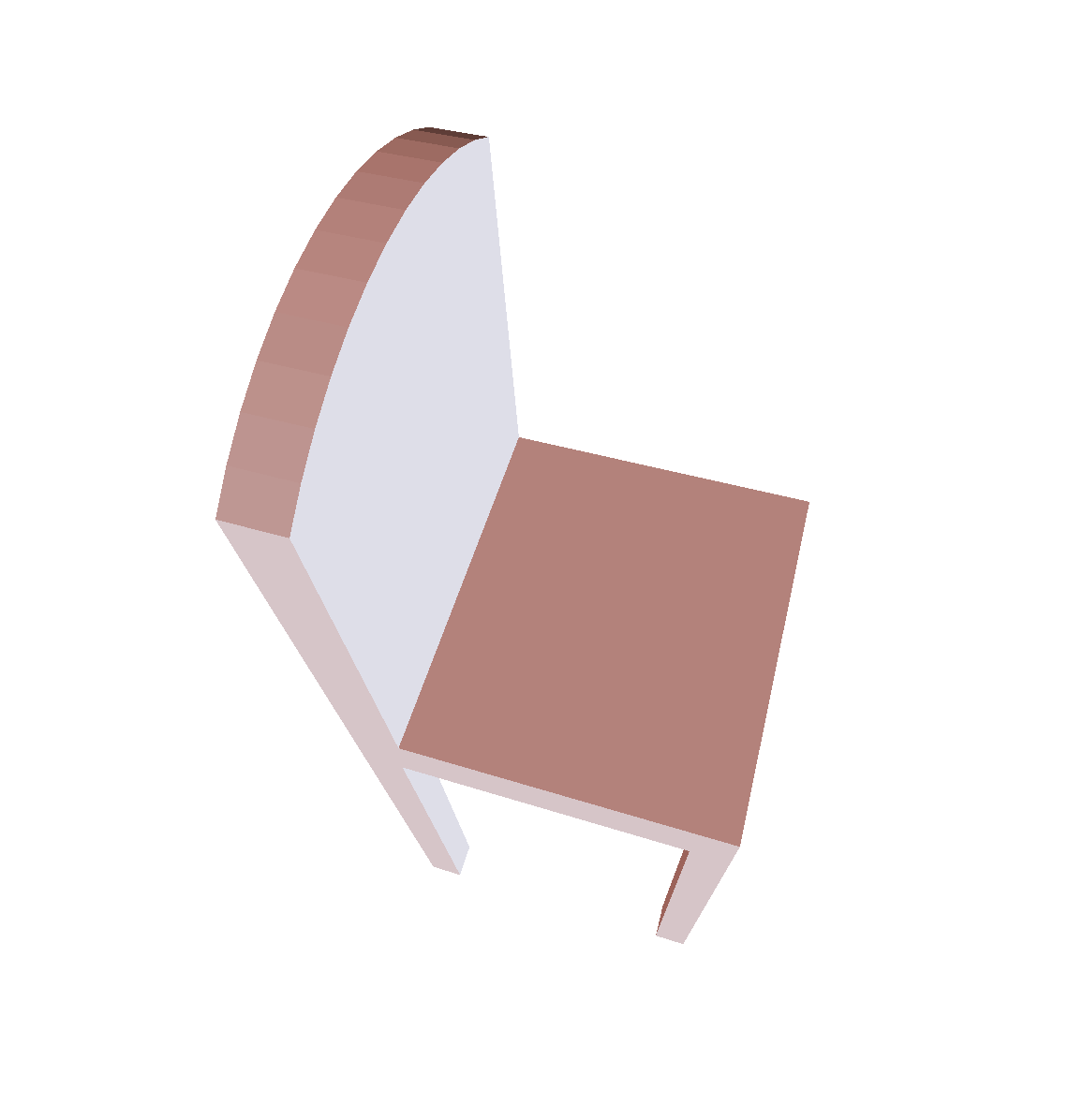}
	\end{subfigure}
	~
	\begin{subfigure}{27.5mm}
		\centering
		\includegraphics[width=27.5mm]{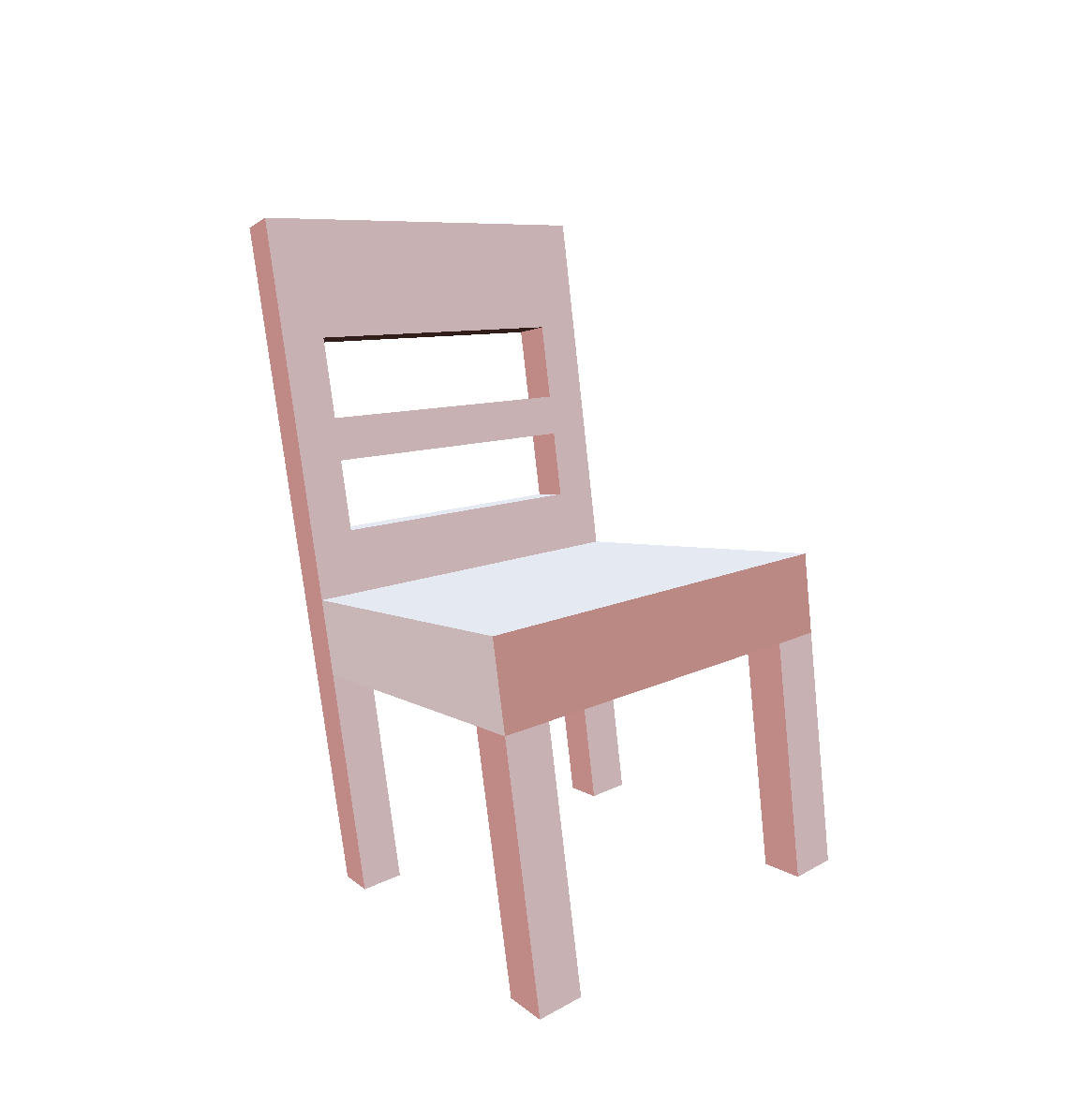}
	\end{subfigure} 
	~
	\begin{subfigure}{27.5mm}
		\centering
		\includegraphics[width=27.5mm]{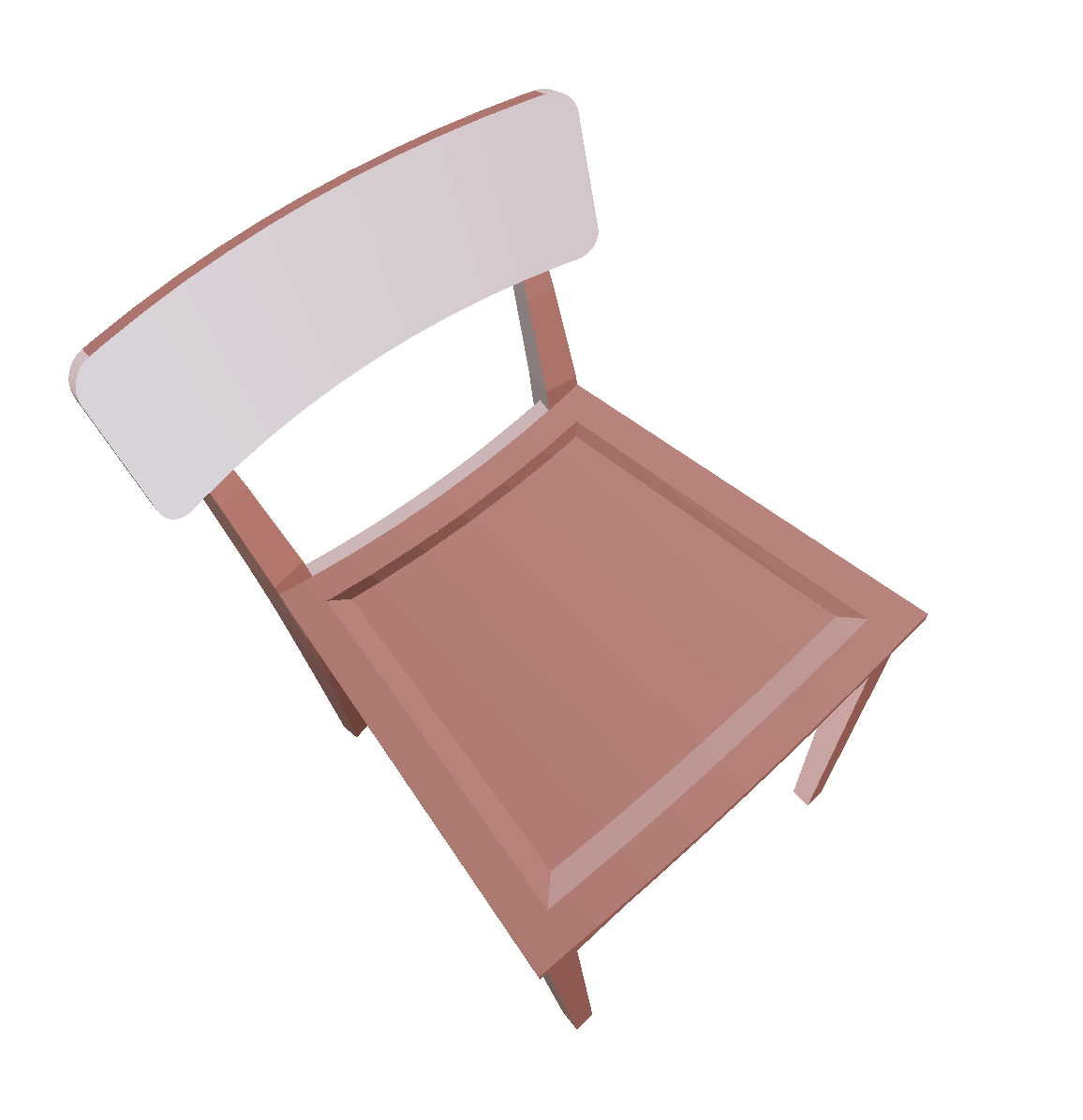}
	\end{subfigure}
	~
	\begin{subfigure}{27.5mm}
		\centering
		\includegraphics[width=27.5mm]{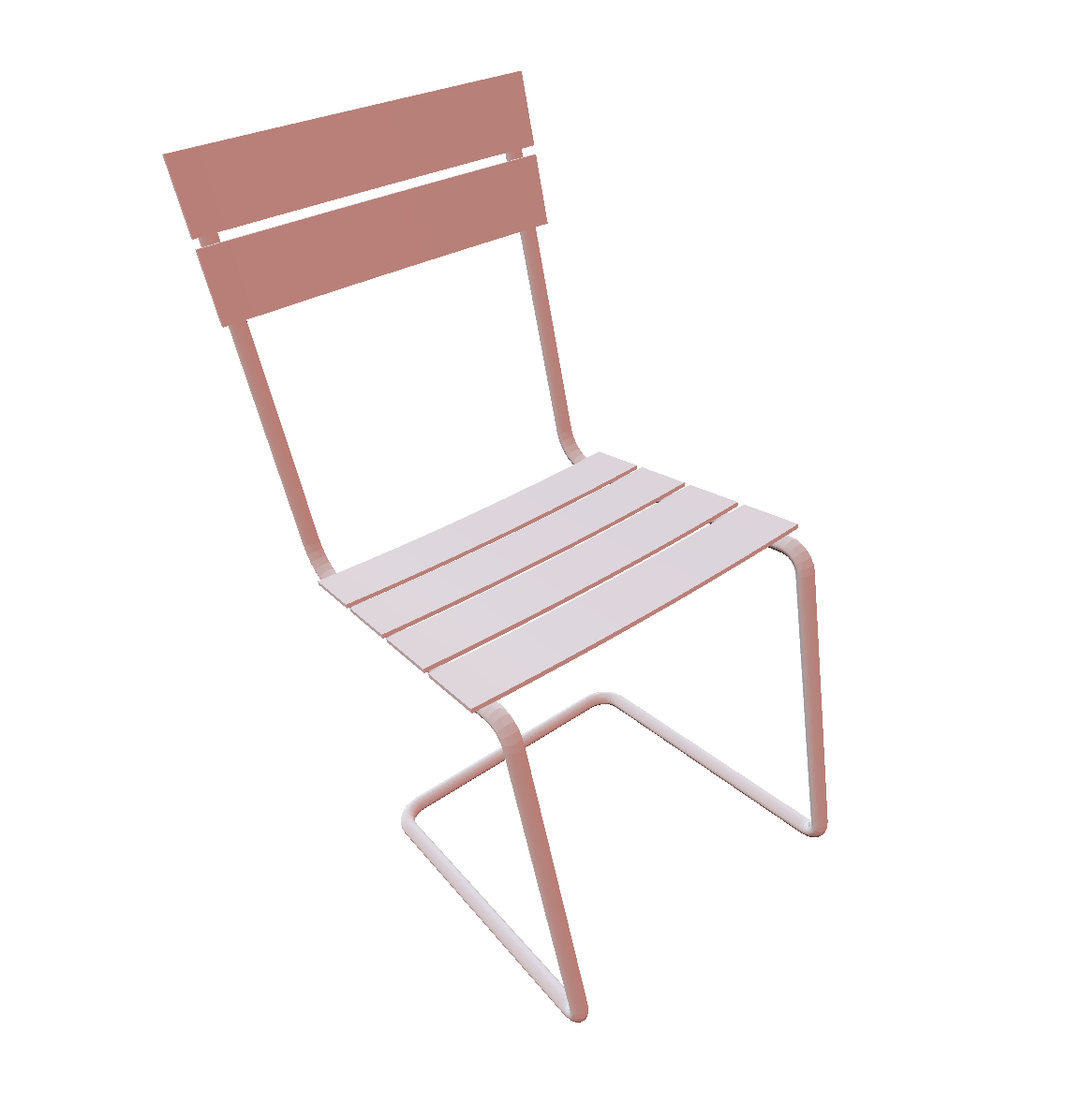}
	\end{subfigure}  \\
	\begin{subfigure}{27.5mm}
		\centering
		\includegraphics[width=27.5mm]{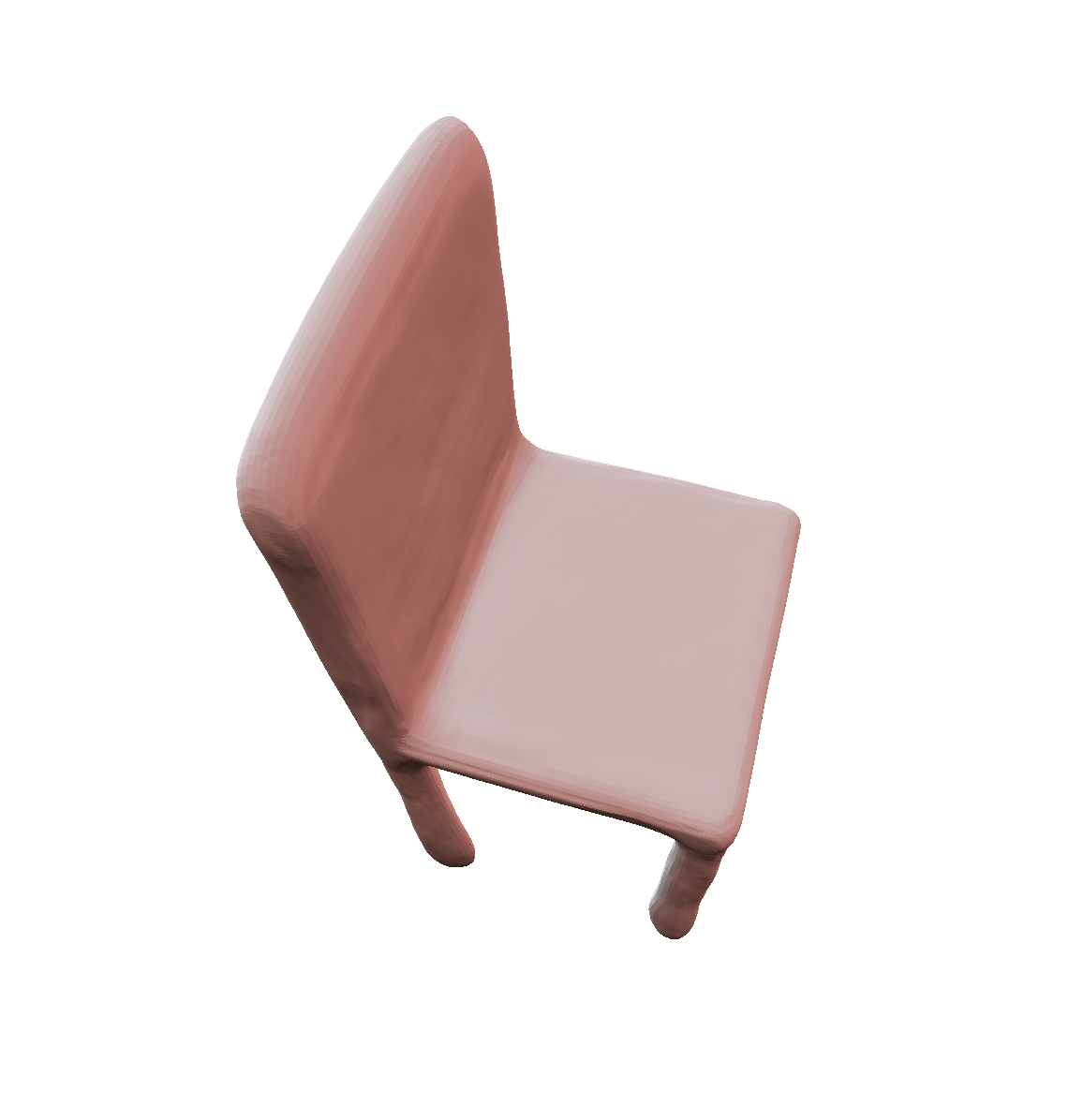}
	\end{subfigure}
	~
	\begin{subfigure}{27.5mm}
		\centering
		\includegraphics[width=27.5mm]{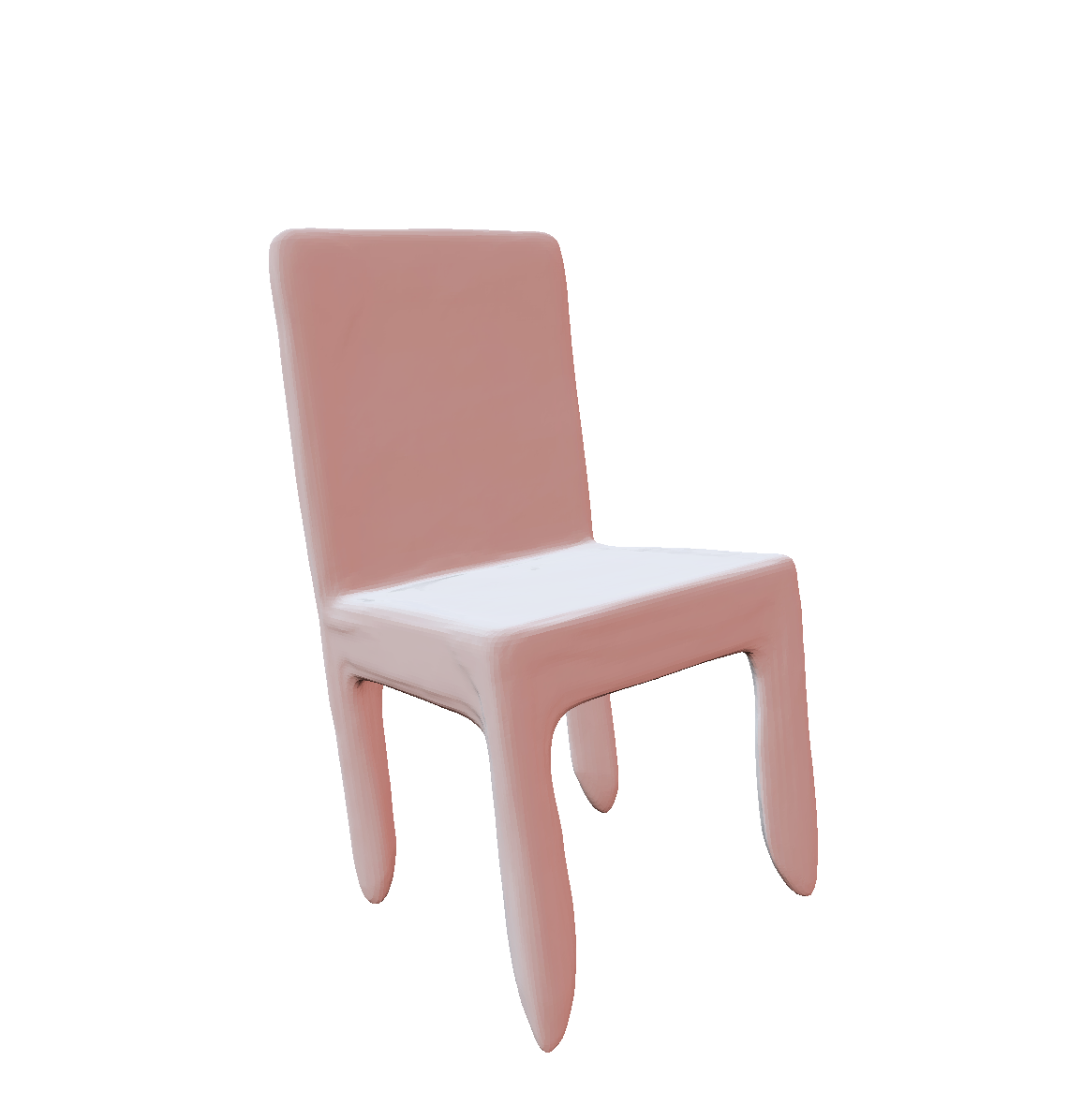}
	\end{subfigure} 
	~
	\begin{subfigure}{27.5mm}
		\centering
		\includegraphics[width=27.5mm]{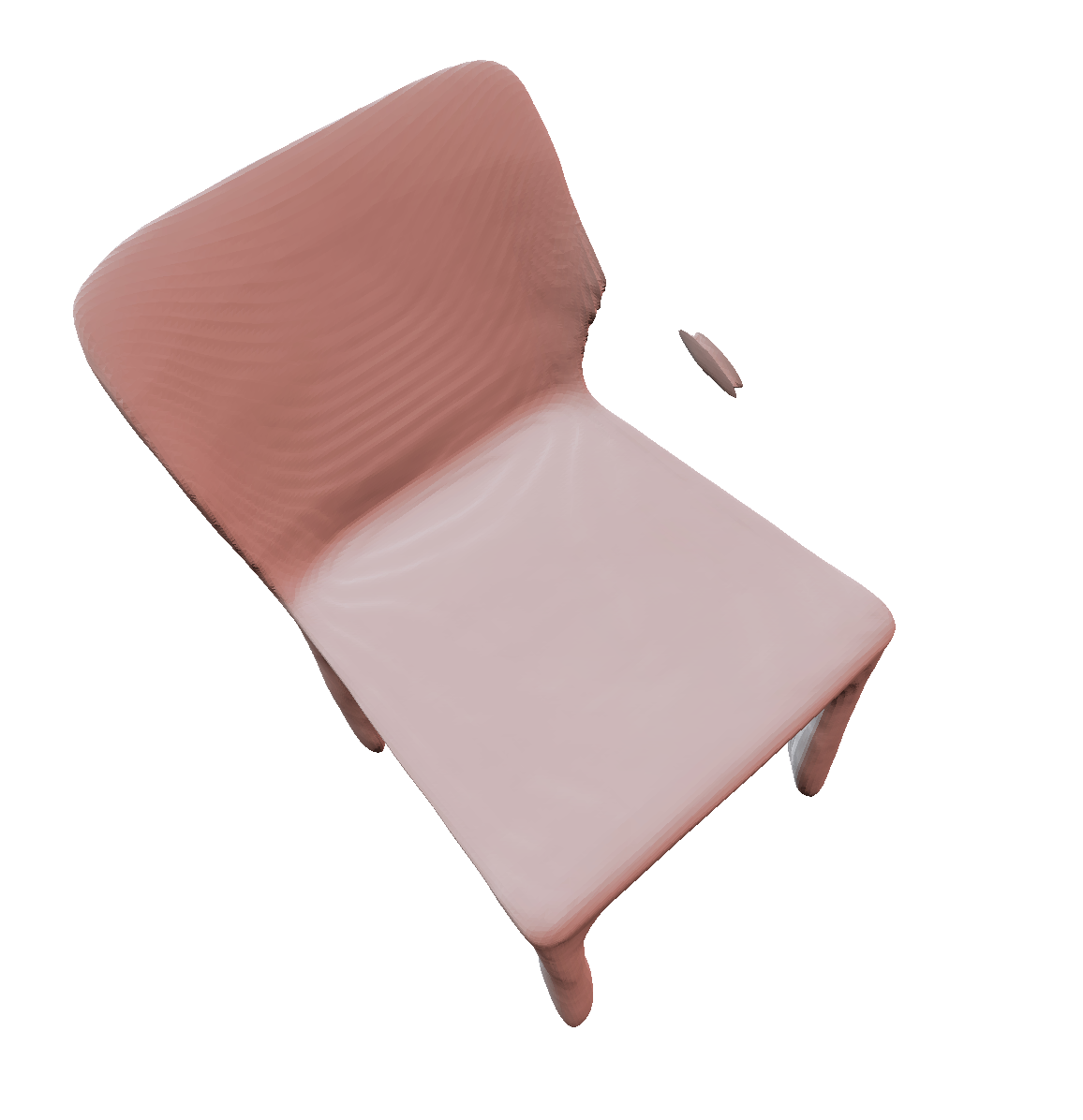}
	\end{subfigure} 
	~
	\begin{subfigure}{27.5mm}
		\centering
		\includegraphics[width=27.5mm]{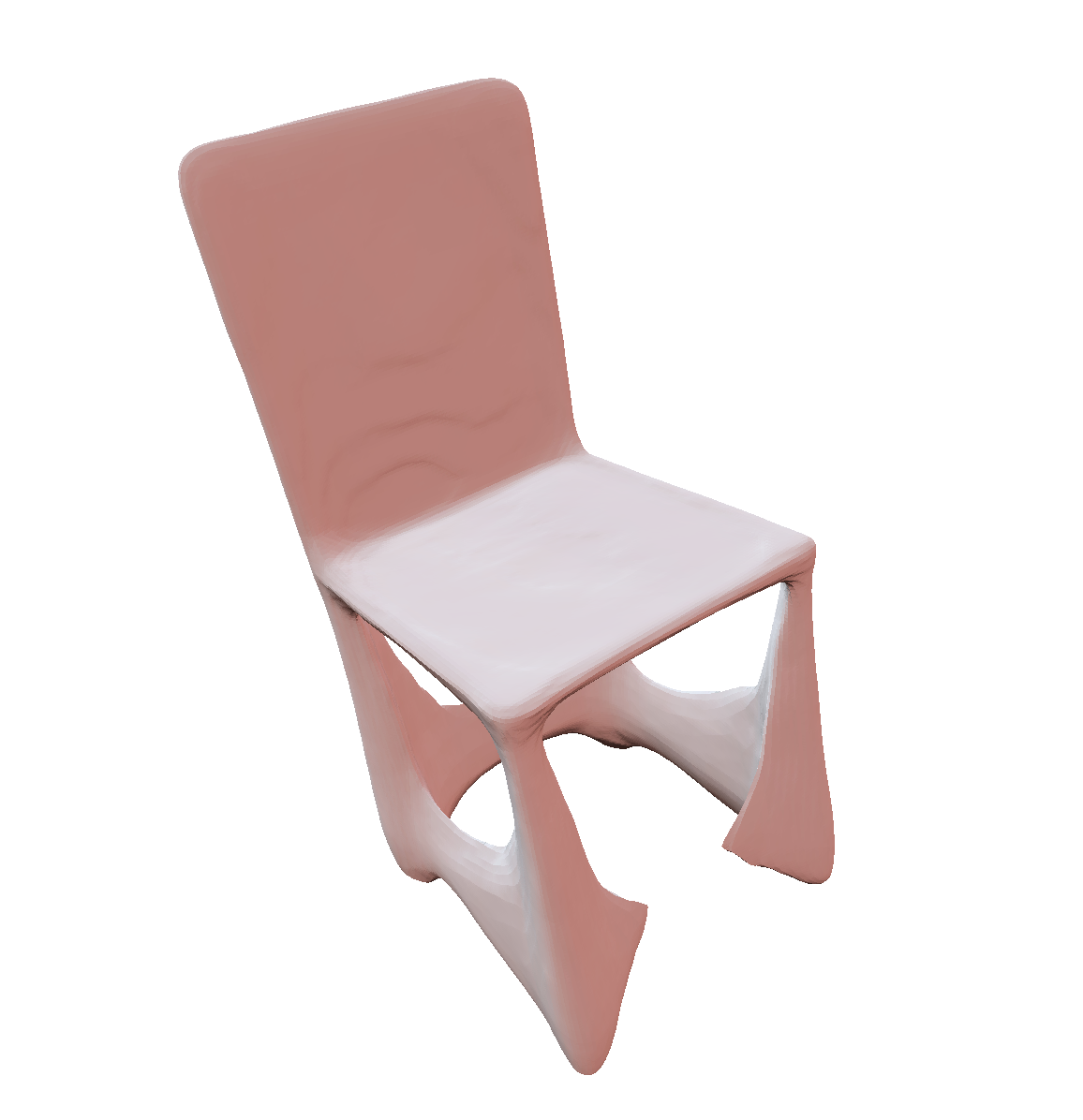}
	\end{subfigure}  \\
	\begin{subfigure}{27.5mm}
		\centering
		\includegraphics[width=27.5mm]{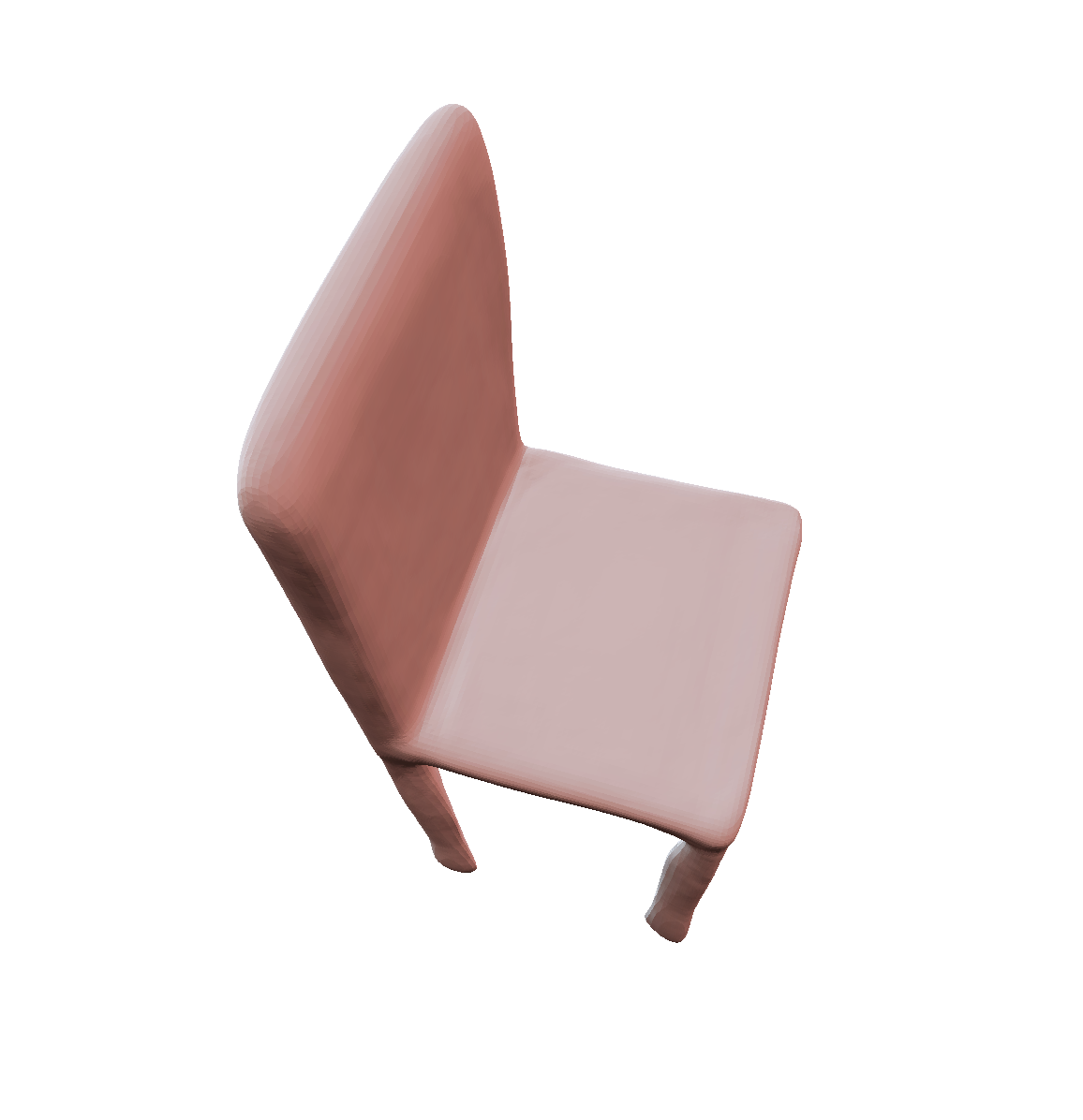}
	\end{subfigure}
	~
	\begin{subfigure}{27.5mm}
		\centering
		\includegraphics[width=27.5mm]{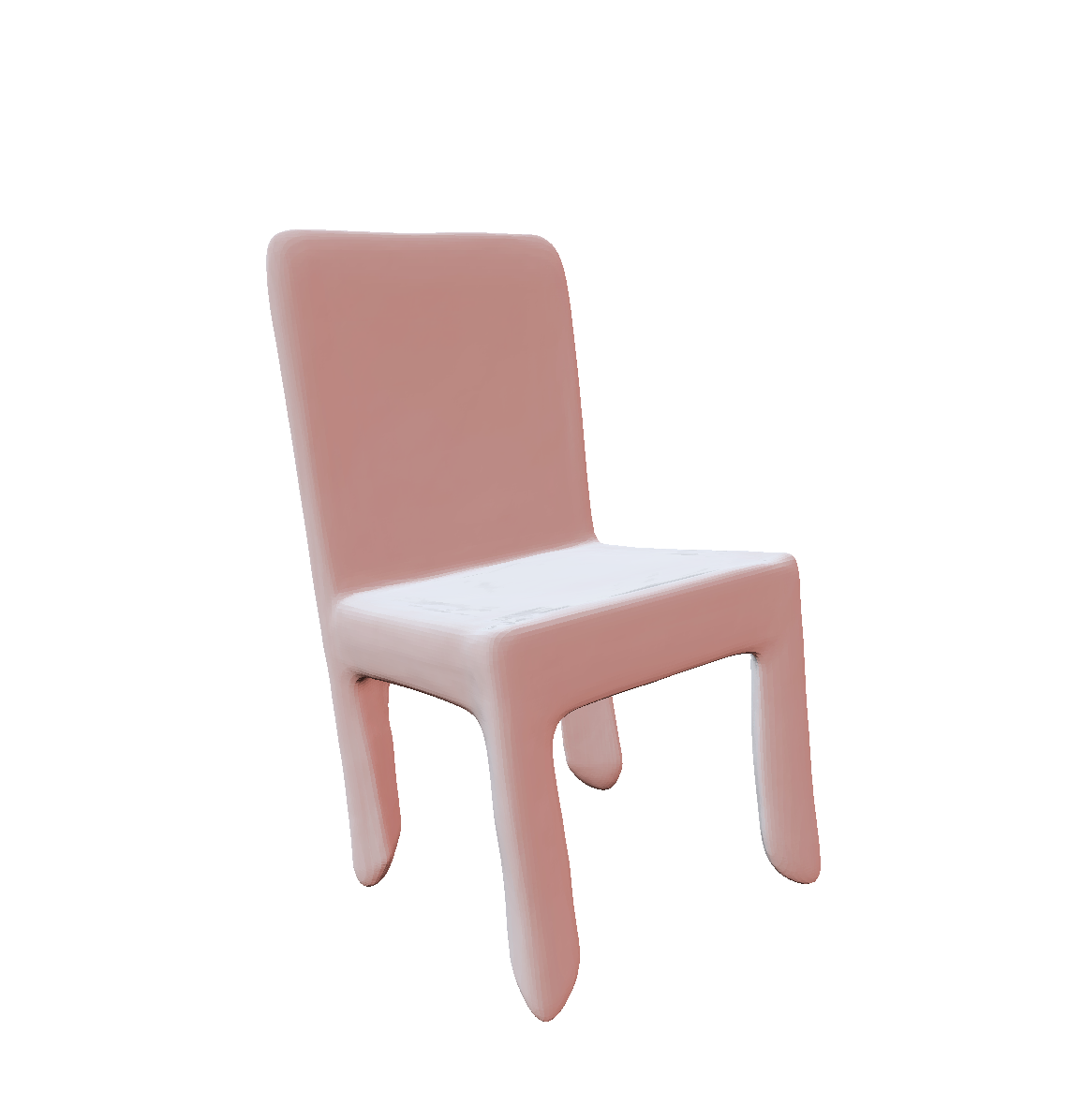}
	\end{subfigure} 
	~
	\begin{subfigure}{27.5mm}
		\centering
		\includegraphics[width=27.5mm]{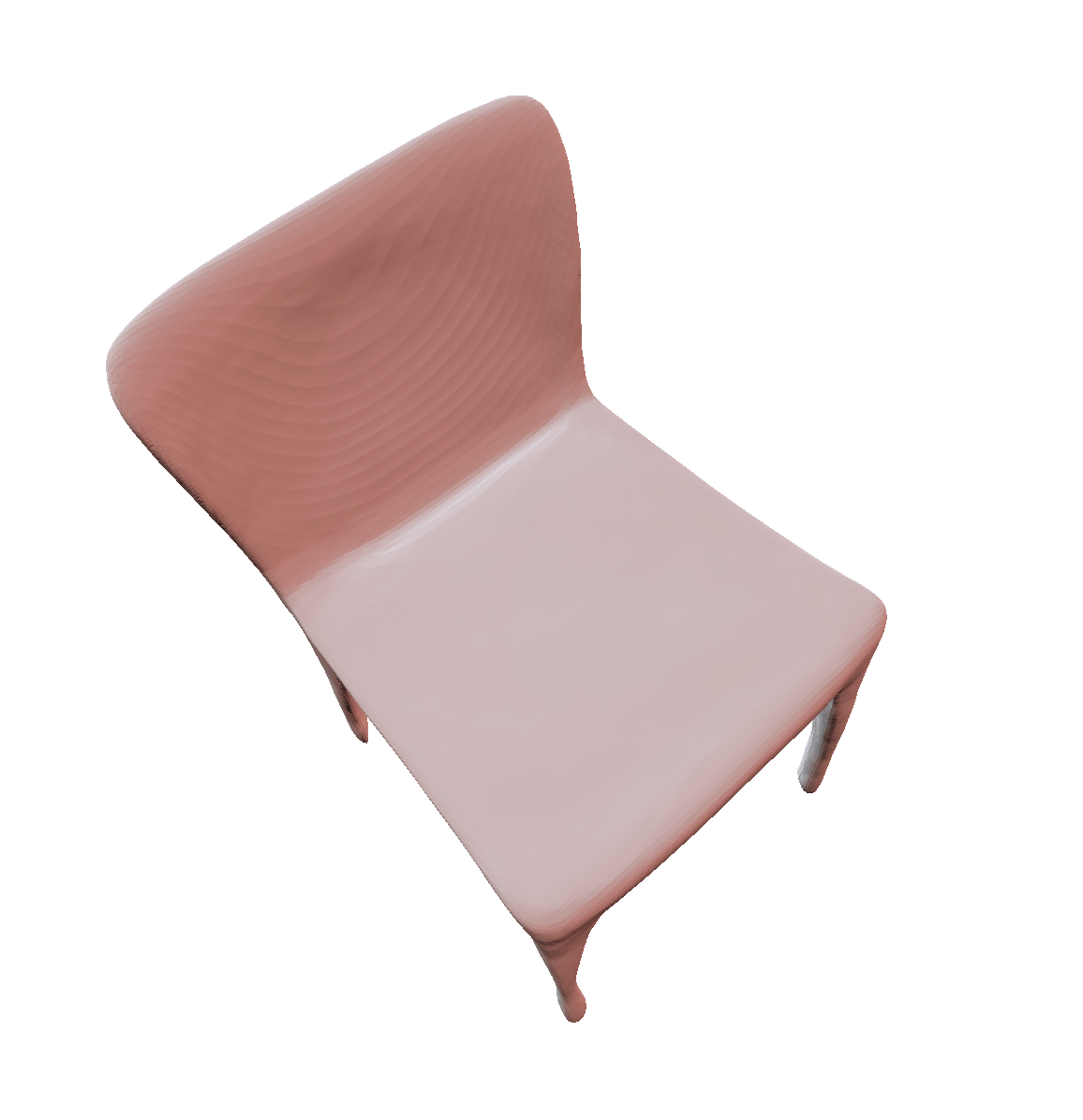}
	\end{subfigure} 
	~
	\begin{subfigure}{27.5mm}
		\centering
		\includegraphics[width=27.5mm]{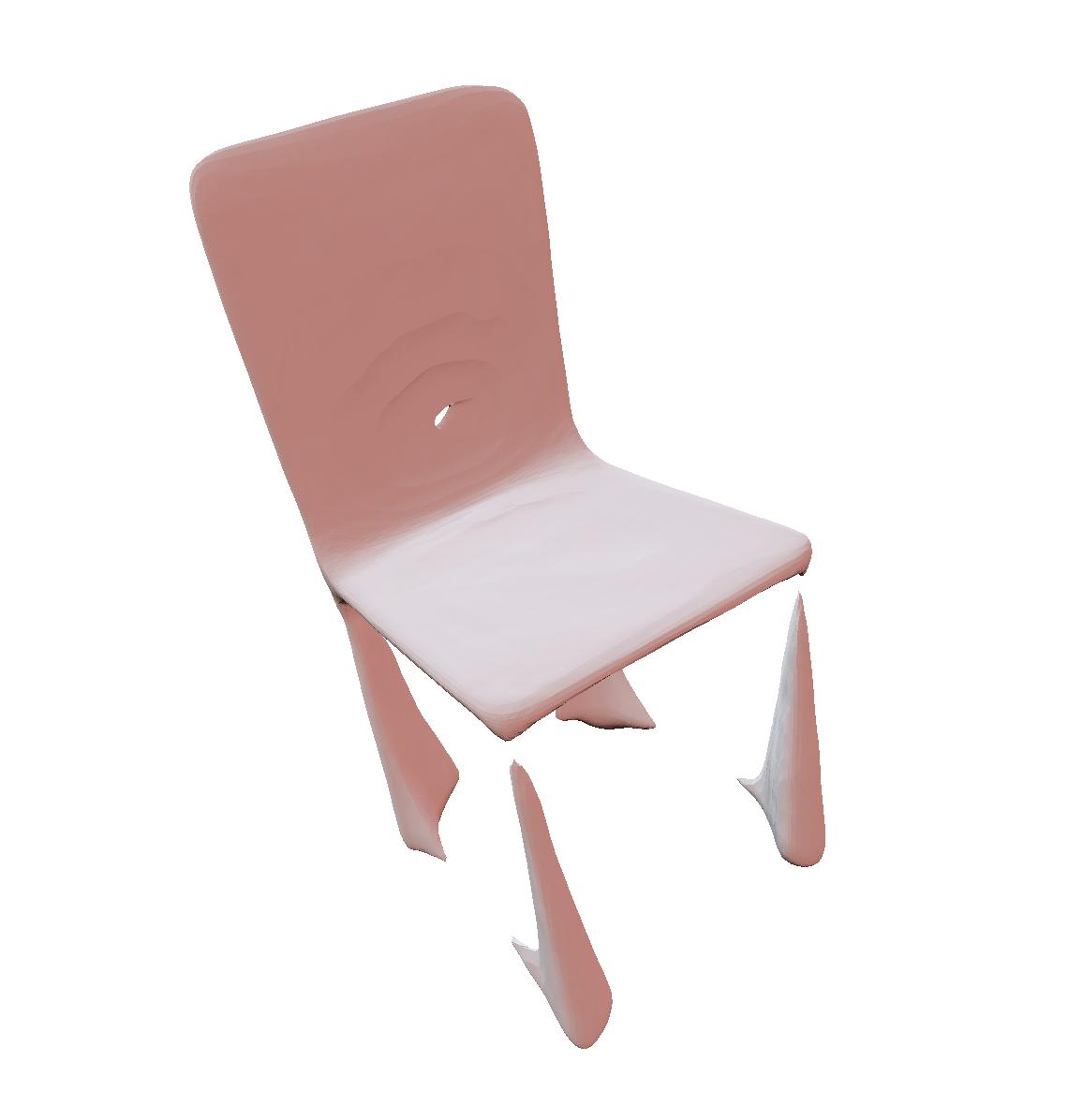}
	\end{subfigure} \\
	\begin{subfigure}{27.5mm}
		\centering
		\includegraphics[width=27.5mm]{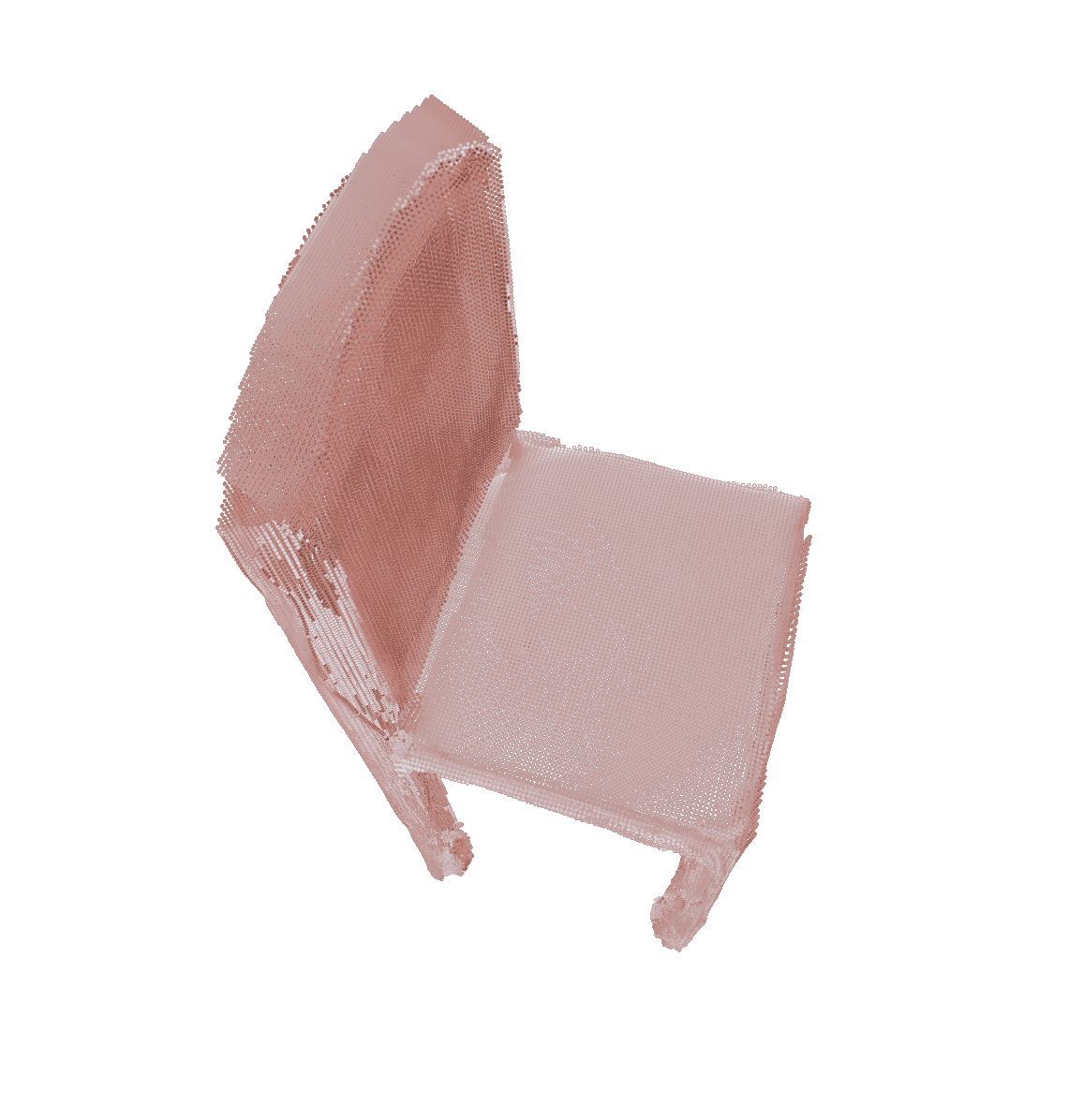}
	\end{subfigure}
	~
	\begin{subfigure}{27.5mm}
		\centering
		\includegraphics[width=27.5mm]{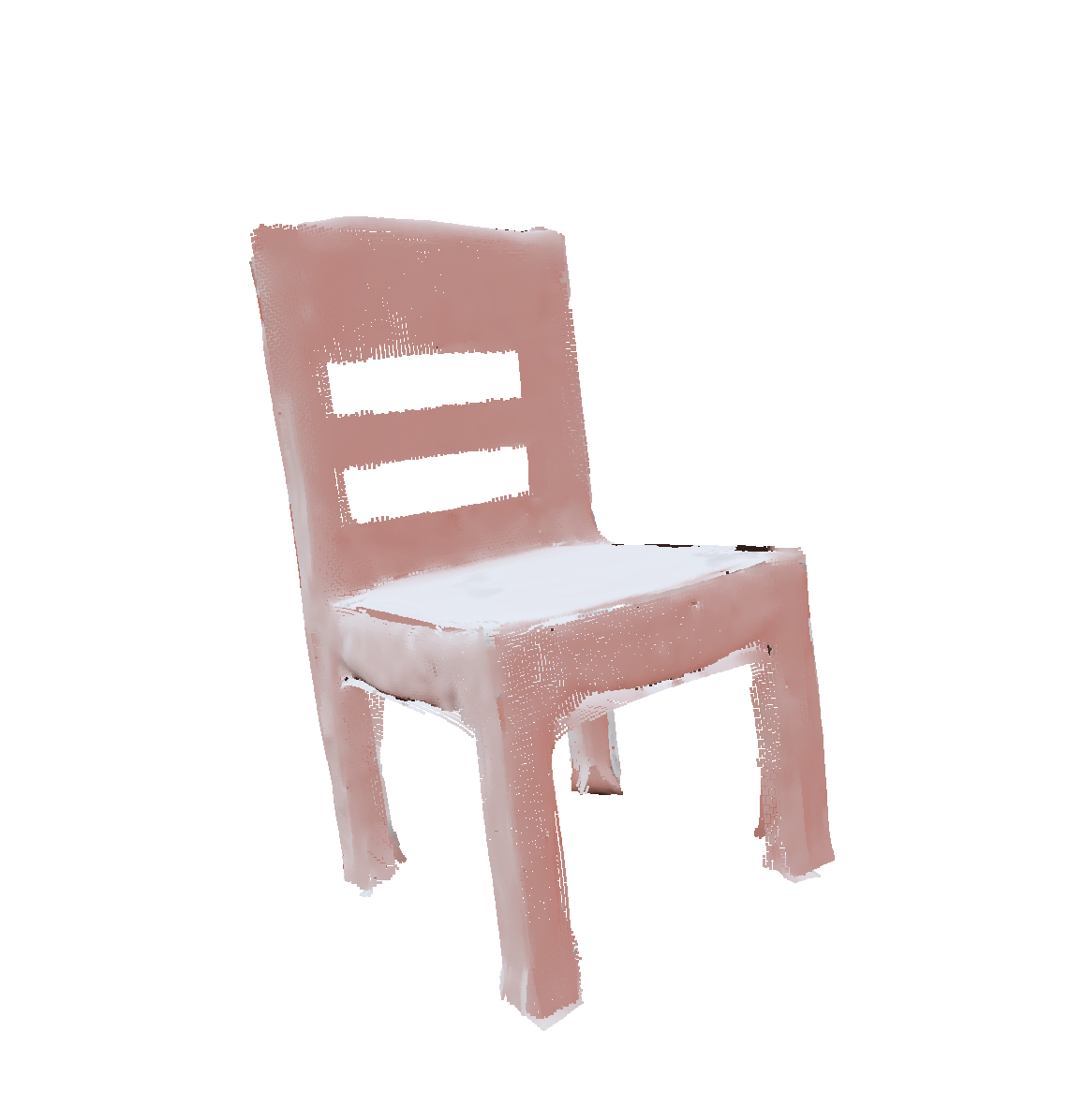}
	\end{subfigure} 
	~
	\begin{subfigure}{27.5mm}
		\centering
		\includegraphics[width=27.5mm]{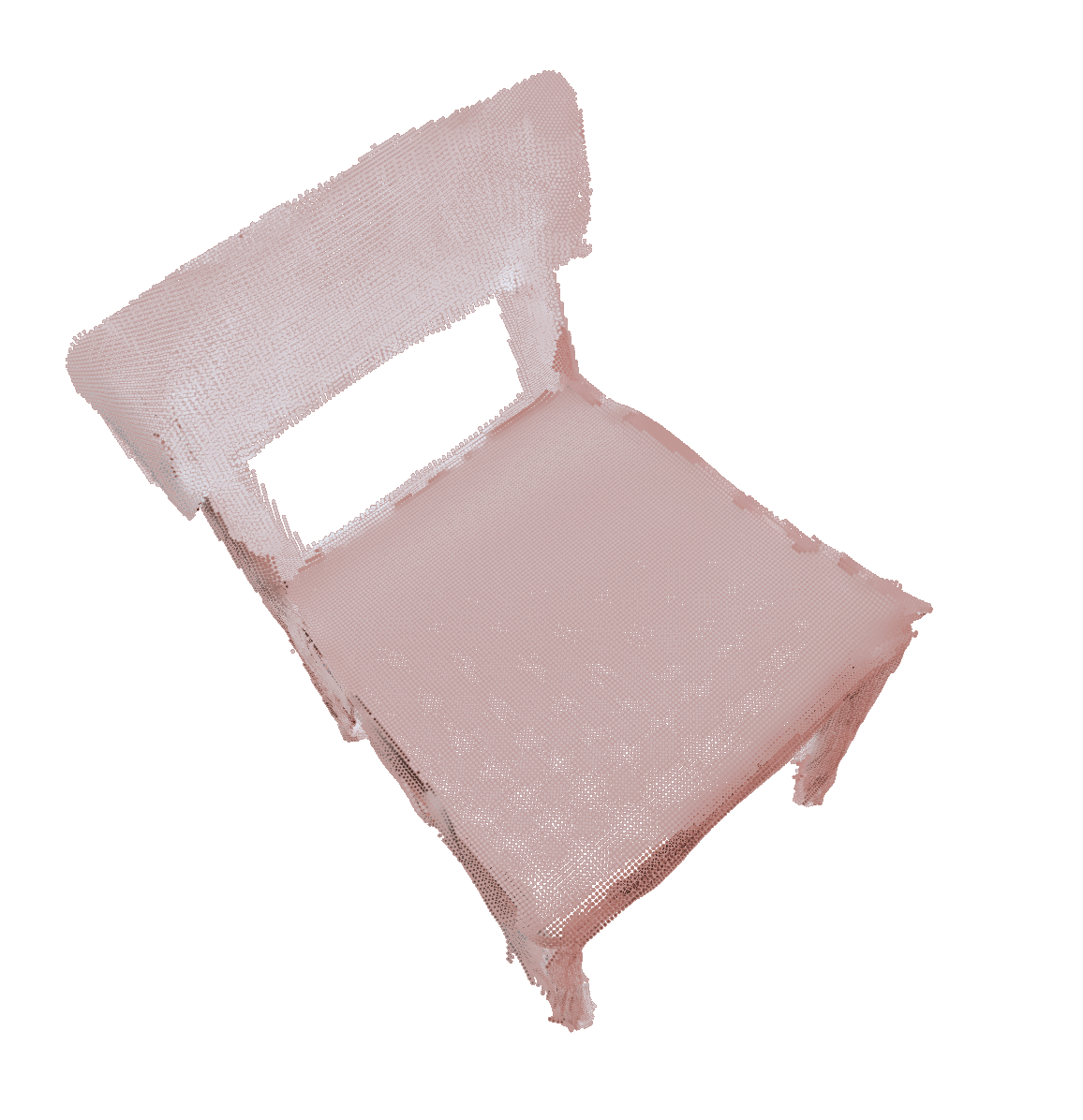}
	\end{subfigure} 
	~
	\begin{subfigure}{27.5mm}
		\centering
		\includegraphics[width=27.5mm]{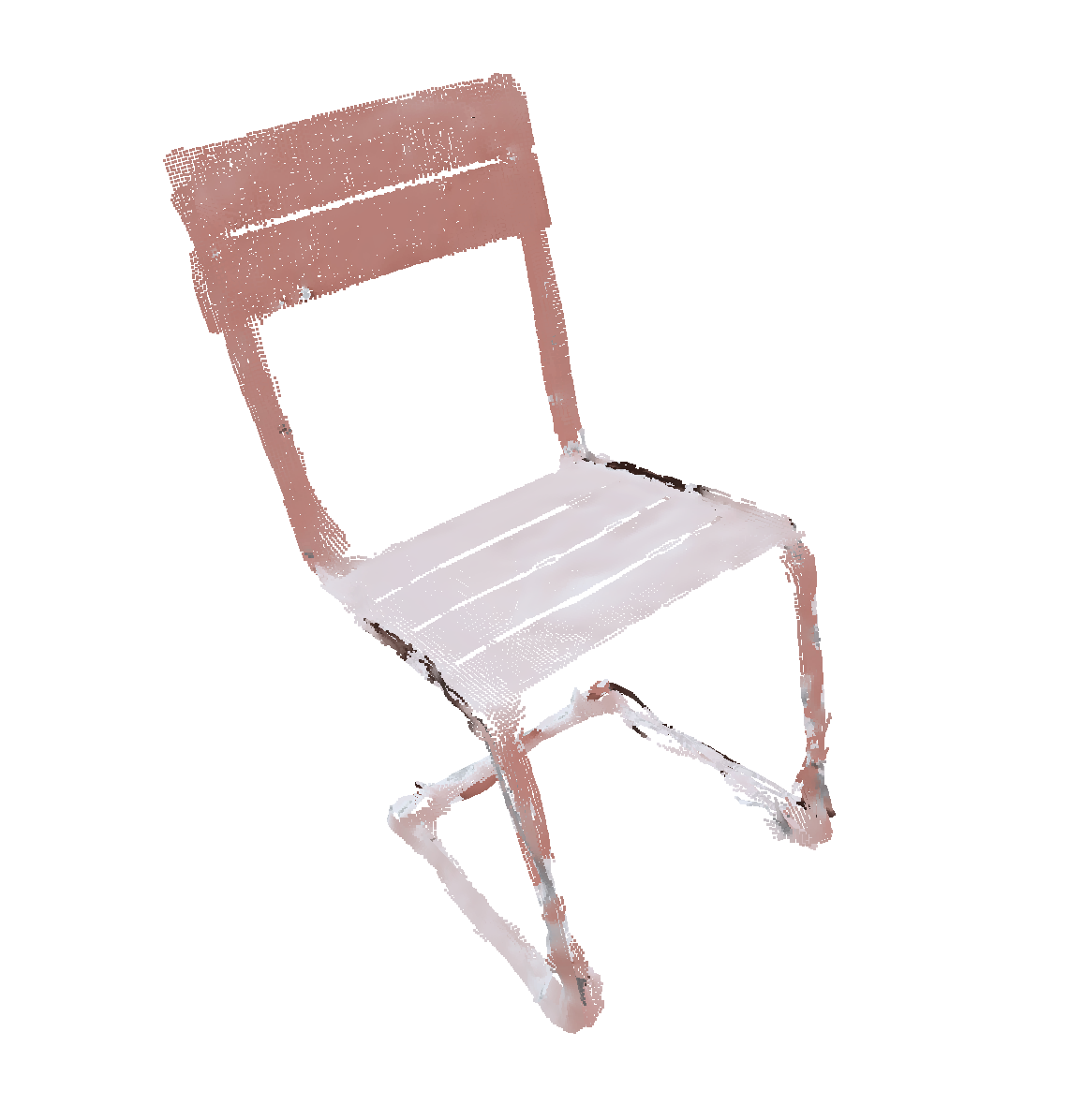}
	\end{subfigure} \\
	\caption{{\bf More Generative Results from Different Categories.} Results from Rows 1 to 4 correspond to {\bf Reference}, {\bf OF}, {\bf SDF}, {\bf PRIF - Mesh}.}
\end{figure}

\begin{figure}[!ht]
    \centering
	\begin{subfigure}{27.5mm}
		\centering
		\includegraphics[width=27.5mm]{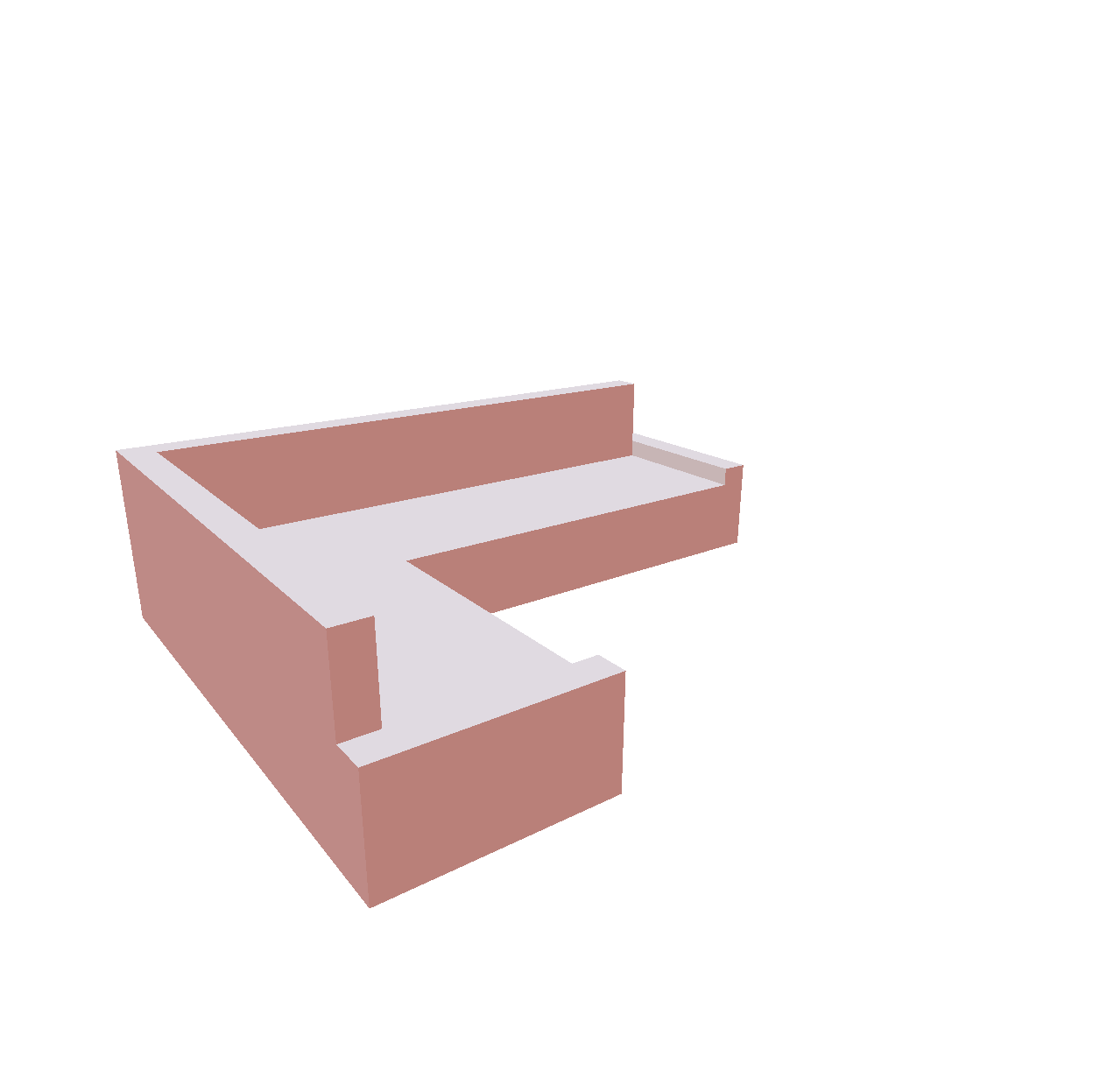}
	\end{subfigure}
	~
	\begin{subfigure}{27.5mm}
		\centering
		\includegraphics[width=27.5mm]{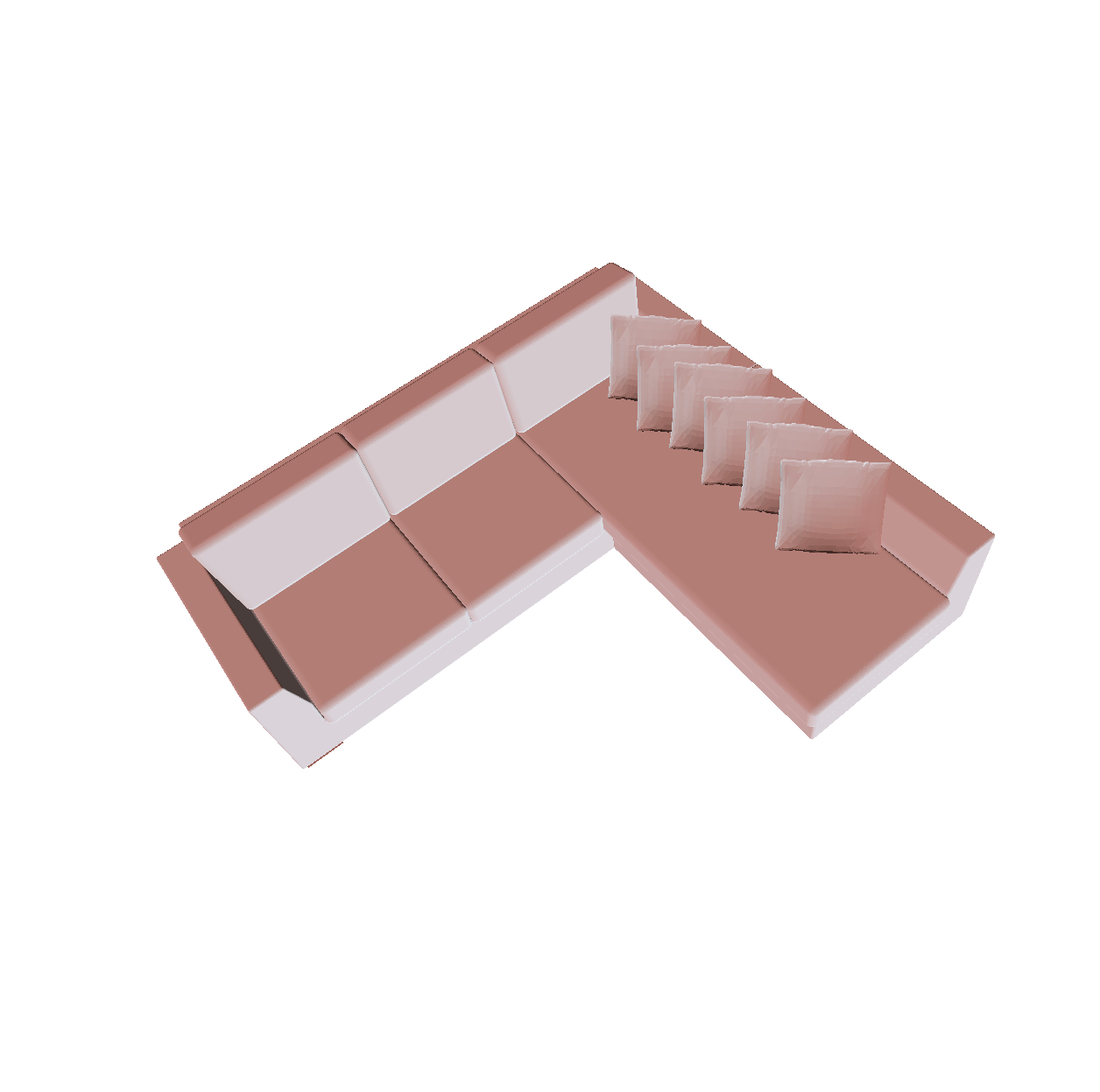}
	\end{subfigure} 
	~
	\begin{subfigure}{27.5mm}
		\centering
		\includegraphics[width=27.5mm]{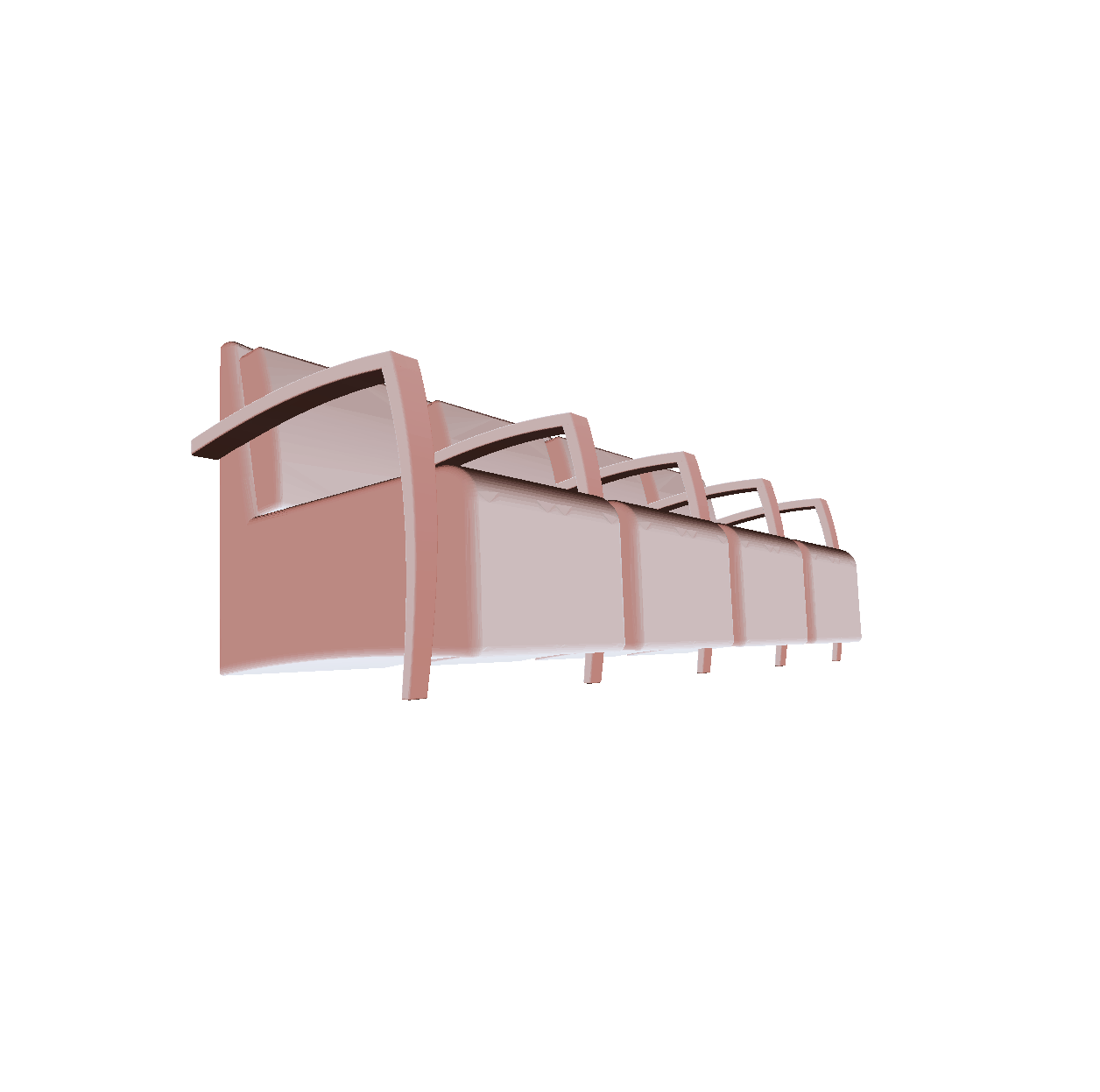}
	\end{subfigure}
	~
	\begin{subfigure}{27.5mm}
		\centering
		\includegraphics[width=27.5mm]{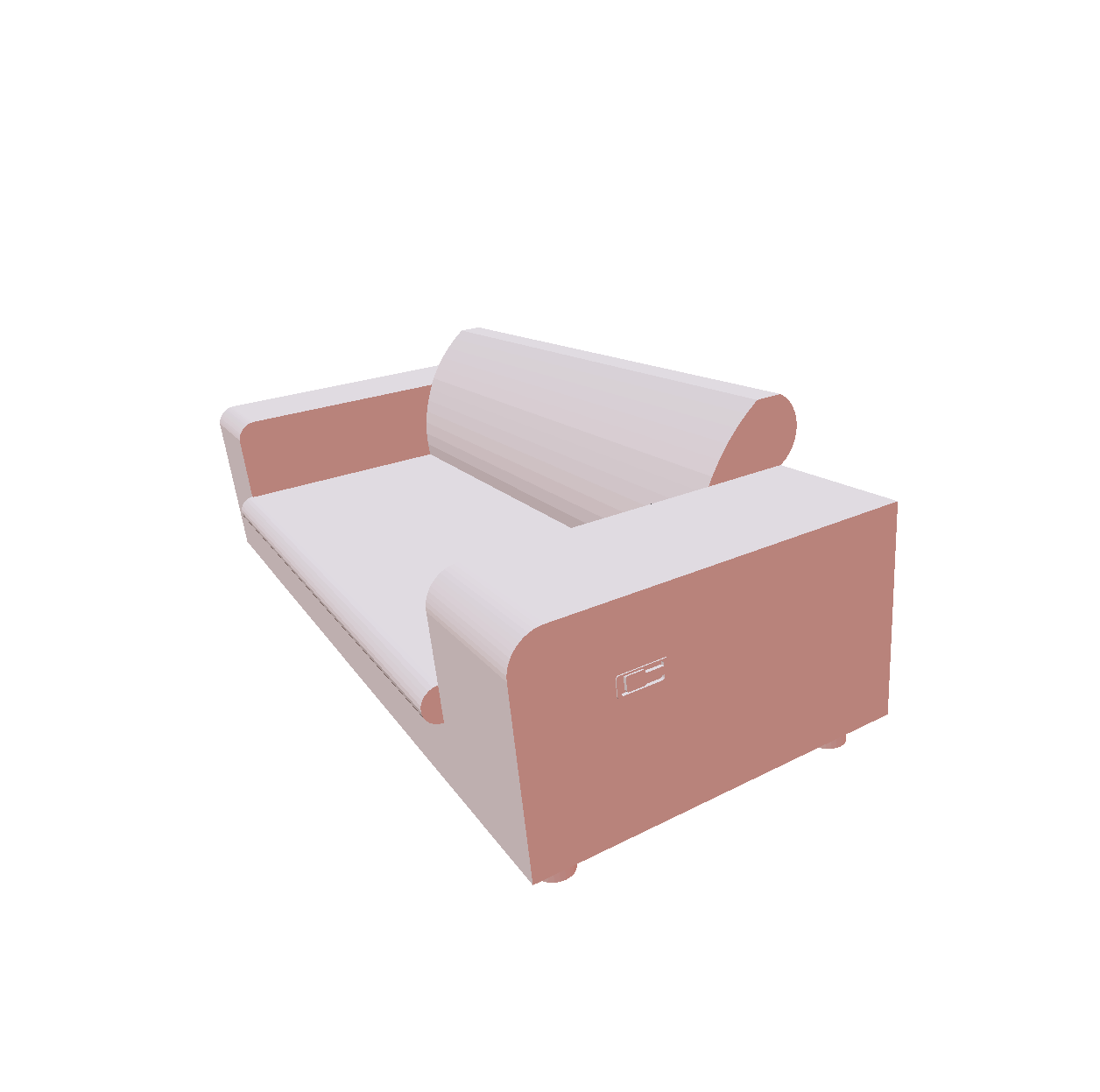}
	\end{subfigure}  \\
	\begin{subfigure}{27.5mm}
		\centering
		\includegraphics[width=27.5mm]{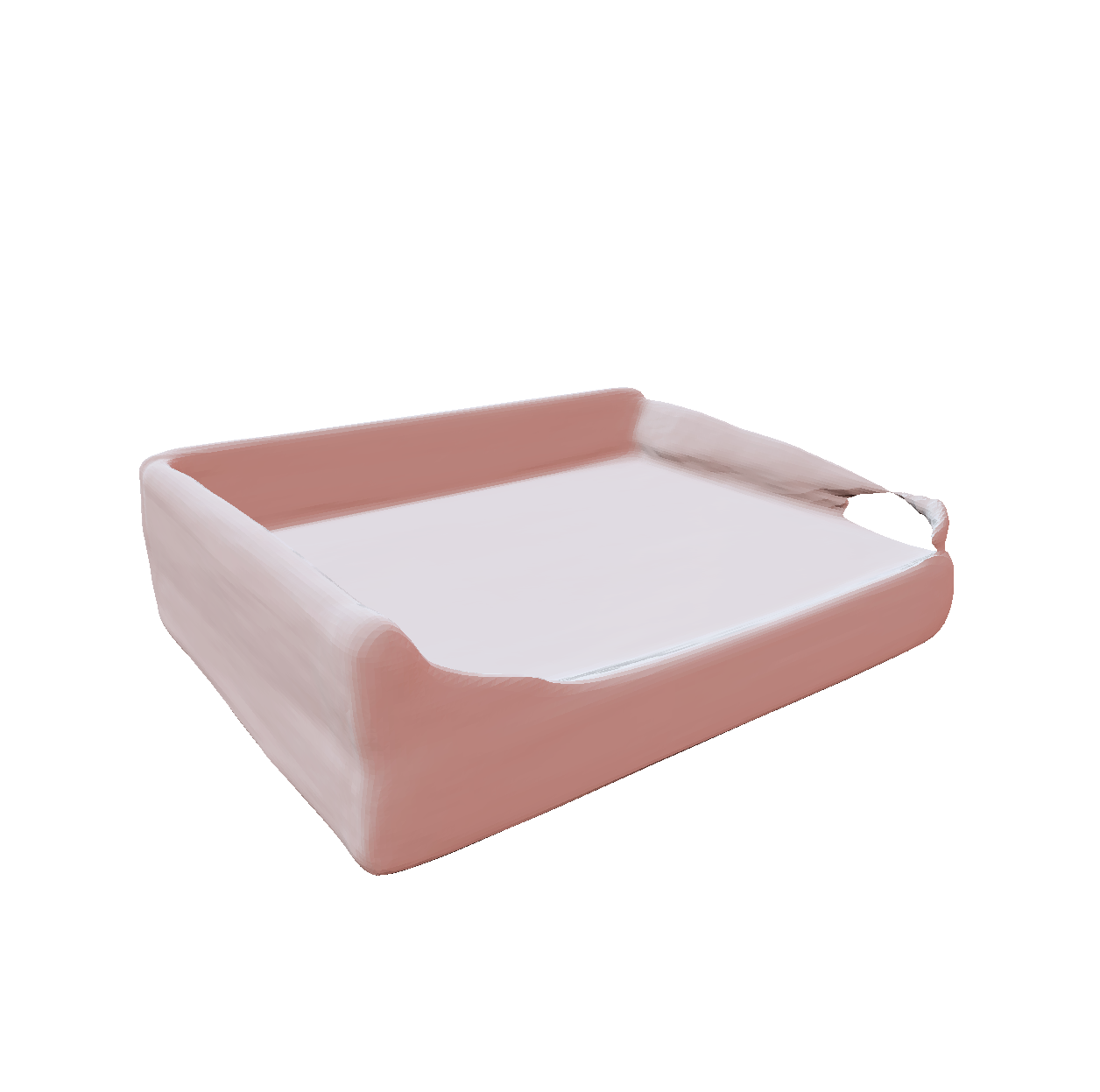}
	\end{subfigure}
	~
	\begin{subfigure}{27.5mm}
		\centering
		\includegraphics[width=27.5mm]{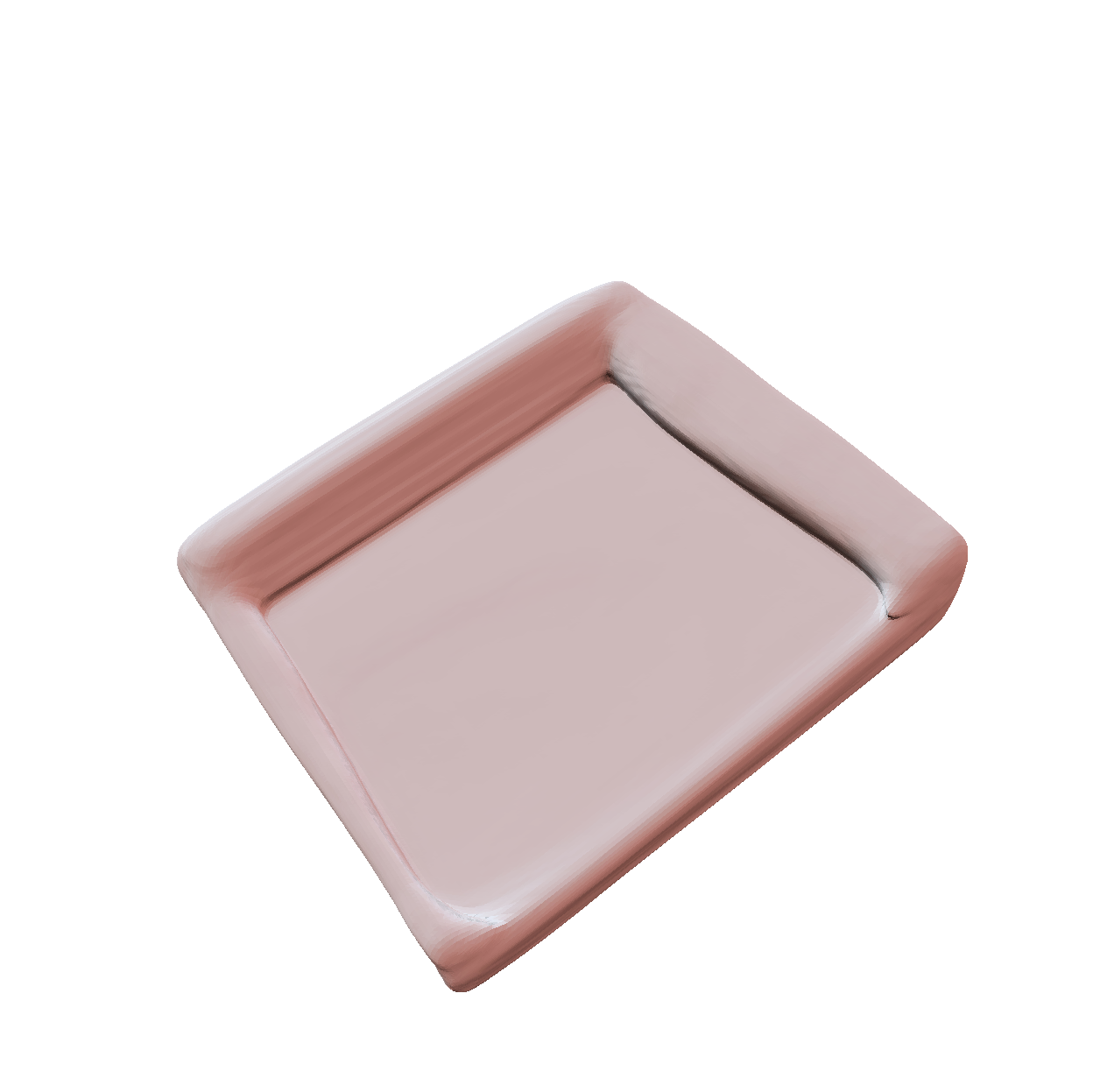}
	\end{subfigure} 
	~
	\begin{subfigure}{27.5mm}
		\centering
		\includegraphics[width=27.5mm]{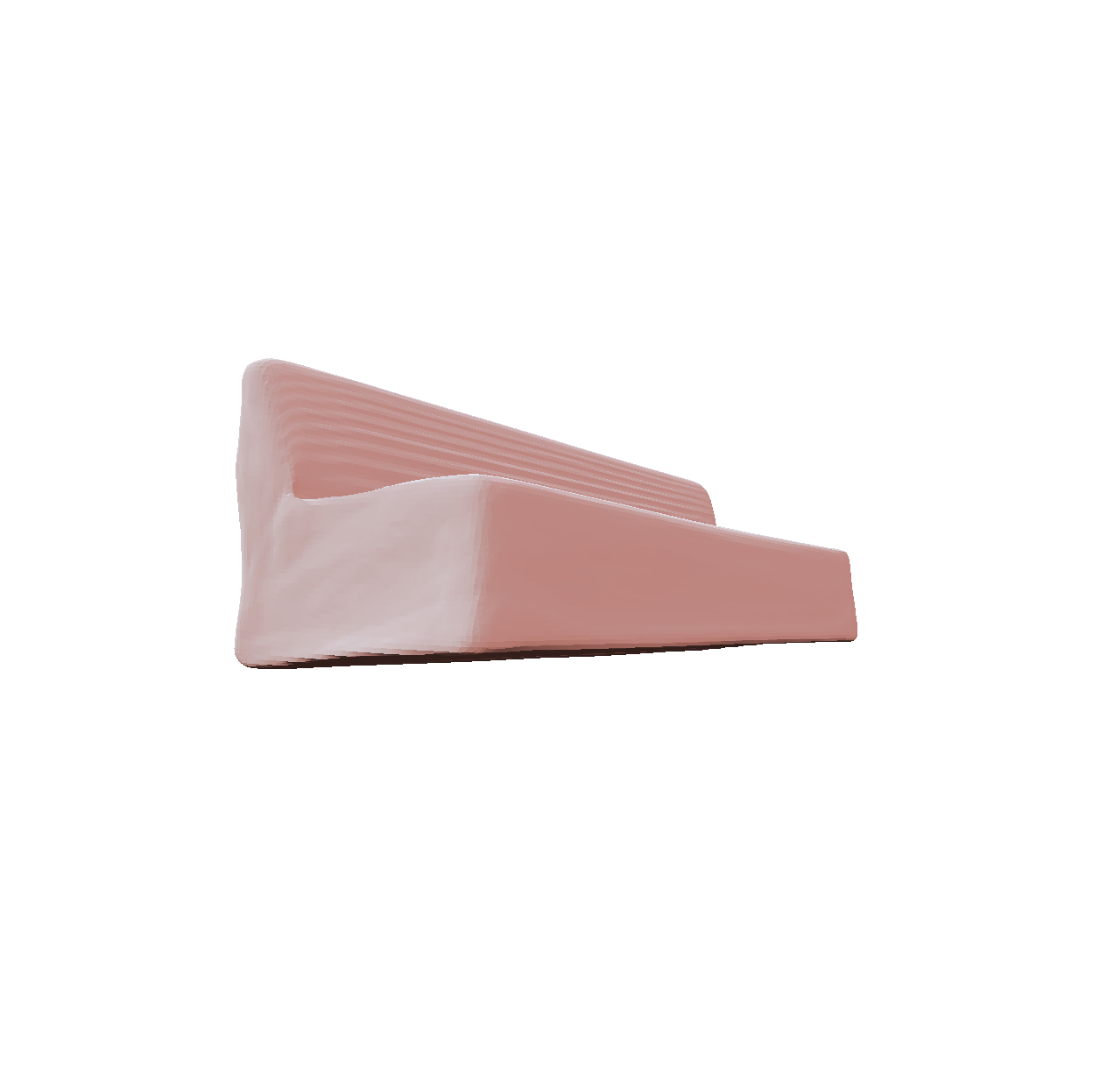}
	\end{subfigure} 
	~
	\begin{subfigure}{27.5mm}
		\centering
		\includegraphics[width=27.5mm]{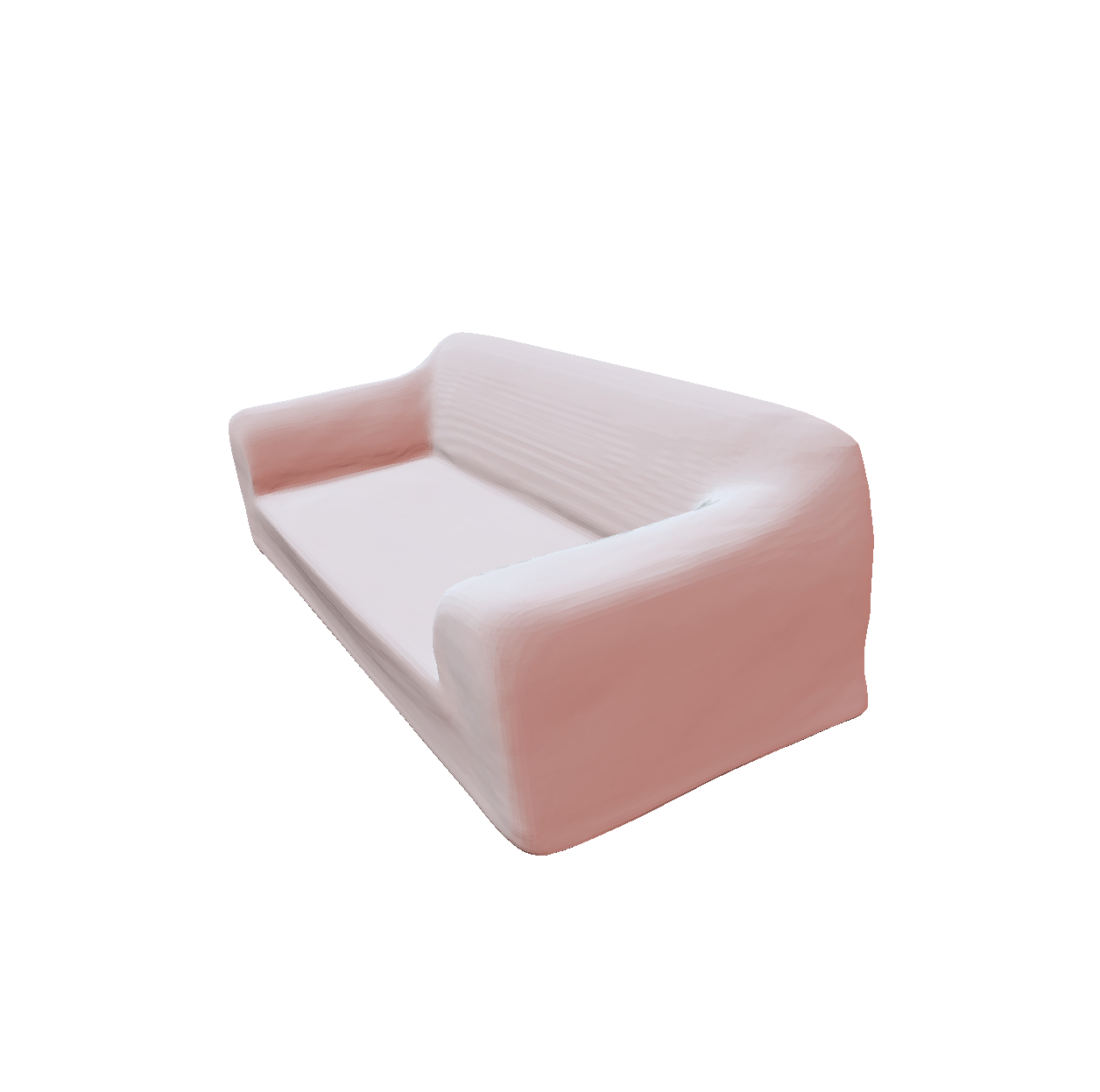}
	\end{subfigure}  \\
	\begin{subfigure}{27.5mm}
		\centering
		\includegraphics[width=27.5mm]{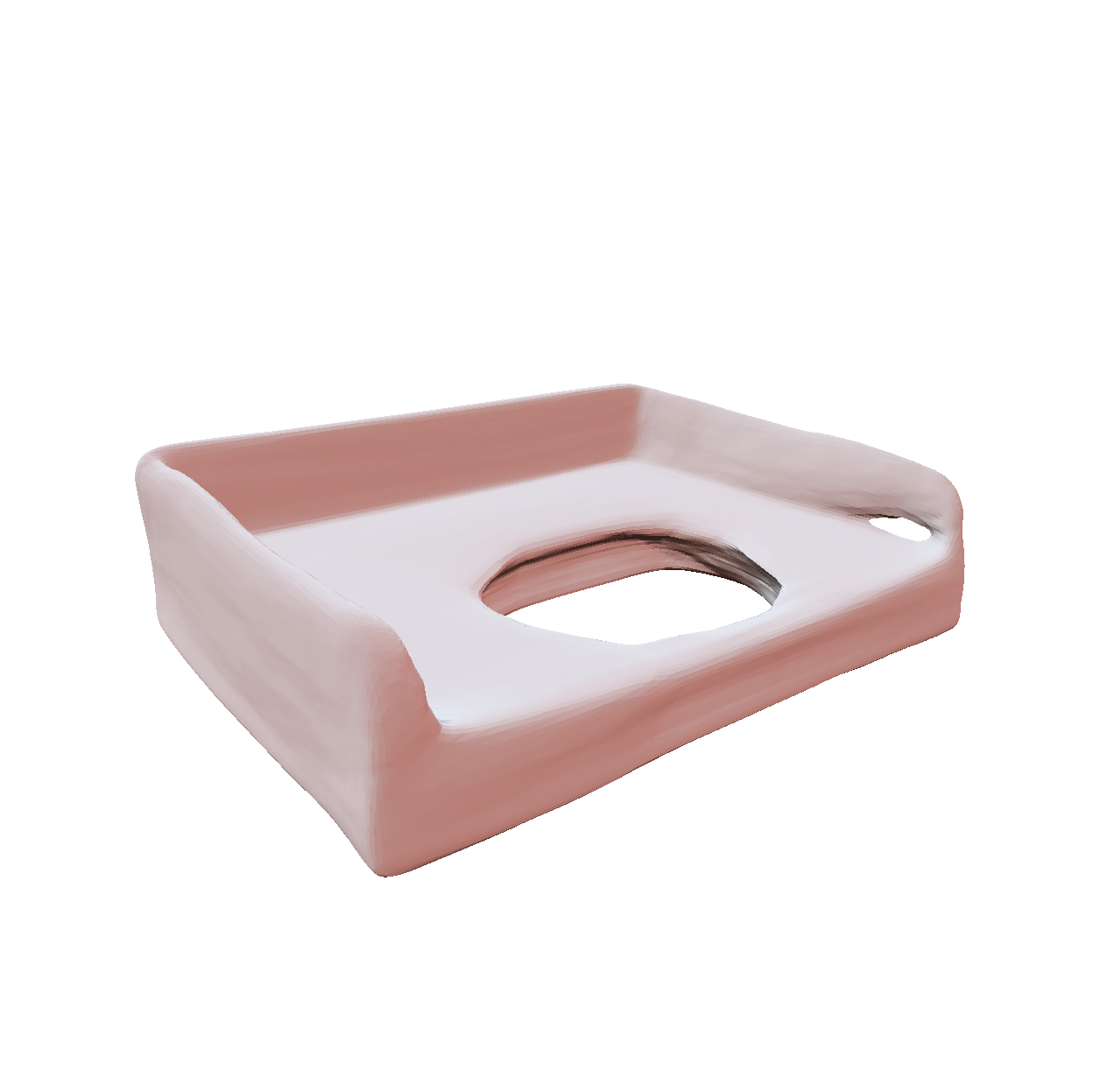}
	\end{subfigure}
	~
	\begin{subfigure}{27.5mm}
		\centering
		\includegraphics[width=27.5mm]{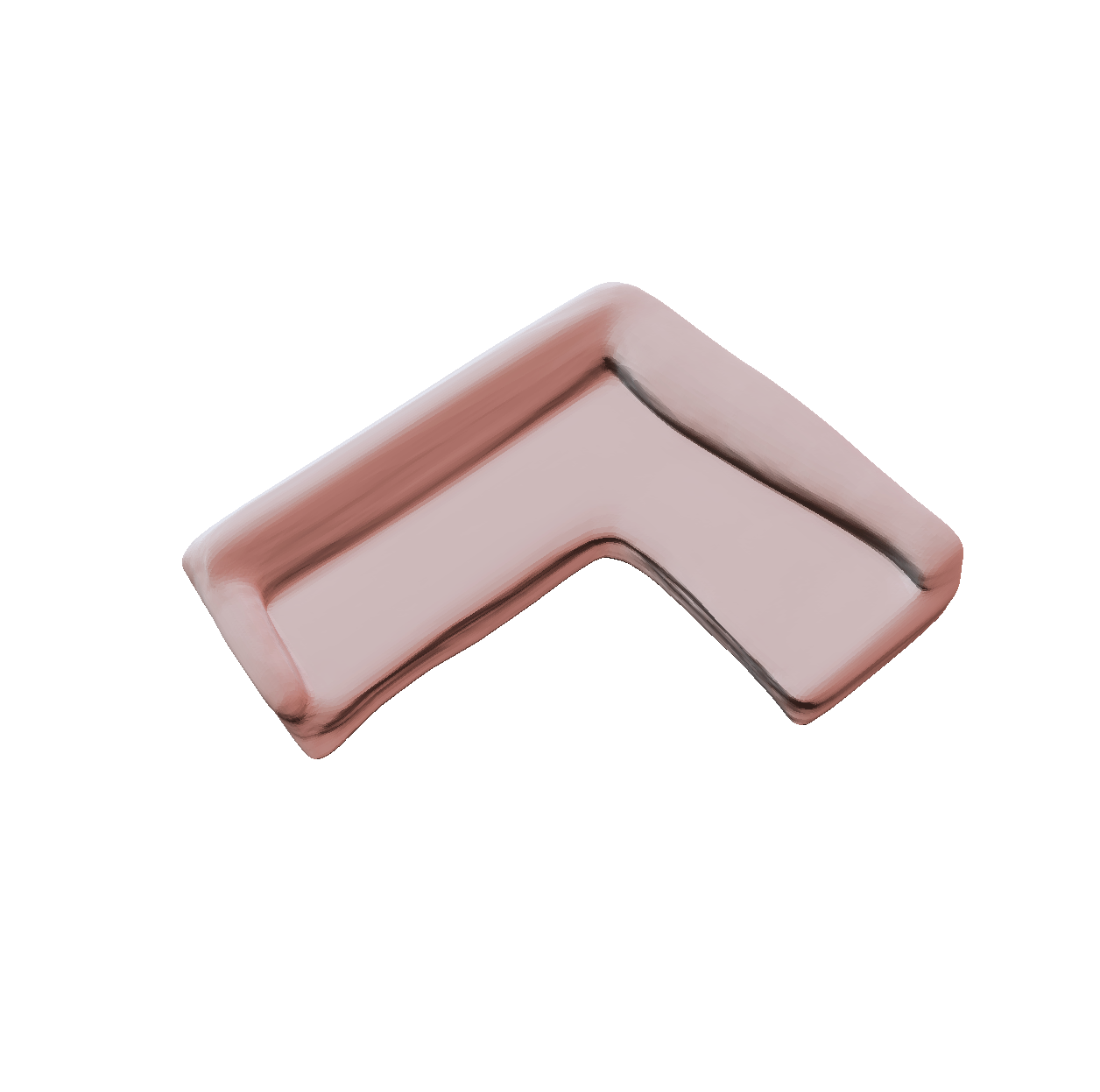}
	\end{subfigure} 
	~
	\begin{subfigure}{27.5mm}
		\centering
		\includegraphics[width=27.5mm]{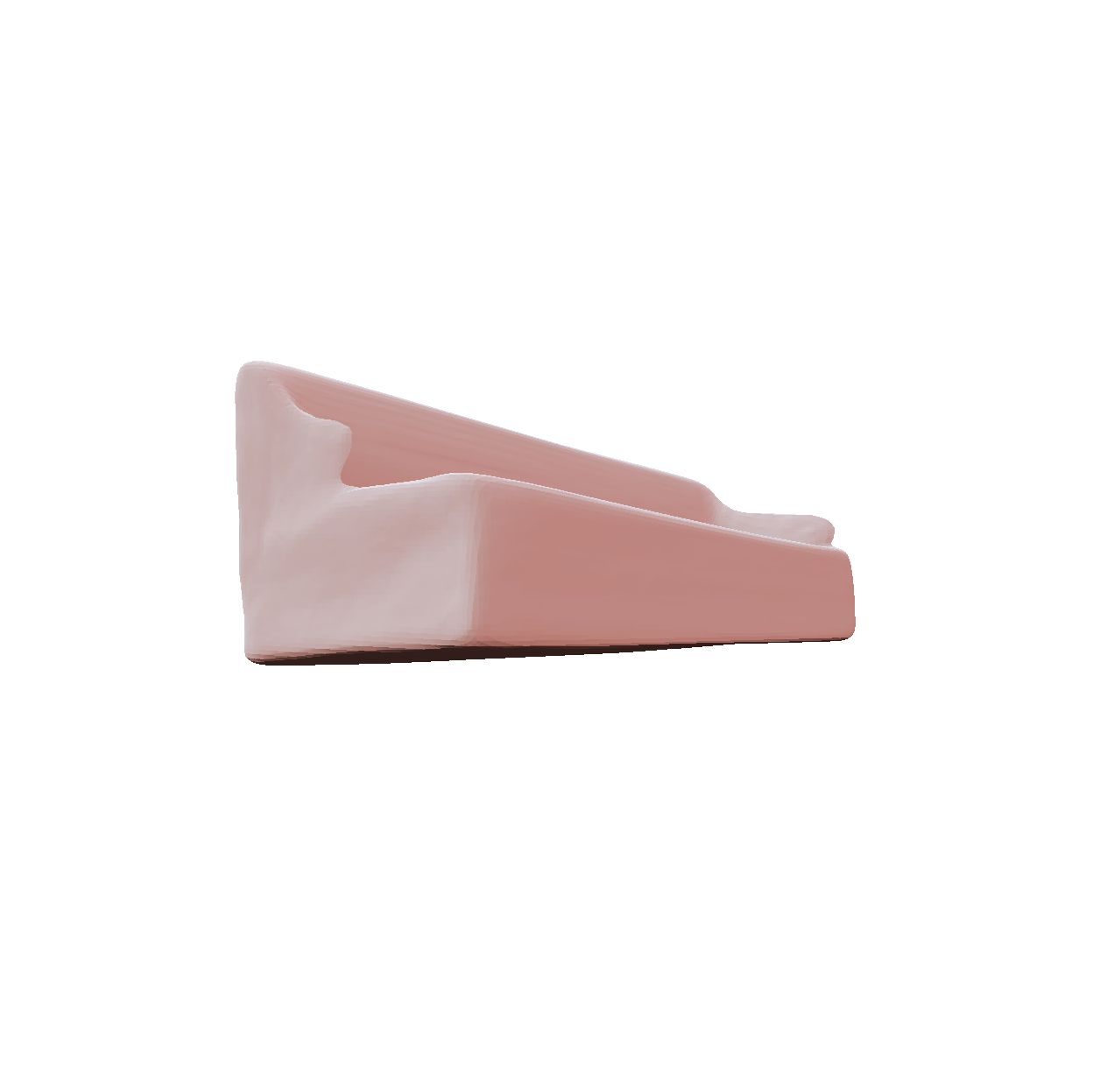}
	\end{subfigure} 
	~
	\begin{subfigure}{27.5mm}
		\centering
		\includegraphics[width=27.5mm]{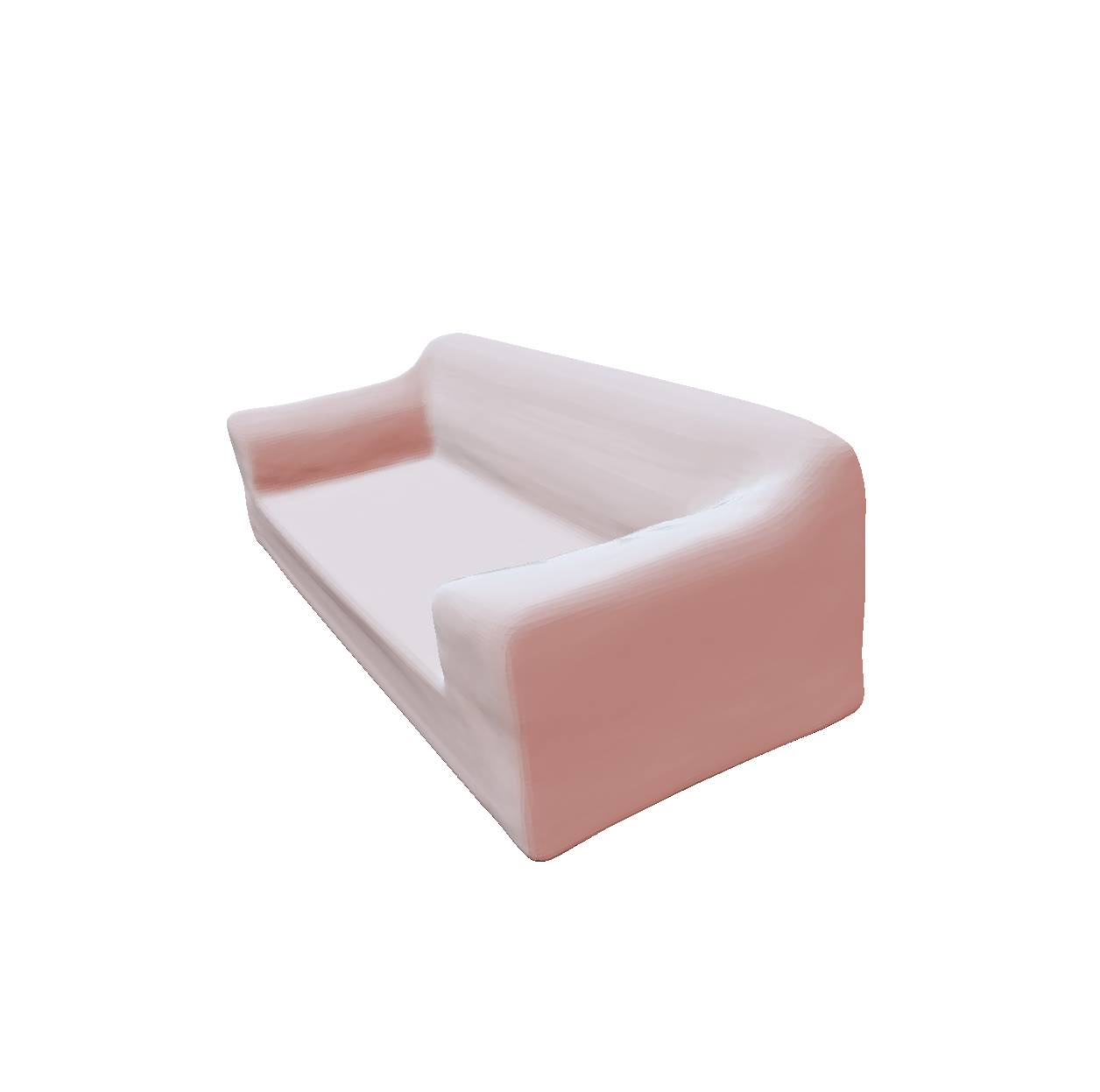}
	\end{subfigure} \\
	\begin{subfigure}{27.5mm}
		\centering
		\includegraphics[width=27.5mm]{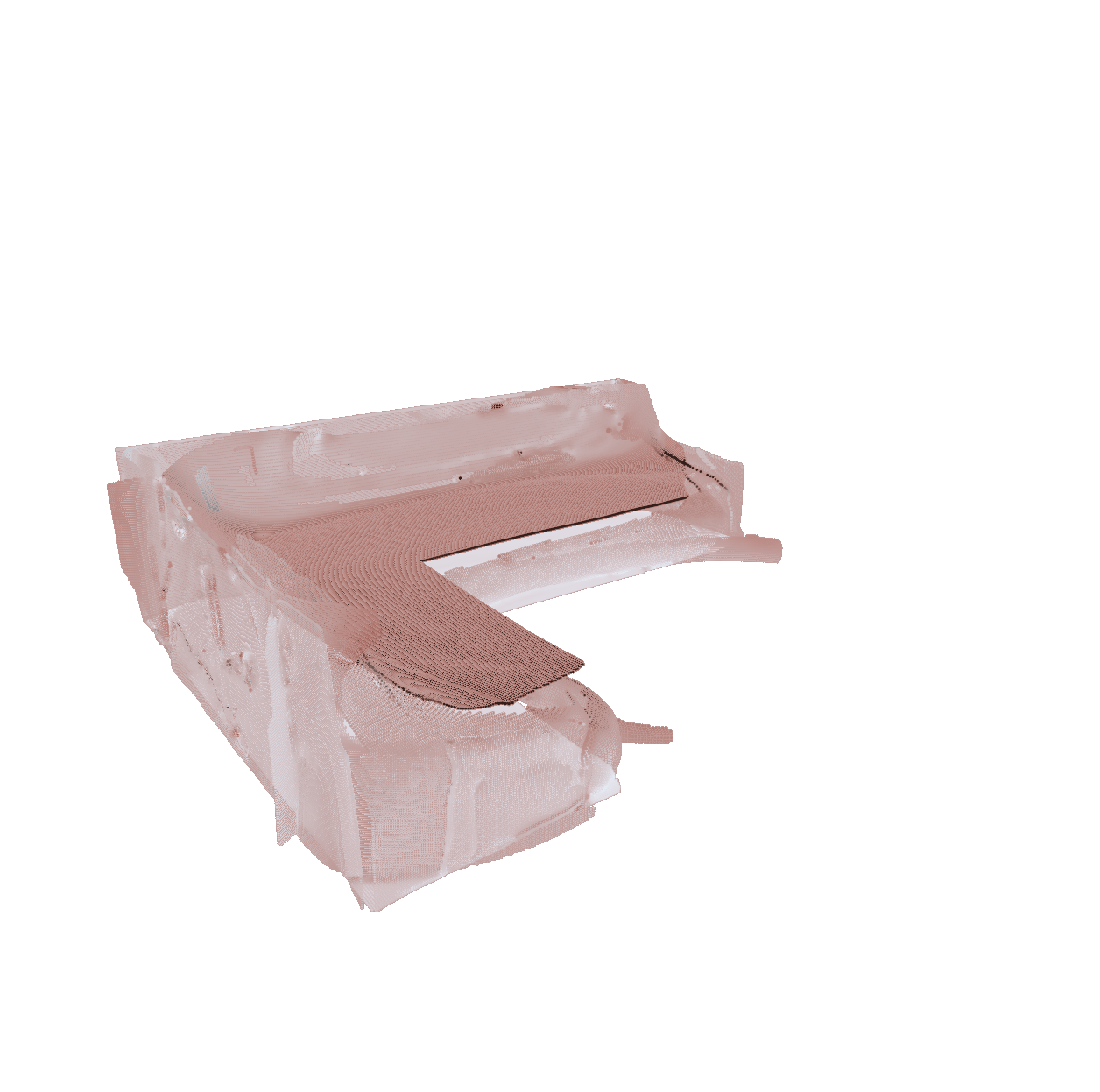}
	\end{subfigure}
	~
	\begin{subfigure}{27.5mm}
		\centering
		\includegraphics[width=27.5mm]{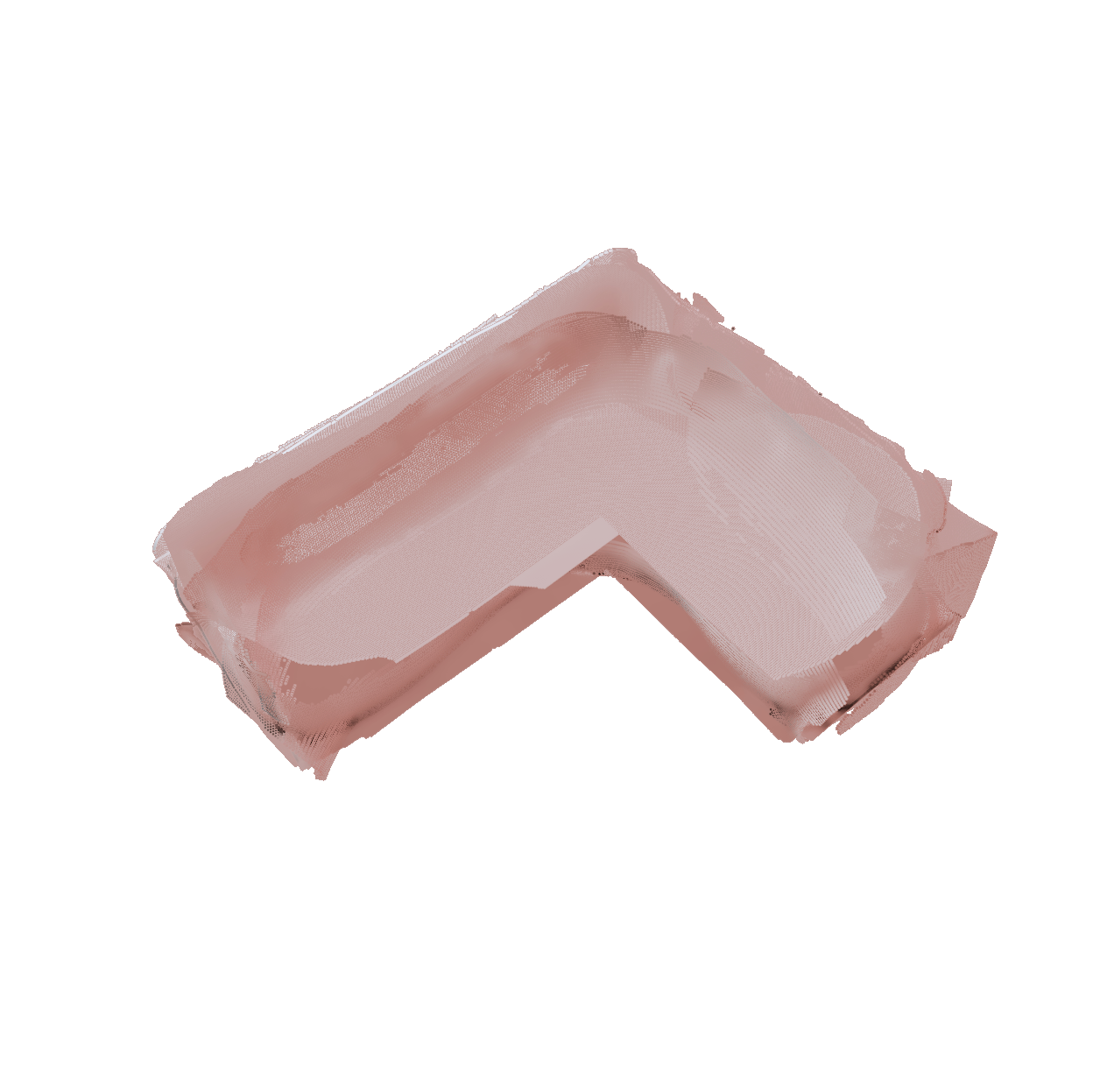}
	\end{subfigure} 
	~
	\begin{subfigure}{27.5mm}
		\centering
		\includegraphics[width=27.5mm]{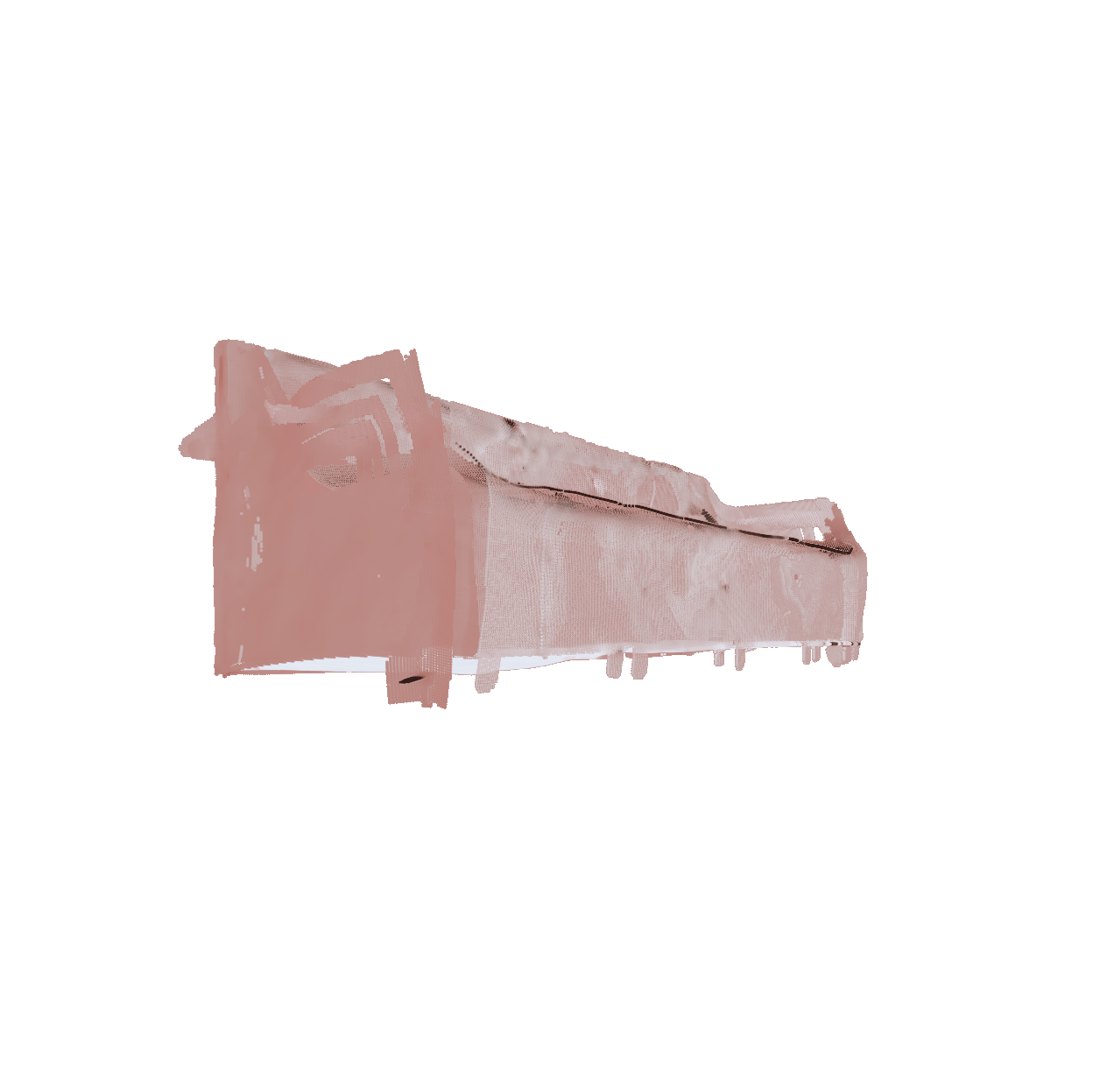}
	\end{subfigure} 
	~
	\begin{subfigure}{27.5mm}
		\centering
		\includegraphics[width=27.5mm]{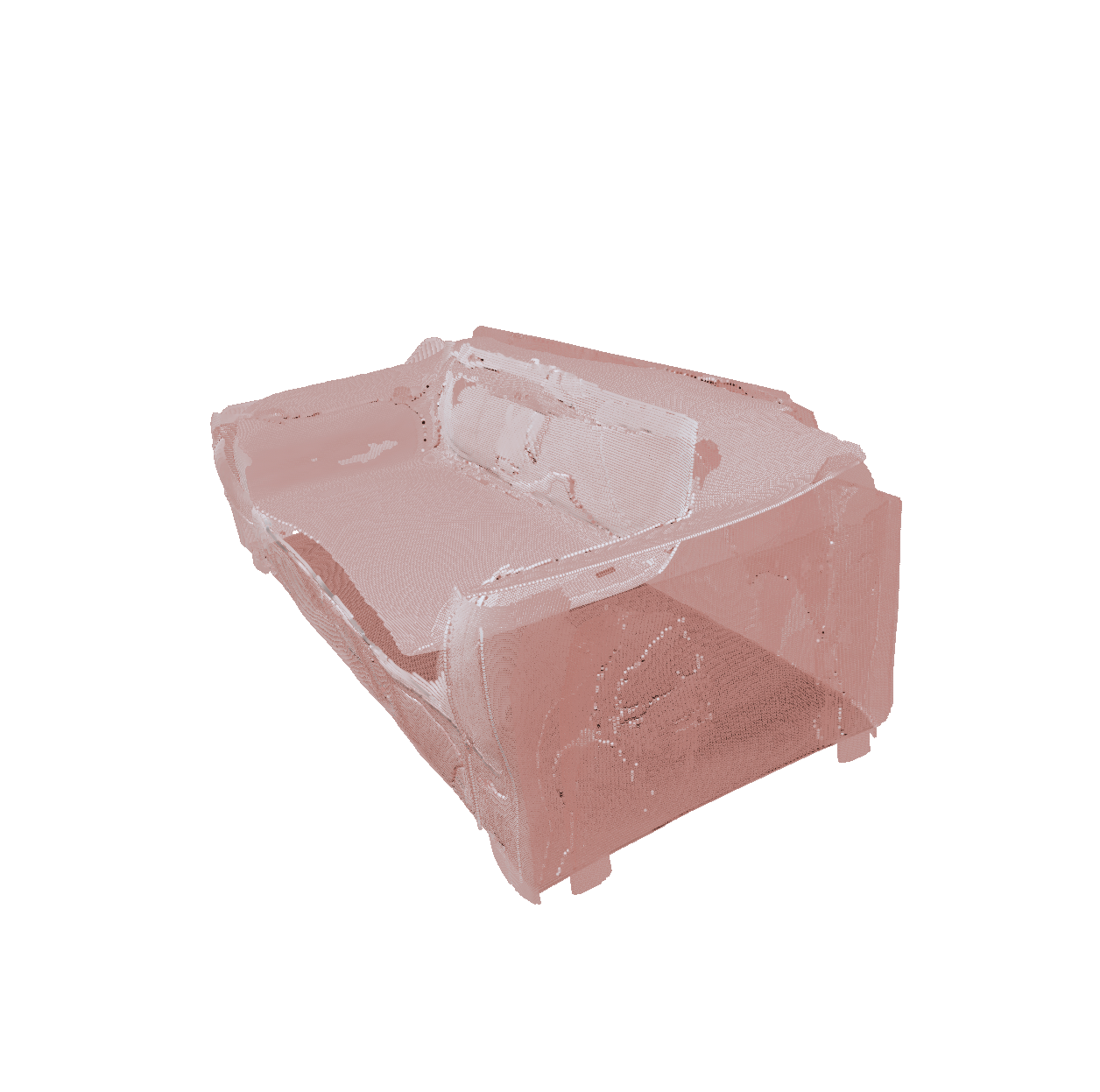}
	\end{subfigure} \\
	\caption{{\bf More Generative Results from Different Categories.} Results from Rows 1 to 4 correspond to {\bf Reference}, {\bf OF}, {\bf SDF}, {\bf PRIF - Mesh}.}
\end{figure}

\begin{figure}[!ht]
    \centering
	\begin{subfigure}{27.5mm}
		\centering
		\includegraphics[width=27.5mm]{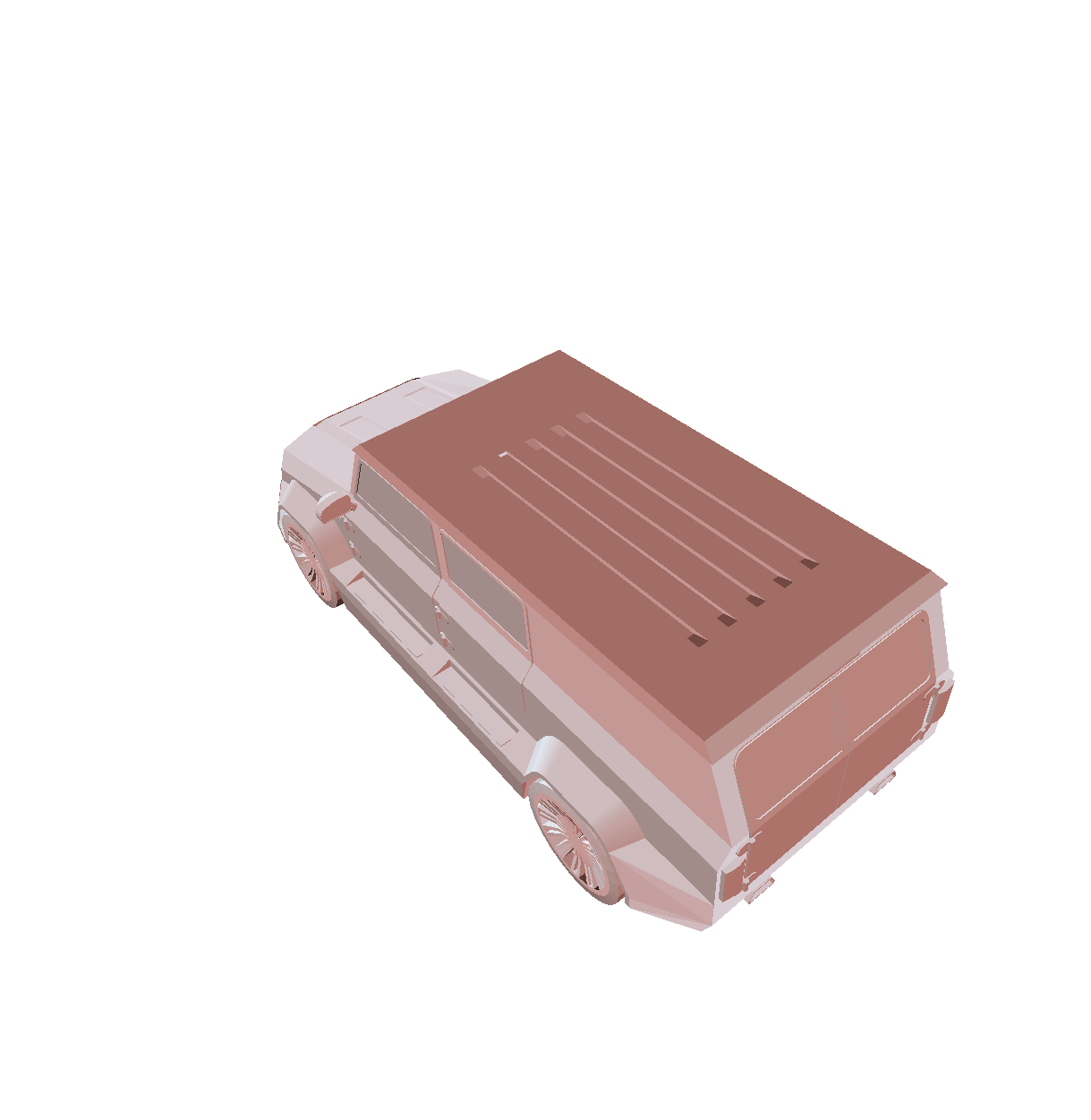}
	\end{subfigure}
	~
	\begin{subfigure}{27.5mm}
		\centering
		\includegraphics[width=27.5mm]{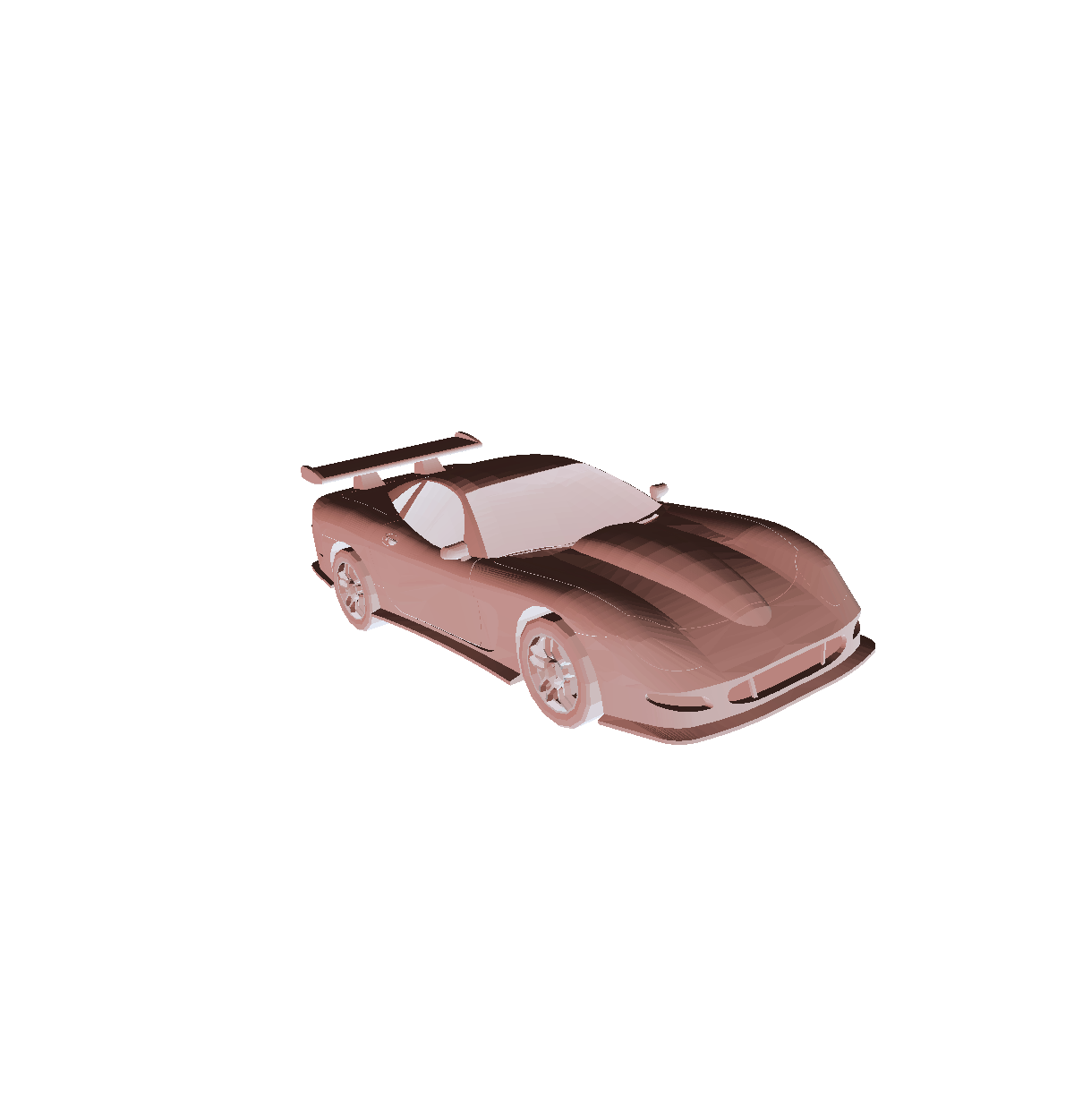}
	\end{subfigure} 
	~
	\begin{subfigure}{27.5mm}
		\centering
		\includegraphics[width=27.5mm]{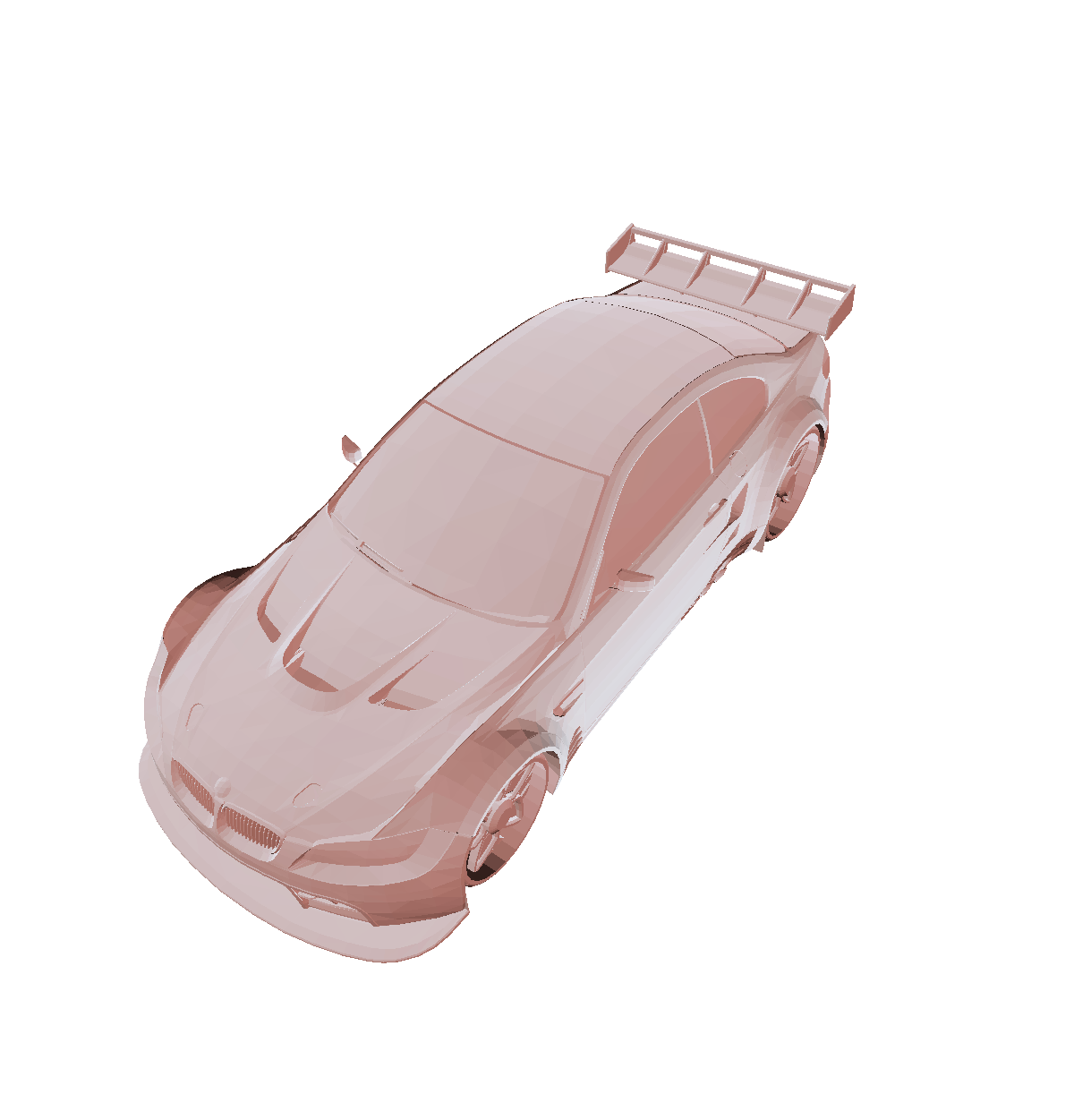}
	\end{subfigure}
	~
	\begin{subfigure}{27.5mm}
		\centering
		\includegraphics[width=27.5mm]{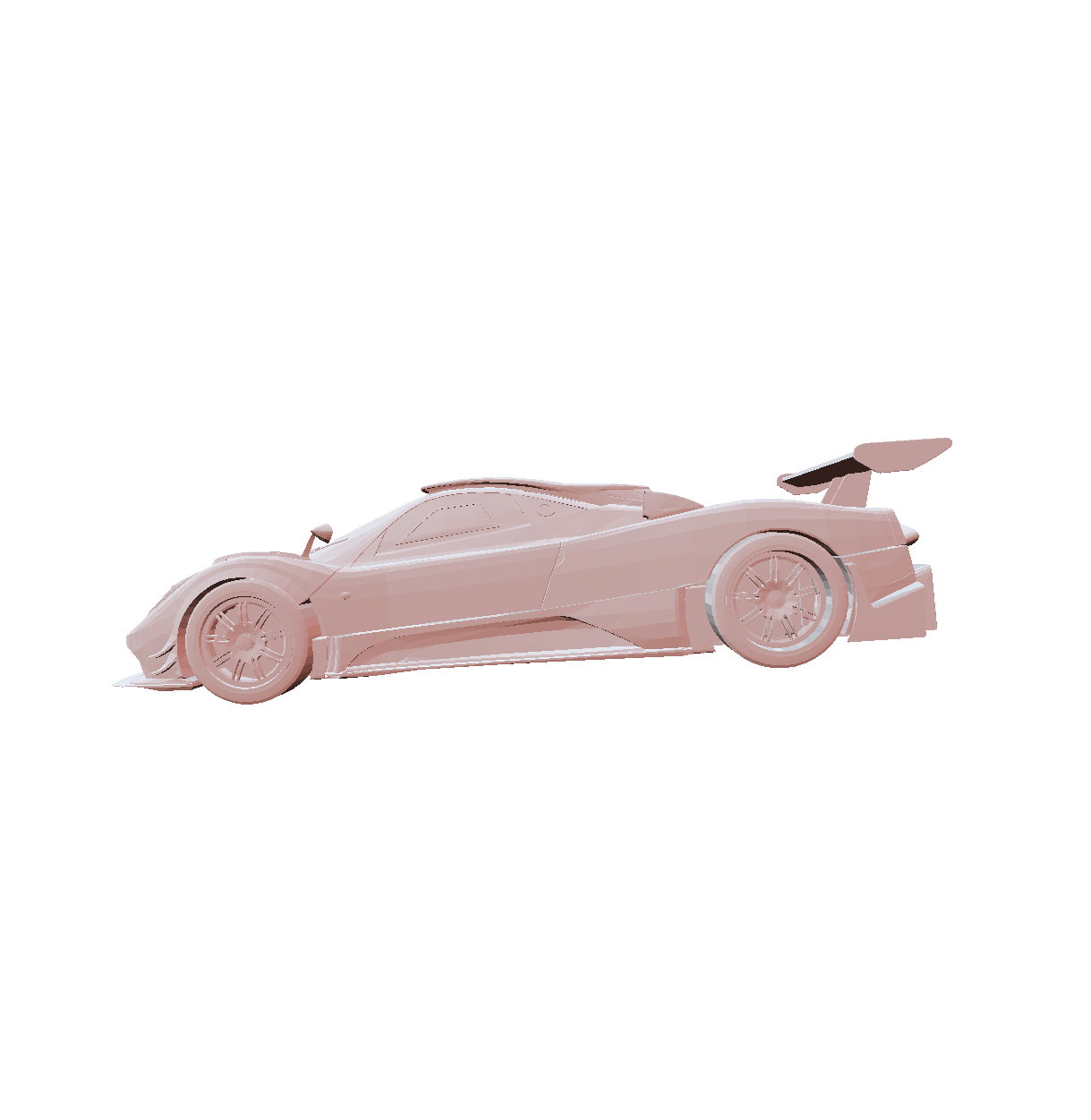}
	\end{subfigure}  \\
	\begin{subfigure}{27.5mm}
		\centering
		\includegraphics[width=27.5mm]{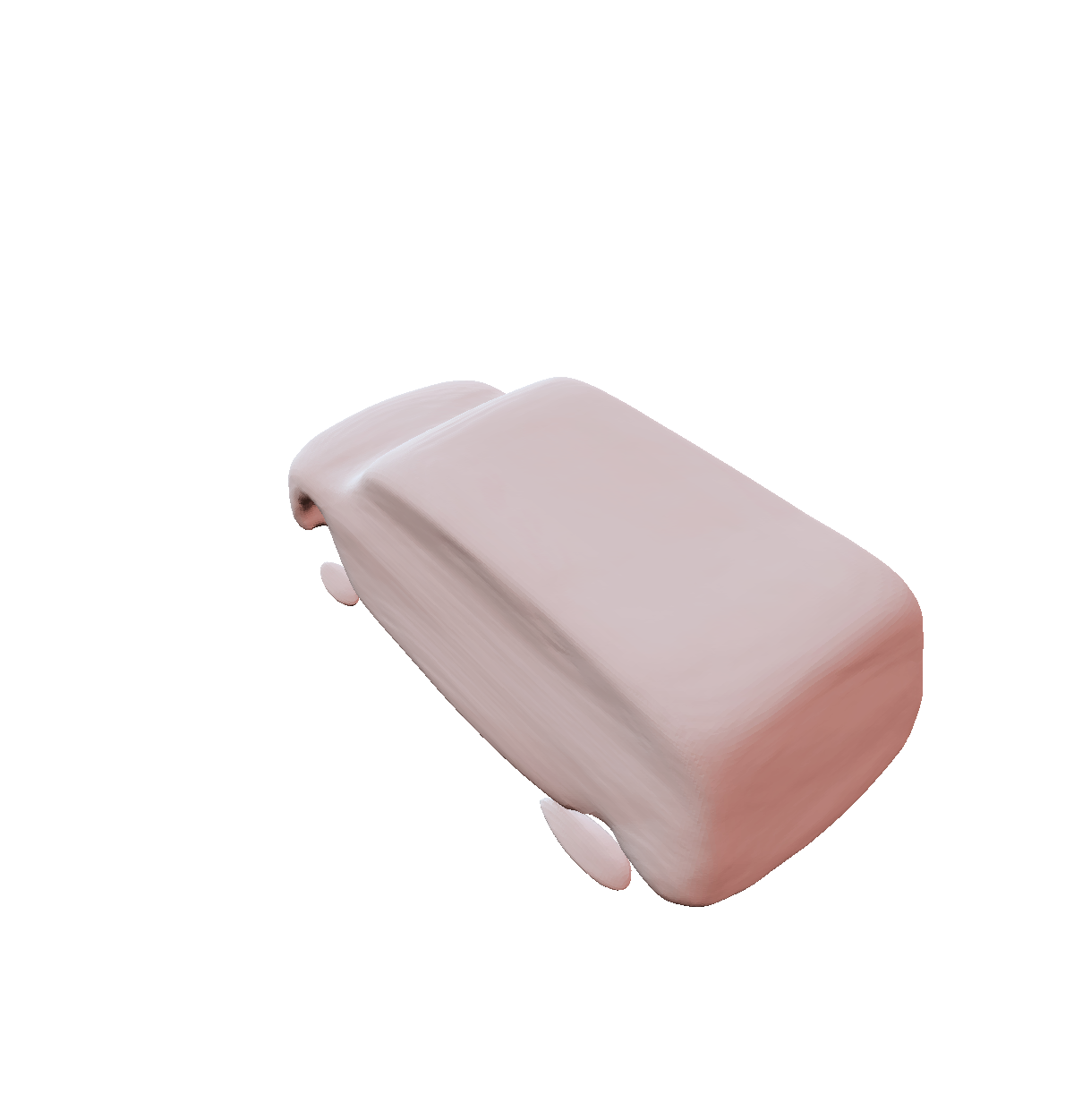}
	\end{subfigure}
	~
	\begin{subfigure}{27.5mm}
		\centering
		\includegraphics[width=27.5mm]{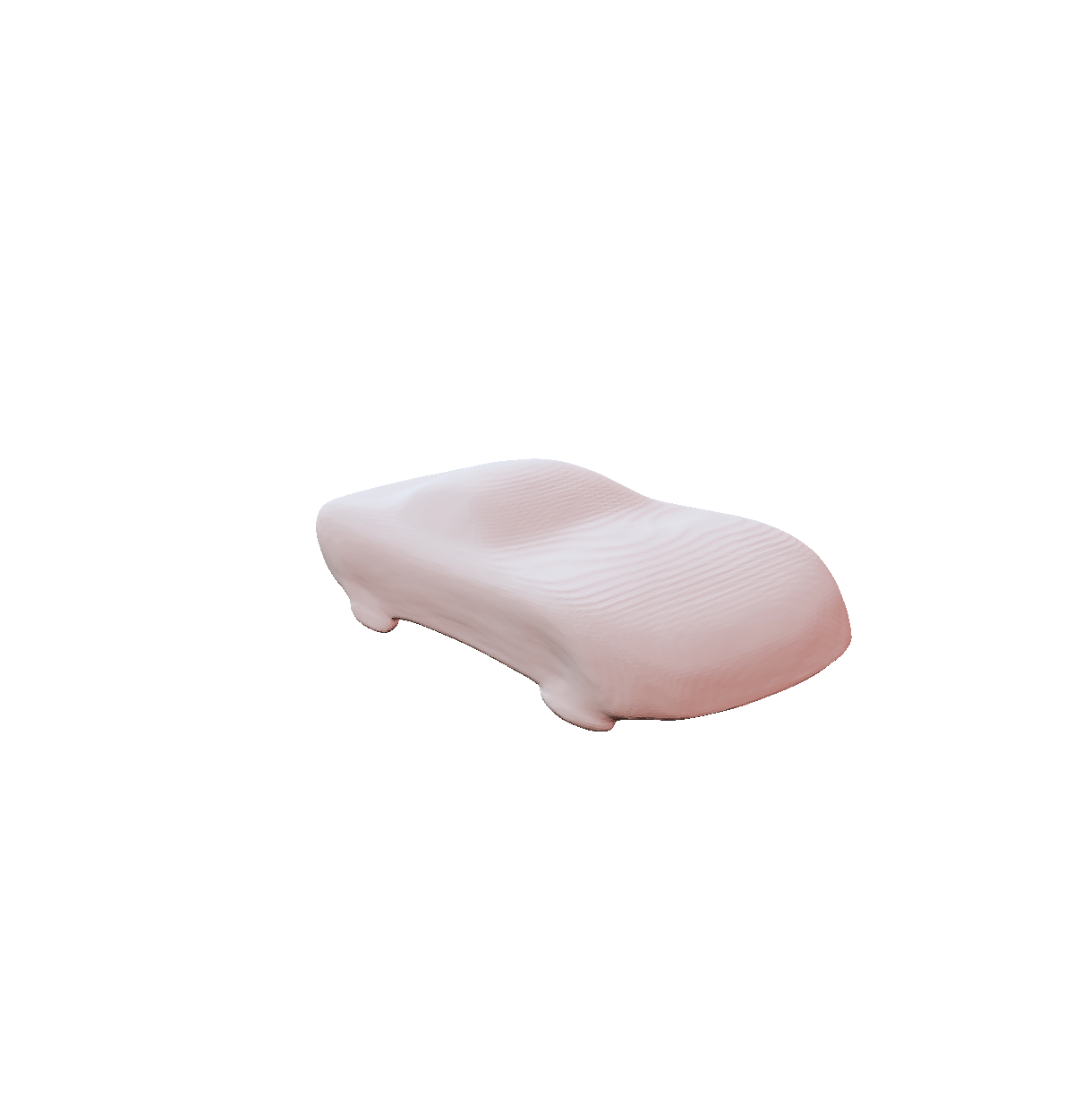}
	\end{subfigure} 
	~
	\begin{subfigure}{27.5mm}
		\centering
		\includegraphics[width=27.5mm]{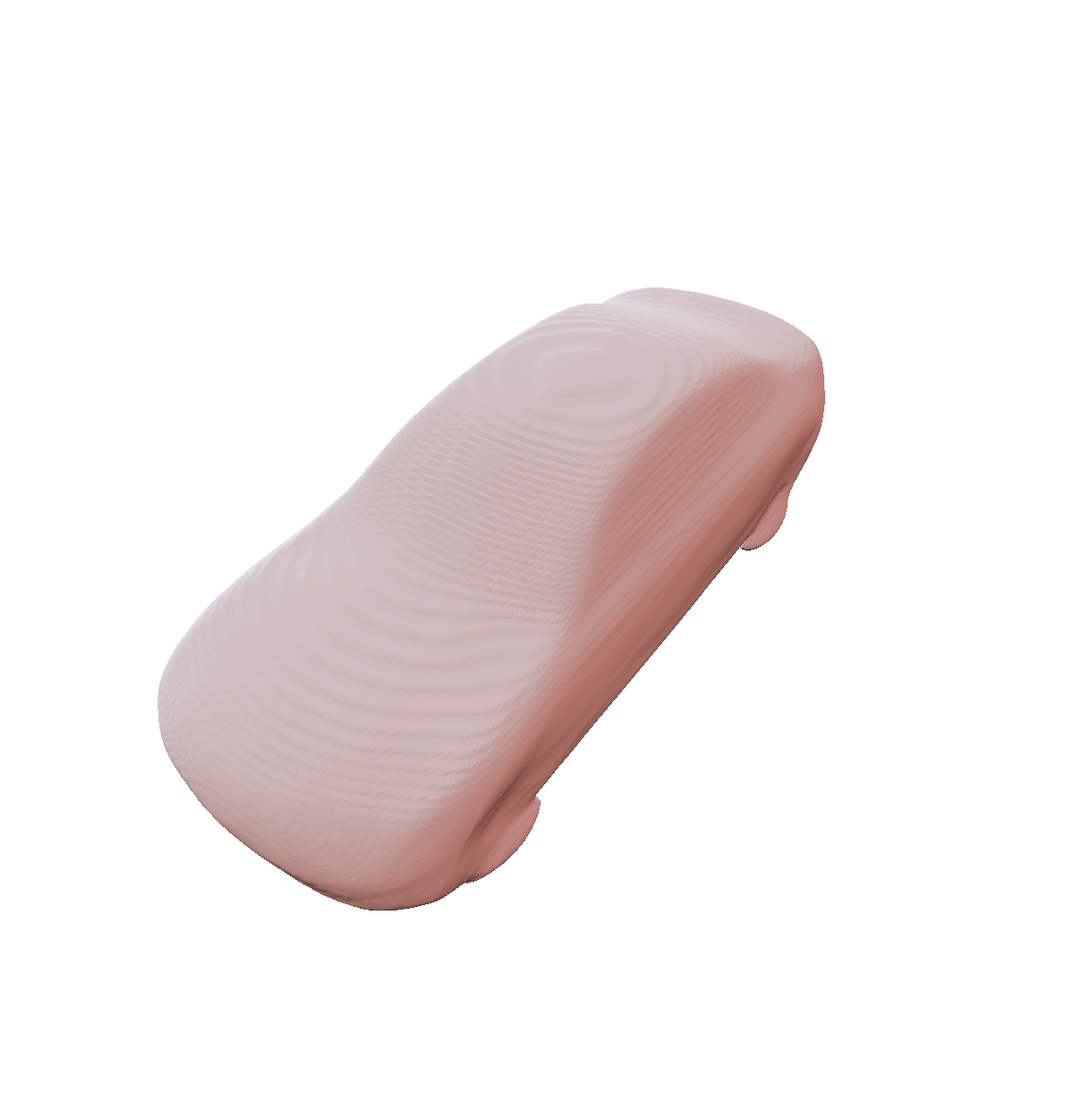}
	\end{subfigure} 
	~
	\begin{subfigure}{27.5mm}
		\centering
		\includegraphics[width=27.5mm]{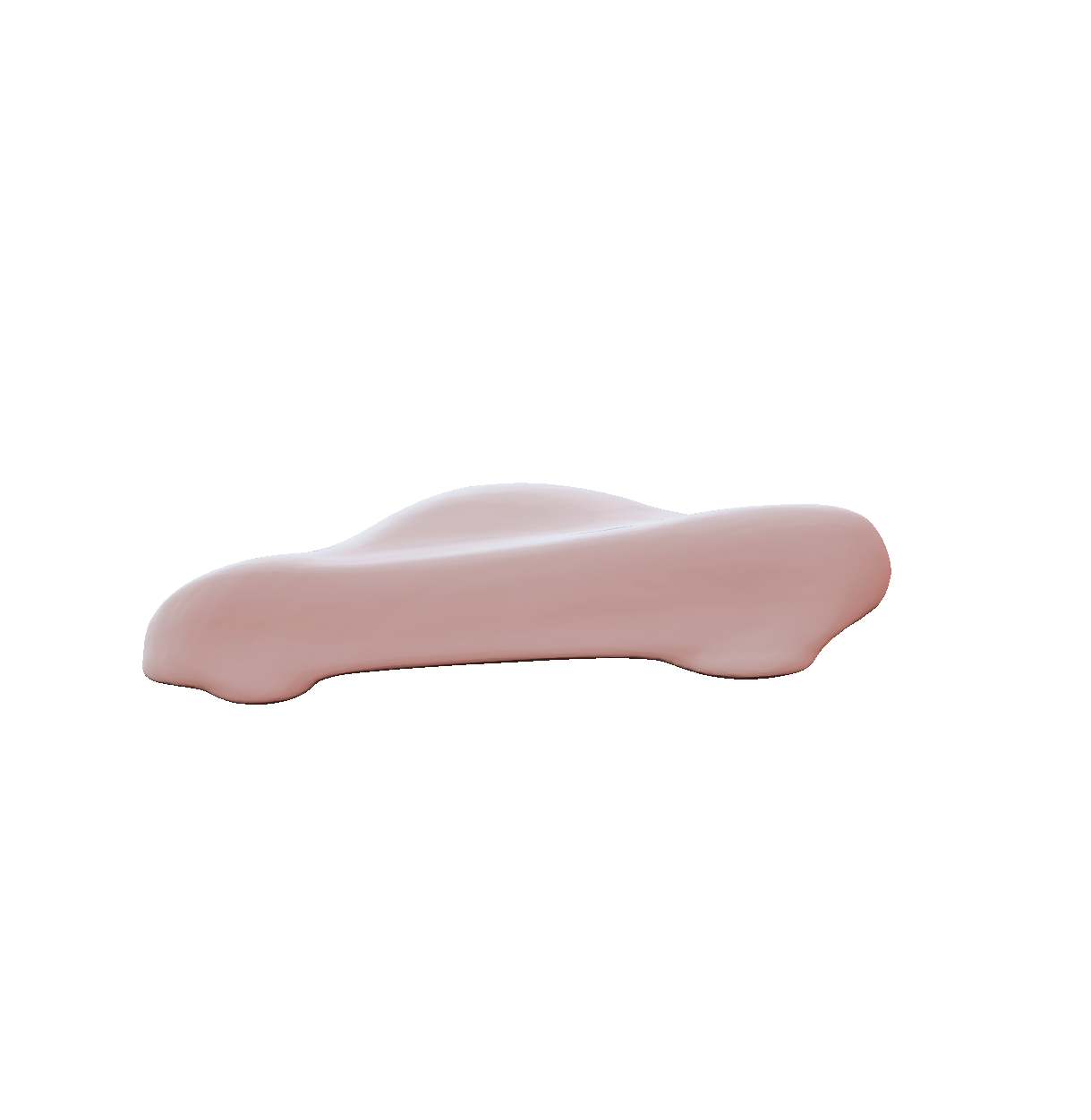}
	\end{subfigure}  \\
	\begin{subfigure}{27.5mm}
		\centering
		\includegraphics[width=27.5mm]{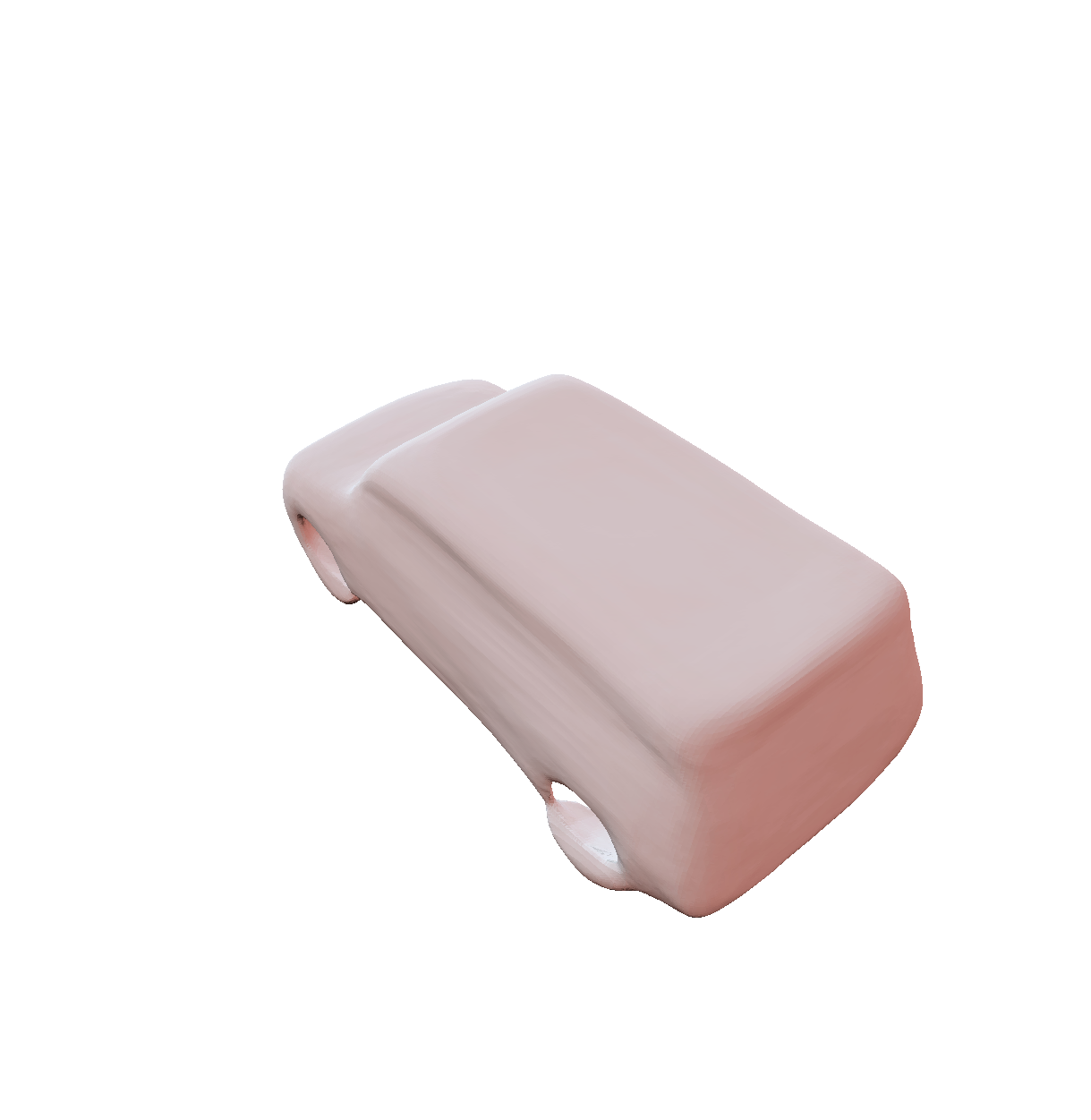}
	\end{subfigure}
	~
	\begin{subfigure}{27.5mm}
		\centering
		\includegraphics[width=27.5mm]{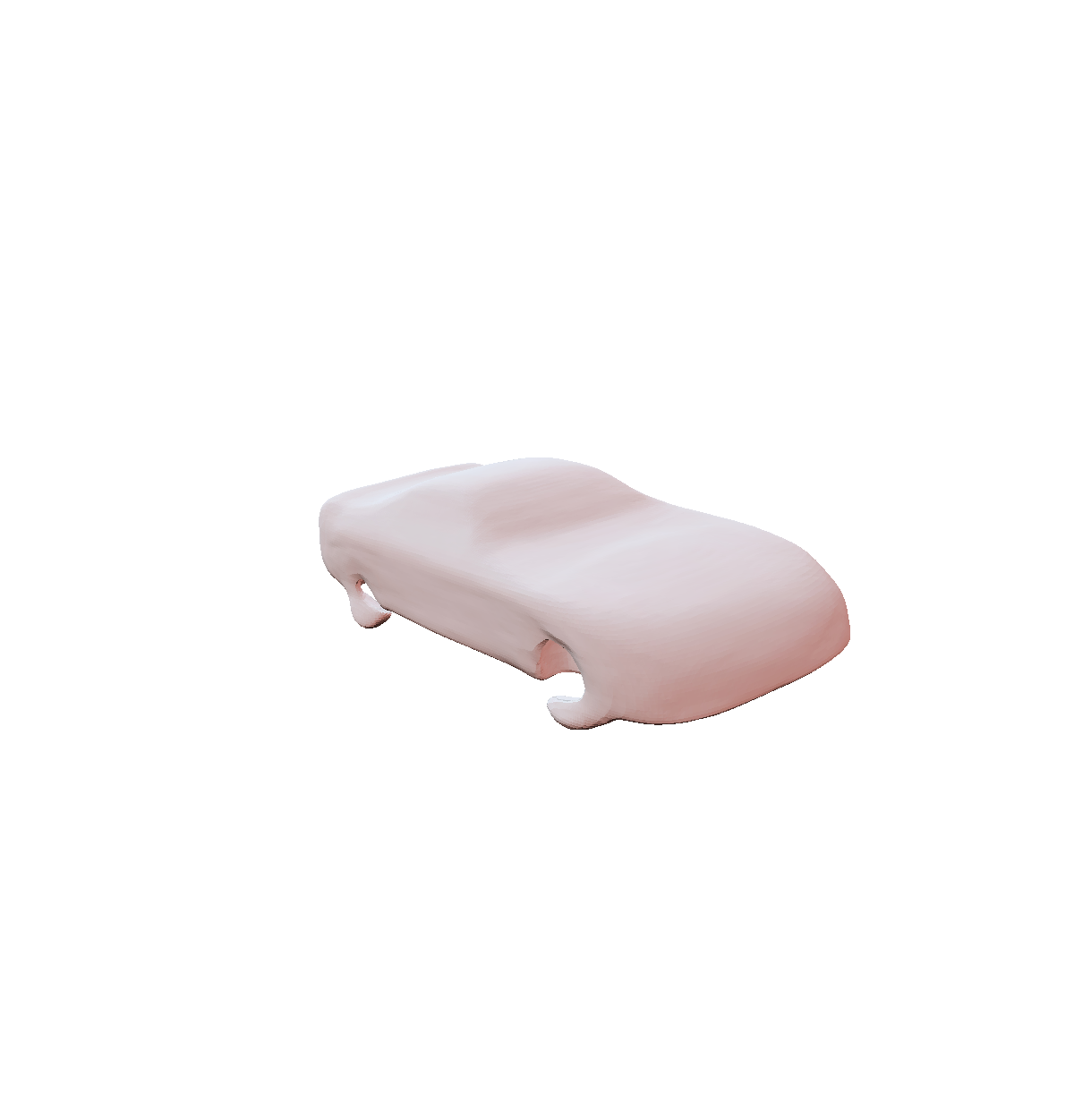}
	\end{subfigure} 
	~
	\begin{subfigure}{27.5mm}
		\centering
		\includegraphics[width=27.5mm]{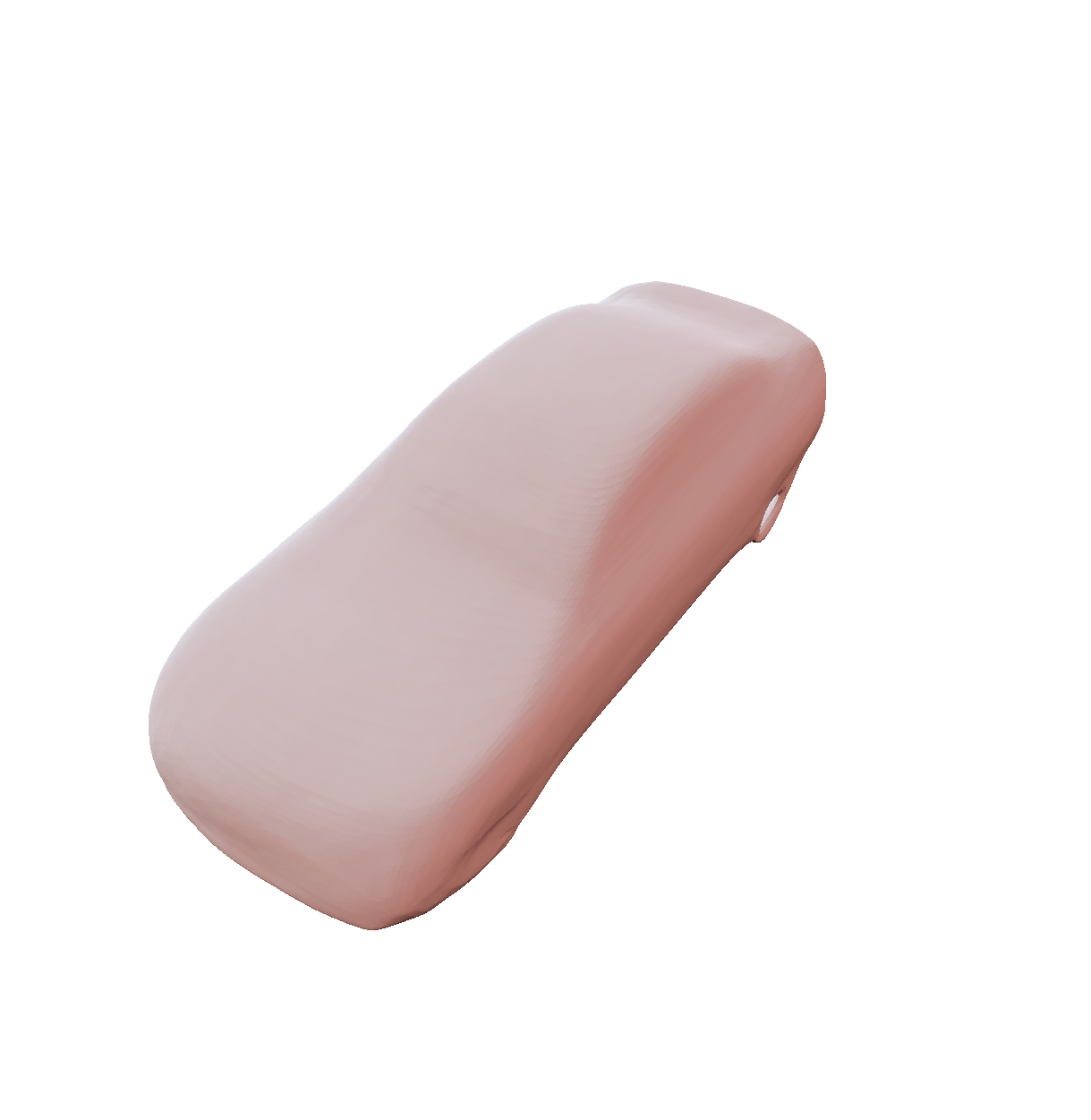}
	\end{subfigure} 
	~
	\begin{subfigure}{27.5mm}
		\centering
		\includegraphics[width=27.5mm]{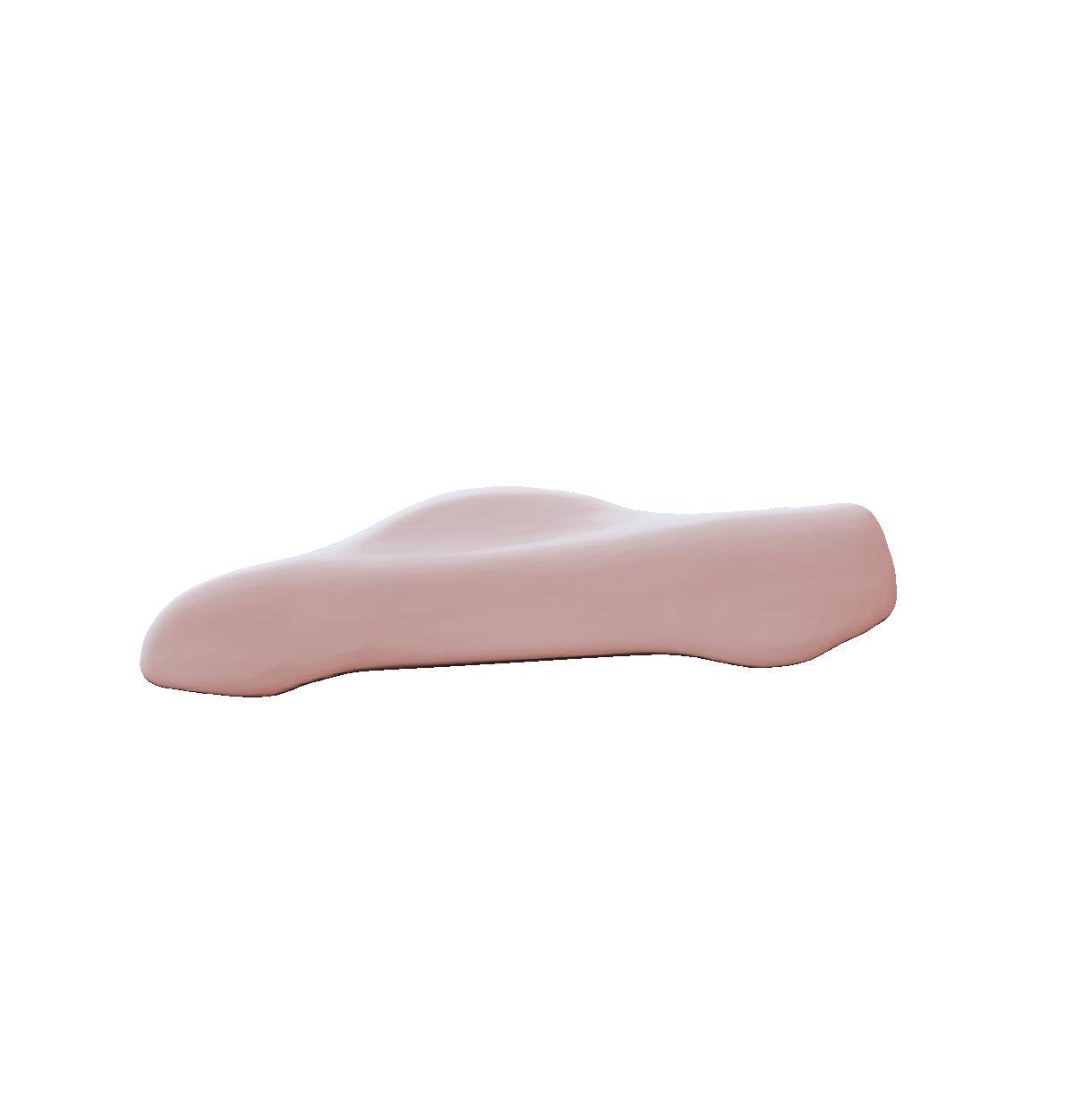}
	\end{subfigure} \\
	\begin{subfigure}{27.5mm}
		\centering
		\includegraphics[width=27.5mm]{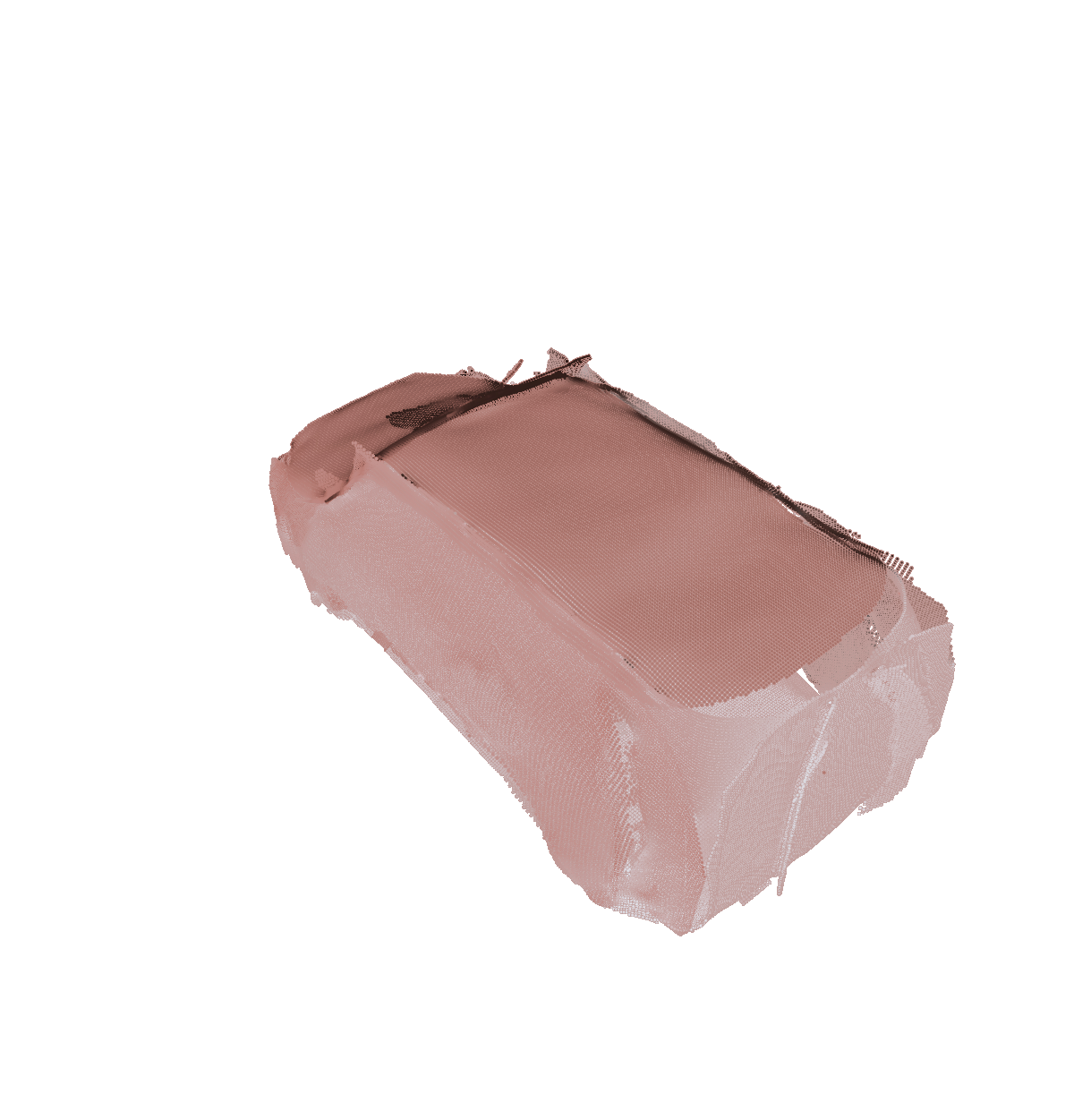}
	\end{subfigure}
	~
	\begin{subfigure}{27.5mm}
		\centering
		\includegraphics[width=27.5mm]{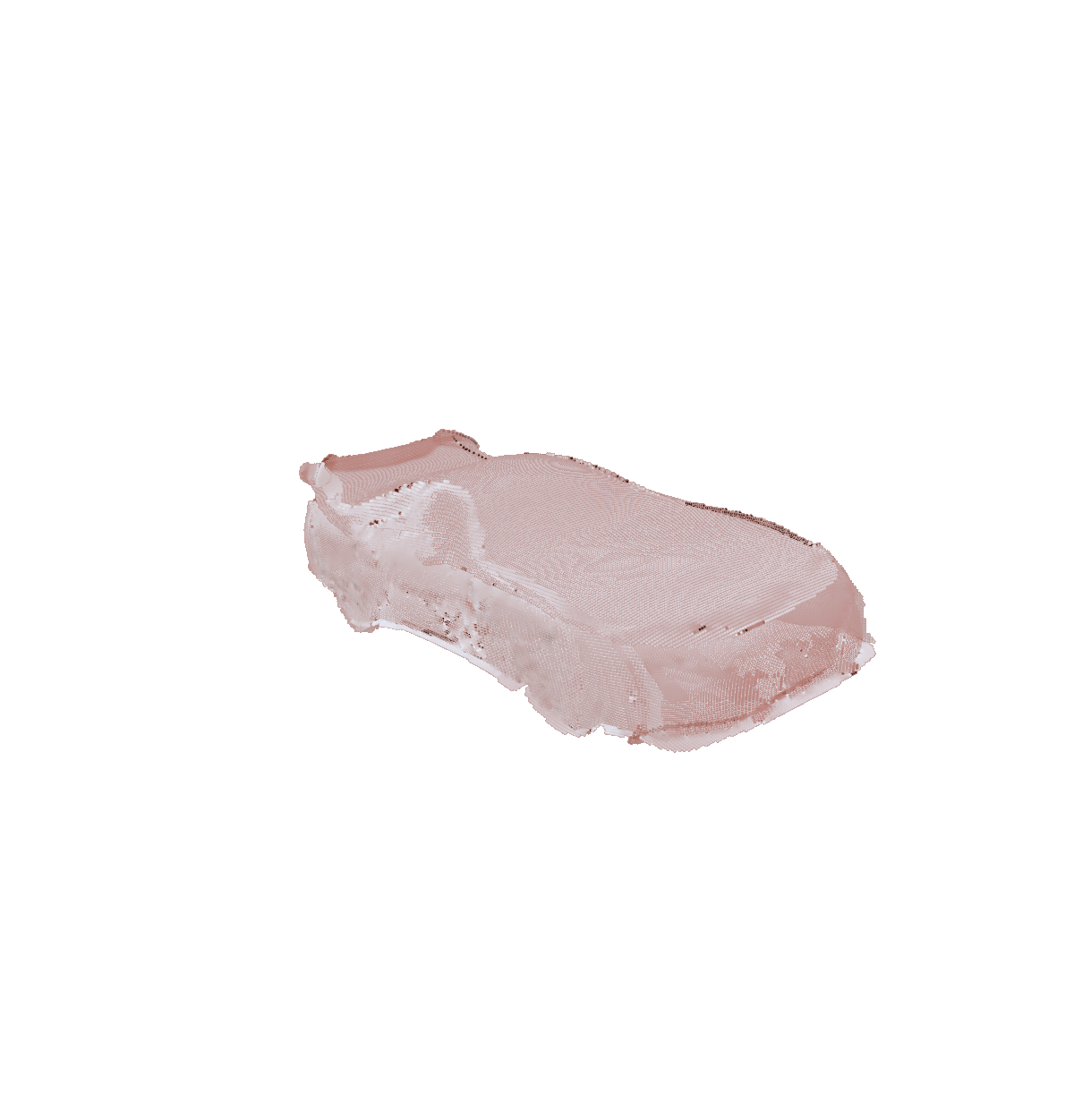}
	\end{subfigure} 
	~
	\begin{subfigure}{27.5mm}
		\centering
		\includegraphics[width=27.5mm]{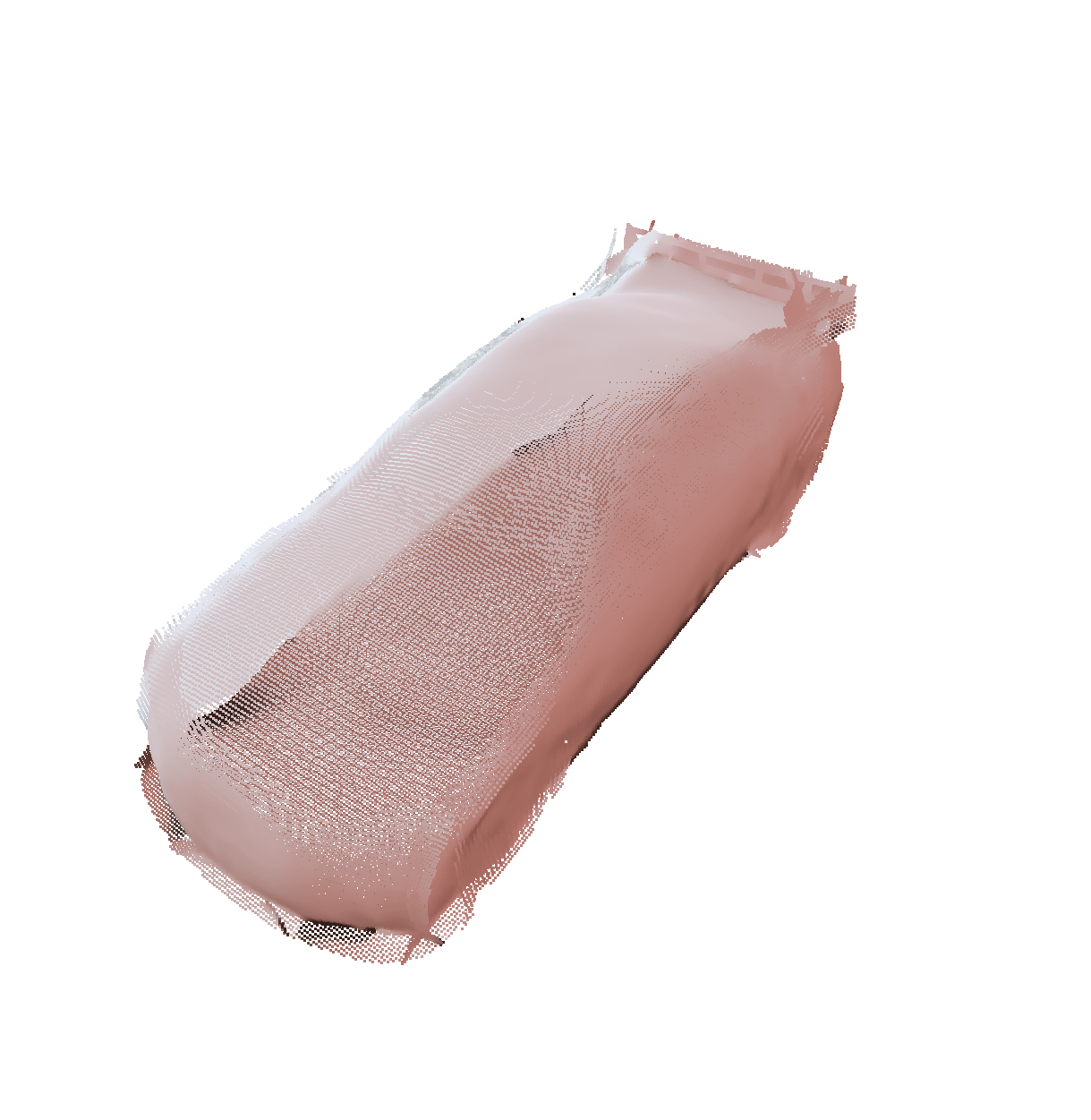}
	\end{subfigure} 
	~
	\begin{subfigure}{27.5mm}
		\centering
		\includegraphics[width=27.5mm]{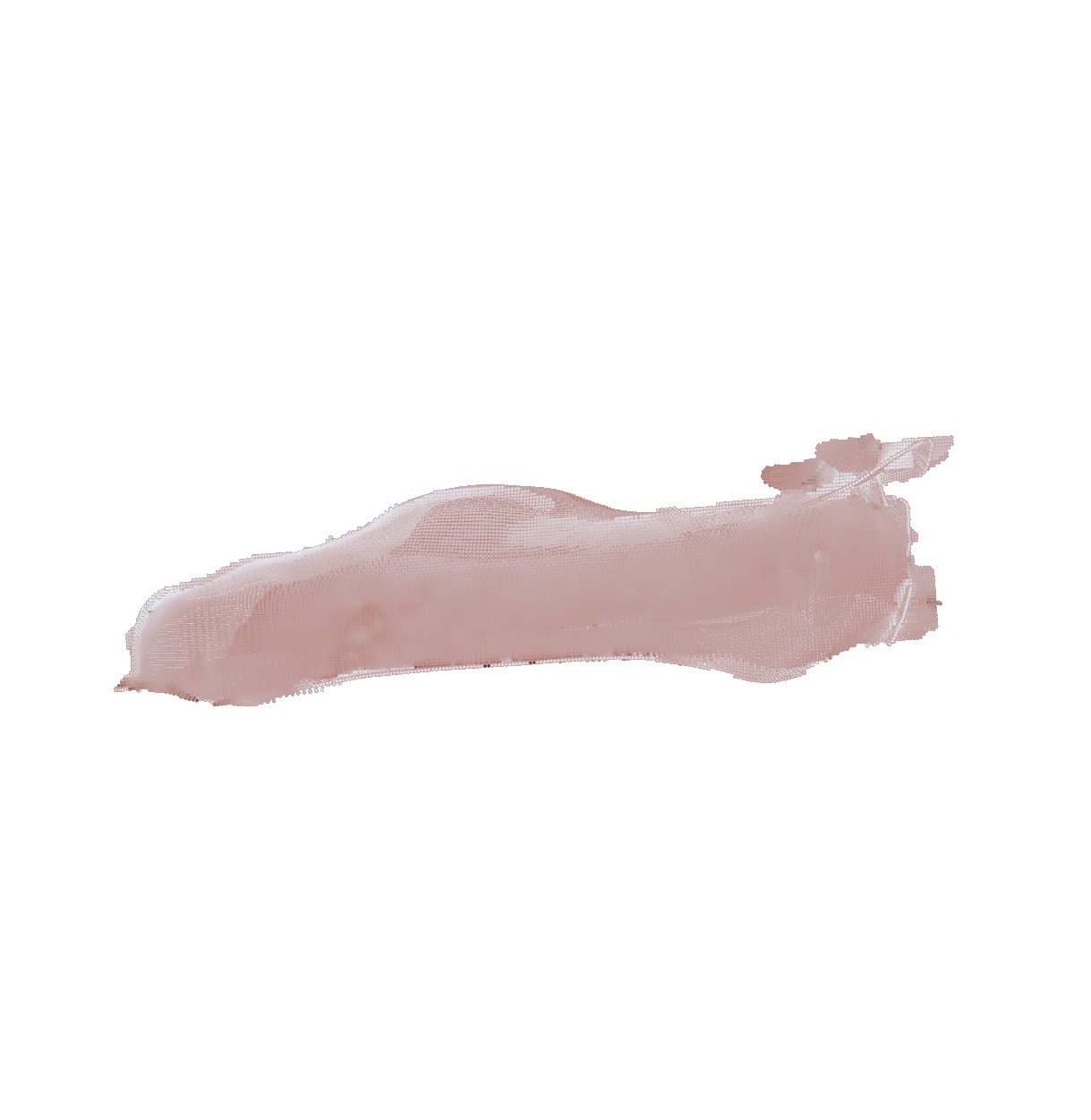}
	\end{subfigure} \\
	\caption{{\bf More Generative Results from Different Categories.} Results from Rows 1 to 4 correspond to {\bf Reference}, {\bf OF}, {\bf SDF}, {\bf PRIF - Mesh}.}
\end{figure}

\begin{figure}[!ht]
    \centering
	\begin{subfigure}{27.5mm}
		\centering
		\includegraphics[width=27.5mm]{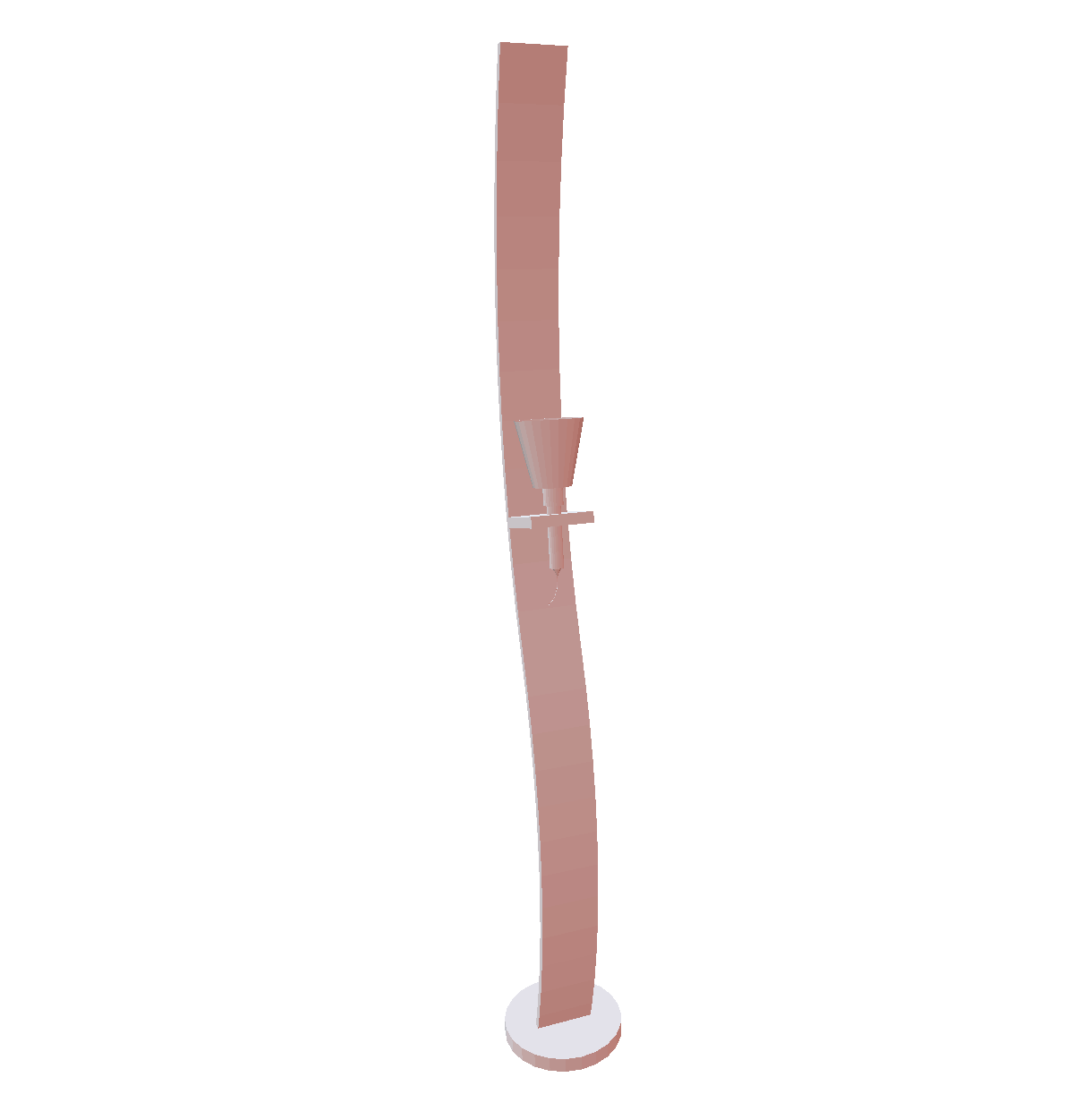}
	\end{subfigure}
	~
	\begin{subfigure}{27.5mm}
		\centering
		\includegraphics[width=27.5mm]{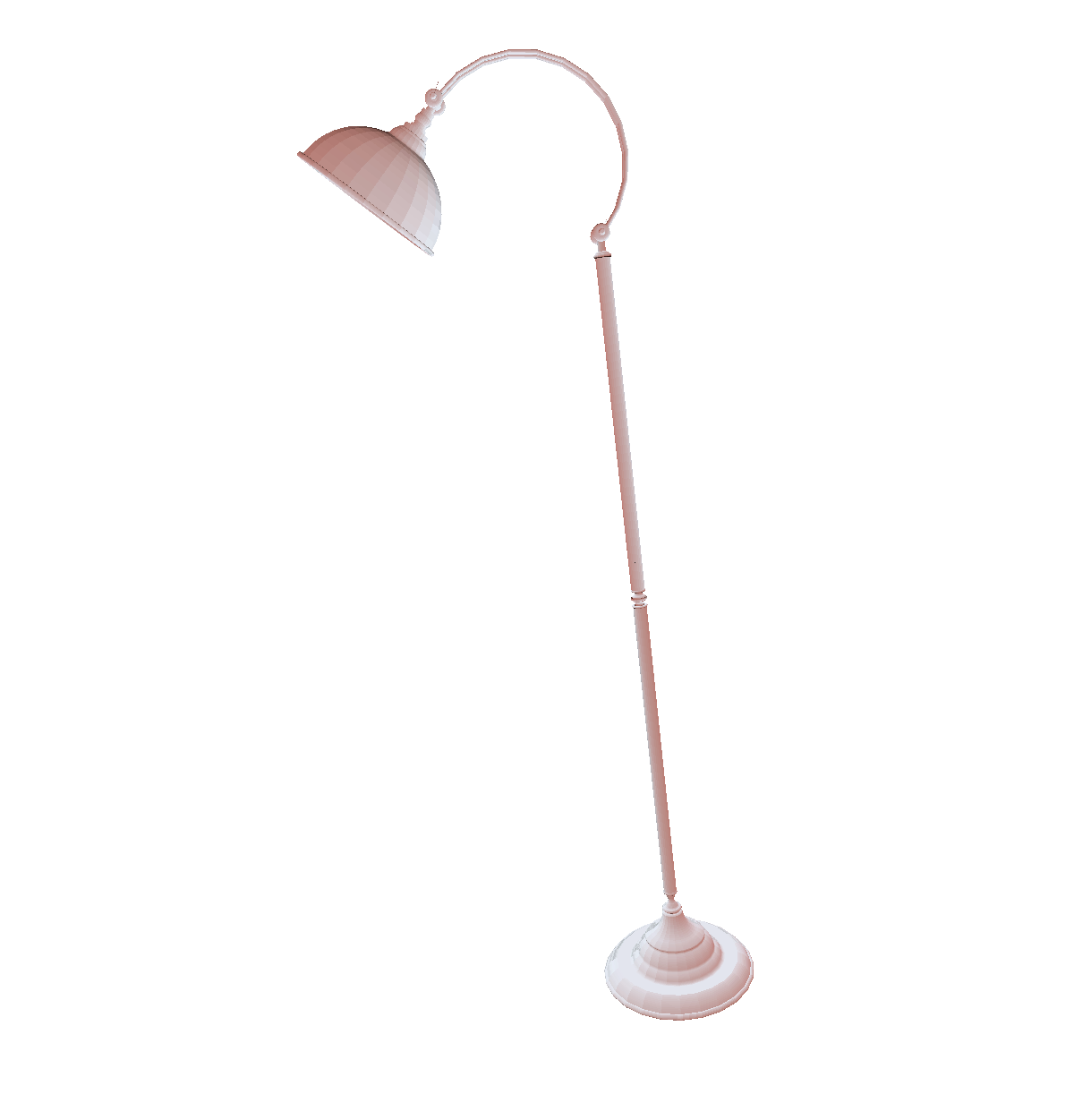}
	\end{subfigure} 
	~
	\begin{subfigure}{27.5mm}
		\centering
		\includegraphics[width=27.5mm]{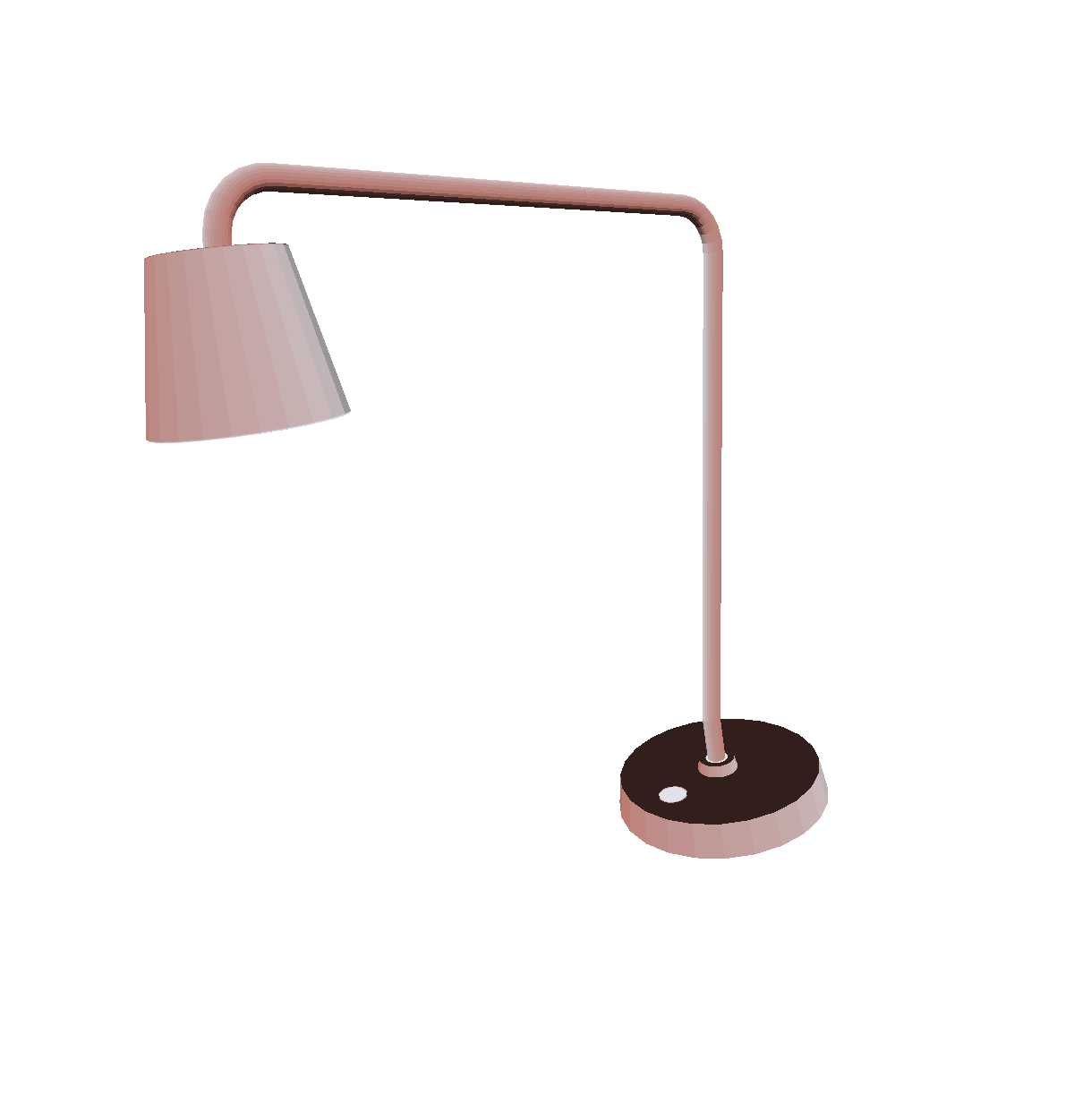}
	\end{subfigure}
	~
	\begin{subfigure}{27.5mm}
		\centering
		\includegraphics[width=27.5mm]{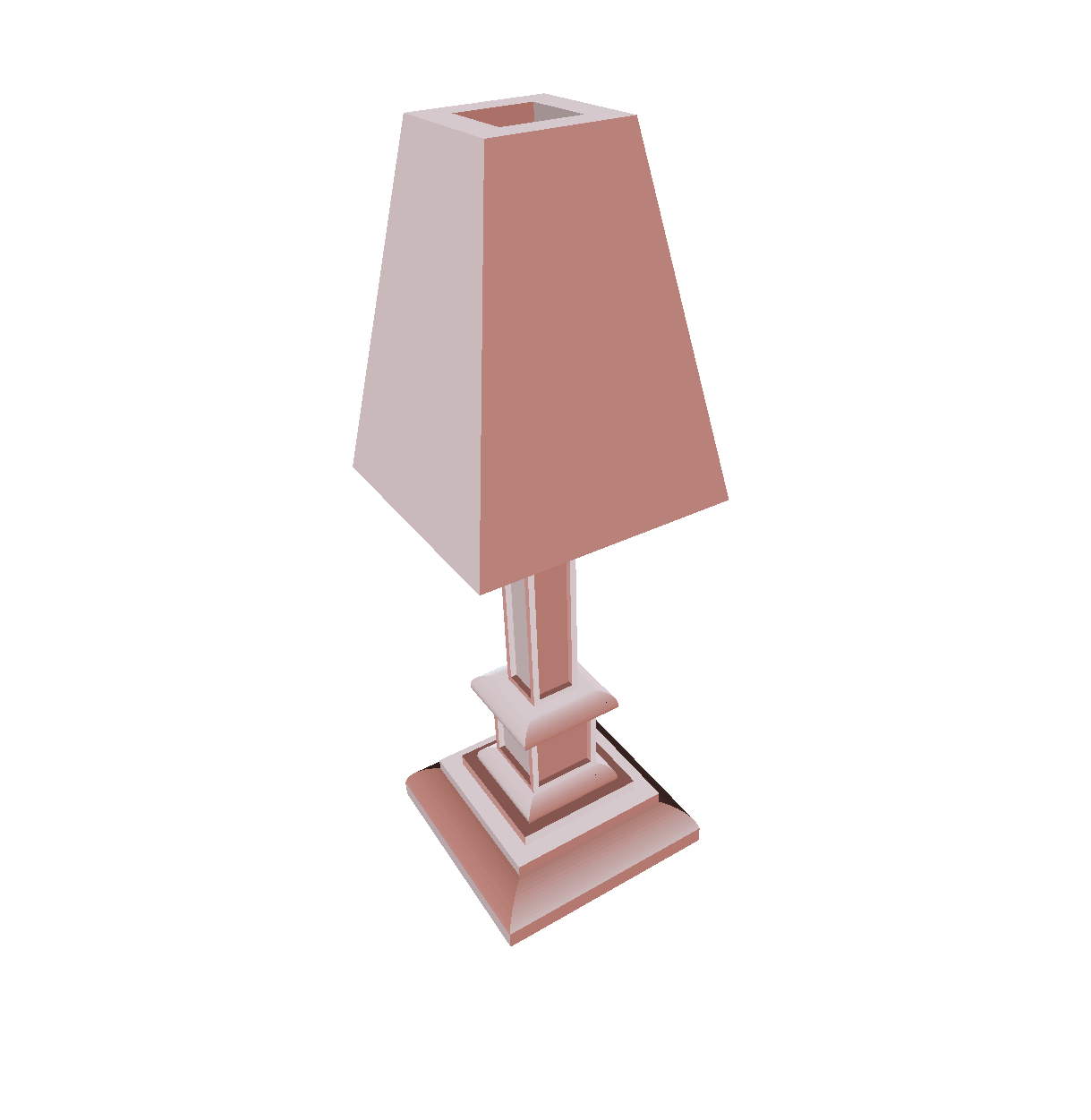}
	\end{subfigure}  \\
	\begin{subfigure}{27.5mm}
		\centering
		\includegraphics[width=27.5mm]{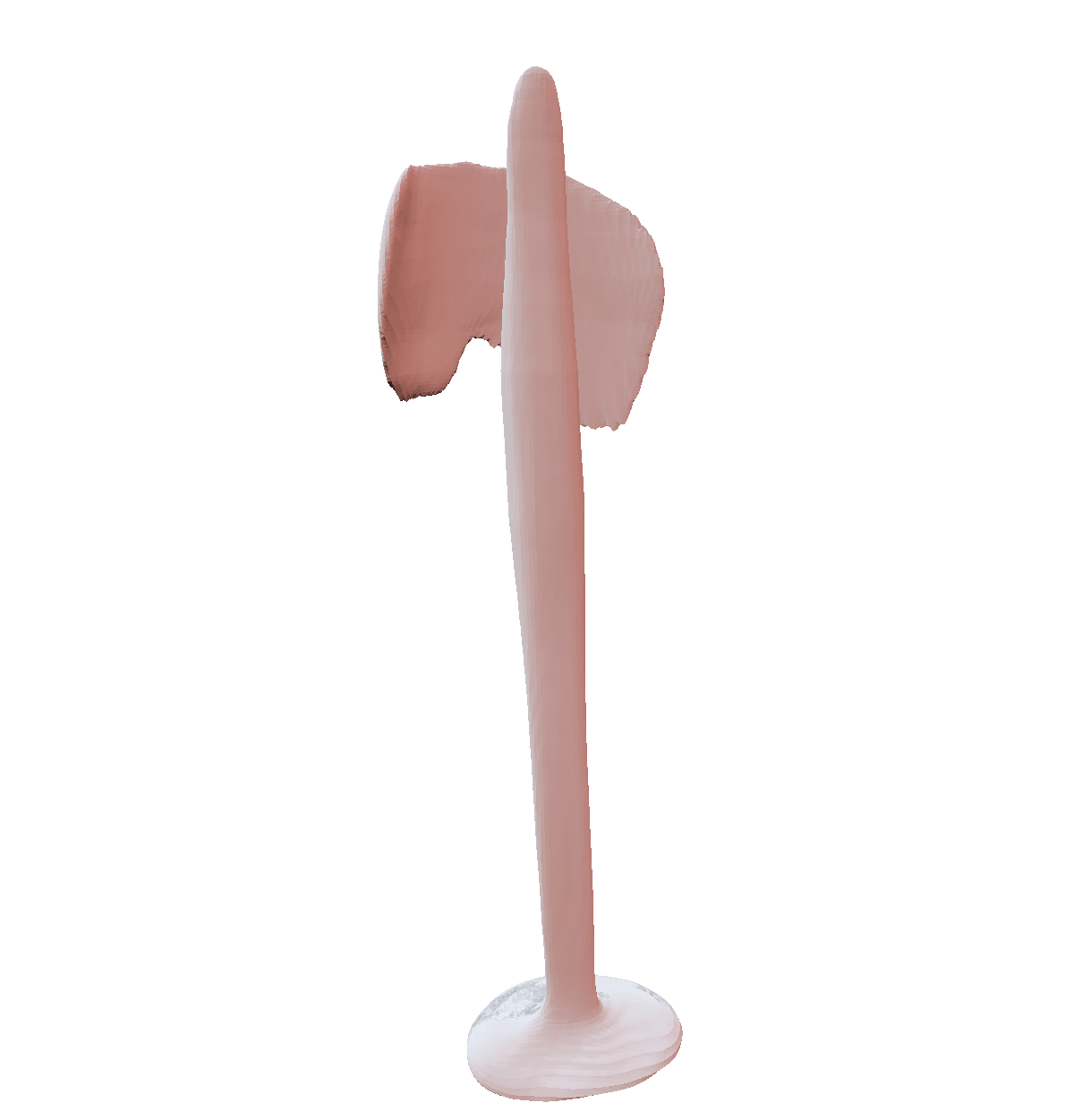}
	\end{subfigure}
	~
	\begin{subfigure}{27.5mm}
		\centering
		\includegraphics[width=27.5mm]{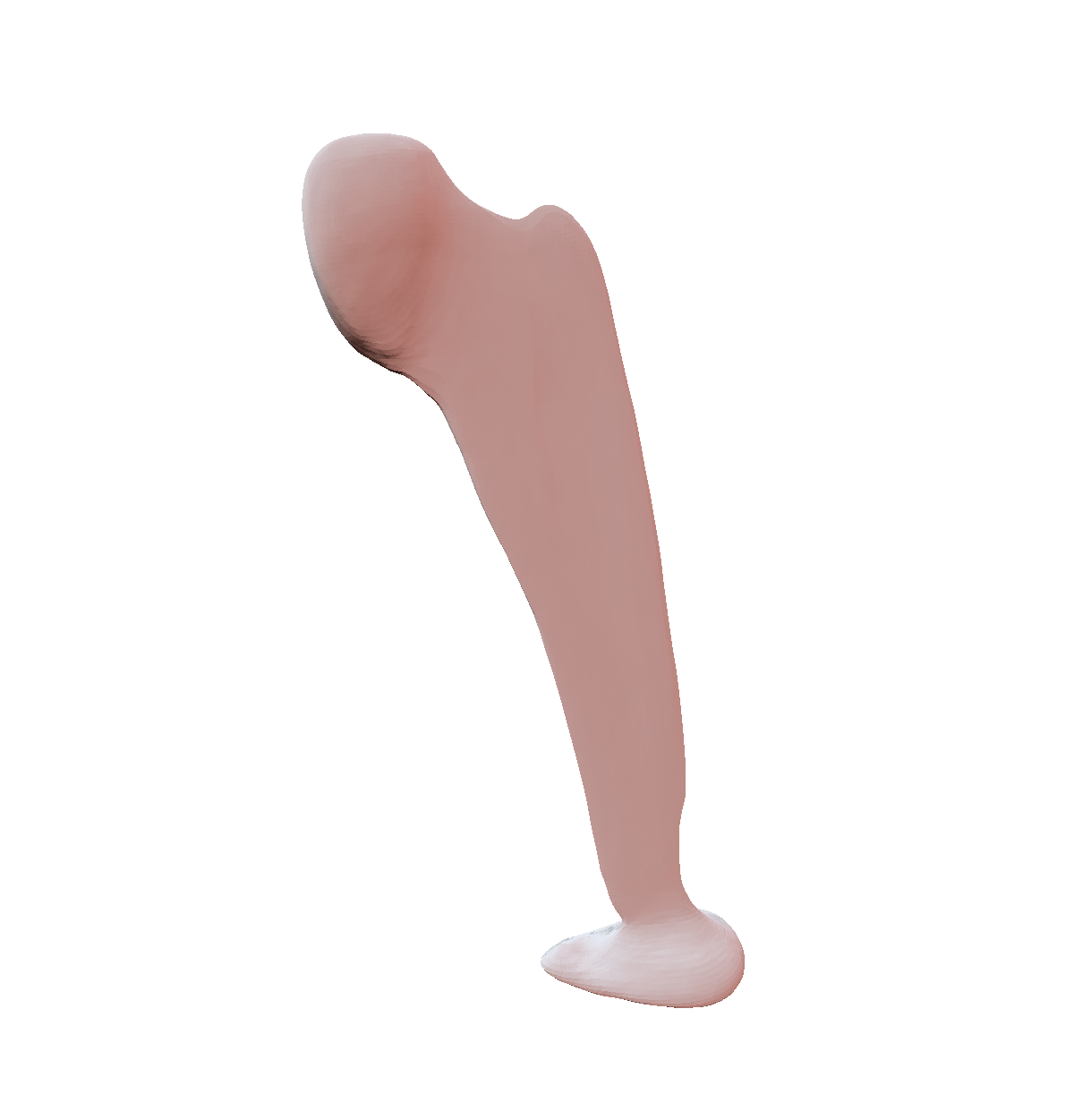}
	\end{subfigure} 
	~
	\begin{subfigure}{27.5mm}
		\centering
		\includegraphics[width=27.5mm]{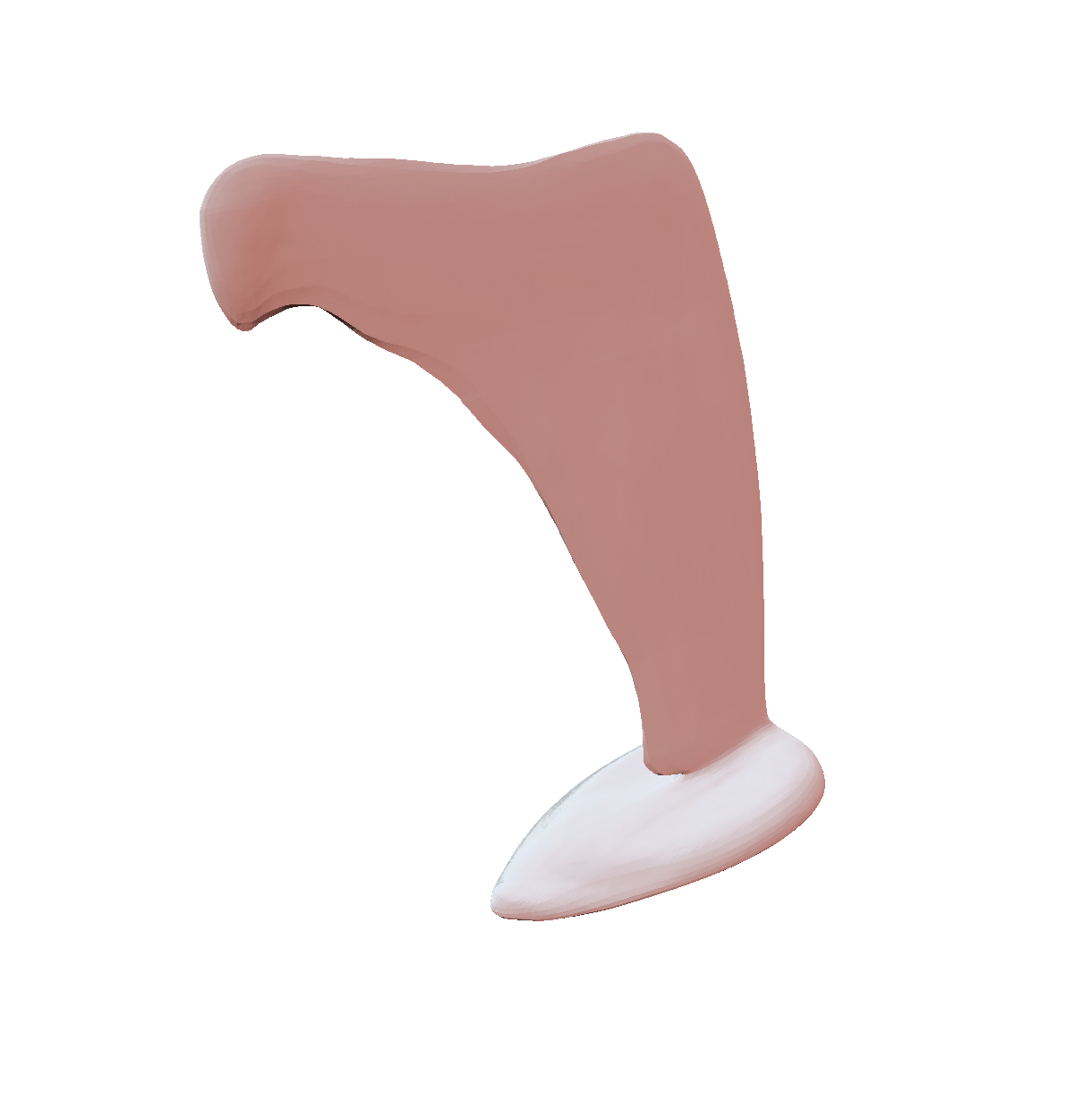}
	\end{subfigure} 
	~
	\begin{subfigure}{27.5mm}
		\centering
		\includegraphics[width=27.5mm]{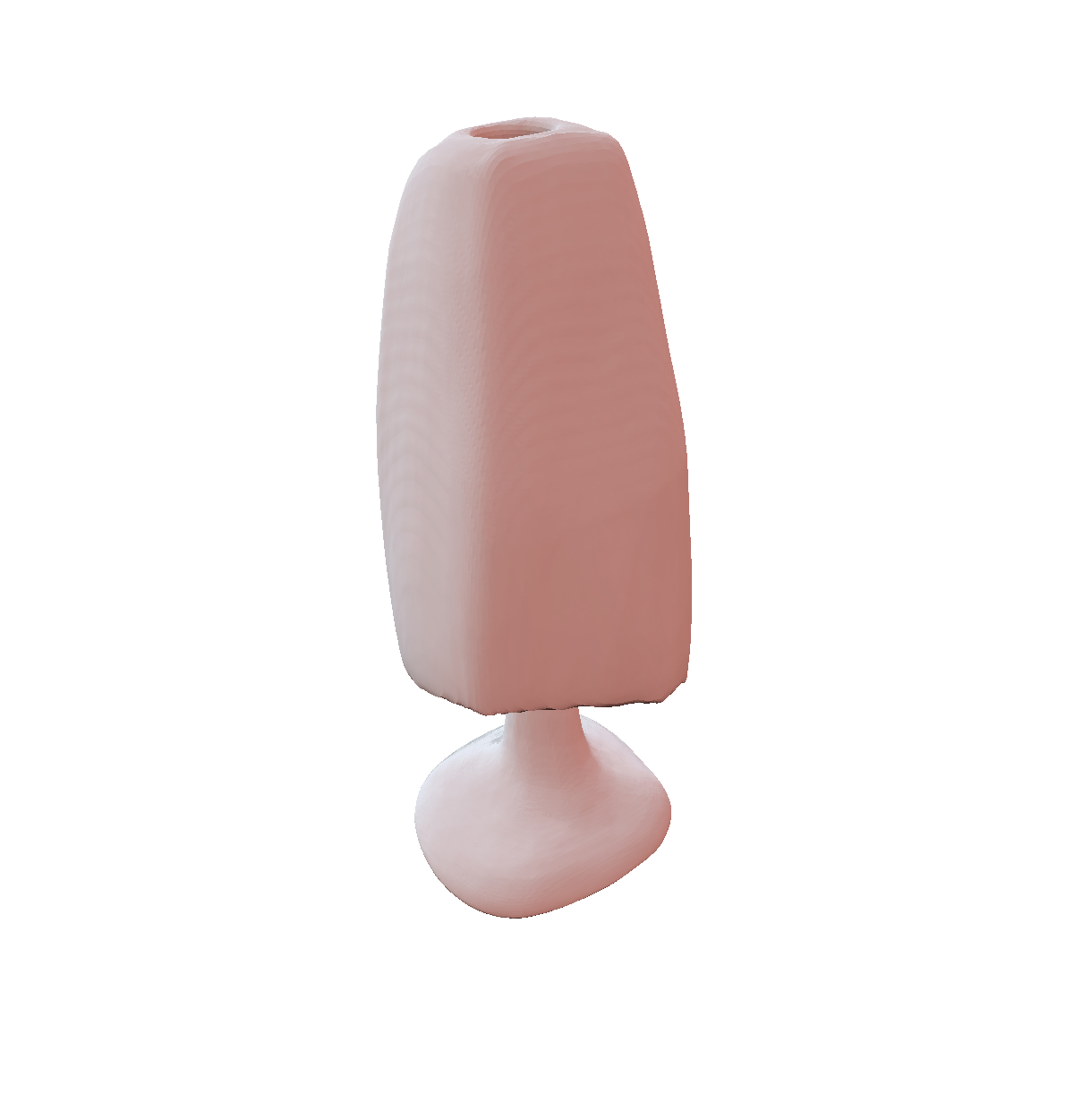}
	\end{subfigure}  \\
	\begin{subfigure}{27.5mm}
		\centering
		\includegraphics[width=27.5mm]{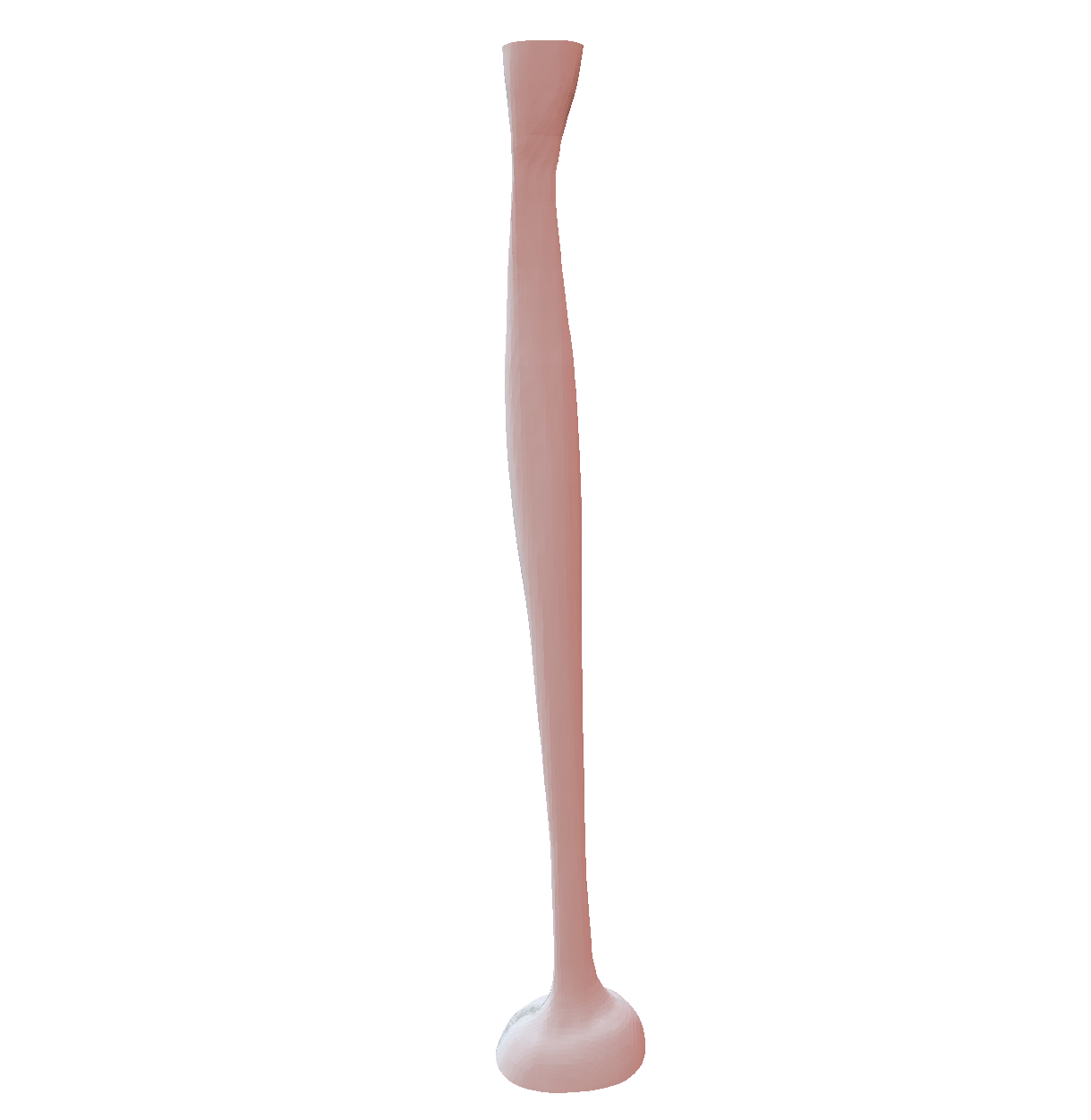}
	\end{subfigure}
	~
	\begin{subfigure}{27.5mm}
		\centering
		\includegraphics[width=27.5mm]{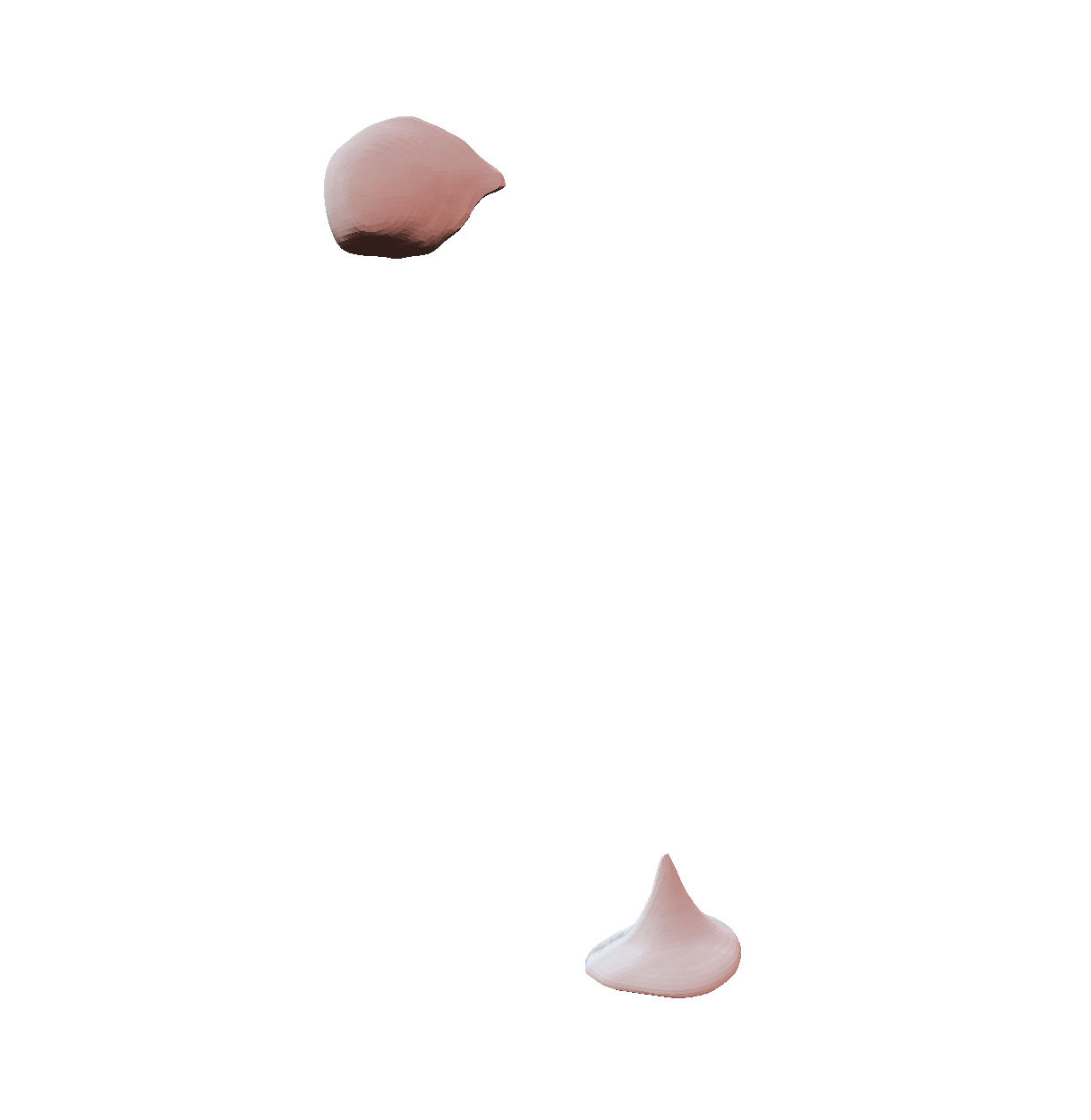}
	\end{subfigure} 
	~
	\begin{subfigure}{27.5mm}
		\centering
		\includegraphics[width=27.5mm]{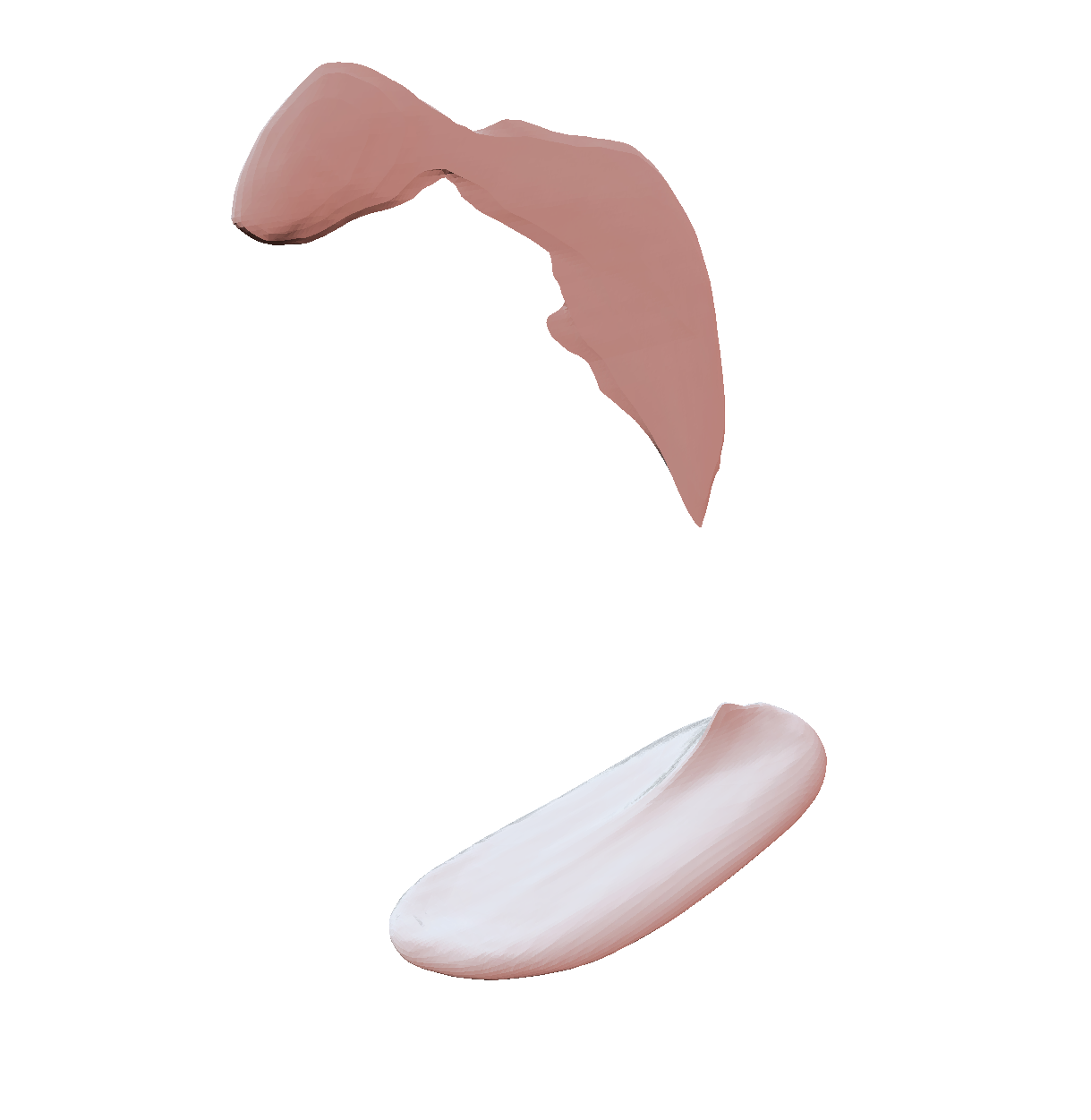}
	\end{subfigure} 
	~
	\begin{subfigure}{27.5mm}
		\centering
		\includegraphics[width=27.5mm]{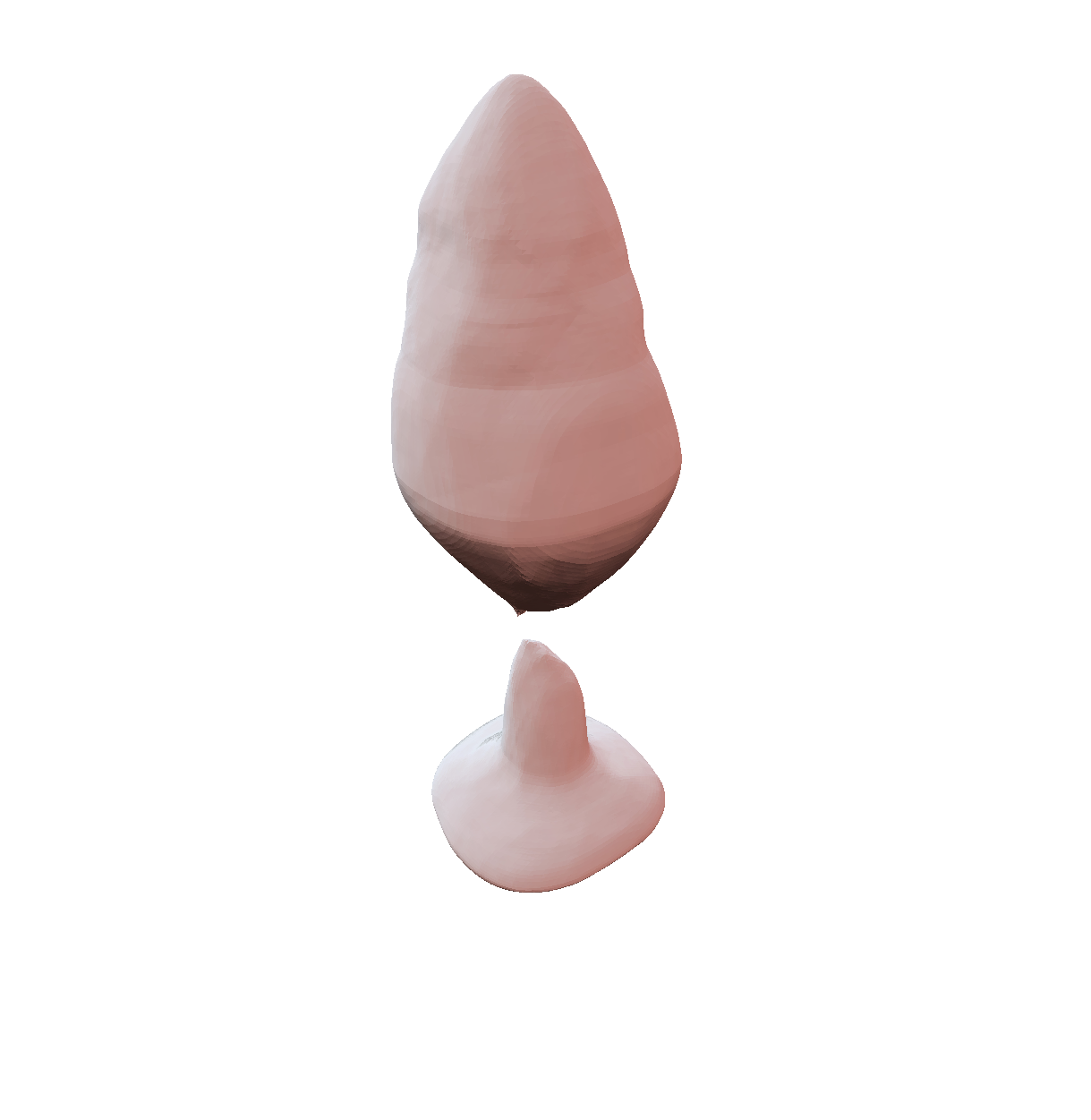}
	\end{subfigure} \\
	\begin{subfigure}{27.5mm}
		\centering
		\includegraphics[width=27.5mm]{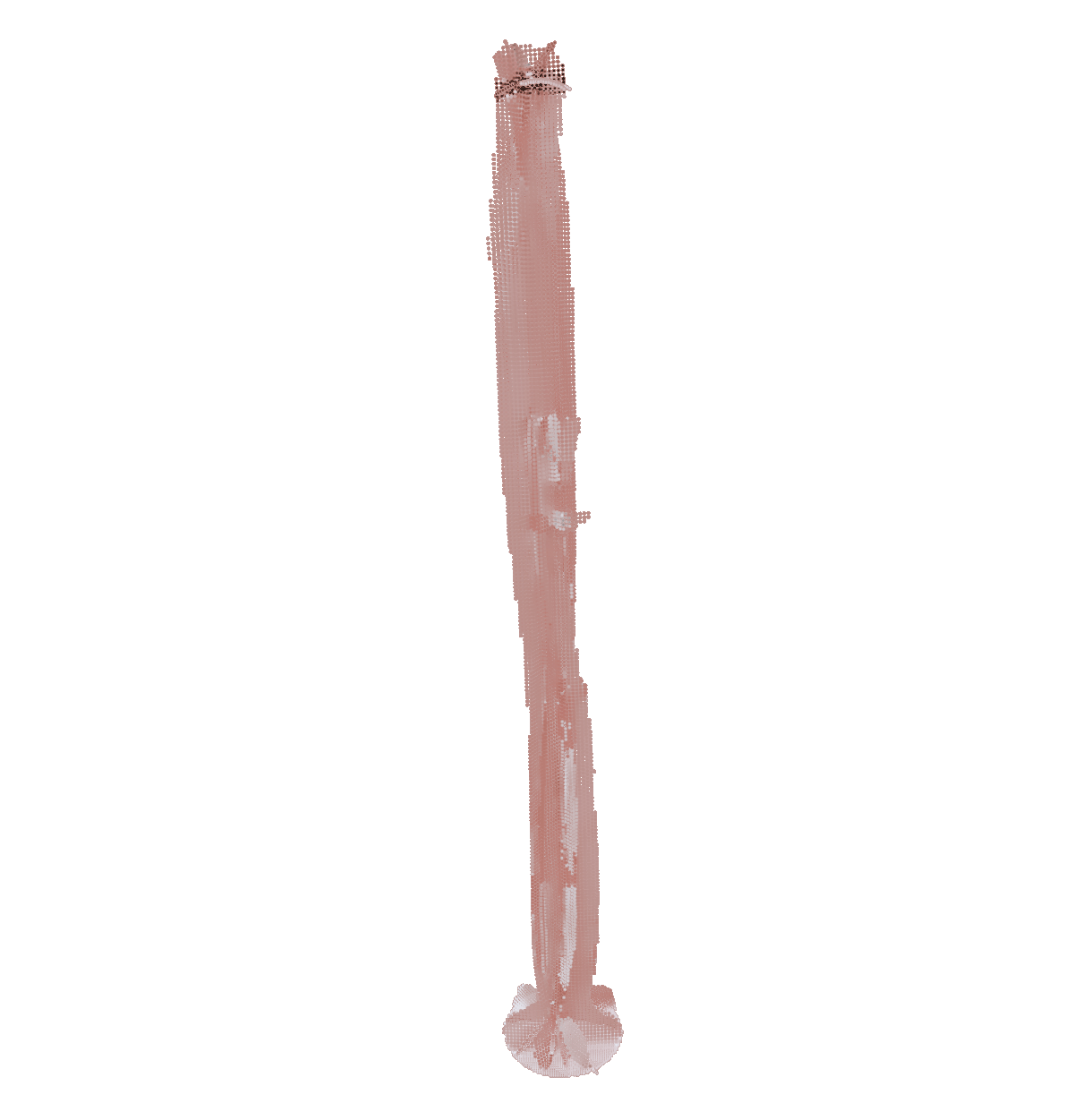}
	\end{subfigure}
	~
	\begin{subfigure}{27.5mm}
		\centering
		\includegraphics[width=27.5mm]{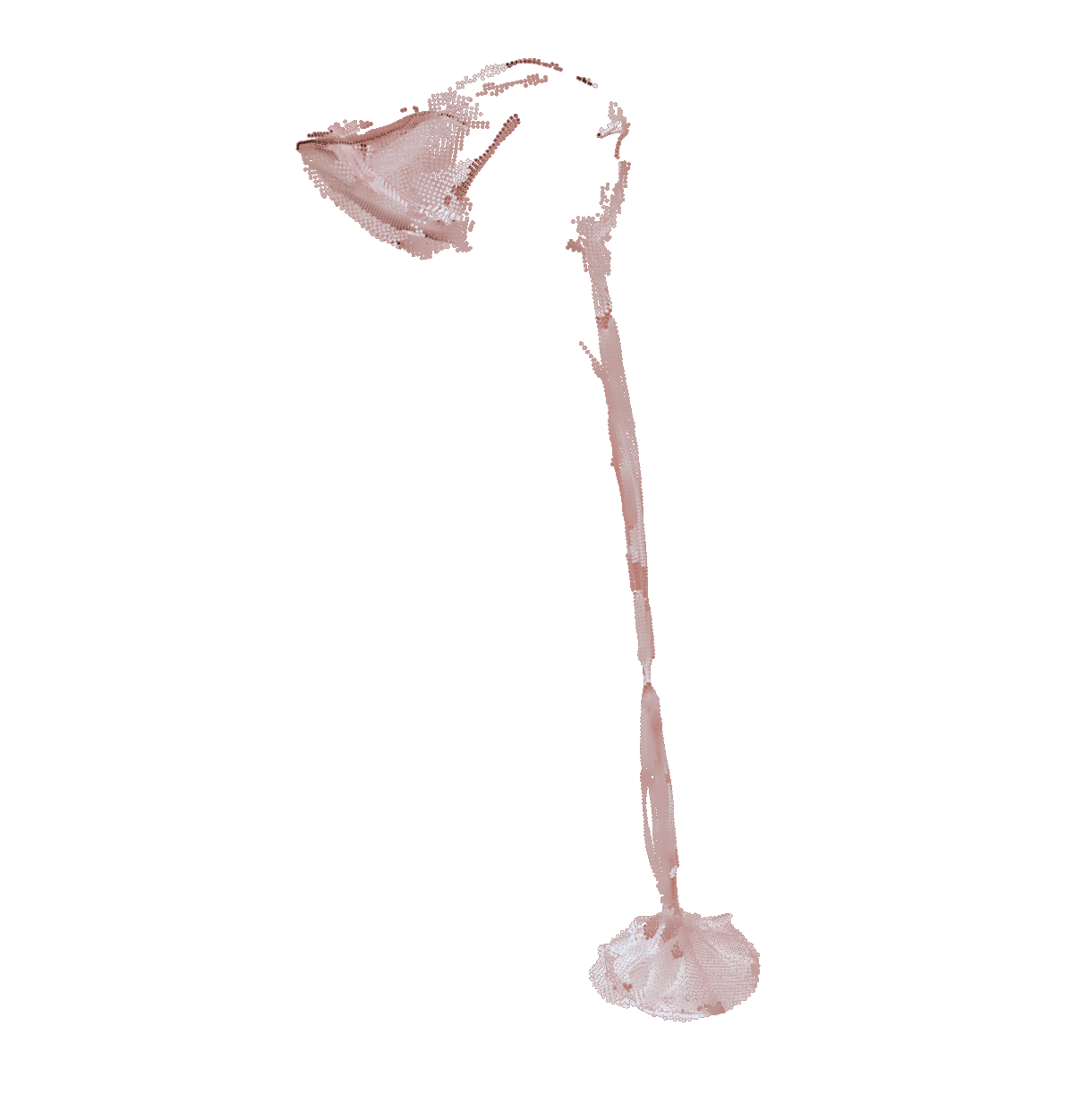}
	\end{subfigure} 
	~
	\begin{subfigure}{27.5mm}
		\centering
		\includegraphics[width=27.5mm]{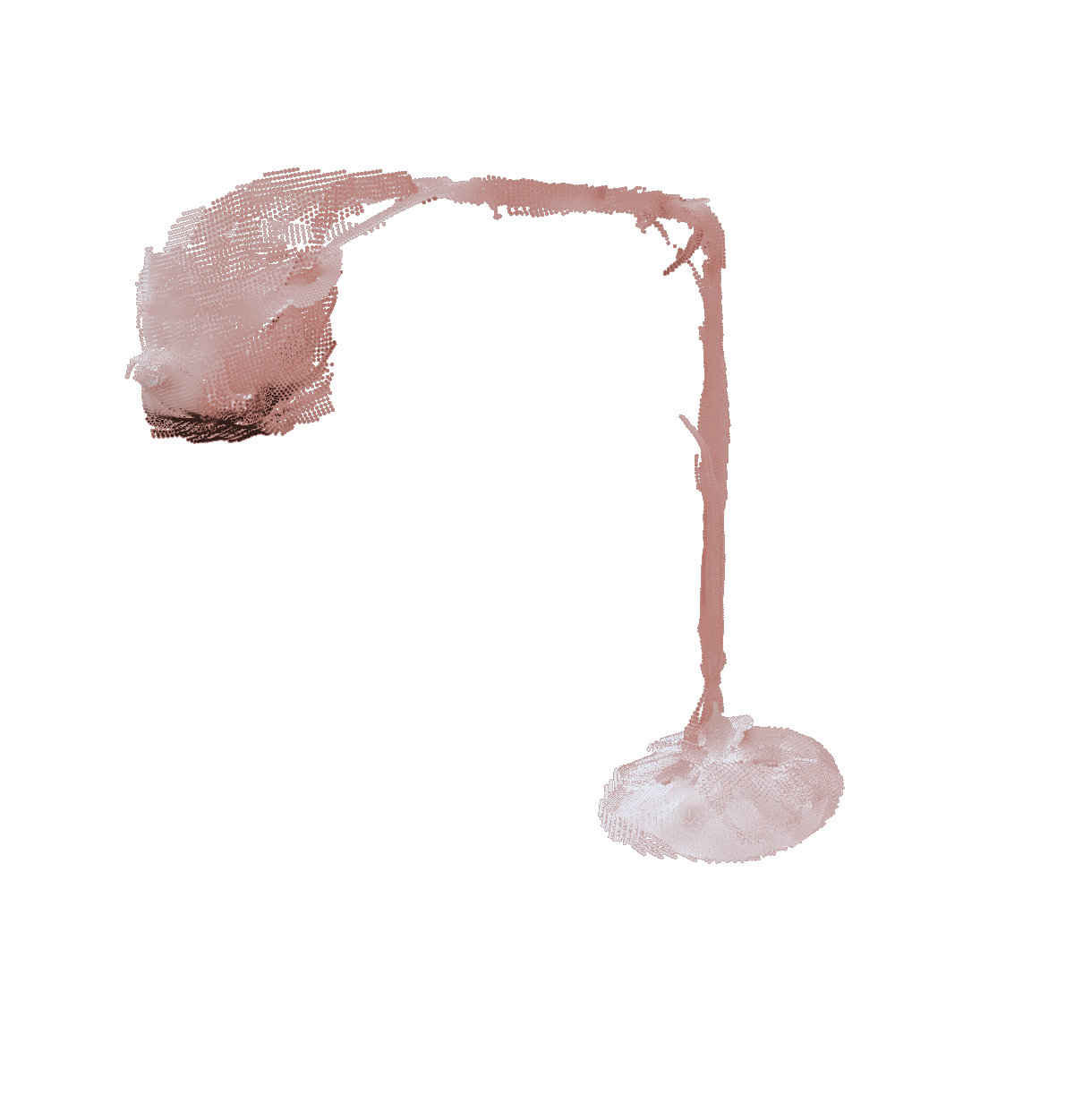}
	\end{subfigure} 
	~
	\begin{subfigure}{27.5mm}
		\centering
		\includegraphics[width=27.5mm]{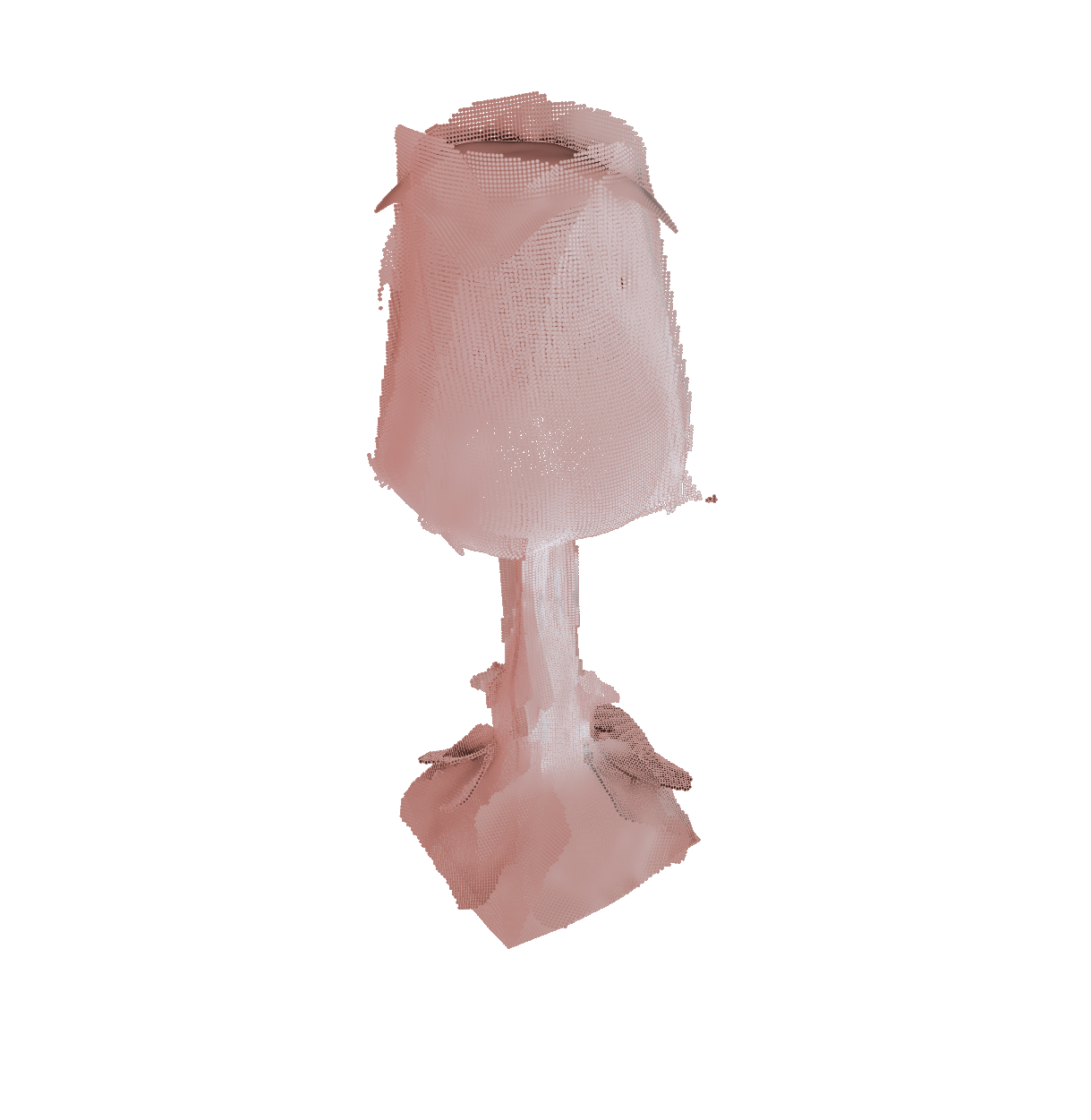}
	\end{subfigure} \\
	\caption{{\bf More Generative Results from Different Categories.} Results from Rows 1 to 4 correspond to {\bf Reference}, {\bf OF}, {\bf SDF}, {\bf PRIF - Mesh}.}
\end{figure}

\begin{figure}[!ht]
    \centering
	\begin{subfigure}{27.5mm}
		\centering
		\includegraphics[width=27.5mm]{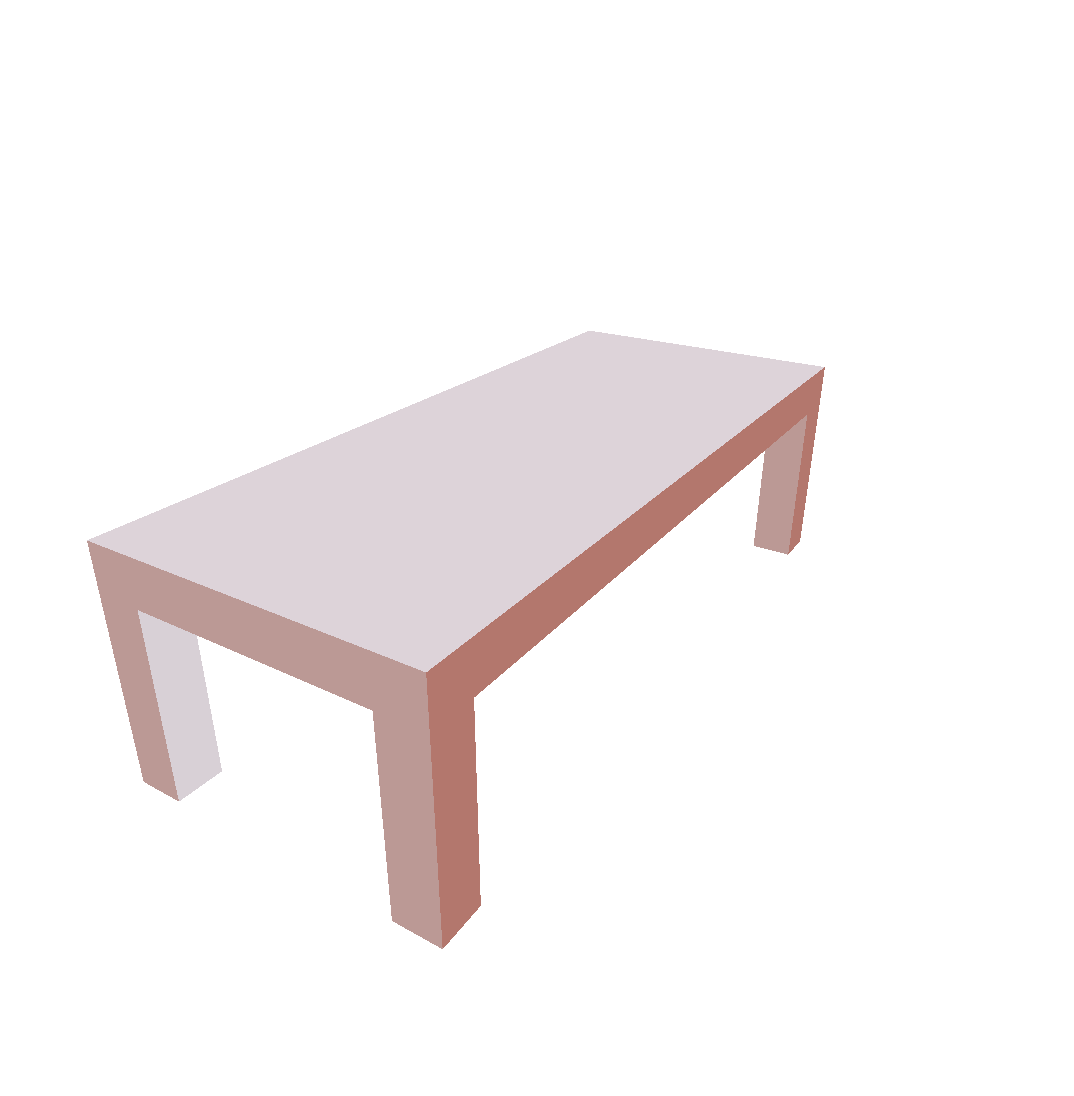}
	\end{subfigure}
	~
	\begin{subfigure}{27.5mm}
		\centering
		\includegraphics[width=27.5mm]{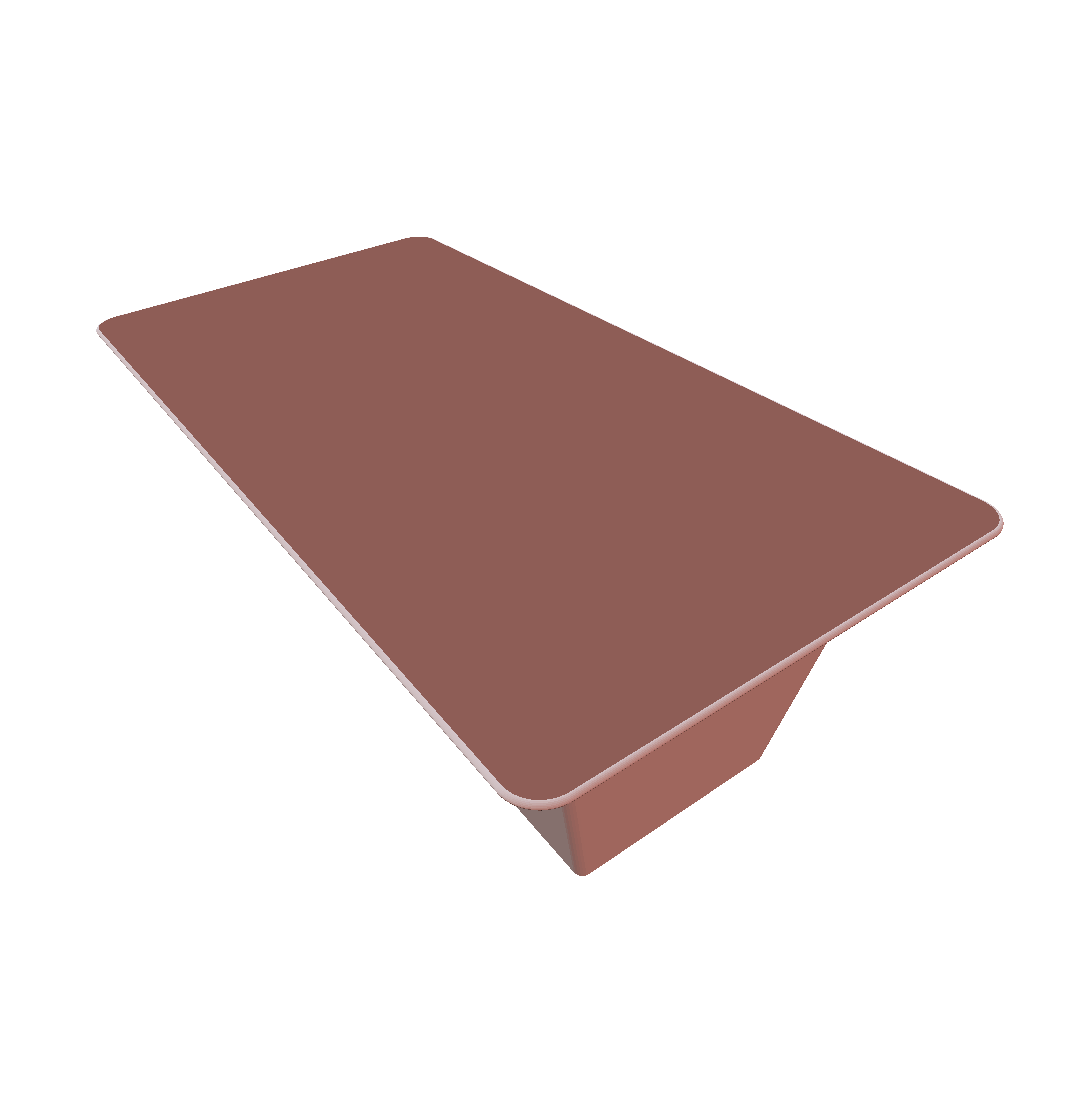}
	\end{subfigure} 
	~
	\begin{subfigure}{27.5mm}
		\centering
		\includegraphics[width=27.5mm]{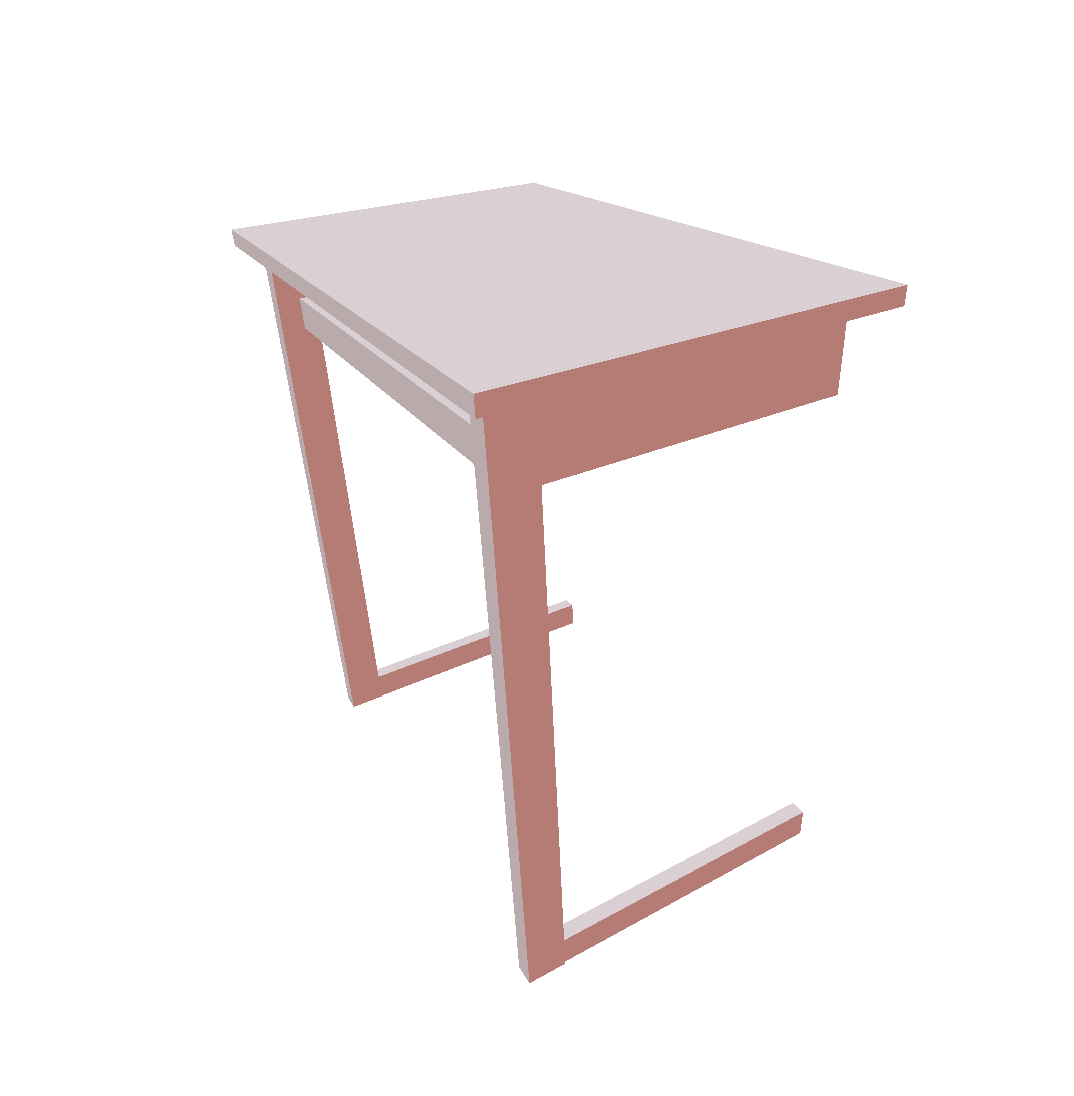}
	\end{subfigure}
	~
	\begin{subfigure}{27.5mm}
		\centering
		\includegraphics[width=27.5mm]{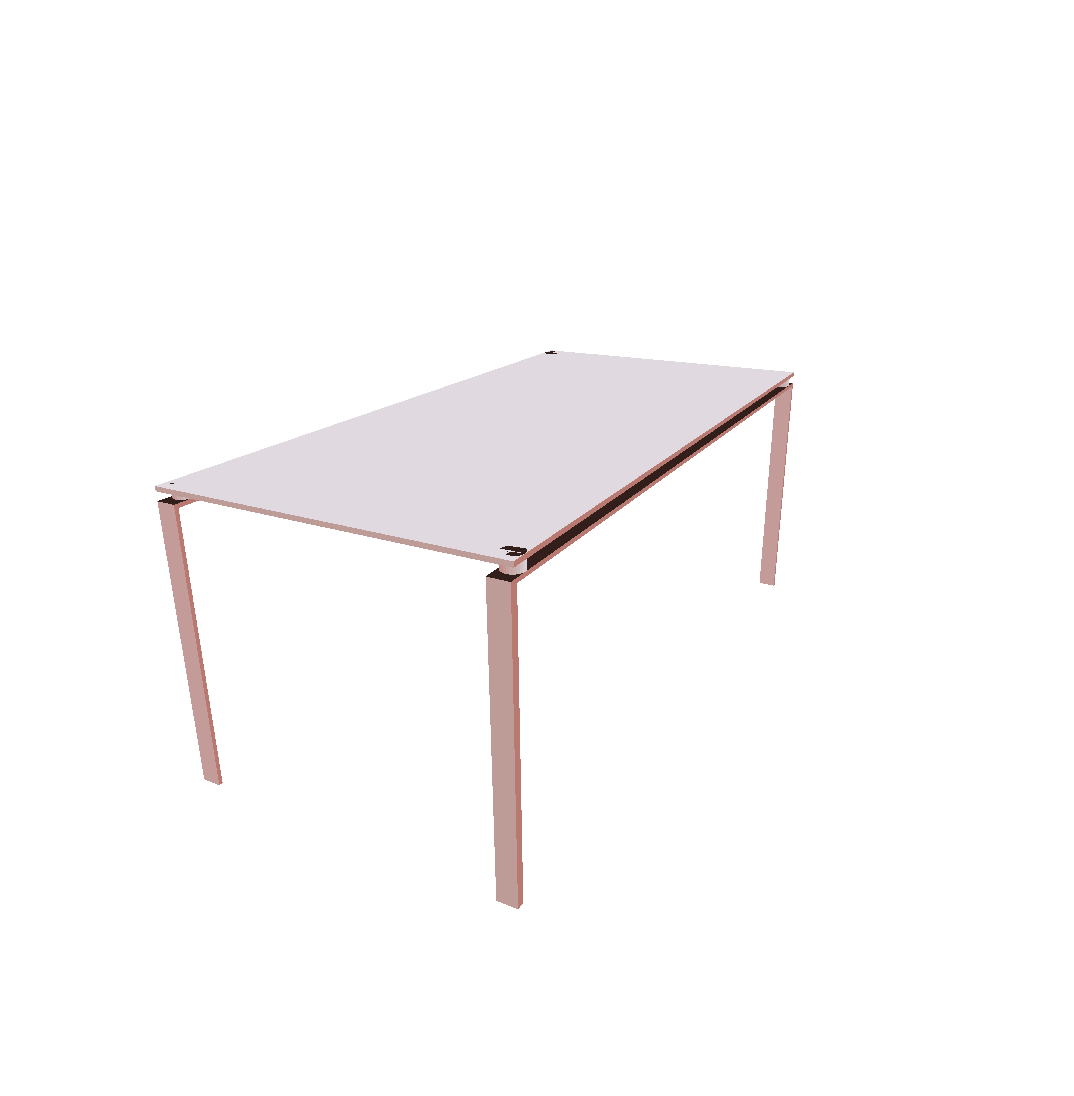}
	\end{subfigure}  \\
	\begin{subfigure}{27.5mm}
		\centering
		\includegraphics[width=27.5mm]{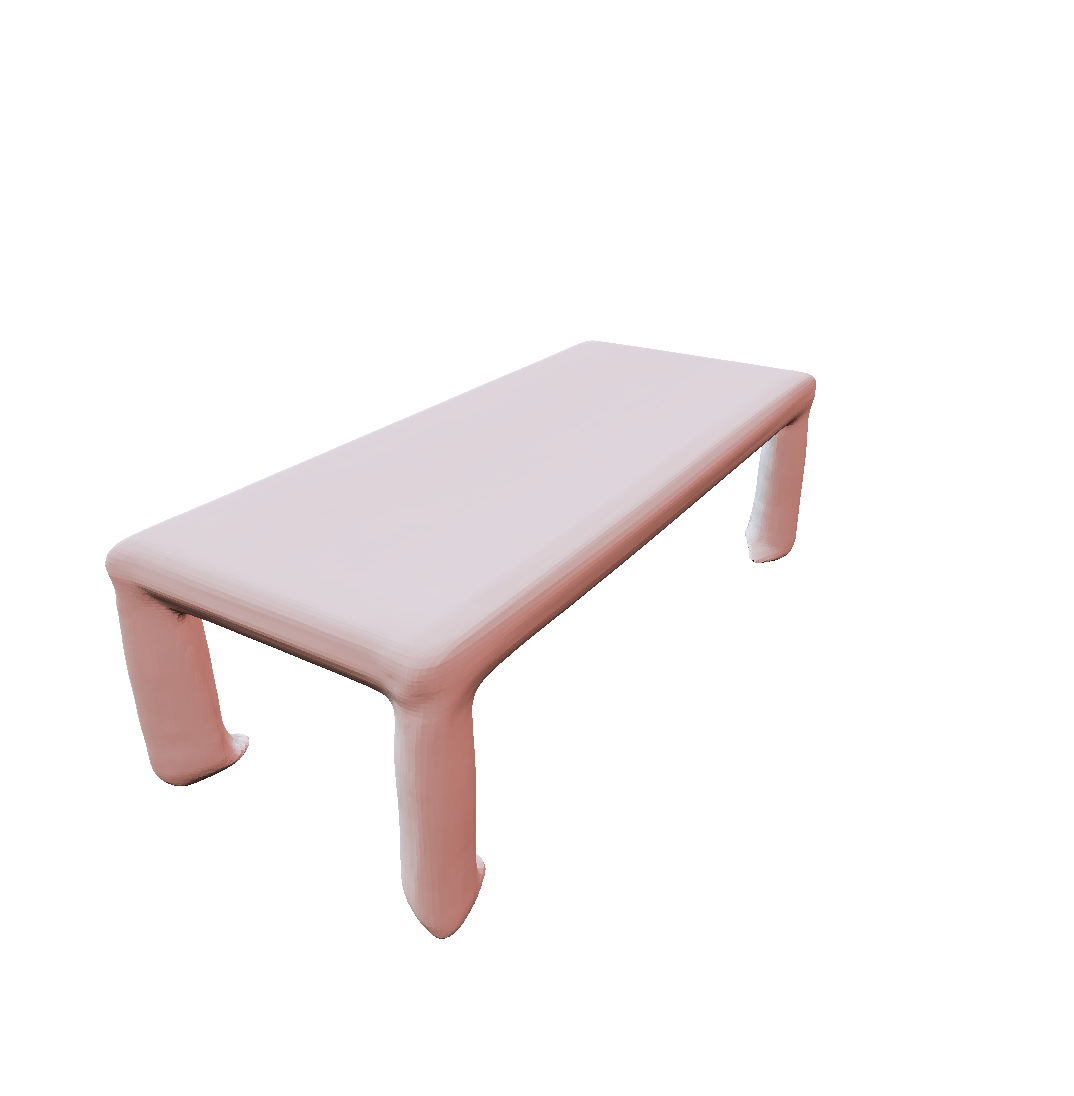}
	\end{subfigure}
	~
	\begin{subfigure}{27.5mm}
		\centering
		\includegraphics[width=27.5mm]{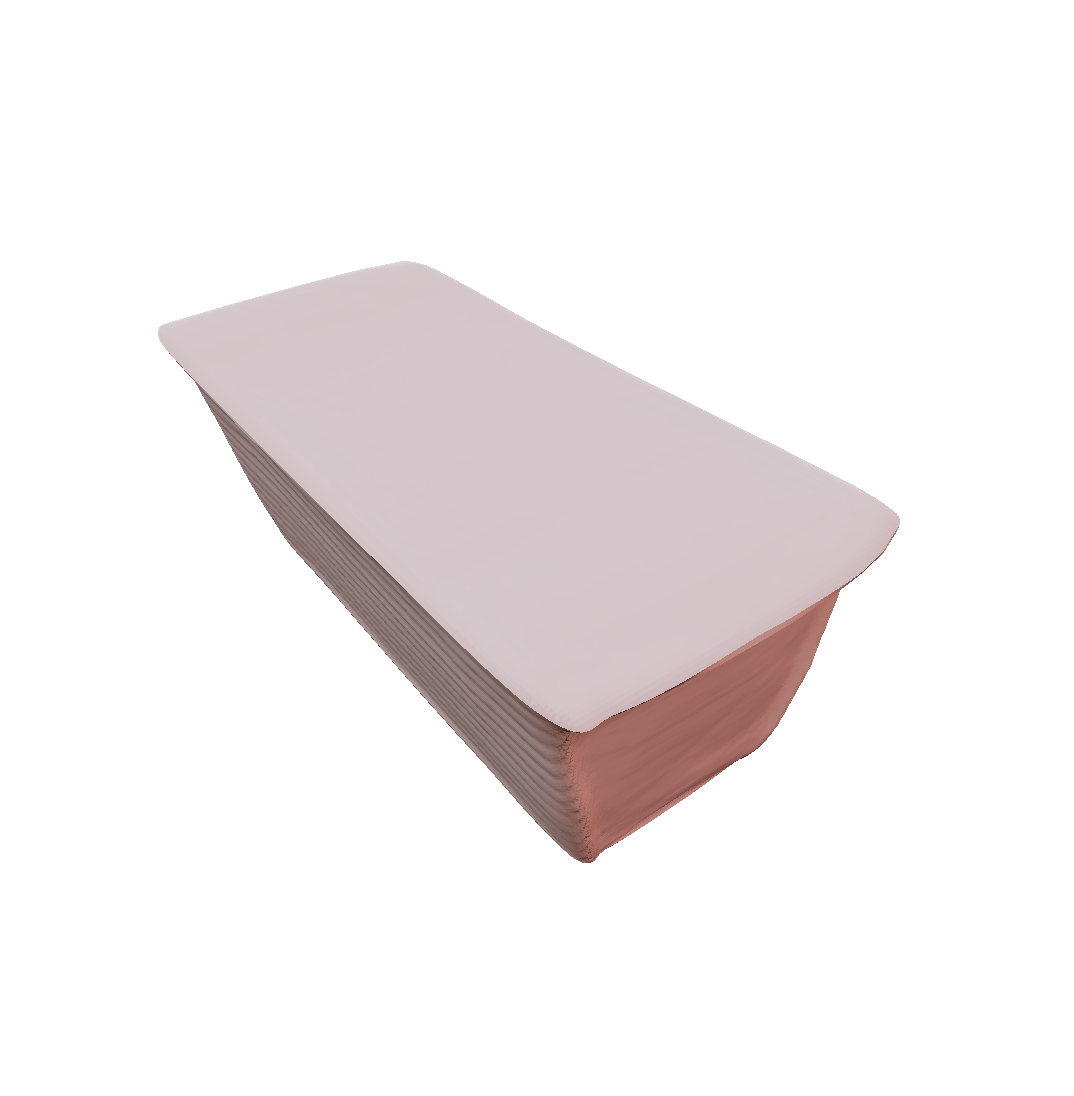}
	\end{subfigure} 
	~
	\begin{subfigure}{27.5mm}
		\centering
		\includegraphics[width=27.5mm]{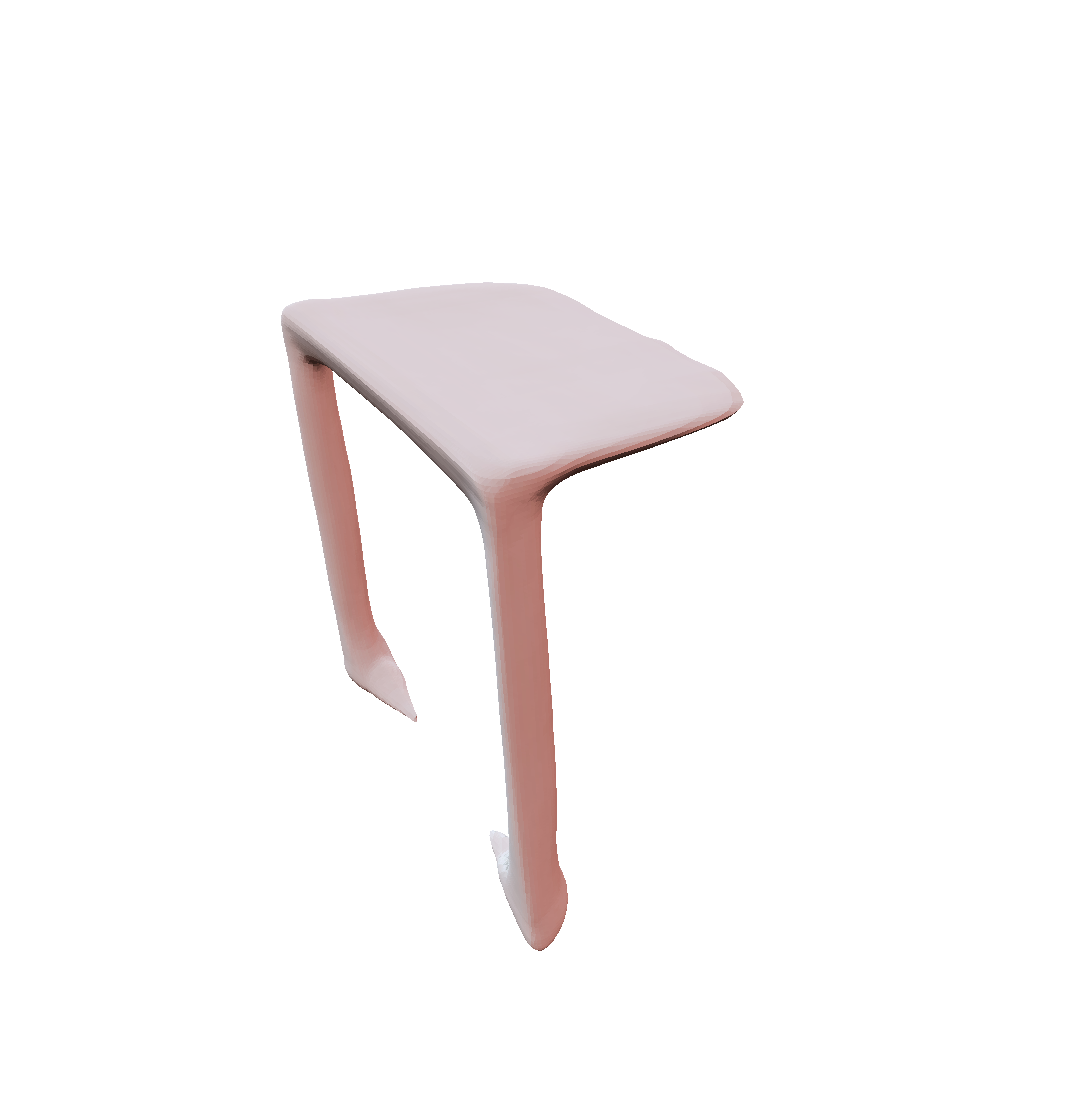}
	\end{subfigure} 
	~
	\begin{subfigure}{27.5mm}
		\centering
		\includegraphics[width=27.5mm]{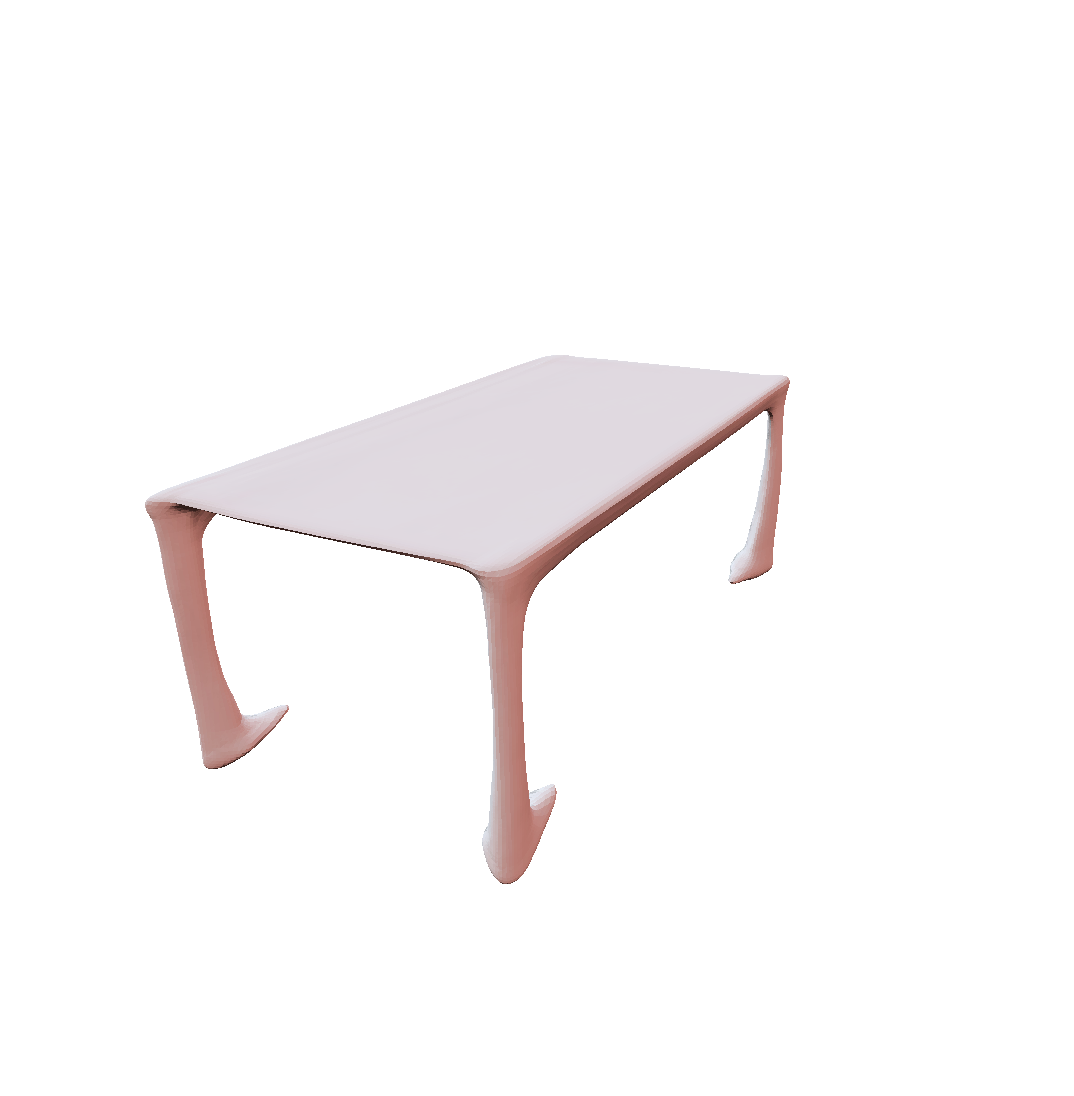}
	\end{subfigure}  \\
	\begin{subfigure}{27.5mm}
		\centering
		\includegraphics[width=27.5mm]{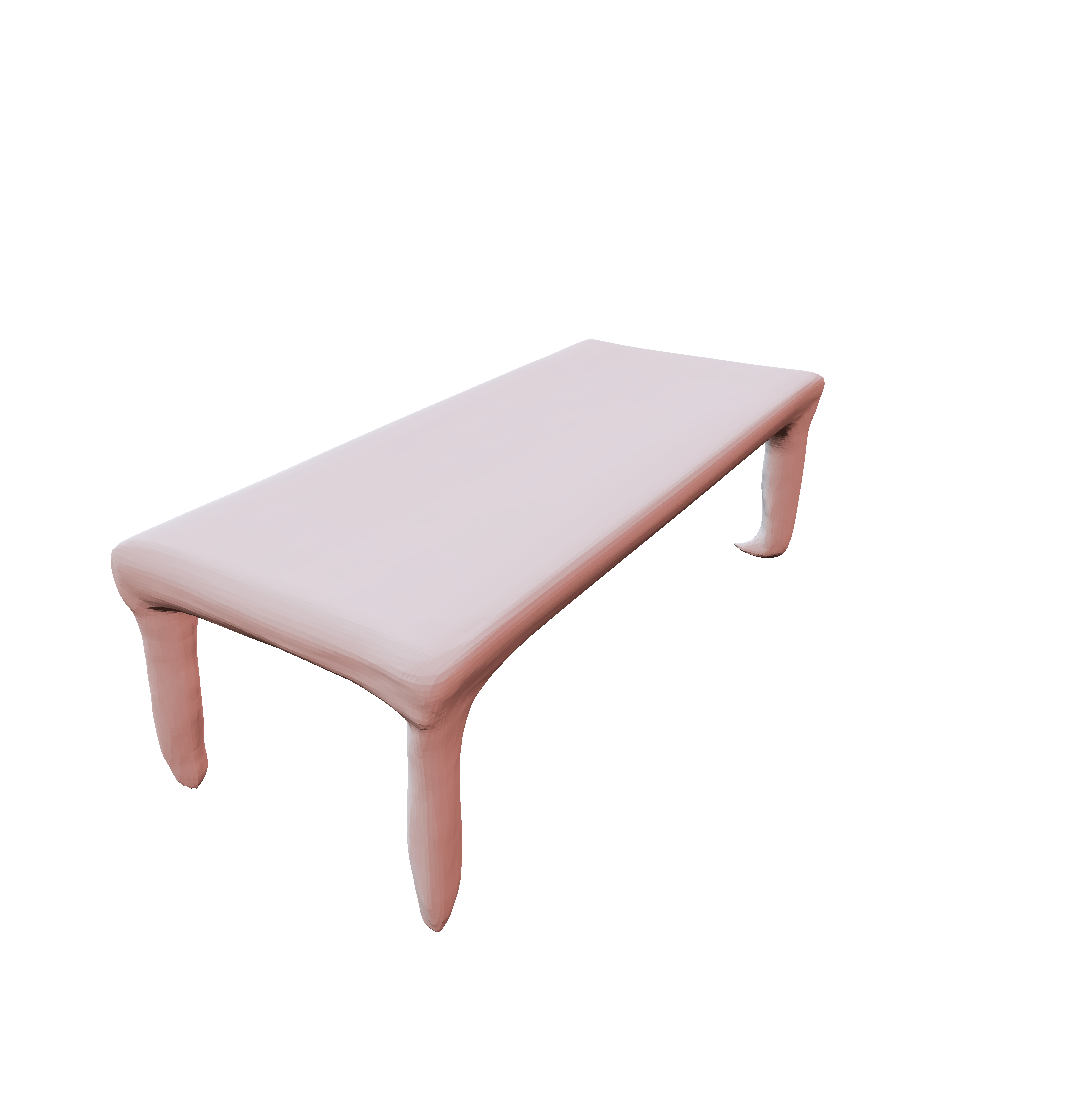}
	\end{subfigure}
	~
	\begin{subfigure}{27.5mm}
		\centering
		\includegraphics[width=27.5mm]{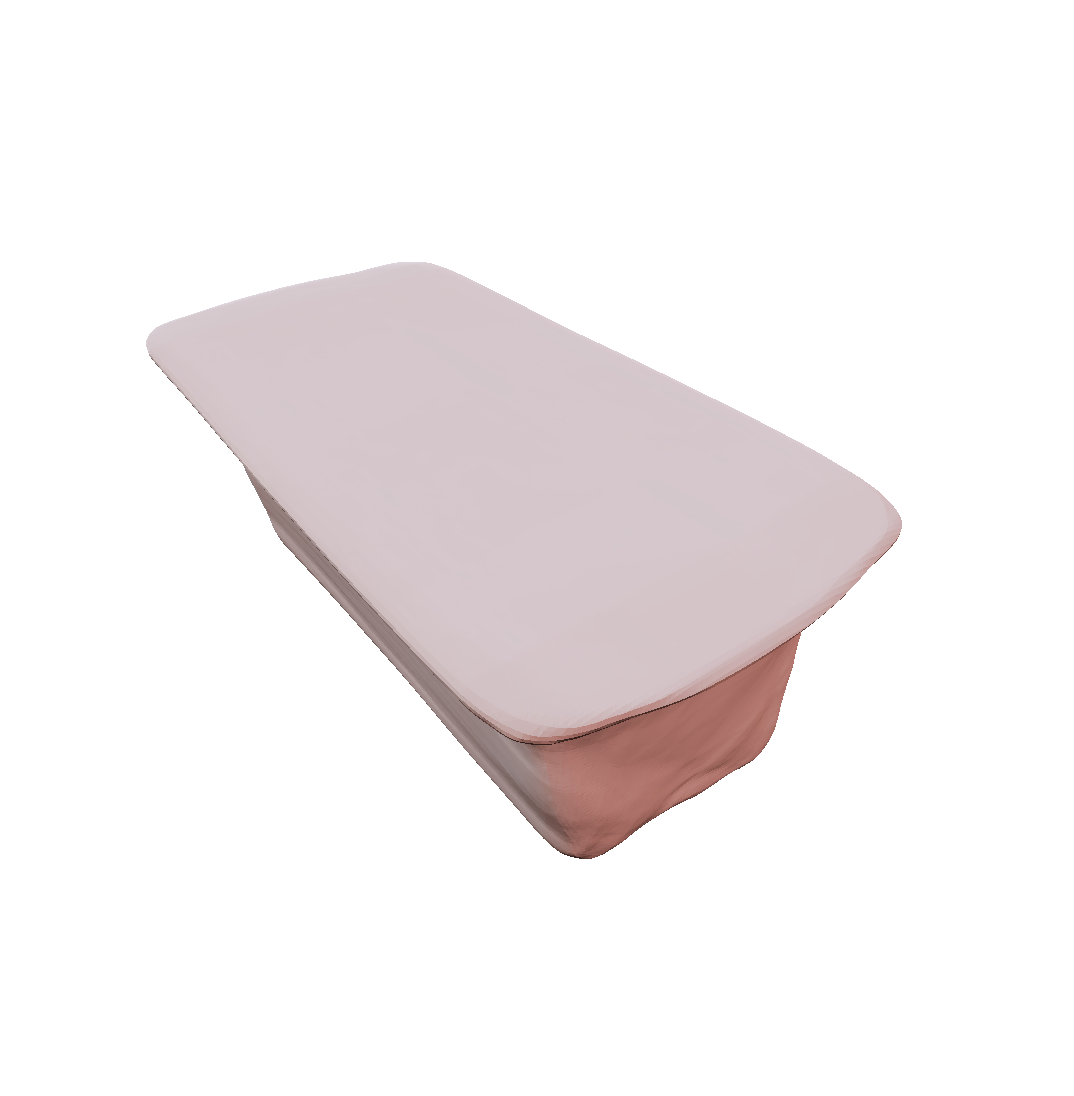}
	\end{subfigure} 
	~
	\begin{subfigure}{27.5mm}
		\centering
		\includegraphics[width=27.5mm]{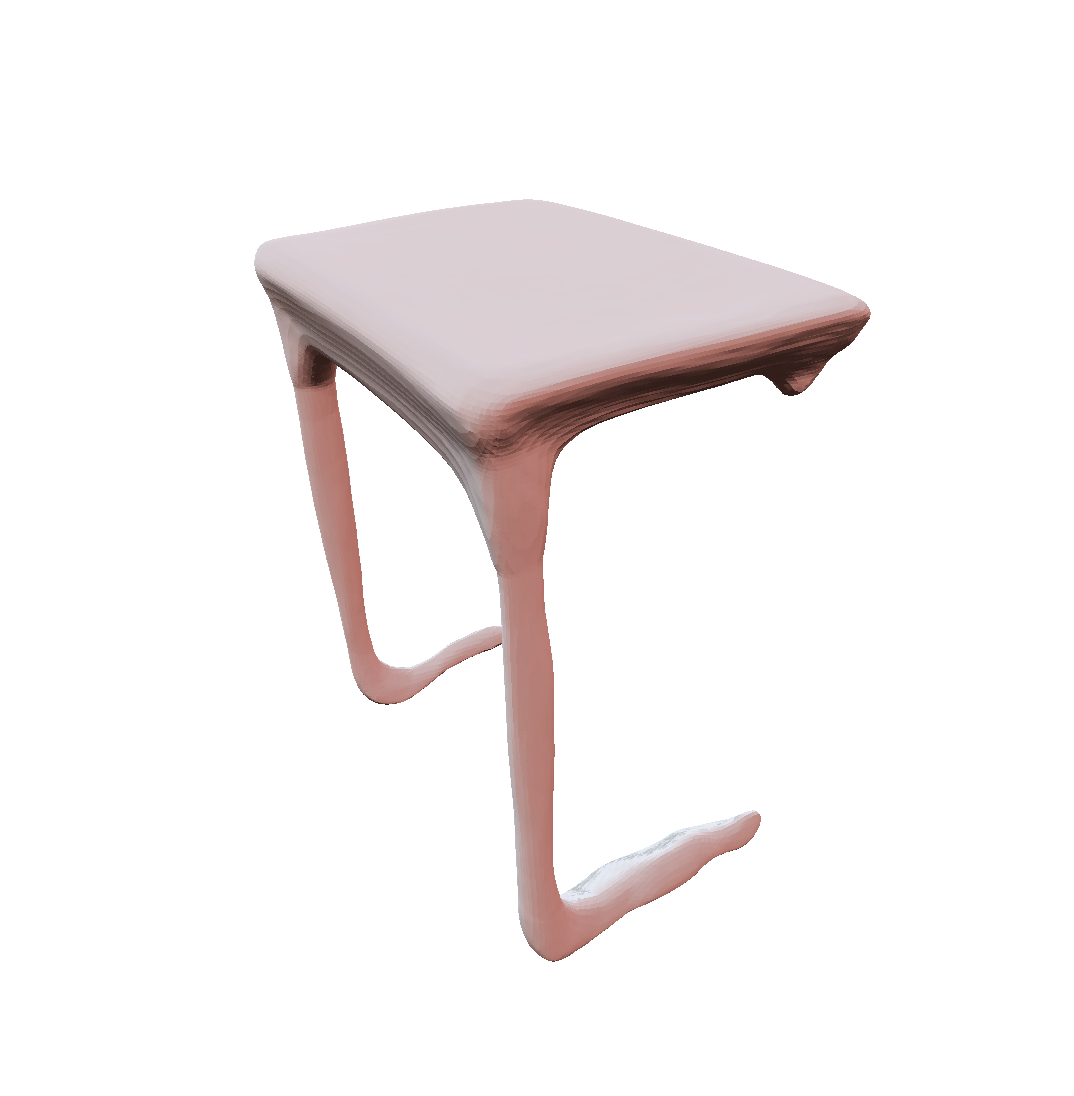}
	\end{subfigure} 
	~
	\begin{subfigure}{27.5mm}
		\centering
		\includegraphics[width=27.5mm]{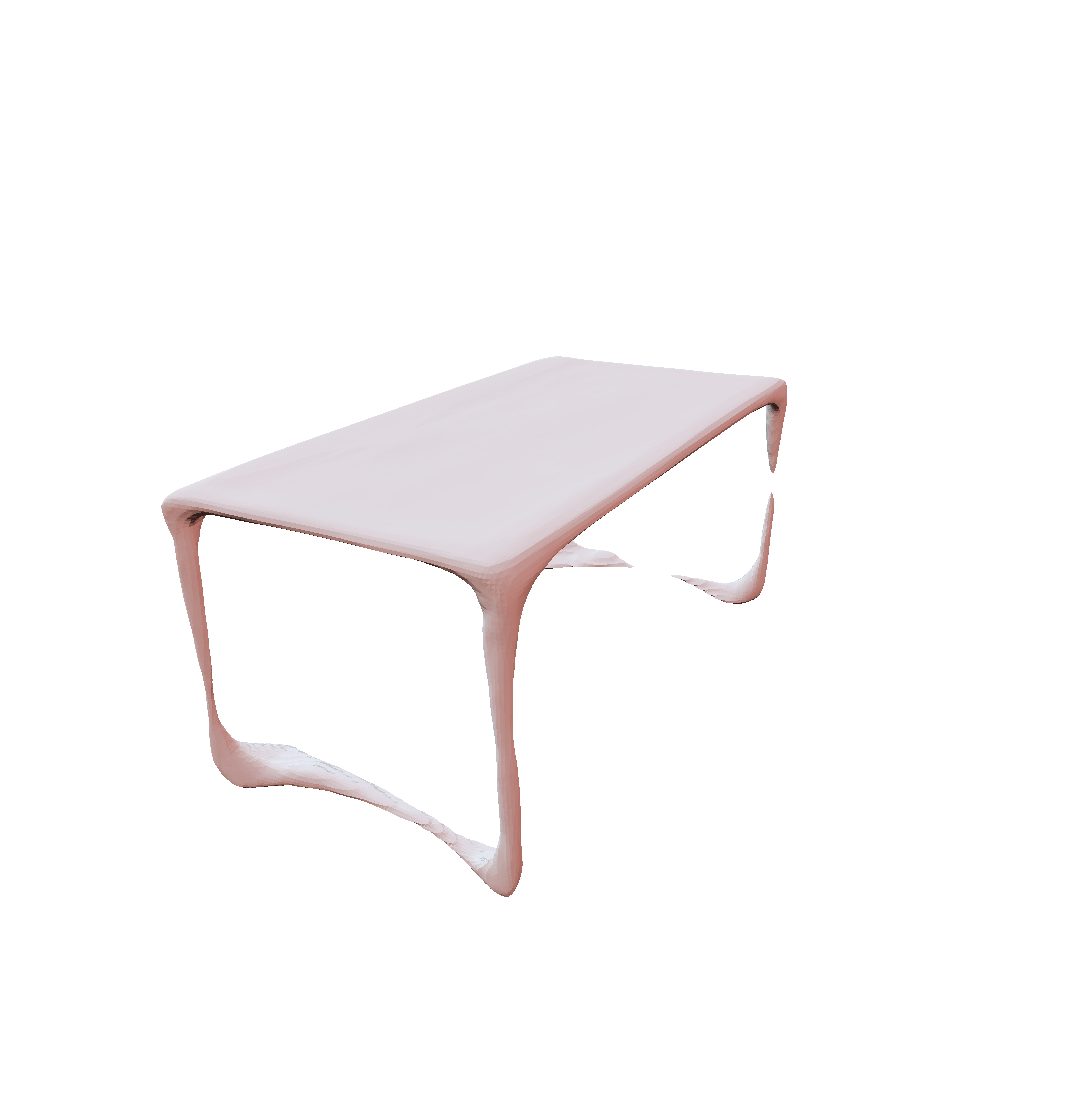}
	\end{subfigure} \\
	\begin{subfigure}{27.5mm}
		\centering
		\includegraphics[width=27.5mm]{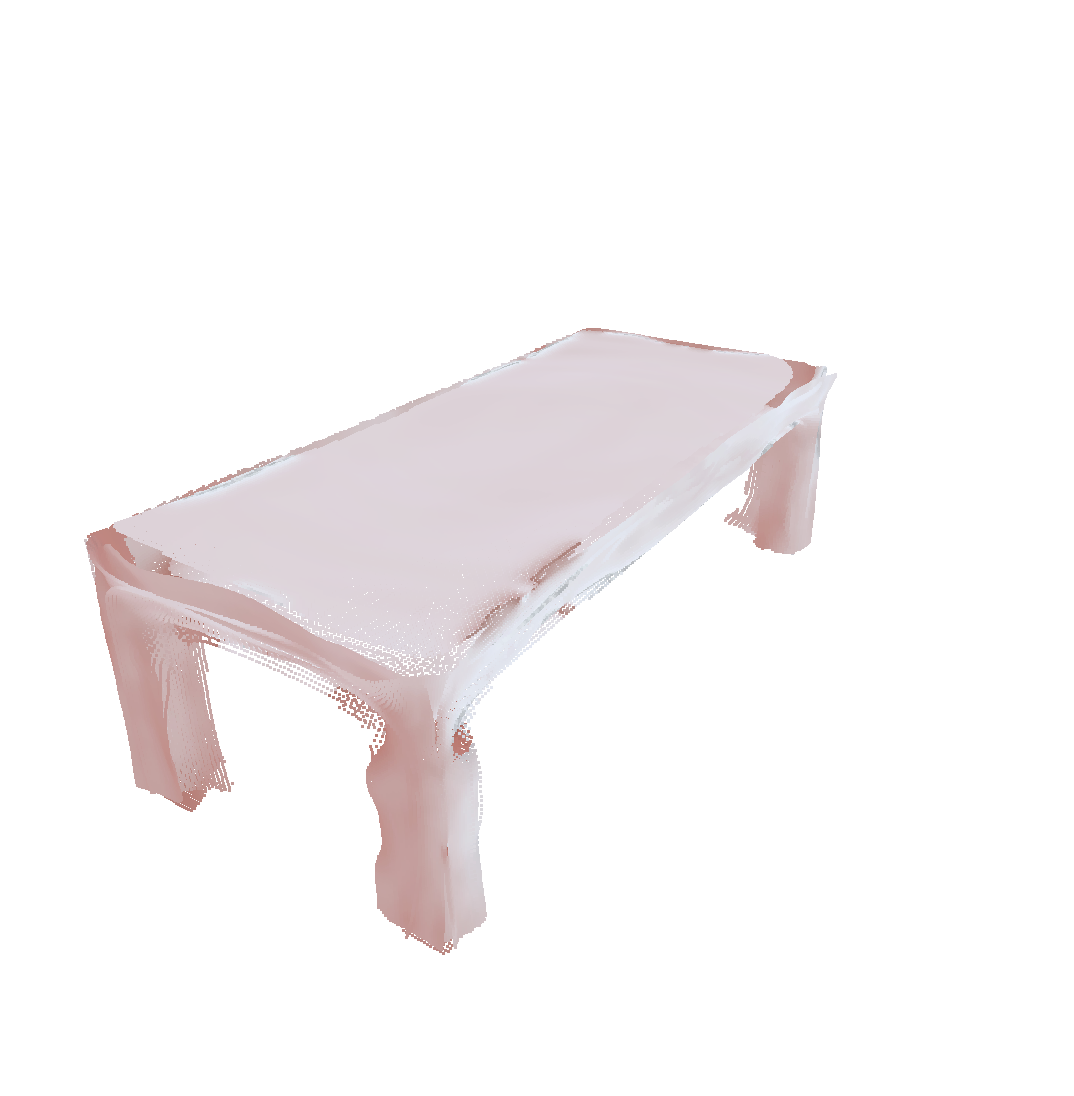}
	\end{subfigure}
	~
	\begin{subfigure}{27.5mm}
		\centering
		\includegraphics[width=27.5mm]{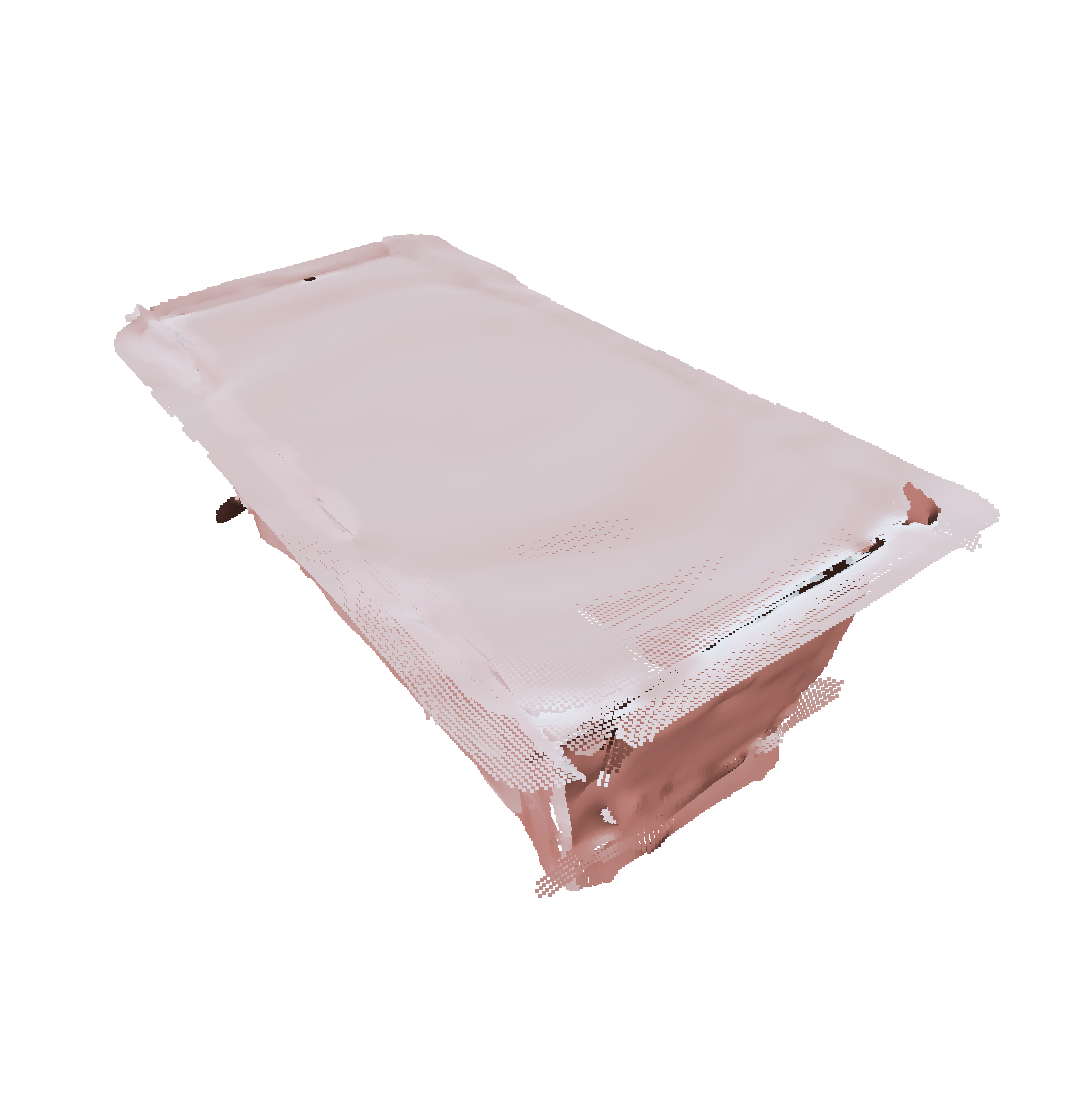}
	\end{subfigure} 
	~
	\begin{subfigure}{27.5mm}
		\centering
		\includegraphics[width=27.5mm]{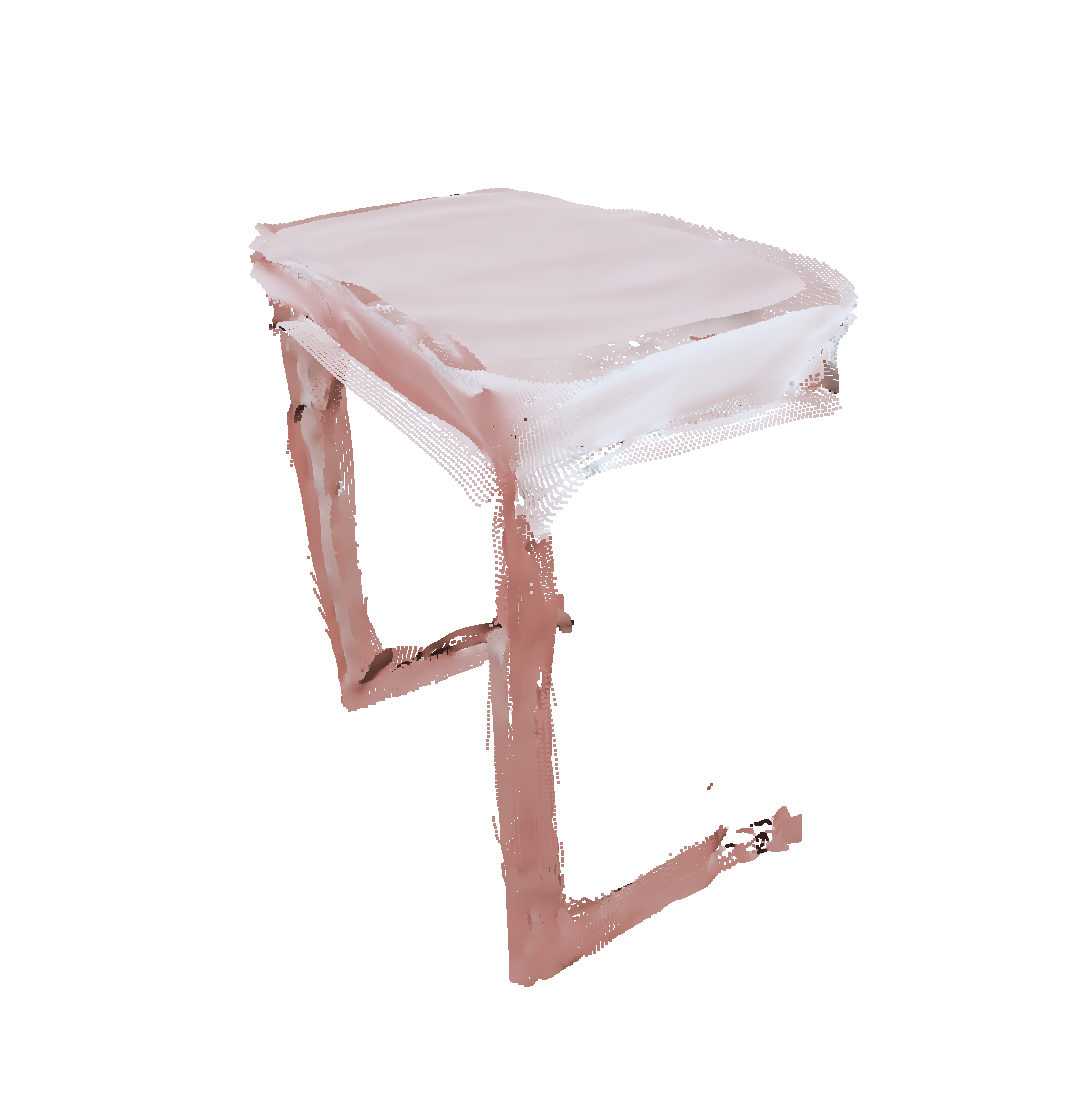}
	\end{subfigure} 
	~
	\begin{subfigure}{27.5mm}
		\centering
		\includegraphics[width=27.5mm]{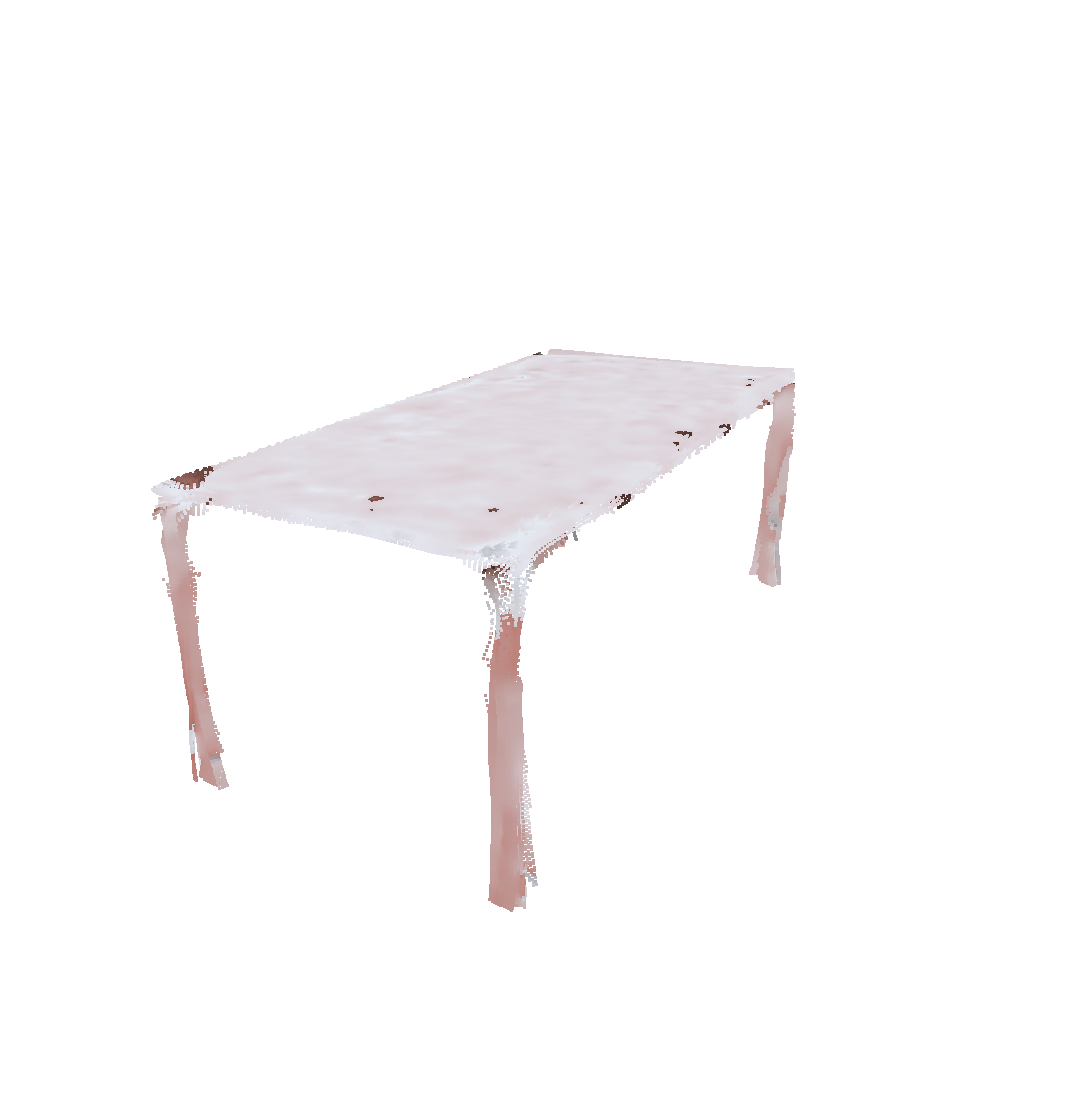}
	\end{subfigure} \\
	\caption{{\bf More Generative Results from Different Categories.} Results from Rows 1 to 4 correspond to {\bf Reference}, {\bf OF}, {\bf SDF}, {\bf PRIF - Mesh}.}
\end{figure}

\begin{figure}[!ht]
    \centering
	\begin{subfigure}{45mm}
		\centering
		\includegraphics[width=45mm]{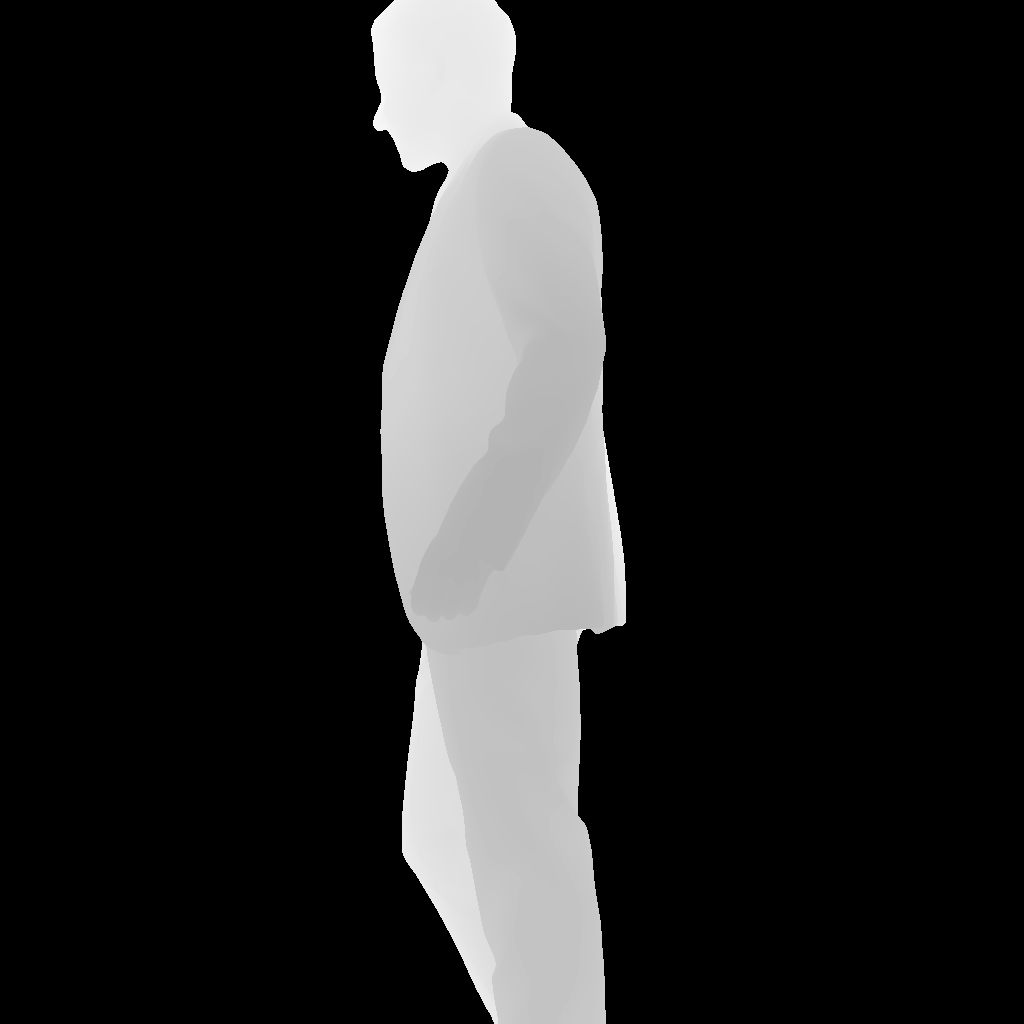}
	\end{subfigure}
	~
	\begin{subfigure}{45mm}
		\centering
		\includegraphics[width=45mm]{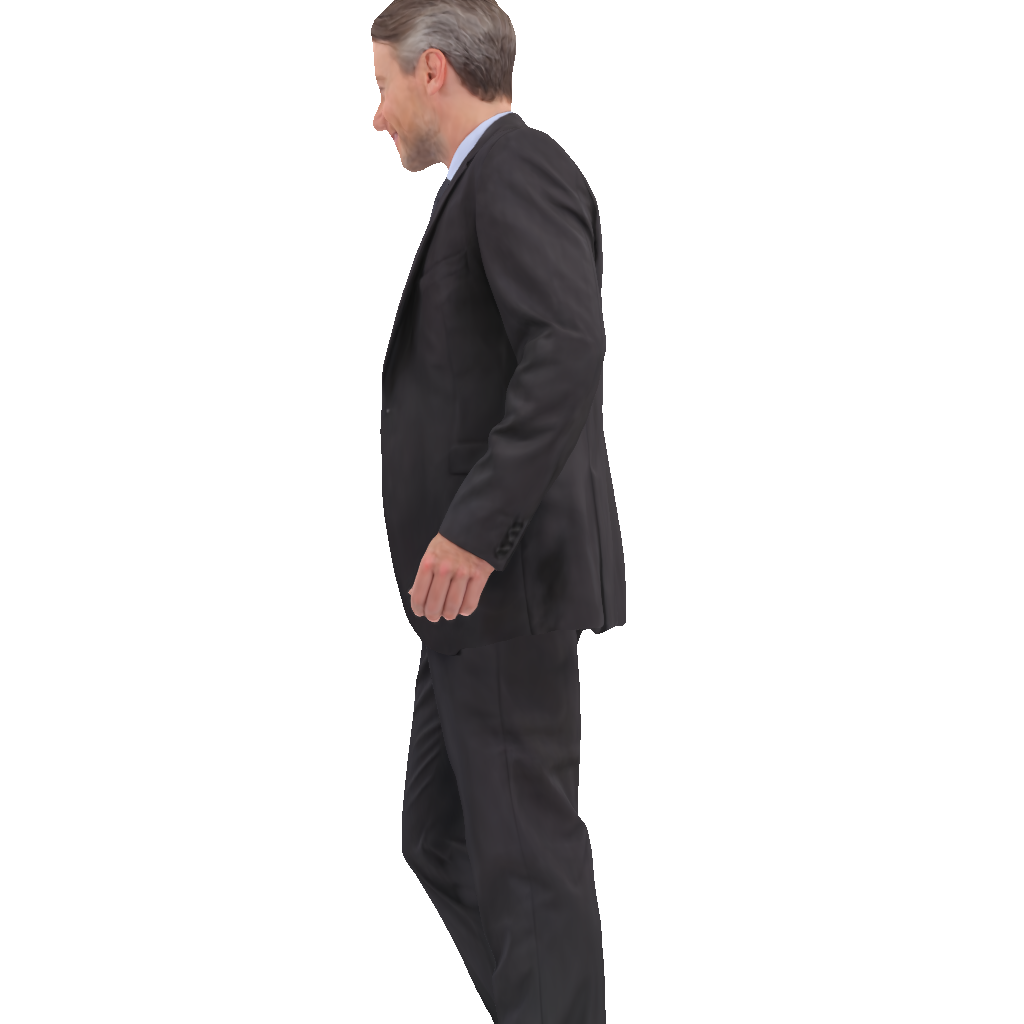}
	\end{subfigure} \\
	\begin{subfigure}{45mm}
		\centering
		\includegraphics[width=45mm]{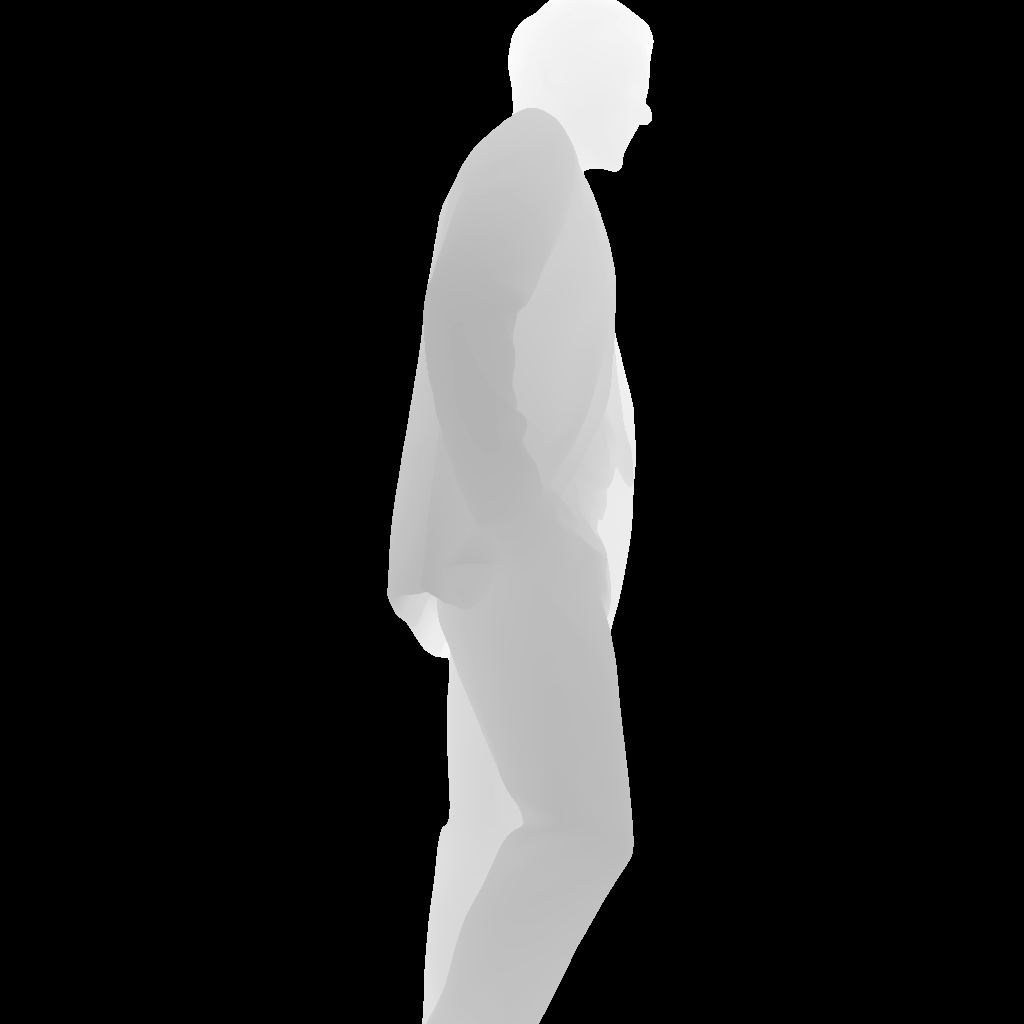}
	\end{subfigure} 	
	~
	\begin{subfigure}{45mm}
		\centering
		\includegraphics[width=45mm]{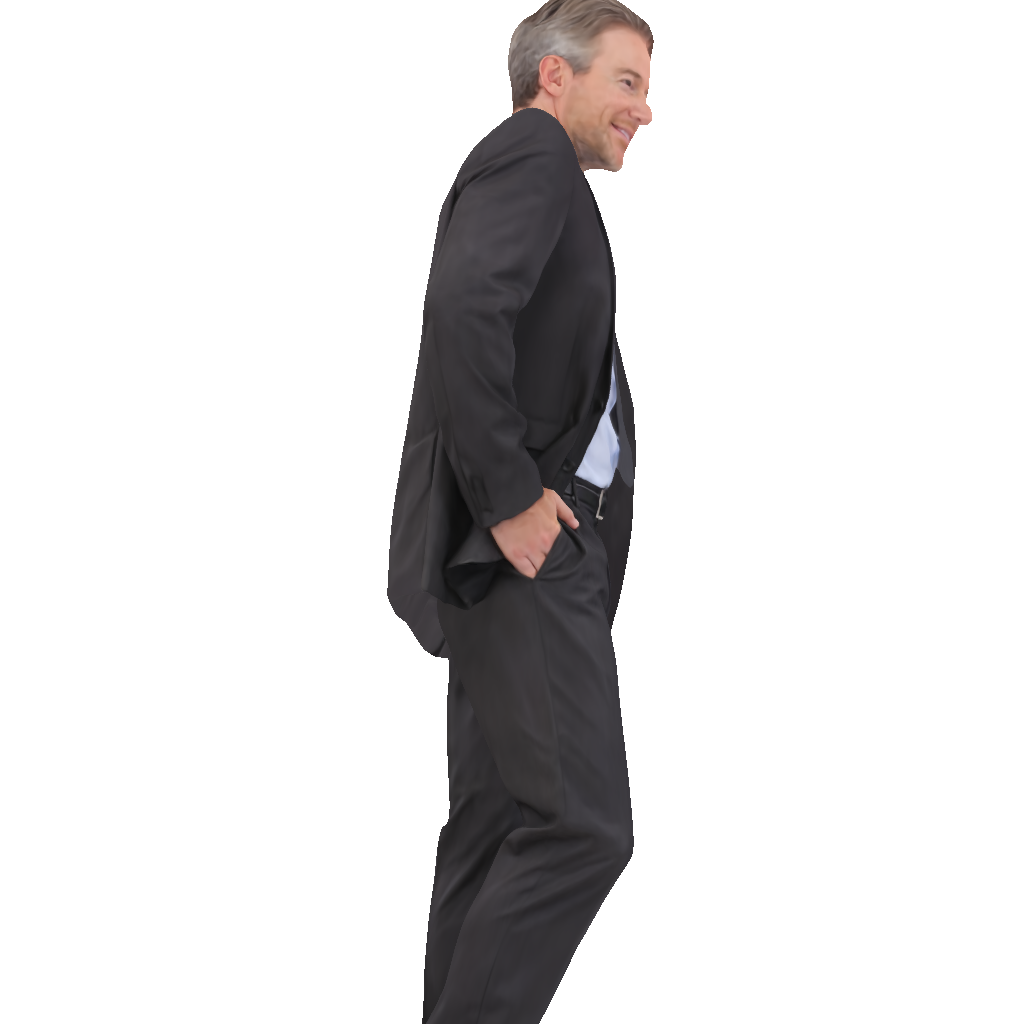}
	\end{subfigure} \\
	\begin{subfigure}{45mm}
		\centering
		\includegraphics[width=45mm]{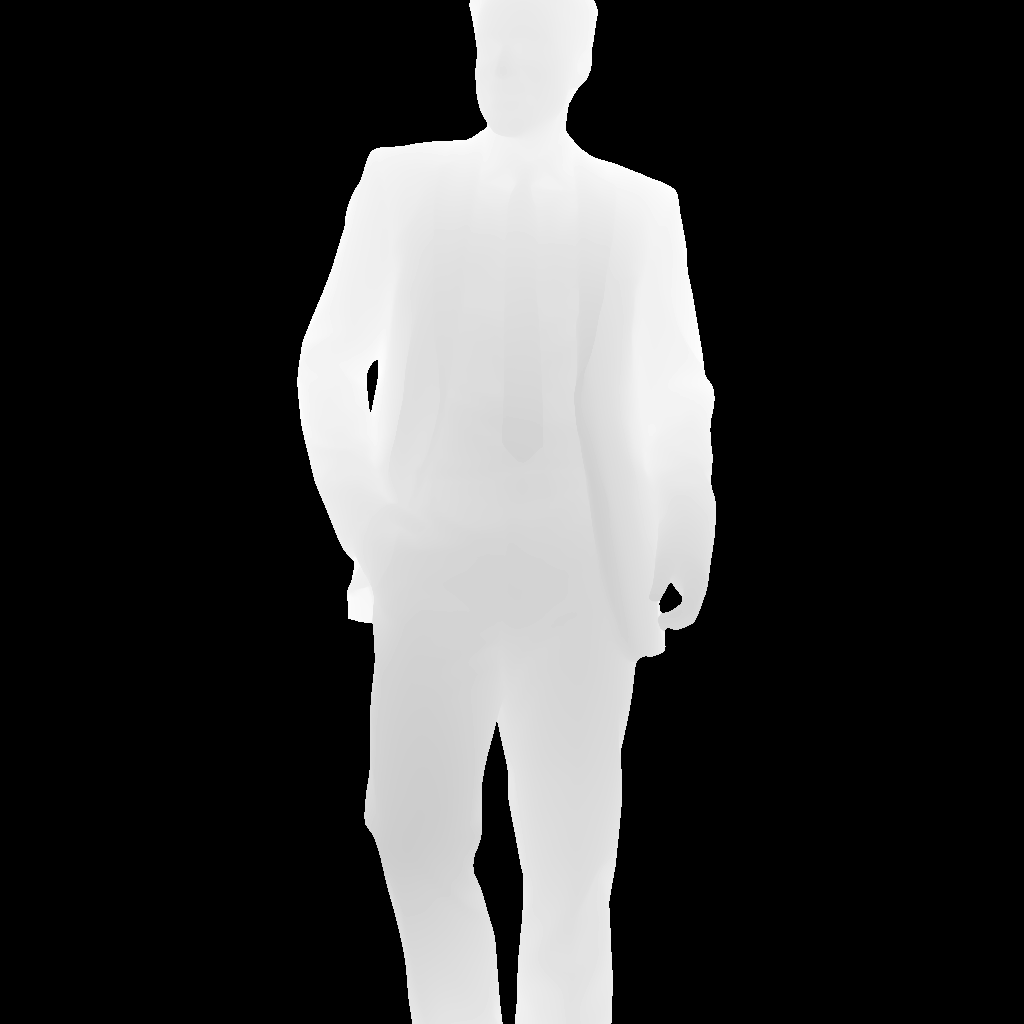}
	\end{subfigure} 
	~
	\begin{subfigure}{45mm}
		\centering
		\includegraphics[width=45mm]{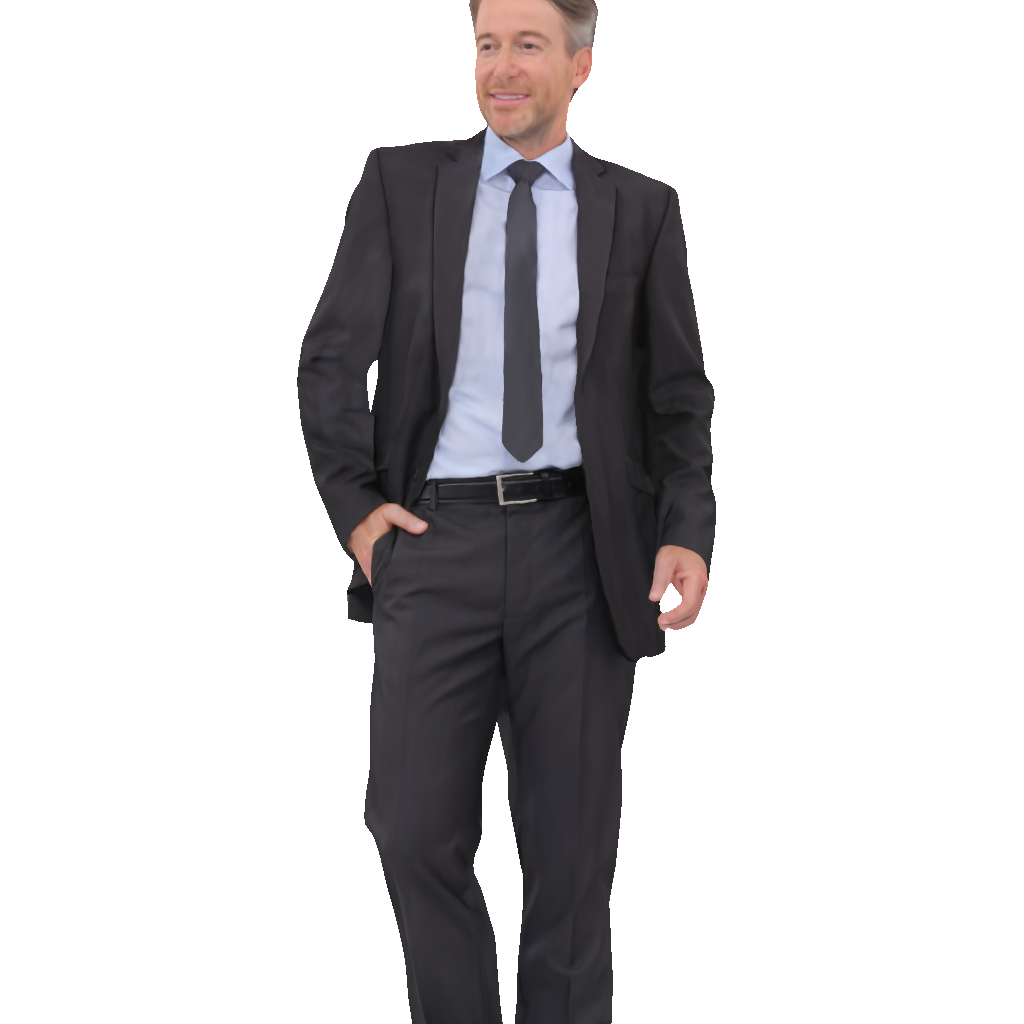}
	\end{subfigure} 
	\caption{{\bf More Color Rendering Results.} Depth (left) and Color (right) renderings from the same view point are shown.}
\end{figure}

\begin{figure}[!ht]
    \centering
	\begin{subfigure}{45mm}
		\centering
		\includegraphics[width=45mm]{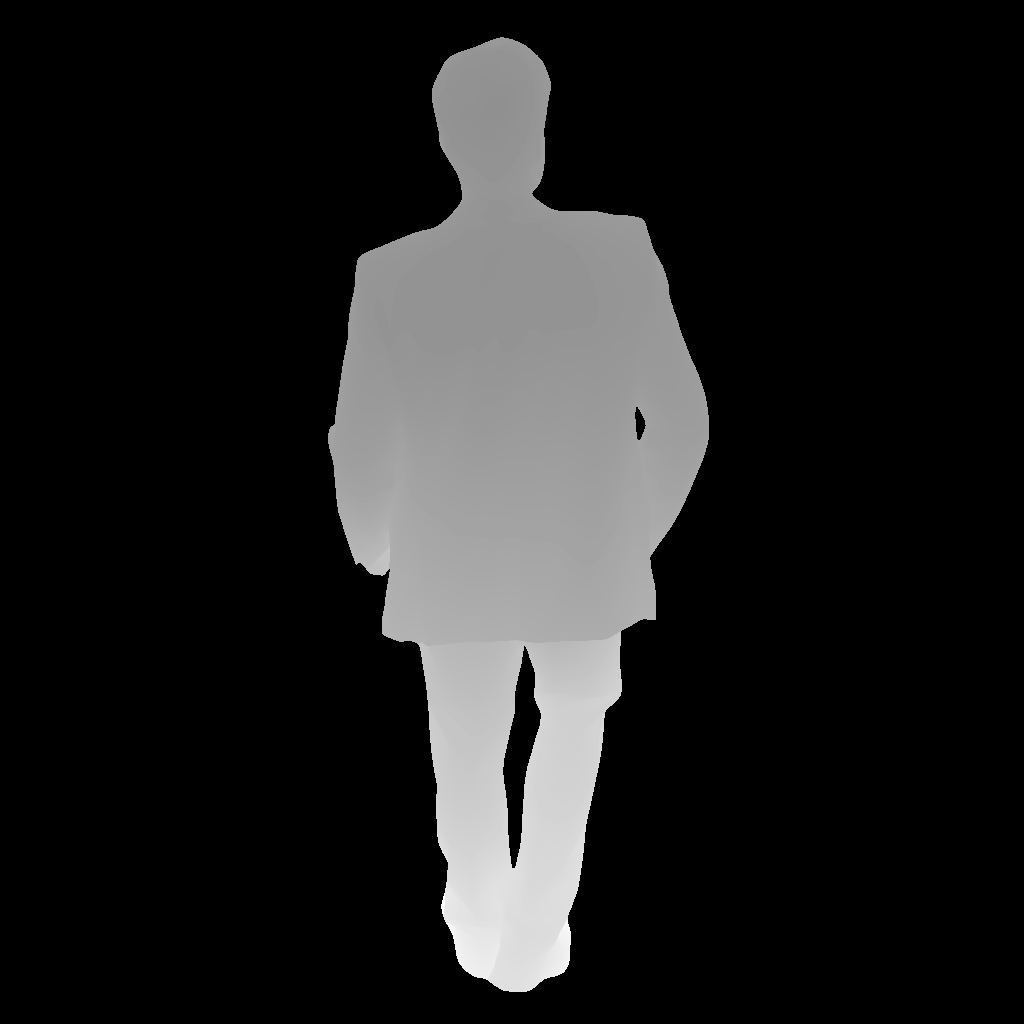}
	\end{subfigure}
	~
	\begin{subfigure}{45mm}
		\centering
		\includegraphics[width=45mm]{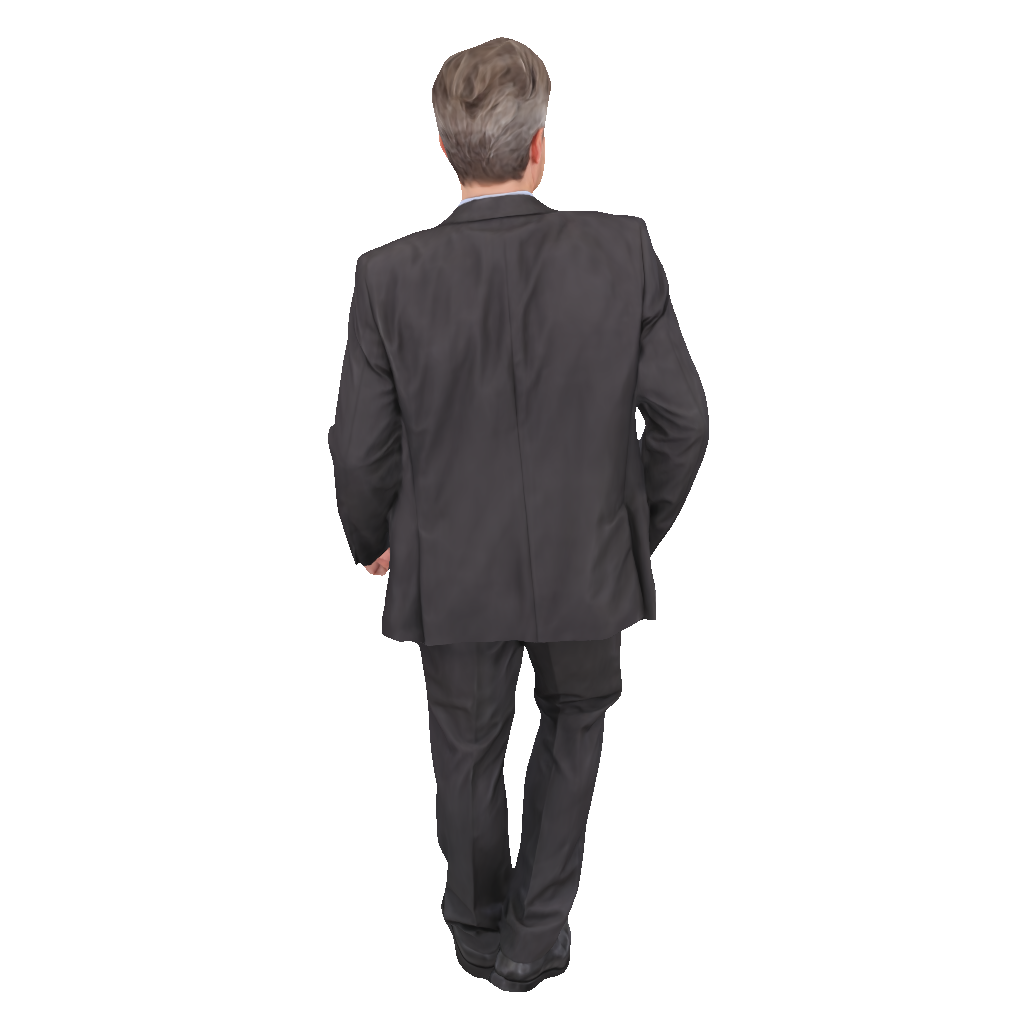}
	\end{subfigure} \\
	\begin{subfigure}{45mm}
		\centering
		\includegraphics[width=45mm]{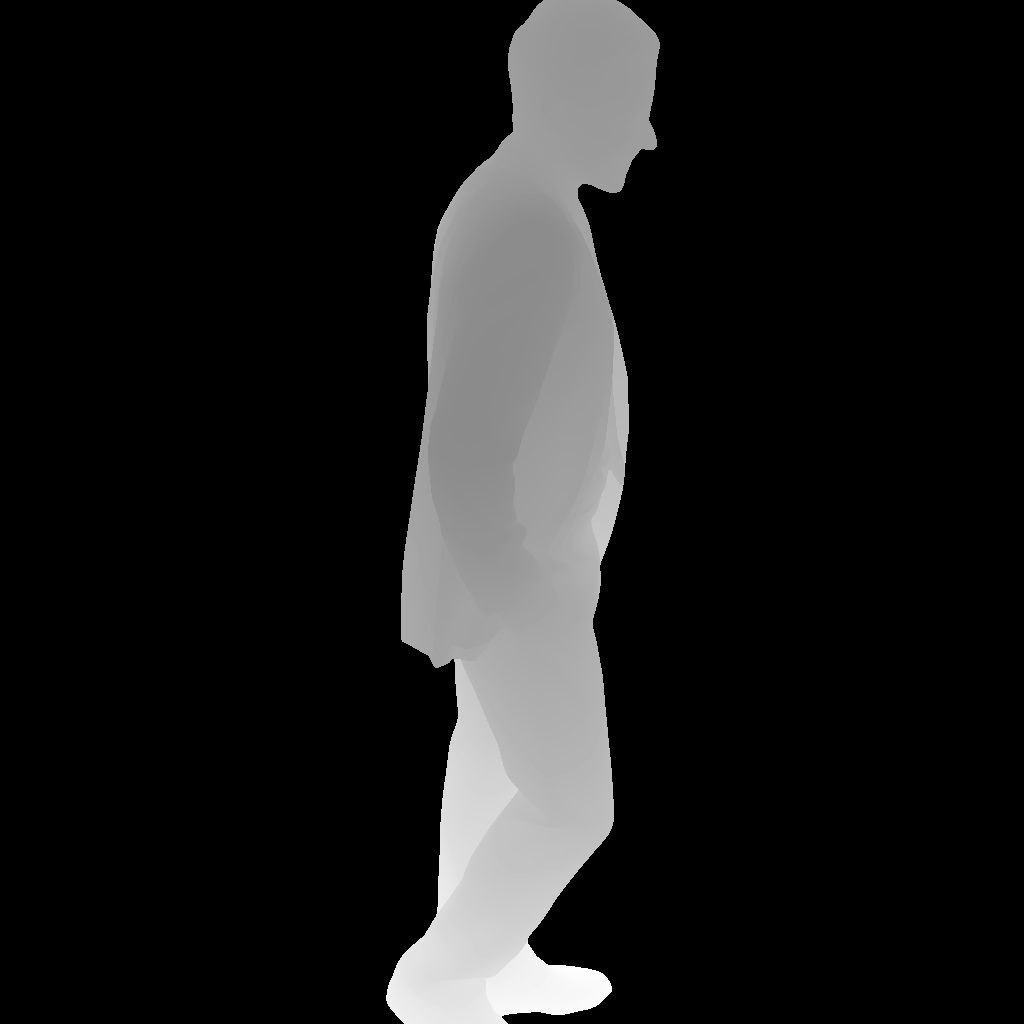}
	\end{subfigure} 	
	~
	\begin{subfigure}{45mm}
		\centering
		\includegraphics[width=45mm]{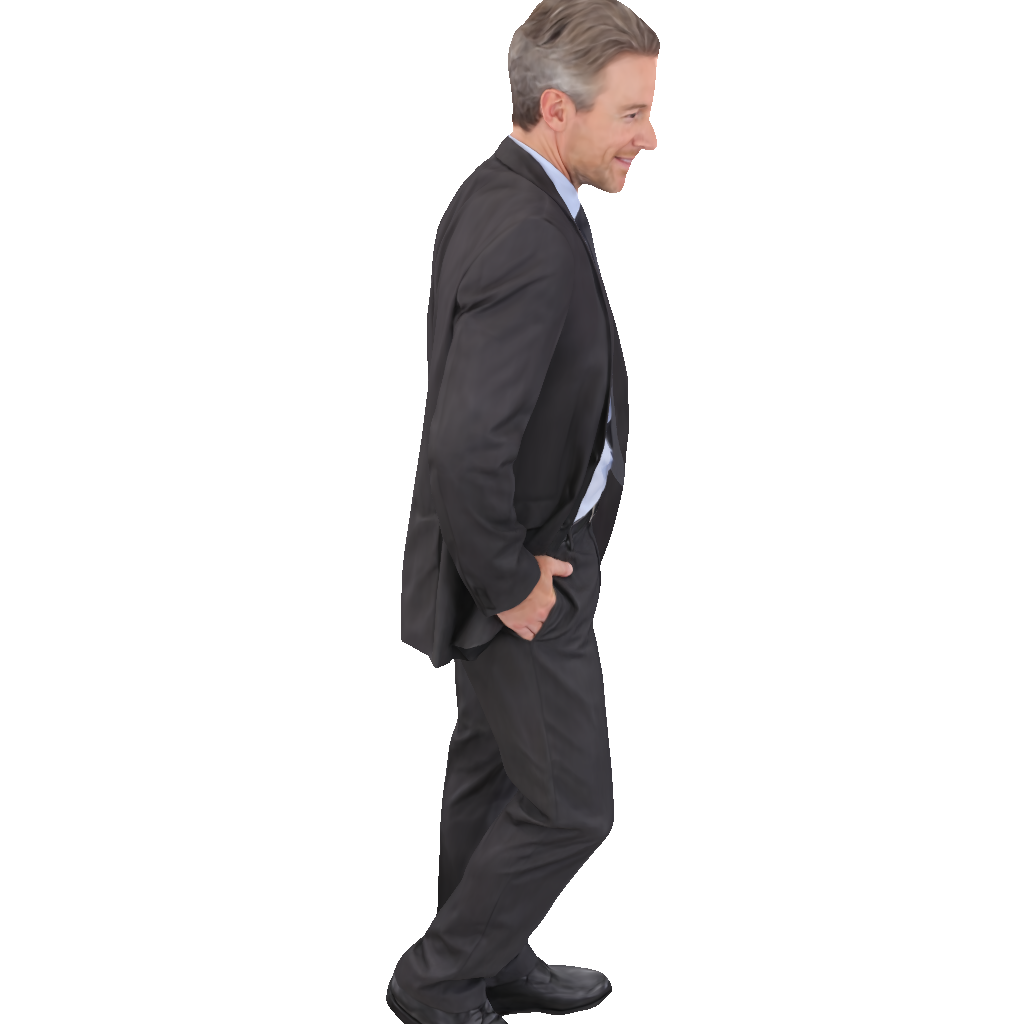}
	\end{subfigure} \\
	\begin{subfigure}{45mm}
		\centering
		\includegraphics[width=45mm]{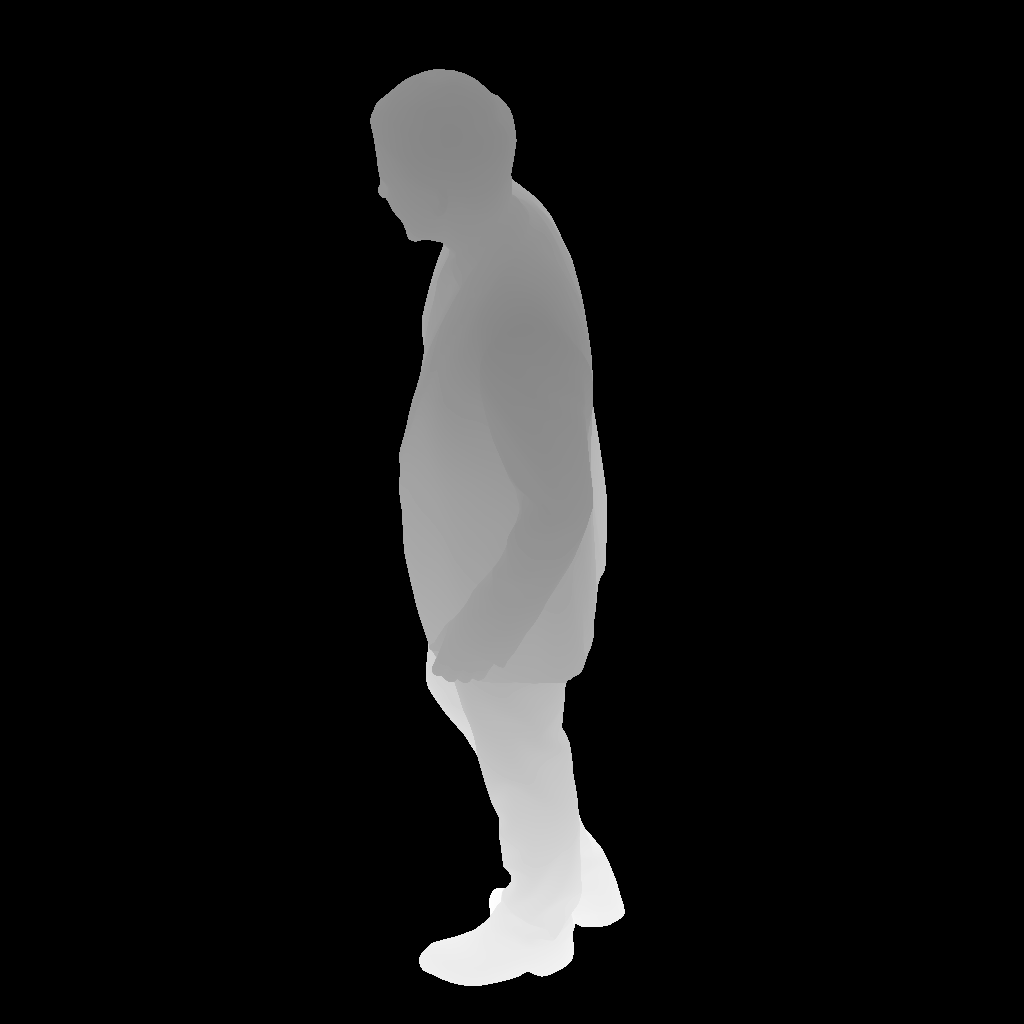}
	\end{subfigure} 
	~
	\begin{subfigure}{45mm}
		\centering
		\includegraphics[width=45mm]{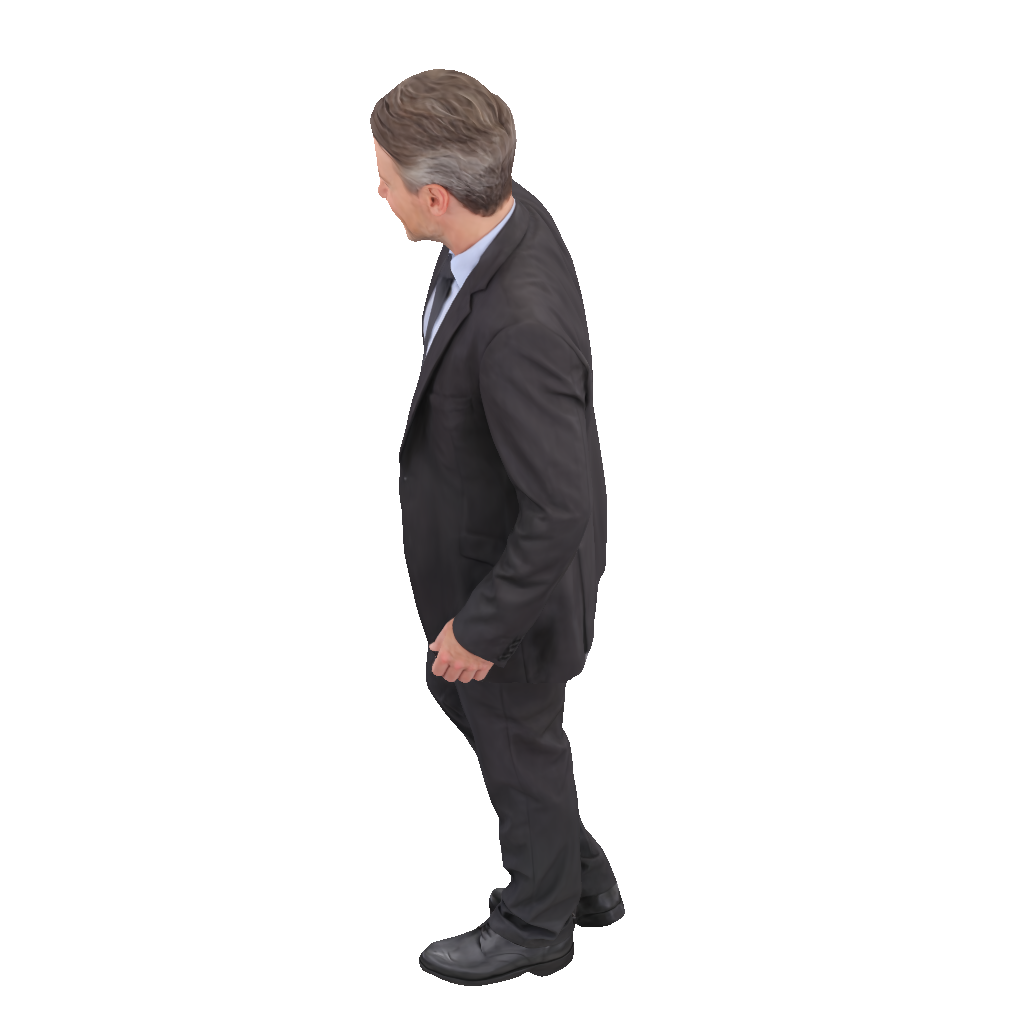}
	\end{subfigure} 
	\caption{{\bf More Color Rendering.} Depth (left) and Color (right) renderings from the same view point are shown.}
\end{figure}

\end{document}